%% file: thesis.tex
\documentclass[a4paper,12pt,times,numbered,print,index, openany]{PhDThesisPSnPDF}

\input{Preamble/preamble}

\input{thesis-info}

\ifdefineAbstract
\fi

\begin{document}

\frontmatter

\maketitle

\include{Declaration/declaration}

\include{Acknowledgement/acknowledgement}
\include{Abstract/abstract}


\tableofcontents

\listoffigures

\listoftables

\printnomenclature[7em]


\mainmatter

\include{Chapter1/chapter1}

\include{Chapter2/chapter2}
\include{Chapter3/chapter3}
\include{Chapter4/chapter4}
\include{Chapter5/chapter5}
\include{Chapter6/chapter6}
\include{Chapter7/chapter7}


\begin{spacing}{0.9}


\bibliographystyle{apalike}
\cleardoublepage
\bibliography{References/thesis} 



\end{spacing}


\begin{appendices} 

\include{Appendix1/appendix1}

\include{Appendix2/appendix2}

\end{appendices}

\printthesisindex 

\end{document}

%% file: thesis-info.tex
\title{Physics-informed Neural Networks for Encoding Dynamics in Real Physical Systems}

\author{Hamza Sharaf F Alsharif}

\dept{Department of Engineering}

\university{University of Cambridge}
\crest{\includegraphics[width=0.2\textwidth]{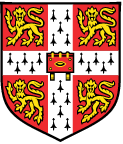}}

\degreetitle{Master of Philosophy}

\college{Girton College}

\subject{LaTeX} \keywords{{LaTeX} {MPhil Thesis} {Engineering} {University of
Cambridge}}

%% file: Declaration/declaration.tex

\begin{declaration}

I hereby declare that except where specific reference is made to the work of 
others, the contents of this dissertation are original and have not been 
submitted in whole or in part for consideration for any other degree or 
qualification in this, or any other university. This dissertation is my own 
work and contains nothing which is the outcome of work done in collaboration 
with others, except as specified in the text and Acknowledgements. This 
dissertation contains fewer than 15,000 words.


\end{declaration}

%% file: Acknowledgement/acknowledgement.tex

\begin{acknowledgements}      

I would like to thank my supervisor Professor Phillip Stanley-Marbell for having faith in me and accepting me into his group, providing academic guidance, support, and encouragement to pursue exciting work. Phillip has been an excellent supervisor and mentor who has given me the freedom to explore and work on topics that interest me whilst also advising me on the most optimal paths to take to achieve positive outcomes. I would also like to thank Professor Suhaib Fahmy for his eagerness in co-supervising my project, for the valuable discussions we've had on FPGAs and hardware, and for providing me with collaboration opportunities that I otherwise would not have had. I would like to thank James for his advice on setting up the parallel heating experiment, Janith for his advice on machine learning, and Chatura for his advice on helping me getting started at the early stages of my project. I would like to thank Vasilis and Orestis for being the first people I got in touch with in the group, and for getting me excited to pursue work here. I would like to thank Hamid and Divya for being the friendly faces from the group who I'd frequently see in the lab/department. The custom copper nib used as the soldering iron heat source was designed by Ady Ginn. The ethernet switch for the parallel heating experiment was provided by Barlow McLeod.

I would like to acknowledge the Saudi Arabian Cultural Bureau for funding my studies here at Cambridge.

Special thanks to the department crew --- Ismail, Alkausil, Ibtisam, Sohail, and Adil for the chill conversations in the department and for the banter over Wednesday cakes. Thank you to Yunwoo for always being around to talk and have dinner with during the late working hours in the department. Special thanks to Faris for our daily lunches and for putting up with me everyday. Big thank you to the Shbeeba/Shabashib especially Alwaleed, Hallamund, Bunyamin, Marwan, Abdulkarim, Moez, Muheeb, Radwan, Ahmad, and Ahmad. You guys made my experience here in Cambridge special so thank you bros. Big special thank you to Omar, my brother away from home. Thank you for being there for the good and hard times, God knows we've struggled through our degrees together. Thank you to Anas, my KFUPM roommate and now brother-in-law for always asking about me, for being there to talk to, and for helping me stay grounded. You've always been by my side as a dear friend, and now you're family. Thank you to my oldest friend Ahmad for always being there for me despite the distance, and for our late night conversations. God knows you've always been there for me from the very beginning brother.

Thank you to my sister Jumana and my brothers Mohammed and Faisal, for your continuous support and encouragement through all of this and for always praying for my success. 

Most of all, thank you to my parents for giving me everything I have in my life, for your consistent support, for the prayers and encouragement, and for your endless love. This dissertation is dedicated to you and I hope I can always make you proud.

Most importantly and before all of this, praise be to God, the Most Gracious, Most Merciful, and Most Benificent. All my successes, blessings, and good fortunes are from Him, and without His mercy, guidance, and support I am nothing.

\end{acknowledgements}

%% file: Abstract/abstract.tex
\begin{abstract}

Predictive data-driven models are gaining widespread attention and are being deployed in embedded systems within physical environments across a wide variety of modern technologies such as robotics, autonomous vehicles, smart manufacturing, and industrial controllers. However, these models have no notion or awareness of the underlying physical principles that govern the dynamics of the physical systems that they exist within. This dissertation studies the encoding of governing differential equations that explain system dynamics, within predictive models that are to be deployed within real physical systems. Based on this, we investigate physics-informed neural networks (PINNs) as candidate models for encoding governing equations, and assess their performance on experimental data from two different systems. The first system is a simple nonlinear pendulum, and the second is 2D heat diffusion across the surface of a metal block. We show that for the pendulum system the PINNs outperformed equivalent uninformed neural networks (NNs) in the ideal data case, with accuracy improvements of $18 \times$ and $6 \times$ for 10 linearly-spaced and 10 uniformly-distributed random training points respectively. In similar test cases with real data collected from an experiment, PINNs outperformed NNs with $9.3 \times$ and $9.1 \times$ accuracy improvements for 67 linearly-spaced and uniformly-distributed random points respectively. For the 2D heat diffusion, we show that both PINNs and NNs do not fare very well in reconstructing the heating regime due to difficulties in optimizing the network parameters over a large domain in both time and space. We highlight that data denoising and smoothing, reducing the size of the optimization problem, and using LBFGS as the optimizer are all ways to improve the accuracy of the predicted solution for both PINNs and NNs. Additionally, we address the viability of deploying physics-informed models within physical systems, and we choose FPGAs as the compute substrate for deployment. In light of this, we perform our experiments using a PYNQ-Z1 FPGA and identify issues related to time-coherent sensing and spatial data alignment. We discuss the insights gained from this work and list future work items based on the proposed architecture for the system that our methods work to develop.

\end{abstract}

%% file: Chapter1/chapter1.tex

\chapter{Incorporating Physics Knowledge in Computation}  

\ifpdf
    \graphicspath{{Chapter1/Figs/Raster/}{Chapter1/Figs/PDF/}{Chapter1/Figs/}}
\else
    \graphicspath{{Chapter1/Figs/Vector/}{Chapter1/Figs/}}
\fi

\section{Introduction} 
\label{sec:intro}
Physical computation refers to computation that affects and is affected by physical quantities in our natural world, such as temperature, pressure, velocity, etc.  This type of computation is prevalent within embedded systems (also referred to as cyber-physical systems) --- self-contained digital devices that are interconnected on a smaller scale than large workstations and servers, and typically process information from real environments. Embedded systems have become ubiquitous in modern society with computers being integrated into objects that we interact with on a daily basis, as well as in modern applications that are increasing in adoption such as autonomous vehicles, or digital manufacturing. However, the computers in these technologies lack a fundamental understanding of the physical nature of the systems and signals that they interact with. In other words, they lack computational abstractions~\cite{Rajkumar2010} that can be used to describe the environments that they exist within. The lack of computational abstractions for physical systems can be classified under four different categories of relevance to this dissertation:

\begin{enumerate}
    \item Contextual information about what physical signals represent~\cite{Abowd1999}.
    \item Information on the physical dimensions and units associated with signals~\cite{wang2019}.
    \item Existence and characterisation of noise~\cite{Meech2022}.
    \item Knowledge of the physical laws and relationships that govern physical quantities~\cite{lim2018}.
\end{enumerate}

We discuss these classifications in greater detail in Section~\ref{sec:lack_physics}. The focus of this MPhil dissertation is on the fourth item. Specifically, we are interested in investigating and developing methods for incorporating differential equations that govern dynamical systems, into predictive models for the purpose of deployment within real physical systems. There are tangible benefits to incorporating physics knowledge into predictive models deployed at the edge. The main benefit is the exploitation of the wealth of information that can be extracted from physical signals captured from different sensors. However, there are numerous challenges associated with developing physics-aware compute systems. This dissertation focuses on two of them.

The first is that it is more difficult to work with real-world data, as opposed to simulated or idealized data, due to the aleatoric uncertainty~\cite{willink_2013} arising from noise. In a real-world setting, noise can come from rapid fluctuations in the measurand, disturbances from the measurement environment, or as a characteristic of the measurement instrument. The amount of noise from each source varies depending on the system under investigation, and the effect it has on model predictions must be considered.

The second challenge relates to the viability of deploying machine learning (ML) models on the edge for embedded inference. ML models, and neural networks (NNs) in particular, tend to have thousands to millions of parameters. AlexNet~\cite{alexnet2012} for example has 60 million parameters and 650,000 neurons. In most cases it is infeasible to store such large models in embedded devices with limited amounts of memory. Additionally, the extensive amounts of processing required for inference raises the issue of power consumption, an important factor to consider for resource-constrained embedded systems running off of batteries.

These two challenges motivate the central research questions of this dissertation, which are as follows:

\begin{enumerate}
\item How well do physics-informed models perform on data captured from real physical setups in terms of predictive accuracy?

\item How viable is it to deploy physics-informed models on edge-based computers within a real physical setup for real-time prediction?

\end{enumerate}

To address the first question, the candidate model that we assess for this dissertation are physics-informed neural networks (PINNs)~\cite{raissi2019}. For the second question, we investigate the use of field-programmable gate arrays (FPGAs) as compute substrates for the models that we develop. 

\section{Dynamical Systems and Differential Equations}

\label{dynamical_systems_and_DEs}

Dynamics is the study of changing and evolving states. Consequently, dynamical systems are ones that are characterised by state evolution. The complex behaviours within these systems are governed and described by differential equations, which are derived through analyses of how these behaviours evolve over time and through space. Differential equations act as physical models for many of the phenomena that we observe in nature as well as in many engineering systems that humans develop.

Mathematically, a differential equation is a relationship between a function of interest and its derivatives with respect to one or more independent variables. Let $u$ be an arbitrary function and $x$ be an independent variable. We define the general form of a differential as follows:

\begin{equation}
\label{general ODE}
    \pazocal{D}(u) = f(x)
\end{equation}

where $\pazocal{D}$ is an arbitrary differential operator, and $f$ is a given function of $x$. A differential equation with the general form in Equation~\ref{general ODE}, where $u$ depends only on $x$ is called an ordinary differential equation (ODE). 

If $u$ depends on more than one independent variable and the equation includes partial derivatives with respect to each, then the equation is a partial differential equation (PDE). Let $\{x_1$, $x_2$, ..., $x_n\}$ be an arbitrary set of independent variables of length $n$. We define the first order general form of a PDE as follows:

\begin{equation}
    \frac{\partial u}{\partial x_1} + \frac{\partial u}{\partial x_2} + ... + \frac{\partial u}{\partial x_n} = f(x_1,x_2,..., x_n)
\end{equation}

Finding solutions to differential equations implies finding the unknown function $u$ in terms of its independent variables In physical contexts it represents finding the expression that describes the relationship between a set of physical quantities.

Differential equations can be classified as either linear or nonlinear. This classification changes how the equation is solved. Let $\{a_1(x)$, $a_2(x)$, ..., $a_n(x)\}$ be a set of coefficient functions of $x$. We define the general form of a linear differential equation as follows~\cite{boyce2017}:

\begin{equation}
    a_1 (x)\,u^{(n)} + a_2 (x)\, u^{(n-1)} + ... + a_n (x)\,u = f(x)
\end{equation}

Specifically, it must be the case that $u$ and all of its derivatives have a power of $1$, and that the coefficients $a_1(x), a_2(x), ..., a_n(x)$, and the function $f(x)$ depend only on the independent variable $x$. On the other hand, if a differential equation does not satisfy these conditions it is considered to be nonlinear. For given initial and boundary conditions, linear differential equations have known methods for finding analytical solutions such as separation of variables, integrating factors, or trial solutions. Nonlinear equations, are much more difficult to deal with and in most cases analytical solutions do not exist for them. In this case, one can only resort to numerical methods for finding solutions.

The following sections list examples of prominent linear and nonlinear differential equations in one dimension. It introduces the equations with brief mention of their physical contexts, and denotes the variable definitions. For all of the equations $x$ and $t$ represent position and time.

\subsection{Linear equations}

The wave equation~\cite{farlow1993partial_wave} for wave-bearing systems:

\begin{equation}
    \frac{\partial^2u}{\partial t^2} = c^2 \, \frac{\partial^2u}{\partial x^2}
\end{equation}

$u$ is the wave amplitude, and $c$ is a constant representing the wave speed.
\\
The diffusion equation~\cite{crank1975mathematics_diffusion} for diffusive processes such as heat conduction or Brownian motion:

\begin{equation}
    \frac{\partial u}{\partial t} = \alpha \, \frac{\partial^2 u}{\partial x^2}
\end{equation}

$u$ is the concentration of the diffusing quantity, and $\alpha$ is the diffusivity coefficient.
\\
The Laplace equation~\cite{evans2010partial} for equilibrium processes and potential field distributions:

\begin{equation}
    \nabla^2 u = 0
\end{equation}

$u$ is the physical quantity under investigation.
\\
The Poisson equation~\cite{evans2010partial}, which is the Laplace equation with a source term:

\begin{equation}
    \nabla^2 u = \sigma (\mathbf{x})
\end{equation}

$\sigma (\mathbf{x})$ is the function representing the source term.

\subsection{Nonlinear equations}

The nonlinear Schrödinger equation~\cite{Ablowitz2008NLS} for wave propagation in quantum mechanics systems, nonlinear optics, Bose-Einstein condensates, and dispersive water waves:

\begin{equation}
    i \, \frac{\partial u}{\partial t} + \Delta u + |u|^2 \, u = 0
\end{equation}

$u$ is the physical quantity under investigation, and $i$ is the imaginary unit where $i^2 = -1$.
\\
The viscous Burgers' equation~\cite{burgers1948}, a simplified model for viscous fluid flow:

\begin{equation}
    \frac{\partial u}{\partial t} + u \, \frac{\partial u}{\partial x} = v \, \frac{\partial^2u}{\partial x^2}
\end{equation}

$u$ is the fluid velocity, and $v$ is a constant representing the diffusivity coefficient.
\\
The Korteweg–De Vries (KdV) equation~\cite{Korteweg_de_vries1895} for shallow water waves and some other dispersive wave systems:

\begin{equation}
    \frac{\partial u}{\partial t} + 6u \, \frac{\partial u}{\partial x} + \frac{\partial^3u}{\partial t^3} = 0
\end{equation}

$u$ is the wave amplitude/displacement.

For predictive models that are deployed within dynamical systems, there is currently no support for programming constructs and abstractions for injecting knowledge about governing differential equations. Developing such methods would be a significant step towards computers that are more aware of the environments that they exist within, making them more robust, adaptable, and reliable.

\section{Lack of Physics Understanding in Computation} \label{sec:lack_physics}

As mentioned in Section~\ref{sec:intro}, we identify four ways in which physics knowledge is absent and can be incorporated within computation.

\subsection{Context of physical signals}
Physical computation deals primarily with data that is sampled from sensors and corresponds to signals captured from real-world environments. The data structures that represent signals are treated as raw collections of integer or floating-point numbers without any context associated to them. As a result, there is no way to specify whether a given data collection in memory corresponds to temperature measurements or accelerometer readings. We know that temperature signals in Kelvin cannot fall below $0$ and that accelerometer readings at rest on Earth should be $9.8$ $m/s^2$ downwards on average. This specification is known and accounted for by the programmer in the algorithmic treatment between different measurements, however if signal context specification was available as a feature of the compute system then algorithmic treatment would be implicit and automatic.

\subsection{Information on dimensions and units}
\label{unit_info}

Any measured physical quantity has dimensions associated with it, if the measurement is not a ratio or relative quantity. The unit of a measurement can be found using dimensional analysis. Many quantities that we can measure are derivations of the seven SI base quantities defined by the Système international~\cite{si-brochure}. Their units are therefore derivations of the SI base units. However, data streams coming from sensors do not contain information about units, and in fact conversion routines must be performed to obtain the unit-calibrated data from raw sensor outputs. Including support for dimensional information within computing would allow for sensory signals to have the proper unit association, regardless of whether the signals are base or derived quantities.

\subsection{Consideration of noise}
Noise is an inherent characteristic in the measurement of physical signals. It could be a dominant property in the measurand itself, arising from the measurement environment, or a prevalent feature of the measurement instrument. Commonly, the existence of noise is attributed to all three sources. Modern compute systems treat signals as collections of definite point values without consideration of the uncertainty arising due to noise. Signal uncertainty is an important factor to keep track of to ensure reliability in computation, especially in real-time systems where uncertain signals result in actuations that affect the real world. Laplace is a microarchitecture that provides bit-level representations for uncertain data types as well as microachitectural components for uncertainty propagation in arithmetic~\cite{tsoutsouras2022}. Such solutions are essential for the move towards trustworthy uncertainty-aware compute systems.

\subsection{Knowledge of physical laws and relationships}
\label{physical_laws_info}
Signals are interrelated with each other through conservation laws, or invariant relationships. As mentioned in section~\ref{dynamical_systems_and_DEs}, dynamical systems are governed by differential equations. Currently however, there is no support within embedded computers for their specification. For example, for the design of a digital controller that regulates the temperature across a flat surface - there is an absence of programmatic structures to specify that the thermal conduction regime is governed by the 2D heat equation~\cite{ahtt5p_heat_conduction}. The state equation structures pertaining to the controller design must therefore be explicitly hard-coded based on the programmer's understanding of the nature of the heating process. 

\section{Proposed Approach}

There have been proposed solutions for encoding physical knowledge based on the aforementioned points. A prominent one is Newton, a specification language for physics~\cite{lim2018}. It features a type system for describing physical signals and their units of measure, as well as the invariant relationships and laws between them.

Newton focuses on dimensional information and physical laws, outlined in Sections~\ref{unit_info} and~\ref{physical_laws_info}. It also serves as a front-end for a back-end process called dimensional function synthesis (DFS)~\cite{wang2019} --- which performs dimensional analysis to find the Pi groups of the Buckingham-Pi theorem~\cite{buckingham1914}. However, for the current implementation of Newton there is no functionality for incorporating differential equations.

One recent popular approach for incorporating differential equations for dynamical systems is physics-informed neural networks (PINNs)~\cite{raissi2019}. PINNs have gained significant attention in the recent scientific machine learning (SciML) literature. Section~\ref{sec:PINNs_background} outlines them in detail. However, the current literature only explores PINNs in their applications to simulations and multi-physics models --- which tend to run in high-performance computers. Instead, we focus on using PINNs as predictive model bases to be deployed within real physical systems. To do this we investigate systems that do not involve complicated dynamics and do not require convoluted experimental setups, yet still provide us data with physical significance. Therefore, we chose two systems: a swinging pendulum, and the heating of a metal block.

We also address the issue of choosing an appropriate compute substrate for the PINNs that we aim to deploy. For this, we propose to use field-programmable gate arrays (FPGAs), due to the numerous advantages that they provide which are discussed in Section~\ref{sec:FPGAs_background}.

\section{Structural Overview}
Chapter~\ref{chapter:background} provides background on PINNs and FPGA-based NN accelerators. Chapter~\ref{chapter:pendulum} presents the first case study system --- a swinging pendulum. We observe how well PINNs perform for predicting the pendulum's angle of oscillation based on a physical setup, and compare the predictive performance against a basic simulation of an ideal pendulum. Chapter~\ref{chapter:heat_diffusion} presents the second case study system --- heat diffusion across a metal block. We assess whether PINNs are able to predict the surface temperatures across the block as it is being heated, and present a method for denoising the thermal data. Chapter~\ref{chapter:hardware} discusses issues related to parallel sensing. Chapter~\ref{chapter:discussion} provides a discussion on the work presented and future research directions. Chapter~\ref{chapter:conclusion} summarises and concludes this dissertation.

%% file: Chapter2/chapter2.tex

\chapter{Background on PINNs and FPGAs}

\label{chapter:background}

\ifpdf
    \graphicspath{{Chapter2/Figs/Raster/}{Chapter2/Figs/PDF/}{Chapter2/Figs/}}
\else
    \graphicspath{{Chapter2/Figs/Vector/}{Chapter2/Figs/}}
\fi

\section{Introduction}

This topic of this dissertation lies at the intersection of two emerging sub-domains: scientific machine learning (SciML)~\cite{baker_workshop2019} and reconfigurable computing~\cite{tessier2015reconfigurable}. For the first we focus on Physics-informed Neural Networks (PINNs), and for the second on FPGAs as a compute substrate for model deployment. The following sections provide outlines on the focus areas, and shed light on some prominent related works.

\section{Physics-informed Neural Networks}

\label{sec:PINNs_background}

\begin{figure}[ht]
    \centering
    \includegraphics[width=\columnwidth]{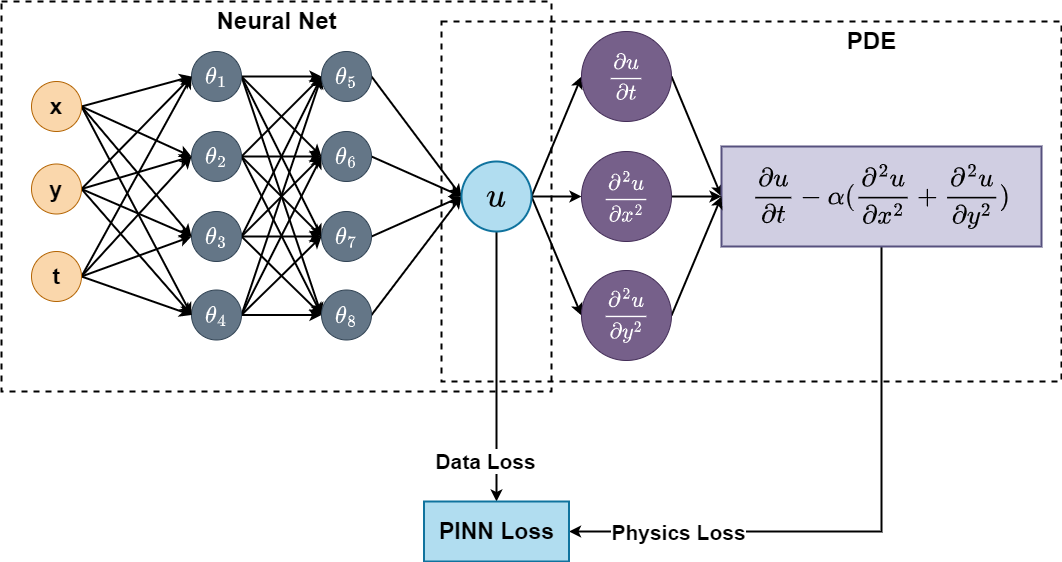}
    \caption{Example of a PINN architecture based on the 2D heat equation using trainable parameters $\theta_n$. The left dashed box shows the neural network which predicts the value of $u$ given the training points to produce the data loss term. The right dashed box shows the PDE residual corresponding to the heat equation, composed from the differential terms. The differential terms are obtained using automatic differentiation. The PDE residual forms the physics loss, which is the distinguishing component of PINNs.}
    \label{fig:pinn_arch}
\end{figure}

PINNs~\cite{raissi2019} are models that incorporate physical laws through the inclusion of terms that correspond to a system's governing differential equation into the loss function. Training a PINN corresponds to finding an approximate solution to the differential equation with the aid of data that is adherent to the equation's solution. From a mathematical perspective, the training process is posed as a constrained optimisation problem where the governing equation loss terms act as a soft constraint that restricts the solution space to physically-plausible ones. From an ML perspective, the physics loss acts as a regularisation term that allows the model to generalise using known physics, often enabling it to predict in regions outside of the training domain. Figure~\ref{fig:pinn_arch} shows an example architecture of a PINN based on the heat equation.

PINNs were first introduced in 1998 by Lagaris et al.~\cite{Lagaris1998}, although they resurfaced after their applicability to real problems was demonstrated by Raissi et al.~\cite{raissi2019}. Their recent widespread popularity in the literature can be attributed to the adoption of ML frameworks such as Tensorflow~\cite{tensorflow2016} and Pytorch~\cite{pytorch2019} which provide automatic differentiation engines (GradientTape and Autograd), as well as the rapid improvement in modern compute infrastructure for network training.

Aside from the promise of generalisability~\cite{karniadakis2021physics} that they offer through the incorporation of physical laws, there are additional reasons that motivate our decision to investigate PINNs as a candidate architecture. These are as follows:

\begin{enumerate}
    \item PINNs are resilient to noise~\cite{ClarkDiLeoni2023}. This makes them a promising choice for deployment within physical environments which are dominated by noise from different sources.
    
    \item PINNs can be trained with less and more sparse data~\cite{arzani2021}, and in some cases no data at all, with the exception of initial and boundary points~\cite{jin2021}. This is advantageous when the sensing capabilities or the amount of data that can be gathered is limited.
    
    \item PINNs are more computationally efficient than traditional numerical solvers, such as finite differences or finite elements, due to not requiring a computational mesh~\cite{Rohrhofer2021}. They are also often more efficient than ordinary feed-forward neural networks since they restrict the solution space to a subset of physically-plausible ones~\cite{karniadakis2021physics}.

    \item PINNs are convenient to implement and flexible, offering the capability of solving forward and inverse problems based on the sample problem formulation, and almost the same code implementation~\cite{cuomo2022scientific}.
\end{enumerate}

\subsection{Mathematical framework}
Let $x$ be a spatial input variable where $x \in \mathbb{R}^n$ and $n$ is the number of dimensions, $t$ be a temporal input variable, and $u$ be a function in terms of both inputs representing the solution of a PDE. PINNs come up with solutions to DEs by posing the following approximation:

\begin{equation}
    u(x, t) \approx f(x, t; \theta)
\end{equation}

where $f$ is a neural network approximation of $u$ based on network parameters $\theta$. Specifically, this approximate form is a solution to partial differential equations (PDEs) of the general form:

\begin{equation}
    u_t + \pazocal{N}(u) = 0
\end{equation}

where $u_t$ is a time derivative of $u$, and $\pazocal{N}$ is a nonlinear differential operator. Therefore, the PINN finds a solution based on the following approximation:

\begin{equation}
\label{eq:PINN_approx_form}
    f_t + \pazocal{N}(f) = 0
\end{equation}

The network loss $\pazocal{L}$ for a PINN can be found using the following equation:

\begin{equation}
    \label{eq:PINN_loss}
    \pazocal{L}(\theta) = \pazocal{L}_d(\theta) + \pazocal{L}_p(\theta)
\end{equation}

$\pazocal{L}_d$ is the data loss, which optimises to fit a set of data points that correspond to the true solution, usually at the initial or boundary conditions. It is traditionally the singular loss term that is used in neural networks. $\pazocal{L}_p$ is the physics loss which places a soft constraint on the network optimisation to obey the governing equation, and is consequently comprised of the equation's differential terms. Let $\{x_d, \,t_d\}$ be a set of $N_d$ data input points for a set of known output values $\{u_d\}$. Additionally, let $\{x_p,\,t_p\}$ be a set of collocation points within the problem domain that are used to evaluate $\pazocal{L}_p$. Therefore, $\pazocal{L}_d$ and $\pazocal{L}_p$ are mean squared error losses denoted by Equations~\ref{eq:data_loss} and~\ref{eq:physics_loss}.

\begin{equation}
\label{eq:data_loss}
    \pazocal{L}_d(\theta) = \frac{1}{N_d}\sum_{i=1}^{N_d}|f(x_d^i, t_d^i;\theta) - u_d^i|^2
\end{equation}

\begin{equation}
\label{eq:physics_loss}
    \pazocal{L}_p(\theta) = \frac{1}{N_p}\sum_{i=1}^{N_p}|f_t(x_p^i, t_p^i; \theta) - \pazocal{N}(f(x_p^i, t_p^i; \theta))|^2
\end{equation}

The squared terms in Equation~\ref{eq:physics_loss} correspond to the left-hand side of Equation~\ref{eq:PINN_approx_form}. The differential terms in Equation~\ref{eq:physics_loss} are computed using automatic differentiation.

\subsection{Related work}

The recent literature on PINNs, and physics-informed machine learning (PIML) in general, is wide and comprehensive, as the methods have successfully been shown to perform well in many different scientific domains. Some of the highlighted reviews on physics-informed machine learning include Karniadakis et al.~\cite{karniadakis2021physics}, Baker et al.~\cite{baker_workshop2019}, and Hao et al.~\cite{hao2023physicsinformed}. Cuomo et al.~\cite{cuomo2022scientific} provide an exhaustive review that focuses specifically on PINNs.

Cai et al.~\cite{cai2021heat_transfer_pinns} review the application of PINNs to a number of inverse problems for heat transfer. They show that PINNs show promising predictive capabilities in a simulation setting for forced and mixed heat convection past a cylinder, as well as for two-phase Stefan problems~\cite{stefan1891theorie}.

A few other scientific domains where PINNs have been investigated include fluid mechanics~\cite{cai2021physics}, power systems~\cite{misyris2020}, cardiac elecrophysiology~\cite{sahli2020physics}, fiber optics~\cite{jiang2022}, laser metal deposition~\cite{li2023_LMD_pinn}, and electromagnetics~\cite{khan2022_EM}. For all of these works, the authors have shown that the embedding of domain-specific governing equations into the training process has proven to improve generalised inference, with better predictions in regions with less or no data.

In addition to surveying PIML methods and providing a categorisation of the different ways in which physics is incorporated into the ML workflow (through the model architecture, loss function, and through hybrid approaches), Ben Moseley's thesis~\cite{moseley2022a} tackles the challenge of using PIML techniques as tools to solve real-world large-scale scientific problems. Moseley investigates the performance of PIML methods by posing the following tasks:

\begin{enumerate}
    \item Using a variational autoencoder (VAE)~\cite{kingma2022autoencoding} to find the physical factors that relate to lunar thermodynamics from temperature measurements of the moon's surface.
    \item Filtering out noise from low-light images for visualisation of permanently-shadowed regions on the lunar surface using a custom-built PIML algorithm.
    \item Simulating complicated seismic wave phenomena using different physics-informed deep learning models.
    \item Investigating the scalability of PINNs to problems with large domains and high-frequency components.
\end{enumerate}

For these tasks, Moseley shows that the fine-tuned PIML techniques generally perform well in terms of their ability to learn physical processes and solve complicated scientific problems. There are still challenges, especially relating to scaling the methods to physical systems with high-frequencies. To alleviate this issue, Moseley et al. proposes finite-basis PINNs (FBPINN)~\cite{moseley2021finite}, a domain decomposition approach for solving large-scale differential equation problems.

\section{FPGAs and Accelerator Architectures}
\label{sec:FPGAs_background}

Field-programmable gate arrays (FPGAs) are reconfigurable computer architectures made up of large collections of digital logic gates, where the connections between the gates can be customised for specific applications. FPGA reconfigurability implies that their hardware designs (and thus the applications that they run) can be configured as many times as is required by the user, and often during runtime through partial reconfiguration (PR)~\cite{vipin2018}. FPGAs have been gaining increased popularity in their deployment as accelerator architectures over the recent years, due to the advantages they provide in running hardware-optimised algorithms for domain-specific computing. These advantages include:

\begin{enumerate}
    \item Design of high throughput architectures, due to the inherent parallelism from synthesizing processing elements across the space of the FPGA's hardware resources.
    
    \item Reconfigurability which allows for rapid design prototyping, enables future hardware design updates, and provides the feature of switching between hardware applications (through partial reconfiguration~\cite{vipin2018}).
    
    \item Low processing latency from running directly on hardware, rather than passing through software abstraction levels such as an operating system.

    \item Energy-efficient architectures, due to the flexibility in synthesizing hardware to perform only the necessary operations with as few overheads as possible. Qasaimeh et al. show that FPGAs outperform CPUs and GPUs for more complicated vision tasks with a 1.2-22.3x energy reduction per frame~\cite{Qasaimeh2019}.
\end{enumerate}

In recent years there has been a shift away from general-purpose computing towards domain-specific architectures~\cite{patterson2018}, with highlight examples such as Google's Tensor Processing Unit (TPU) for accelerating deep neural networks (DNNs)~\cite{Jouppi2017}. This is mainly attributed to the struggle of modern transistors to keep up with Moore's law~\cite{eeckhout2017moore} and the end of conventional Dennard Scaling~\cite{kuhn2009}. Therefore, the new approach is to design accelerator architectures for performing specialised tasks rather than relying on general-purpose CPUs. FPGAs thrive in this new specialisation-focused compute paradigm, and are therefore becoming more mainstream both in research and in commercial settings.

\subsection{Neural networks on FPGAs}

This section highlights some of the relevant research involving NN acceleration on FPGAs. Implementing neural networks on FPGAs is not a novel idea with one of the earliest implementations, GANGLION, dating back to 1992~\cite{cox1992_ganglion}. GANGLION is a fully-connected network architecture with 12, 14, 4 input, hidden, and output layer neurons implemented on 40 Xilinx XC3000 series logic cell array (LCA) chips (36 XC3090s and 4 XC3042s), and 18 2KB PROMs as lookup tables. This is in stark contrast to modern FPGA NN designs, which are implemented on a single chip. The GANGLION architecture used different techniques to efficiently conserve hardware, as well as to increase data throughput. These include splitting 8-bit multiplications into summations of two 8-bit by 4-bit partial products, using carry-save and three-to-two reduction adders to avoid long carry propagation datapaths, and scaling down 20-bit accumulation results to 11-bits for the activation function inputs. For image segmentation tasks it achieves a data processing rate at the inputs of 240 MB/s corresponding to 20 million pixels per second.

Early FPGA NN implementations~\cite{Arroyo1999,  bade1994, cloutier1996, Ferrer2004, van_daalen1993}, faced similar problems: limited hardware resources due to the low number of configurable logic blocks (CLB) of early FPGAs, complicated and large multiplication circuitry, and in some cases insufficient routing control within high-level design software. Since then FPGA designs have advanced tremendously, with modern implementations accommodating for state-of-the-art NN architectures with millions of parameters (AlexNet~\cite{alexnet2012}, VGG-16~\cite{simonyan_vgg16}, GoogLeNet~\cite{szegedy_googlenet2015}, etc.), and can perform hundreds of billions of operations per second (OPS)~\cite{Guo2019}. Some designs reach peaks of around 40 teraOPS (TOPS)~\cite{moss2017}.

FPGA NN inference accelerators focus on optimising for speed and energy efficiency. Optimizing for speed enables for accelerators that can perform many inferences with real-time performance. Energy efficiency is critical given that accelerators often run either within small-scale embedded systems where the energy cost is constrained, or within large scale data centers serving thousands of clients, where the energy cost is multiplicative. For a given FPGA, an accelerator design is constrained by the amount of available hardware resources, so area-efficient designs are favourable. 

The following sections highlight two techniques for optimizing NN architectures for FPGAs: quantization and weight pruning. The final section presents FINN~\cite{Blott2018}, an end-to-end framework for NN deployment on FPGAs.

\subsubsection{Quantization}
Quantization aims at reducing the size of the computation units, as well as the memory and bandwidth requirements by narrowing the bit-width of the data, i.e. the weights and activations. The trade-off here is between the accuracy degradation due to the loss of precision, and the performance gain due to the quantization scheme. 

Quantization is used in conjunction with fixed-point rather than floating-point data representations. Quantized NN architectures often use fully 16-bit layers~\cite{Guan2017, Shen2018, fpgaConvNet_Venieris2016, Xiao2017, Zhang2016}, fully 8-bit layers~\cite{Ding2023, angel-eye_Guo2018, Jia2022}, or a mix of 8-bit and 16-bit layers~\cite{Liu2016, Ma2017, Suda2016}. Additionally, many accelerators~\cite{Liang2018, moss2017, Nakahara2017, Zhang2022} implement Binarized NN (BNN) architectures with 1-bit representations for all of the layers~\cite{Hubara2016}. 

Qiu et al. introduce a dynamic approach to quantization where different layers and feature map sets can have different fractional bit-lengths, based on an optimal quantization configuration strategy~\cite{Qiu2016}. However, for a given configuration the bit-widths are fixed, and in their experiments they use 16-bit, 8-bit, and a mix of 8 and 4 bits for the weights in the convolutional (CONV) and fully-connected (FC) layers respectively. They show that using dynamic precision with 8-bits can restore the top-1 and top-5 accuracies to values marginally less than the single-precision floating-point benchmark (1.52\% loss for top-1, and 0.62\% for top-5 accuracies), as opposed to static precision 8-bits which suffered from high accuracy degradation.

Abd El-Maksoud et al. use a 4-bit weight quantization in their GoogLeNet FPGA accelerator~\cite{AbdElkMaksoud2021}. They use incremental network quantization (INQ)~\cite{zhou2017incremental}, a post-training quantization method which is partitions the weights into two groups, one to be quantized and the other to be retrained. The two groups are switched afterwards. This is done iteratively until the accuracy requirement is met. Abd El-Maksoud et al. show that using INQ in addition to weights pruning allows them to reduce the CNN model by 57.6x, with their accelerator achieving a classification rate of 25.1 FPS with 3.92 W of power~\cite{AbdElkMaksoud2021}.

\subsubsection{Weight pruning}

Denil et al. have shown that NN models are over-parameterized, and that in many cases only a few of the weights are required to predict all the rest~\cite{Denil2013}. For NN models that are to be deployed in embedded systems, this redundancy results in a waste of storage and computation requirements. Weight pruning, or weight reduction, seeks to resolve the over-parameterization of NN models by removing zero or small absolute value weights. This results in compact NN models that require less storage and utilize less hardware resources, resulting in lower power consumption. It also frees up space for additional processing for optimizing other areas of NN accelerators. Song et al. present a three-step process for pruning NNs by first training the network to identify the important connections, removing the unimportant connections, and then retraining the network after the parameter reduction~\cite{Song2015}. They show that their method achieved a $9 \times$ and $13 \times$ reduction in parameters for AlexNet~\cite{alexnet2012} and VGG-16~\cite{simonyan_vgg16} respectively, without loss of accuracy. Denton et al. apply compression techniques that exploit the linear redundancy within CNNs to reduce the number of weights~\cite{denton2014}. They use matrix singular value decomposition (SVD) to approximate high order tensors into tensors with lower dimensions. They achieved a $2-3 \times$ memory reduction for the first two layers of a CNN, and $5-13 \times$ for the fully conected layers.

\subsubsection{FINN framework}

\label{sec:FINN}

FINN\footnote{Named after the cat of one of the authors.} is a full end-to-end framework for NN deployment on FPGAs~\cite{Blott2018}. It provides features for a design-space exploration of mixed bit precisions for network weights, biases, and activations. The front-end for FINN is a Pytorch library named Brevitas~\cite{brevitas}, that supports post-training quantization and quantization-aware training. A user can set different bit-widths for different layers and activations by defining a quantized version of a given network. The user trains the quantized network, and benchmarks its accuracy against the floating-point version, tweaking the bit-widths until the network accuracy is sufficient. The quantized network can then be exported as an ONNX object~\cite{bai2019}, which can then be converted into a dataflow accelerator hardware design using the FINN compiler. Blott et al. have shown that NNs compiled by FINN have achieved 5 TOPS on an embedded platform, and 50 TOPS for a datacenter implementation~\cite{Blott2018}.

%% file: Chapter3/chapter3.tex

\chapter{Predicting the Oscillation Angle of a Swinging Pendulum}

\label{chapter:pendulum}

\ifpdf
    \graphicspath{{Chapter3/Figs/Raster/}{Chapter3/Figs/PDF/}{Chapter3/Figs/}}
\else
    \graphicspath{{Chapter3/Figs/Vector/}{Chapter3/Figs/}}
\fi

\section{Introduction}

This chapter investigates the predictive capability of PINNs for a simple pendulum system. First it introduces the dynamics of the nonlinear pendulum. Then it covers tests on the predictive performance on an idealized version of the system using numerically-generated data. This idealized case acts as a reference for the best-case accuracy. Then it outlines a real experimental setup of the system, and discusses results obtained from it. For both of these cases a PINN is benchmarked against a standard uninformed NN. Finally it closes with a discussion on the results and insights gained from them.

\section{Pendulum Dynamics}

\begin{figure}[ht!]
    \centering
    \includegraphics[scale=2.5]{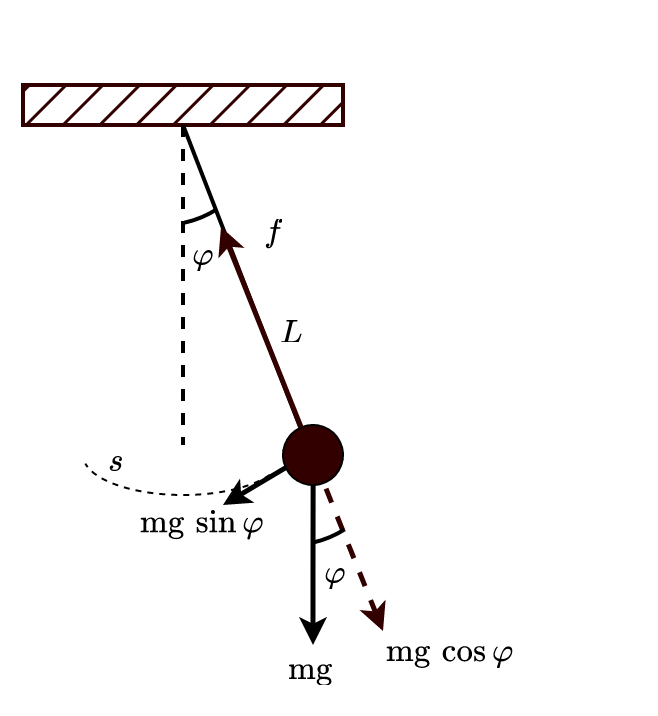}
    \caption{Illustrative diagram of a pendulum system.}
    \label{fig:pend_diagram}
\end{figure}

In the study of differential equations, the pendulum is often one of the first examples students are introduced to of a dynamical system governed by a nonlinear differential equation~\cite{boyce2017}. Figure~\ref{fig:pend_diagram} shows an illustration of a pendulum system. $m$ is the mass of the pendulum bob, $L$ is the length of the rod, $f$ is the force from the rod, $\varphi$ is the angle of oscillation, $s$ is the oscillation arc length, and $g$ is gravitational acceleration. We can derive the differential equation for the motion of the pendulum by applying Newton's second law along the tangent of the oscillation arc. In the following $F$ is the summation of forces along the tangential axis, and $a$ is the linear acceleration:

\begin{align}
    F &= m \, a \nonumber \\
    -mg \, \sin \varphi &= m \, a \nonumber \\
    a &= - g \, \sin \varphi \label{eq:accel_int}
\end{align}

The negative sign in Equation~\ref{eq:accel_int} indicates that the pendulum is decelerating as it moves towards the top of the arc. By using the equation for the arc length $s$ we get:

\begin{align}
 s &= L \, \varphi \nonumber \\ \nonumber \\
 a &= \frac{d^2s}{dt^2} \nonumber \\ \nonumber \\
 a &= L \, \frac{d^2 \varphi}{dt^2} \label{eq:accel_deriv}
\end{align}

Substituting Equation~\ref{eq:accel_deriv} into Equation~\ref{eq:accel_int} gives us the differential equation for the simple nonlinear pendulum, Equation~\ref{eq:simple_nonlinear_pend}:

\begin{gather}
L \, \frac{d^2 \varphi}{dt^2} = - g \, \sin \varphi \nonumber \\ \nonumber \\
\frac{d^2 \varphi}{dt^2} + \frac{g}{L} \, \sin \varphi = 0 \label{eq:simple_nonlinear_pend}
\end{gather}

\section{Ideal Pendulum Simulation}

\label{sec:ideal_pendulum}

This section covers a formulation of an idealized setup based on a numerical solution of Equation~\ref{eq:simple_nonlinear_pend}. The equation is difficult to solve due to the nonlinearity introduced by $\sin \varphi$~\cite{belendez2007exact}, and its exact solution is expressed in terms of elliptic integrals~\cite{abramowitz1948handbook}. The equation becomes even more complicated when one takes air resistance into consideration~\cite{dahmen2015pendulums}, which is a critical factor to consider for real systems. Therefore, we generate a simplified solution using the Euler-Cromer method~\cite{Cromer1981}.

\subsection{Data generation}

We begin by defining a discretization scheme. Let $\{t_1, t_2, ..., t_n\}$ be a discrete set of time points where $t \in \mathbb{R^+}$, and $\{\varphi_1, \varphi_2, ..., \varphi_n\}$ be the corresponding set of angular displacements where $\varphi \in [-2\pi,2\pi]$. Additionally, we define the index set $i = \{1, 2, ..., n\}$. A time frame $\Delta t$ is the difference between  $t_{i+1}$ and $t_i$. By defining a discretized expression for the angular acceleration we obtain the following:

\begin{gather}
    \frac{\dot{\varphi}_{i+1} - \dot{\varphi_i}}{\Delta t} = - \frac{g}{L} \, \sin \varphi_i \nonumber \\ \nonumber \\
    \dot{\varphi}_{i+1} = \dot{\varphi_i} - \frac{g}{L} \sin \varphi_i \, \Delta t \label{eq:ang_veloc_update}
\end{gather}

Equation~\ref{eq:ang_veloc_update} is the approximate discretized solution for the angular velocity. A more exact expression would necessitate that we take the sine of $\varphi_{i+1}$, but at this stage we would not have computed it yet. Fortunately for a small enough $\Delta t$, $\varphi_{i+1} \approx \varphi_i$. 

To get the angular displacement, we follow a similar approach based on discretization of the angular velocity:

\begin{gather}
    \dot{\varphi}_{i+1} = \frac{\varphi_{i+1} - \varphi_i}{\Delta t} \nonumber \\ \nonumber \\
    \varphi_{i+1} = \varphi_i + \dot{\varphi}_{i+1} \, \Delta t \label{eq:ang_displ_update}
\end{gather}

Equation~\ref{eq:ang_displ_update} is the approximate discretized equation for the angular displacement. We apply the Euler-Cromer method~\cite{Cromer1981} through the usage of $\dot{\varphi}_{i+1}$ in Equation~\ref{eq:ang_displ_update} instead of $\dot{\varphi}_{i}$, which allows the solution to maintain energy conservation. 

\begin{figure}[ht]
    \centering
    \includegraphics[scale=0.6]{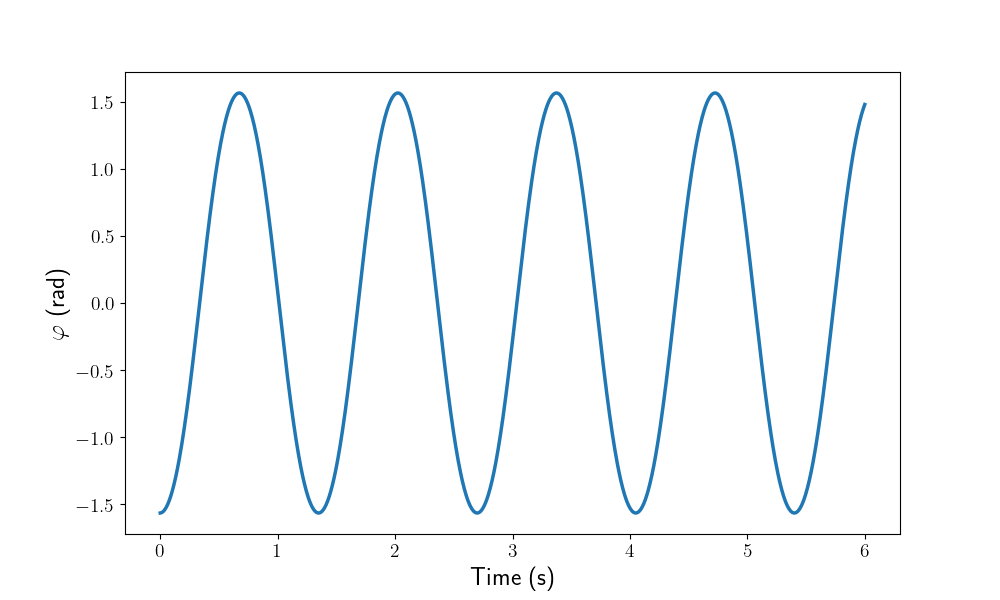}
    \caption{Numerical solution of a pendulum system generated using Equations~\ref{eq:ang_veloc_update} and~\ref{eq:ang_displ_update}.}
    \label{fig:theta_frictionless_sol}
\end{figure}

To generate a solution that is similar to a real experiment, we generate 1500 linearly spaced time points in the interval $[0, 6]$ ($\Delta t = 0.004$ s), set the initial conditions to be $\varphi_1 = - \frac{\pi}{2}$~rad and $\dot{\varphi}_1 = 0$~rad/s, and set $g = 9.8 \, \text{m/s}^2$ and $L = 0.325$~m --- the length of the rod that we use for our experiment in Section~\ref{sec:real_pendulum}. Figure~\ref{fig:theta_frictionless_sol} shows the numerical solution of the angular displacement based on this setup.

The next step is to account for air resistance. The exact amount of air resistance acting on the pendulum mass and the rod is dependant on many factors such as speed, surface roughness, air density, and the object's geometry. In our case, we use a simple air resistance model where we assume that the drag force is linearly proportional to the object's speed with a constant of proportionality $b$. Therefore the model becomes:

\begin{equation}
    \label{eq:simple_nonlinear_pend_air_resist}
    \frac{d^2 \varphi}{dt^2} + b \, \frac{d\varphi}{dt} + \frac{g}{L} \, \sin \varphi = 0
\end{equation}

Additionally, Equation~\ref{eq:ang_veloc_update} becomes:

\begin{equation}
    \label{eq:ang_veloc_update_air_resist}
    \dot{\varphi}_{i+1} = \dot{\varphi_i} - b \, \dot{\varphi_i} - \frac{g}{L} \sin \varphi_i \, \Delta t
\end{equation}

\begin{figure}[ht]
    \centering
    \includegraphics[scale=0.6]{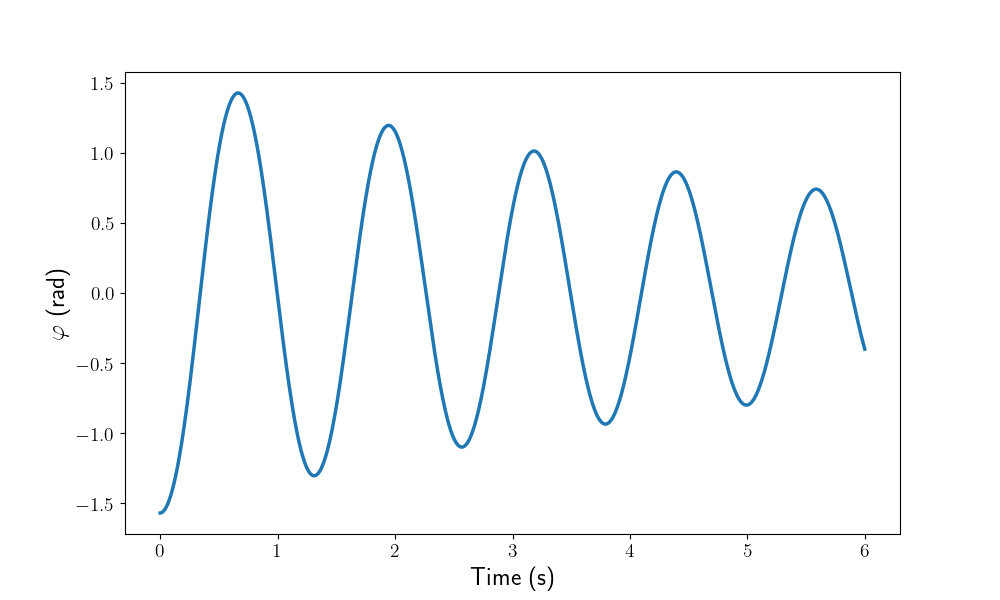}
    \caption{Numerical solution of a pendulum system generated using Equations~\ref{eq:ang_veloc_update_air_resist} and~\ref{eq:ang_displ_update}, taking air resistance into consideration. A more realistic solution would consider a smaller amount of damping over a longer interval, but for our purposes this solution is sufficient.}
    \label{fig:theta_friction_sol}
\end{figure}

We arbitrarily take $b$ to be $0.001$ since we do not have an exact value to rely on. This gives us the data that we will use for training, shown in Figure~\ref{fig:theta_friction_sol}.

\subsection{Training setup}

\label{sec:ideal_pend_train_setup}

The essential element of PINNs is the inclusion of a physics-informed component in the loss function based on the differential equation residual. For the pendulum data that we generated, this corresponds to Equation~\ref{eq:simple_nonlinear_pend_air_resist}. Thus, based on Equation~\ref{eq:PINN_loss} we get Equation~\ref{eq:pend_PINN_loss} as the loss function. $\lambda_p$ is a configurable hyperparameter that we introduce to enforce the strength of the physics loss constraint. We fix its value to be $0.001$.

\begin{equation}
\label{eq:pend_PINN_loss}
    \pazocal{L} = \frac{1}{N_d}\sum_{i=1}^{N_d}|f_{\varphi} - \varphi_d^i|^2 + \frac{\lambda_p}{N_p}\sum_{i=1}^{N_p}|\ddot{\varphi}_f^i + b \, \dot{\varphi}_f^i + \frac{g}{L} \, \text{sin} \varphi_f^i|^2
\end{equation}

We compare PINNs against ordinary NNs that do not include the physics loss component. For the NN, we eliminate the second term in Equation~\ref{eq:pend_PINN_loss}. The code implementation we develop for our evaluations is based partly on open-source repositories developed by Moseley~\cite{Moseley2021} and Bhustali~\cite{Bhustali2021}, although we adapt the implementation to suit our particular problem. For all our training cases we use a multi-layer perceptron (MLP) architecture~\cite{Bishop2006} with different network parameters, and we assess the predictive performance based on a variation of these parameters. We run the training on a workstation running an Intel i7-7820X 16-core CPU, and an NVIDIA Quadro P1000 4 GB GPU.

Based on tests with different activation functions, we found that the sine function performed the best for the pendulum system and so we use it for all of our training cases. Additionally, we found that the limited-memory Broyden–Fletcher–Goldfarb–Shanno (LBFGS) algorithm~\cite{liu1989lbfgs} performed best for PINNs, and so we use it as our default optimizer. This is corroborated by the usage of LBFGS in the PINN paper by Raissi et al.~\cite{raissi2019}. We use the default tolerance values in Pytorch for the termination, so LBFGS stops when there is nothing left to learn based on these tolerances. We fix the learning rate at $0.01$.

\begin{figure}[ht]
    \centering
    \includegraphics[scale=0.05]{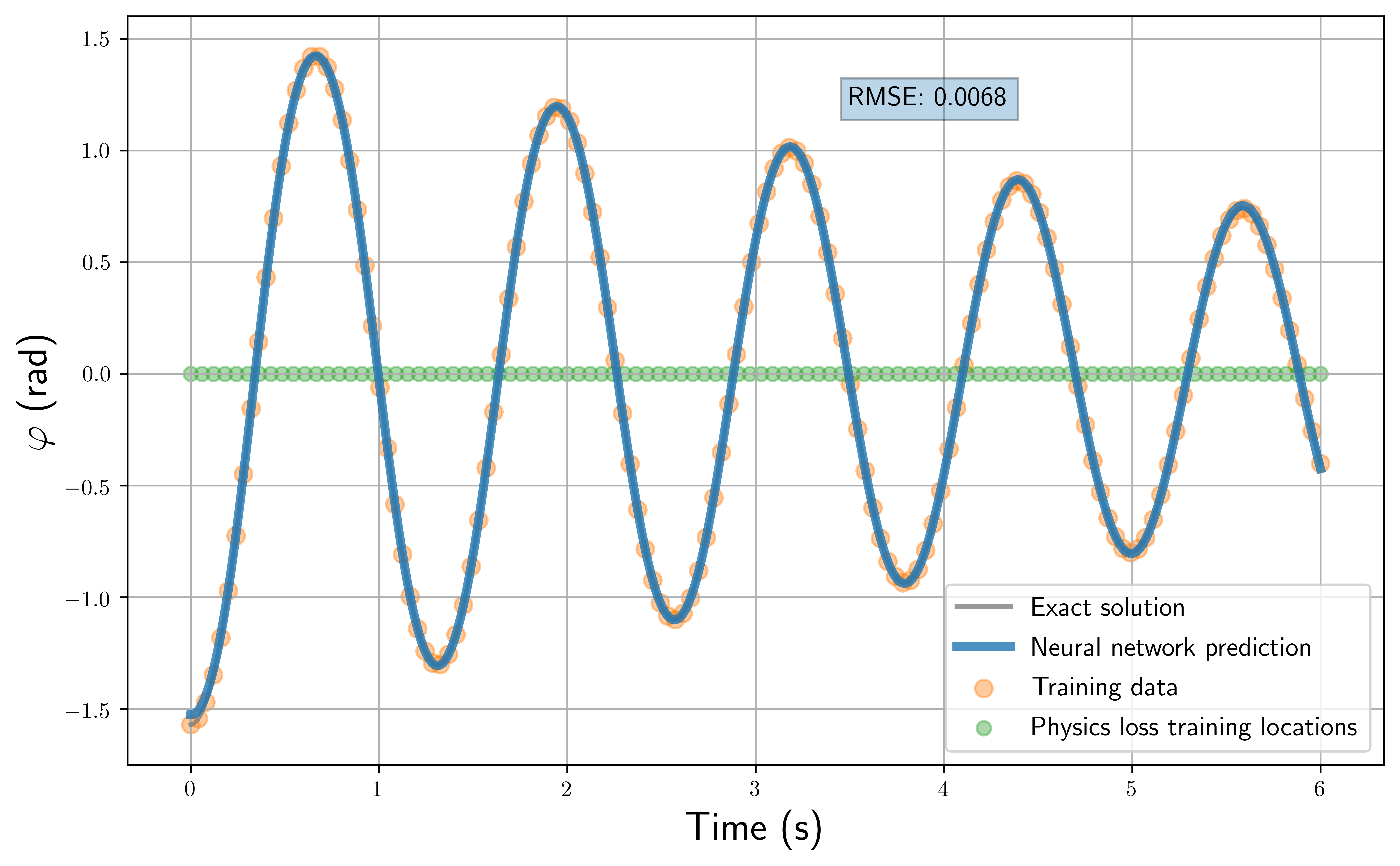}
    \caption{PINN predictions on the synthetic data pendulum given 150 training points. PINN architecture: 3 FC hidden layers with 32 neurons each. RMSE = 0.0068 as the PINN has no trouble fitting the data given a perfect setup.}
    \label{fig:synth_pend_32-32-32_physinformed}
\end{figure}

For the training data, we test four different training variations: linearly-spaced, uniformly-distributed random, adjacent, and noisy data. For the first three cases we vary the number of training points, and in the fourth we vary the amount of noise.

\begin{figure}[ht]
    \centering
    \includegraphics[width=\textwidth]{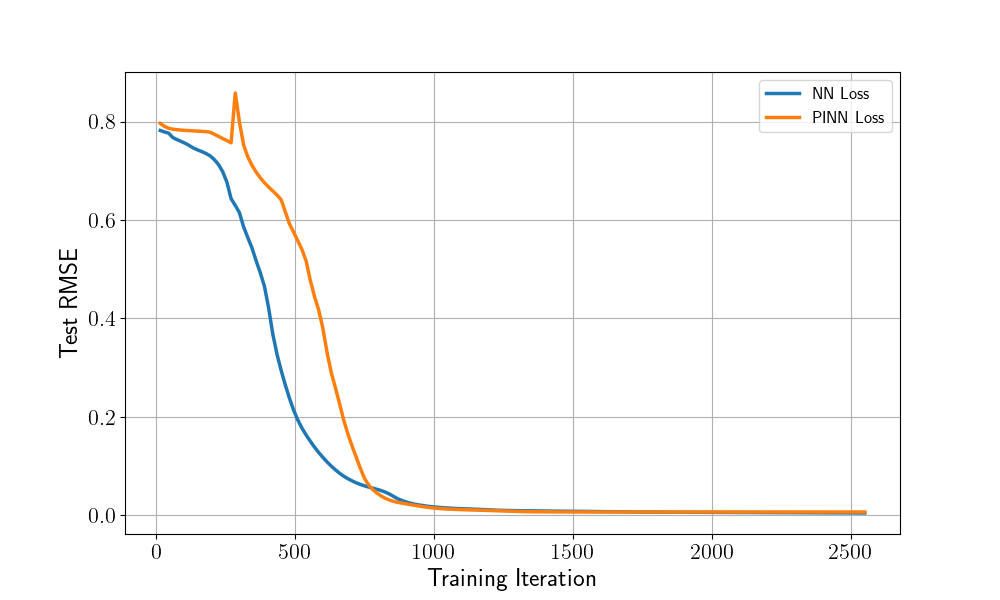}
    \caption{Test RMSE values against training iterations of a PINN and an equivalent NN, given 150 linearly spaced training points. Both models converge to an accurate solution after approximately 1250 iterations, although the PINN solution faces a spike that it overcomes during the optimization near the 300 iteration mark.}
    \label{fig:loss_graph_32-32-32}
\end{figure}

Figure~\ref{fig:synth_pend_32-32-32_physinformed} shows the predictions of a 3-layer PINN with 32 neurons in each layer, given 150 linearly spaced training points ($N_d = 150$) and 100 linearly spaced collocation points ($N_p = 100$). In the best possible case, one where we have enough data available and a sufficiently expressive network architecture, the PINN is able to predict the entire pendulum solution with high accuracy (RMSE $ = 6.8 \times 10^{-3}$). Figure~\ref{fig:loss_graph_32-32-32} shows the test RMSE plotted against the training iterations for the PINN and the NN. We can see that the NN also has no trouble fitting the data due to its abundance, with an RMSE of $4.0 \times 10^{-3}$. In the coming sections we gradually make the training setup more difficult and compare the PINN's performance against that of an equivalent NN.

\subsection{Results}

\label{sec:ideal_pend_results}

The first thing to determine for both a PINN and an NN, is the smallest possible network architecture size so that we can fix it for training. Tables~\ref{tab:pend_pinn_loss_network_arch_variation} and~\ref{tab:pend_nn_loss_network_arch_variation} show the test RMSE values for the PINN and NN respectively based on a variation of the number of hidden layers and number of units in each layer, after 2000 iterations. The blow-up (BU) entries correspond to instances where the training failed due to exploding gradients, and the early termination (ET) entries correspond to instances where the training ends due to the termination condition of LBFGS. The first number in those entries is the last valid training iteration before blow-up, and the second number is the final reported RMSE value. For both the PINN and the NN, the network maintains a high level of accuracy even with small architectures. Therefore we fix the architecture to be 3 layers with 5 units each, to keep the network small whilst maintaining expressiveness for more difficult training cases.

\begin{table}[ht]
\centering
\begin{tabular}{|c|c|c|c|c|c|}
\hline
\backslashbox{\textbf{Units}}{\textbf{Layers}}    & \textbf{1} & \textbf{2} & \textbf{3} & \textbf{4}      & \textbf{5}      \\ \hline
\textbf{32} & 0.2914     & 0.0126     & 0.0066     & 0.0067          & 0.0076          \\ \hline
\textbf{16} & 0.6960     & 0.0134     & 0.0067     & 0.0067          & 0.0067          \\ \hline
\textbf{8}  & 0.6987     & 0.0233     & 0.0133     & 0.0072          & 0.0070          \\ \hline
\textbf{5}  & 0.6536     & 0.2190     & 0.0184     & \begin{tabular}[c]{@{}c@{}}BU: 435\\ L: 0.7278 \end{tabular}  & \begin{tabular}[c]{@{}c@{}}BU: 300\\ L: 0.7732 \end{tabular} \\ \hline
\textbf{4}  & 0.6957     & 0.0361     & 0.0086     & \begin{tabular}[c]{@{}c@{}}BU: 784\\ L: 0.6862 \end{tabular}  & 0.2391          \\ \hline
\textbf{3}  & 0.7448     & 0.0256     & 0.0286     & \begin{tabular}[c]{@{}c@{}}BU: 835\\ L: 0.6757 \end{tabular}  & 0.0333          \\ \hline
\end{tabular}
\caption{PINN RMSE values for different variations of hidden layers and units in each layer.}
\label{tab:pend_pinn_loss_network_arch_variation}
\end{table}

\begin{table}[ht]
\centering
\begin{tabular}{|c|c|c|c|c|c|}
\hline
\backslashbox{\textbf{Units}}{\textbf{Layers}}    & \textbf{1}                                                   & \textbf{2}                                                   & \textbf{3} & \textbf{4}                                                   & \textbf{5}                                                   \\ \hline
\textbf{32} & \begin{tabular}[c]{@{}c@{}}BU: 636\\ L: 0.7376\end{tabular}  & \begin{tabular}[c]{@{}c@{}}ET: 1914\\ L: 0.0066\end{tabular} & 0.0059     & \begin{tabular}[c]{@{}c@{}}ET: 1423\\ L: 0.0037\end{tabular} & 0.0037                                                       \\ \hline
\textbf{16} & 0.6615                                                       & 0.0087                                                       & 0.0033     & 0.0071                                                       & 0.0054                                                       \\ \hline
\textbf{8}  & 0.6610                                                       & 0.0101                                                       & 0.0173     & 0.0067                                                       & 0.0056                                                       \\ \hline
\textbf{5}  & \begin{tabular}[c]{@{}c@{}}ET: 1681\\ L: 0.6626\end{tabular} & 0.0227                                                       & 0.0084     & 0.0044                                                       & \begin{tabular}[c]{@{}c@{}}ET: 1763\\ L: 0.0031\end{tabular} \\ \hline
\textbf{4}  & \begin{tabular}[c]{@{}c@{}}ET: 1560\\ L: 0.7240\end{tabular} & 0.0168                                                       & 0.0125     & 0.00158                                                      & 0.0092                                                       \\ \hline
\textbf{3}  & \begin{tabular}[c]{@{}c@{}}ET: 1915\\ L: 0.7692\end{tabular} & \begin{tabular}[c]{@{}c@{}}BU: 273\\ L: 0.7277\end{tabular}  & 0.0473     & 0.0141                                                       & \begin{tabular}[c]{@{}c@{}}BU: 266\\ L: 0.7786\end{tabular}  \\ \hline
\end{tabular}
\caption{NN RMSE values for different variations of hidden layers and units in each layer.}
\label{tab:pend_nn_loss_network_arch_variation}
\end{table}

\subsubsection{Linearly-spaced data}

\label{sec:ideal_data_linspace}

\begin{table}[ht]
\centering
\begin{subtable}{0.45\textwidth}
\centering
\begin{tabular}{|c|c|c|}
\hline
$\mathbf{N_d}$     & \textbf{NN}                                                  & \textbf{PINN} \\ \hline
\textbf{100} & 0.0051                                                       & 0.0177        \\ \hline
\textbf{50}  & 0.0044                                                       & 0.0246        \\ \hline
\textbf{25}  & 0.0136                                                       & 0.0084        \\ \hline
\textbf{15}  & \begin{tabular}[c]{@{}c@{}}ET: 1131\\ L: 0.2904\end{tabular} & 0.0109        \\ \hline
\textbf{10}  & \begin{tabular}[c]{@{}c@{}}ET: 1021\\ L: 1.3906\end{tabular} & 0.0756        \\ \hline
\textbf{5}   & \begin{tabular}[c]{@{}c@{}}ET: 612\\ L: 1.1184\end{tabular}  & 0.0470        \\ \hline
\end{tabular}
\caption{Linearly-spaced points.}
\label{tab:pend_loss_linspace_points}
\end{subtable}
\hfill
\begin{subtable}{0.45\textwidth}
\centering
\begin{tabular}{|c|c|c|}
\hline
$\mathbf{N_d}$     & \textbf{NN}                                                  & \textbf{PINN} \\ \hline
\textbf{100} & 0.0093                                                       & 0.0118       \\ \hline
\textbf{50}  & 0.0181                                                       & 0.0084        \\ \hline
\textbf{25}  & 0.5434                                                       & 0.0424        \\ \hline
\textbf{15}  & \begin{tabular}[c]{@{}c@{}}ET: 1468\\ L: 0.5097\end{tabular} & 0.0292        \\ \hline
\textbf{10}  & \begin{tabular}[c]{@{}c@{}}ET: 152\\ L: 1.0219\end{tabular}  & 0.1488        \\ \hline
\textbf{5}   & \begin{tabular}[c]{@{}c@{}}ET: 956\\ L: 0.9285\end{tabular}  & 0.4922        \\ \hline
\end{tabular}
\caption{Uniformly-distributed points.}
\label{tab:pend_loss_uniform-dist_points}

\end{subtable}
\caption{RMSE values for variations of numbers of variably-spaced training points.}
\label{tab:pend_loss_var-space_points}

\end{table}

\begin{figure}[ht]
    \centering
    \includegraphics[width=\textwidth]{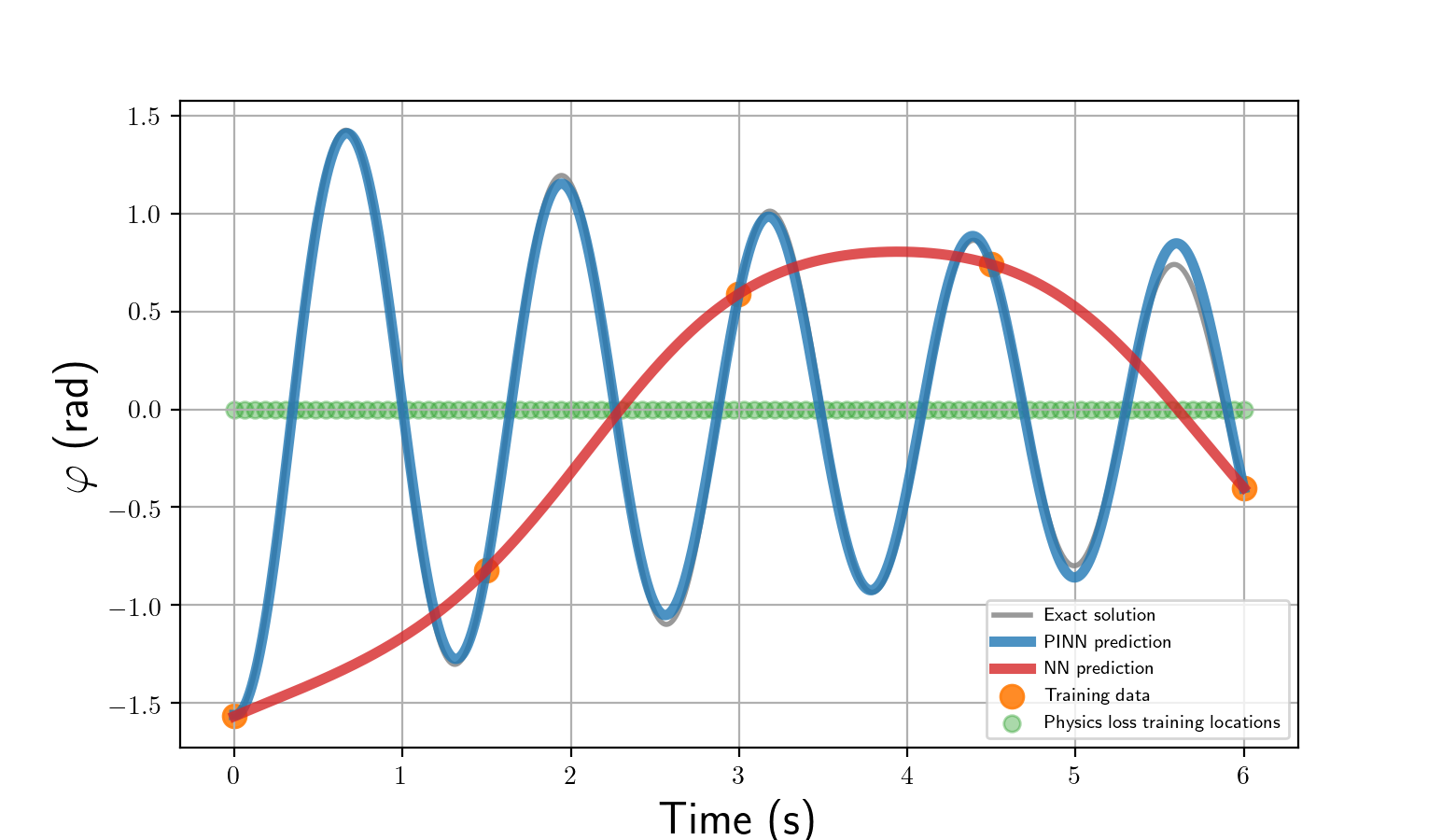}
    \caption{PINN and NN predictions on the data for the idealized pendulum using 5 linearly-spaced training points. The PINN is able to predict the correct solution based on the physics loss, whereas the NN is only able to fit the training data.}
    \label{fig:ideal_pend_preds_5-5-5_points5}
\end{figure}

Table~\ref{tab:pend_loss_linspace_points} shows the RMSE values for different numbers of linearly spaced data points based on the training configuration. The NN maintains good accuracy down to 25 points. Lower than that, the RMSE values begin to suffer considerably. In contrast, the PINN maintains good accuracy even with only 5 linearly-spaced training points. Figure~\ref{fig:ideal_pend_preds_5-5-5_points5} shows that the physics loss term allows the network to accurately generalise across the entire domain, whilst the uninformed NN is only able to fit the data.

\subsubsection{Uniformly-distributed random data}

\begin{figure}[ht]
    \centering
    \includegraphics[width=\textwidth]{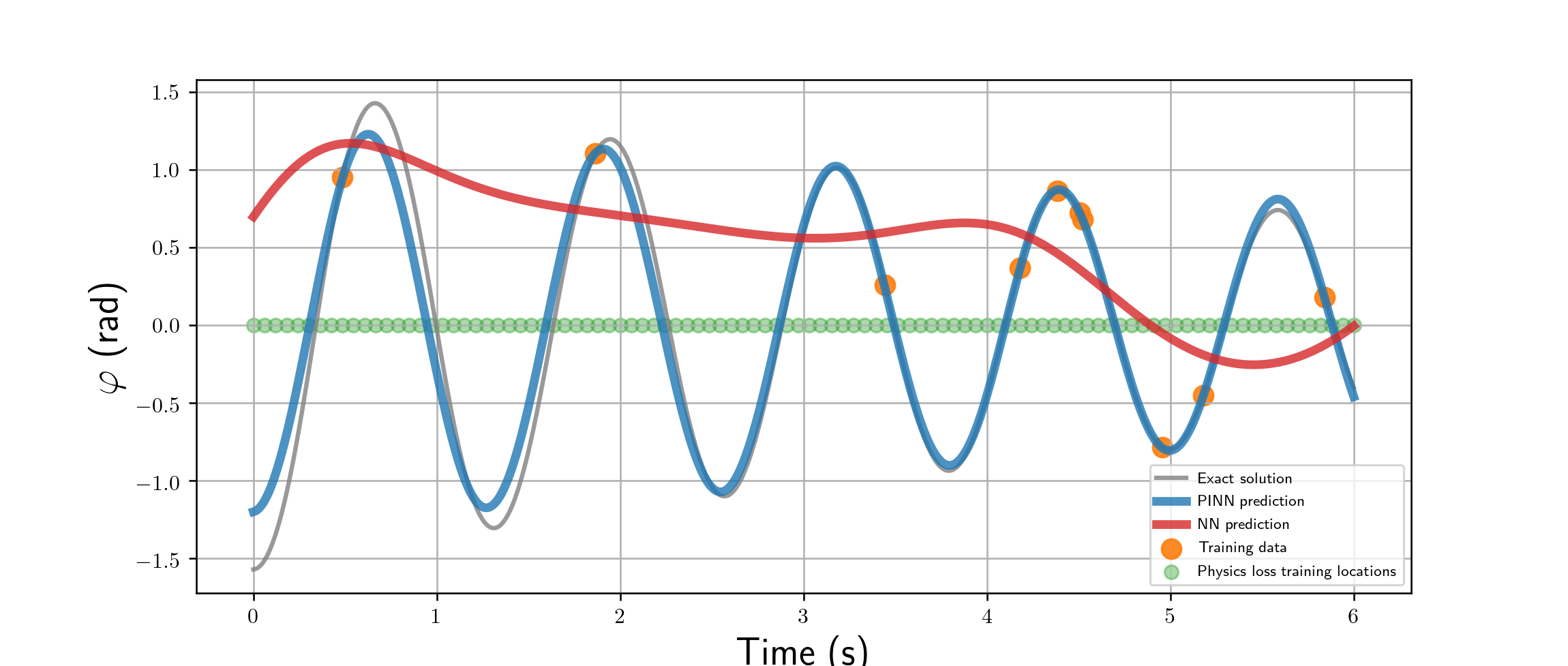}
    \caption{PINN and NN predictions on the synthetic data pendulum using 10 training points for uniformly-distributed random data. The NN is trained for 150 iterations --- its final state before predictions became unstable. The PINN is trained for 2000 iterations. The PINN maintains a reasonable fit of the data while the NN struggles due to the data's irregularity.}
    \label{fig:ideal_pend_preds_5-5-5_points10_uniform-random}
\end{figure}

Here we randomly sample training points from a uniform distribution. We evaluate training in a similar fashion to the linearly-spaced data case. Table~\ref{tab:pend_loss_uniform-dist_points} shows the results. The NN accuracy drops significantly below the 50 point mark, and for 10 points and less the optimizer stops early as the network is unable to learn anything. The equivalent PINN maintains good accuracy for 2000 iterations down to 15 points. We repeated the PINN training runs for 10 and 5 points but this time allowing them to stop based the LBFGS termination condition. We found that for 10 points the training stops after 5161 iterations with an RMSE of $0.1056$, and for 5 points it stops at 4083 iterations with an RMSE of $0.4432$. By comparing Figure~\ref{fig:ideal_pend_preds_5-5-5_points5} with Figure~\ref{fig:ideal_pend_preds_5-5-5_points10_uniform-random}, we can see that the data irregularity degrades the accuracy of both models, although the PINN is still able to maintain a reasonable prediction of the true solution. 

\subsubsection{Adjacent points}
\begin{figure}[ht]
    \centering
    \includegraphics[width=\textwidth]{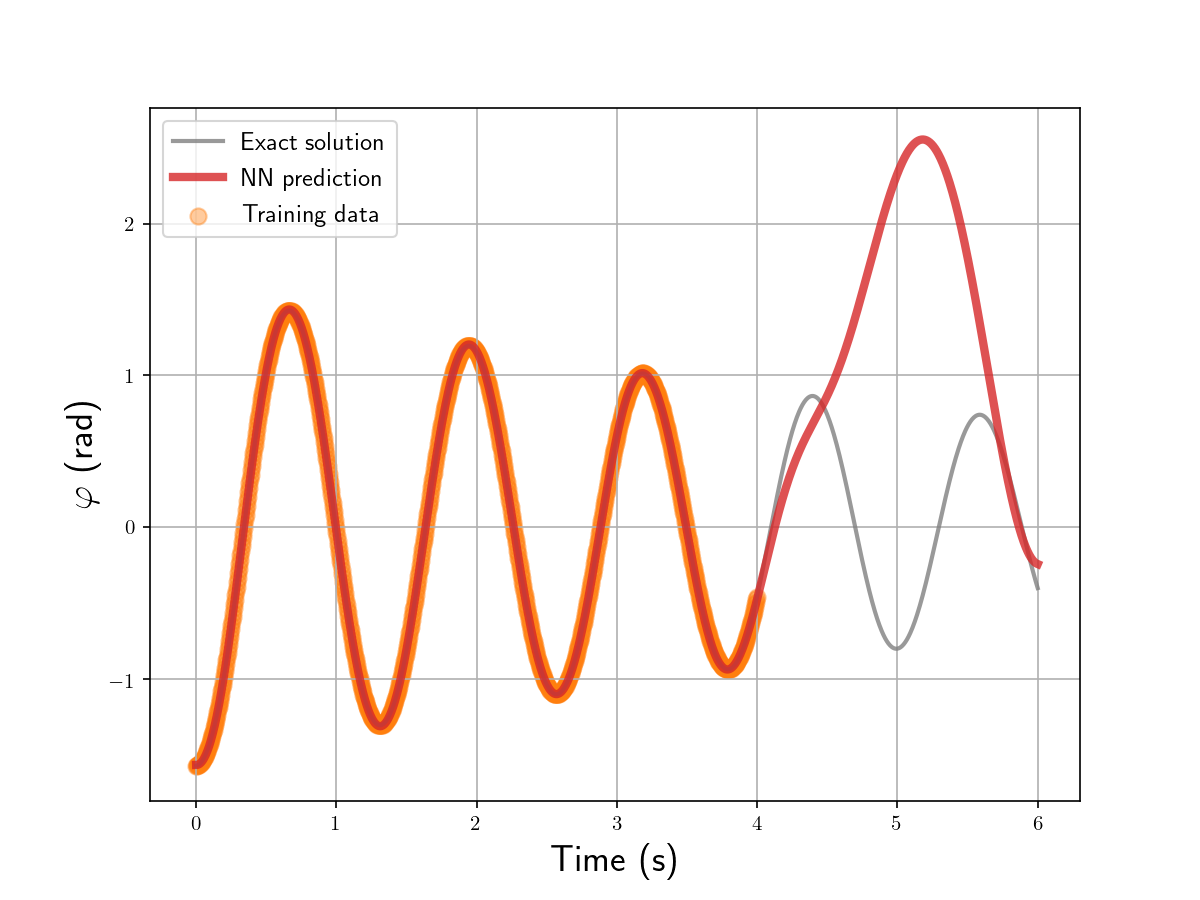}
    \caption{NN prediction when trained with 1000 adjacent points. The NN fails to extrapolate the accurately predicted solution on the training points to the last 500 test points.}
    \label{fig:ideal_pend_preds_nn_5-5-5_failure}
\end{figure}

\begin{figure}[ht!]
    \centering
    \includegraphics[width=\textwidth]{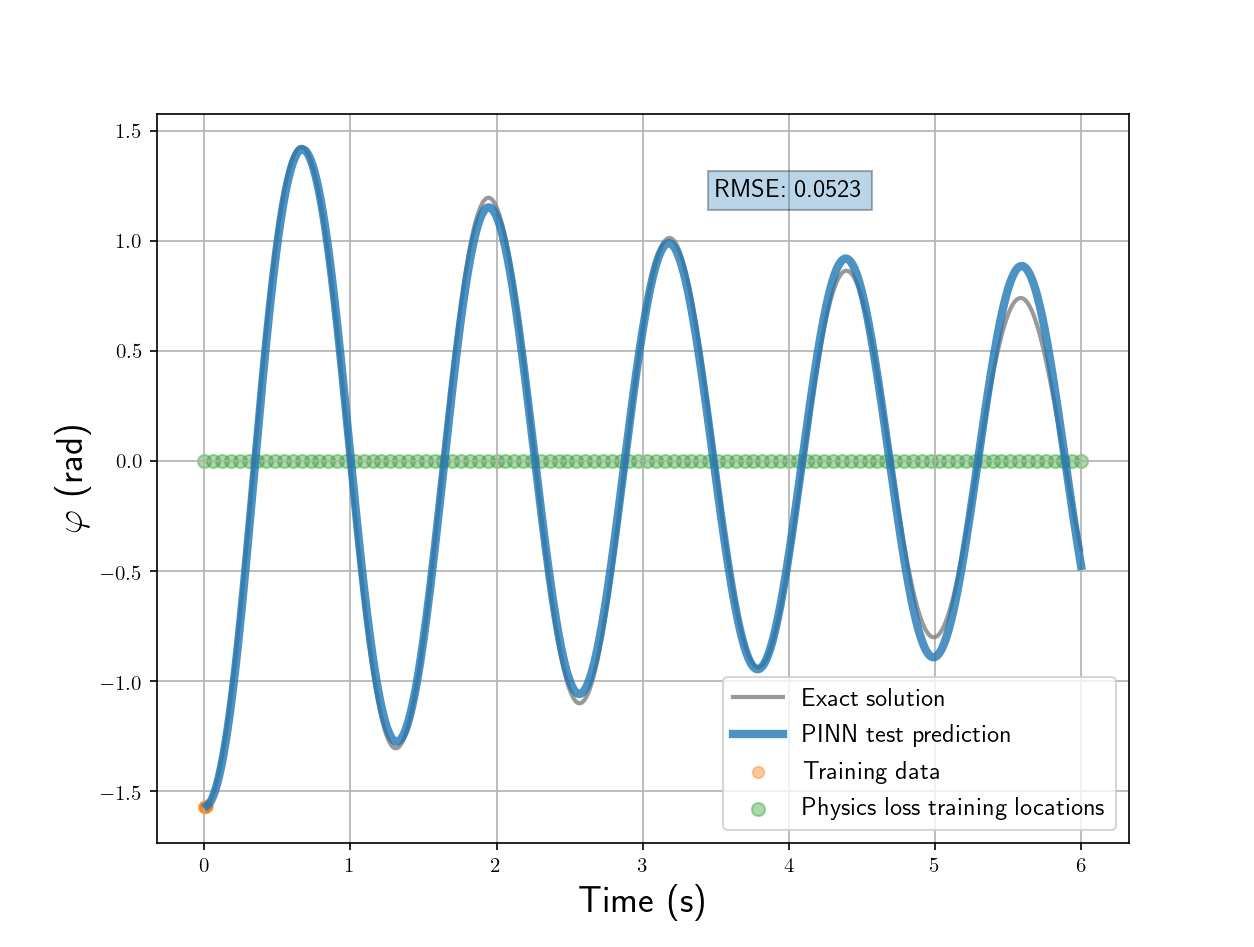}
    \caption{PINN predicted solution based on the first 5 points of the numerical solution. The PINN consists of two hidden layers with 12 units in the first and 9 in the second --- corresponding to the 9th entry in Table~\ref{tab:pend_loss_adj_points}. Remarkably, the PINN is able to accurately predict the solution despite being trained with only the first 5 points.}
    \label{fig:ideal_pend_pinn_preds_adj-points5}
\end{figure}

For adjacent training points, we take a different approach to PINN evaluation. We do not compare against an uninformed NN, as we have already shown that they fail in data-absent regions. Figure~\ref{fig:ideal_pend_preds_nn_5-5-5_failure} shows a further example for this. Instead, we focus on the PINN for the test cases. Due to the difficulty of predicting outside of the training distribution even with the aid of physics knowledge, we deviate away from the default training configuration. Instead we configure the network parameters to show that given the right conditions it is possible for a PINN to predict outside of the training distribution. Additionally, for all of the configurations we either allow the network to run until the termination condition, or stop the training early once a satisfactory test accuracy is achieved. Table~\ref{tab:pend_loss_adj_points} reports the configurations, losses, and iteration numbers. Based on the data, we observe that given the right architecture it is possible for PINNs to make predictions outside of the training set. The accuracy of the prediction shown in Figure~\ref{fig:ideal_pend_pinn_preds_adj-points5} further emphasizes this point. Additionally, all of the architectures that have been proven to be successful have a low memory footprint, consisting of 2-3 hidden layers with less than 10 units each for most cases. 

\begin{table}[ht]

\begin{subtable}[t]{0.45\linewidth}
\centering
\begin{tabular}[t]{|c|c|c|c|}
\hline
\textbf{Points}     & \textbf{Loss} & \textbf{Arch} & \textbf{Iters} \\ \hline
\textbf{750} & 0.0927        & 5-5                   & 2852                \\ \hline
\textbf{500} & 0.0672        & 5-5                   & 5558                \\ \hline
\textbf{300} & 0.0659        & 5-5                   & 3205                \\ \hline
\textbf{200} & 0.0910        & 4-4                   & 2417                \\ \hline
\textbf{100} & 0.0364        & 6-6-4                 & 1432                \\ \hline
\textbf{50}  & 0.0471        & 5-6-5                 & 1768                \\ \hline
\textbf{25}  & 0.0338        & 10-6                  & 3912                \\ \hline
\textbf{10}  & 0.0582        & 12-9                  & 5572                \\ \hline
\textbf{5}   & 0.0523        & 12-9                  & 5134                \\ \hline
\textbf{3}   & 0.0714        & 12-9                  & 6664                \\ \hline
\textbf{1}   & 0.0458        & 5-9-3                 & 1480              \\ \hline
\end{tabular}
\caption{PINN evaluation runs for a decreasing number of adjacent points. The numbers in the architecture column refer to the units in each layer, so 6-6-4 means 6 units in the first and second layer and 4 in the third.}
\label{tab:pend_loss_adj_points}
\end{subtable}
\hfill
\begin{subtable}[t]{0.45\linewidth}
\centering
\begin{tabular}[t]{|c|c|c|}
\hline
\textbf{Std. Dev.} & \textbf{NN}                                           & \textbf{PINN}                                         \\ \hline
\textbf{1.5}       & \begin{tabular}[c]{@{}c@{}}576\\ 0.6188\end{tabular}  & \begin{tabular}[c]{@{}c@{}}978\\ 0.4698\end{tabular}  \\ \hline
\textbf{1.0}       & \begin{tabular}[c]{@{}c@{}}615\\ 0.4024\end{tabular}  & \begin{tabular}[c]{@{}c@{}}742\\ 0.3112\end{tabular}  \\ \hline
\textbf{0.7}       & \begin{tabular}[c]{@{}c@{}}716\\ 0.2598\end{tabular}  & \begin{tabular}[c]{@{}c@{}}976\\ 0.2126\end{tabular}  \\ \hline
\textbf{0.5}       & \begin{tabular}[c]{@{}c@{}}1696\\ 0.1791\end{tabular} & \begin{tabular}[c]{@{}c@{}}4022\\ 0.1583\end{tabular} \\ \hline
\textbf{0.3}       & \begin{tabular}[c]{@{}c@{}}1342\\ 0.1269\end{tabular} & \begin{tabular}[c]{@{}c@{}}3070\\ 0.0940\end{tabular} \\ \hline
\textbf{0.2}       & \begin{tabular}[c]{@{}c@{}}997\\ 0.0791\end{tabular}  & \begin{tabular}[c]{@{}c@{}}2192\\ 0.0624\end{tabular} \\ \hline
\textbf{0.1}       & \begin{tabular}[c]{@{}c@{}}2104\\ 0.0432\end{tabular} & \begin{tabular}[c]{@{}c@{}}2692\\ 0.0336\end{tabular} \\ \hline
\end{tabular}
\caption{Loss values for the NN and PINN in the case of noisy data. The additive noise values are sampled from a Gaussian distribution with zero mean and variable standard deviations shown in the first column. For each entry, the first number indicates the optimal iteration and the second number indicates the corresponding RMSE.}

\label{tab:pend_loss_noisy_data}
    
\end{subtable}
\caption{}
\label{tab:random_cap}
\end{table}

\subsubsection{Noisy data}

\begin{figure}[ht!]
    \centering
    \includegraphics[width=\textwidth]{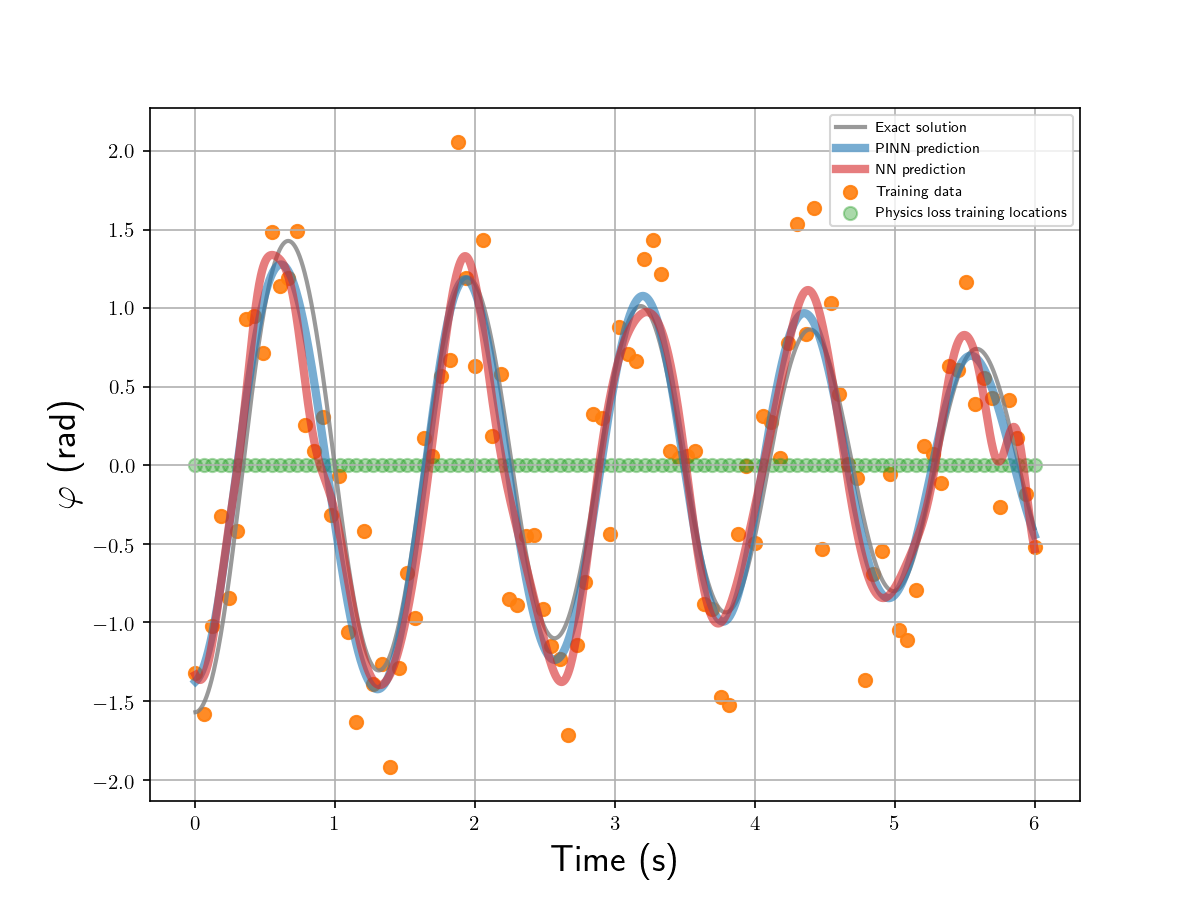}
    \caption{PINN vs NN predictions on 100 linearly-spaced points with added Gaussian noise with a mean of 0 and a standard deviation of 0.5. The PINN and NN solutions are similar, although the PINN is slightly less impacted by the noise.}
    \label{fig:ideal_pend_pinn_preds_noisy_sigma0.5}
\end{figure}

This training case involves taking 100 linearly-spaced points and adding Gaussian noise to them. The Gaussian has a mean of 0 and variable standard deviation. We fix the architecture at 3 hidden layers with 5 units each, as per the default configuration. Table~\ref{tab:pend_loss_noisy_data} reports the RMSEs and iteration numbers. The PINN outperforms the NN in all cases, but only marginally so in cases with less noise thresholds. In general, we can see that both the PINN and NN tend to be impacted with large amounts of noise, although the PINN is better at adapting to it. The predictions in Figure~\ref{fig:ideal_pend_pinn_preds_noisy_sigma0.5} show this to be the case, where the shape of the solution for the NN deviates away from a sinusoid, whereas the PINN solution maintains its sinusoidal behaviour. The difficulty of making accurate predictions in the presence of noise puts forward a strong case for finding efficient ways to denoise data or to model the uncertainty arising from noise.

\clearpage
\section{Real Pendulum Experiment}

\label{sec:real_pendulum}

This section presents an experiment evaluating PINNs on real-world data for the pendulum system. 

\section{Hardware Block Design}

\begin{figure}[ht]
    \centering
    \includegraphics[width=\textwidth]{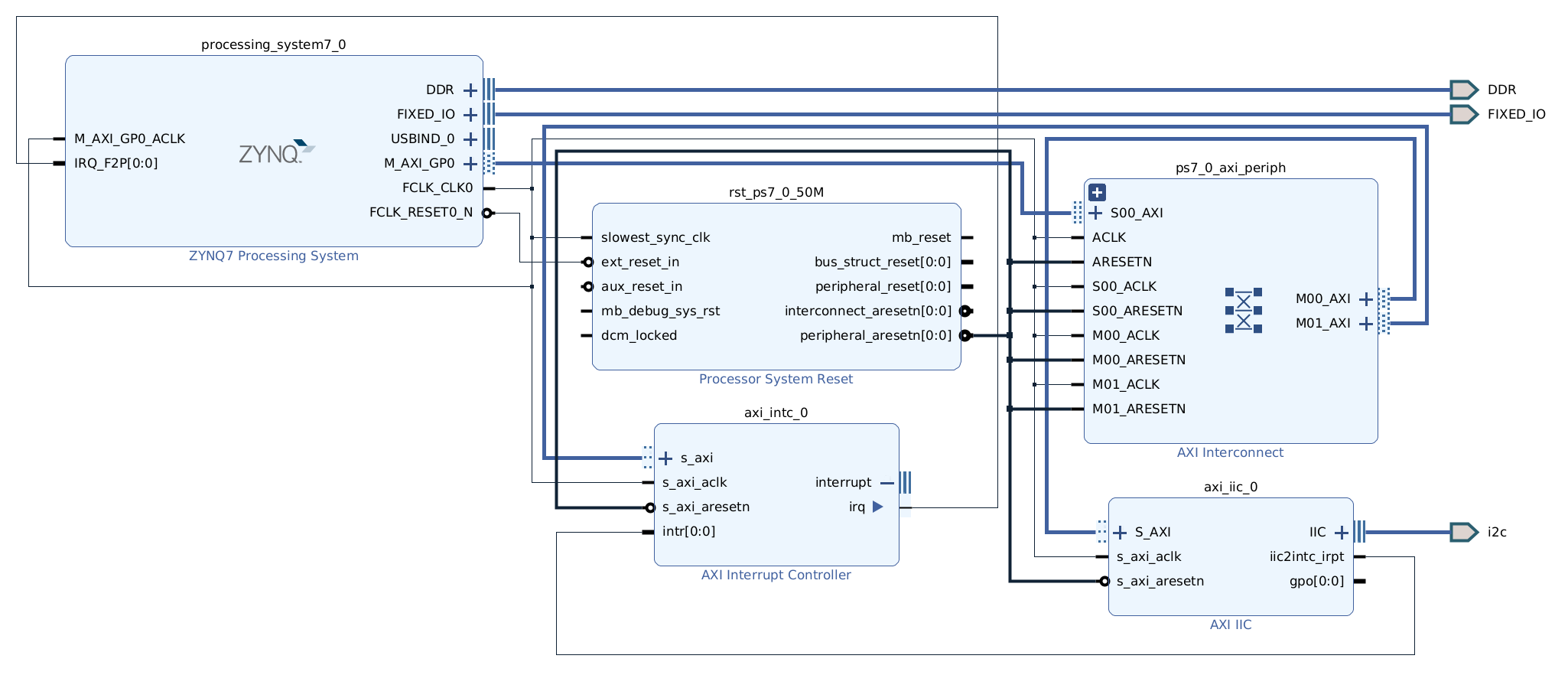}
    \caption{AXI IIC block design that we use for our experiments. The ZYNQ7 processing system is the operating system side of the system-on-chip (SoC) FPGA, on which the Python layer runs. The AXI IIC block is the direct interface with the sensors through I2C. We configure the I2C clock frequency to be 1000 KHz. The AXI IIC block interfaces with the processing system through the AXI interconnect block. Read and write commands are issued to the AXI interconnect through the Python driver API.}
\label{fig:AXI_IIC_block_design}
\end{figure}

We chose to use PYNQ as the framework of choice for the pendulum experiment outlined in this section as well as the heat diffusion experiment in Section~\ref{sec:heating_diff_exp}. Based on this choice, we used the PYNQ-Z1 board as the hardware platform. The advantage of this is that we were able to conveniently interface with the hardware block design from the high-level Python-based API, without the need to dive deep into the low-level hardware. Figure~\ref{fig:AXI_IIC_block_design} shows the block design.

\subsection{Experimental setup}

\begin{figure}[ht!]
    \centering
    \includegraphics[scale=0.12, angle=-90]{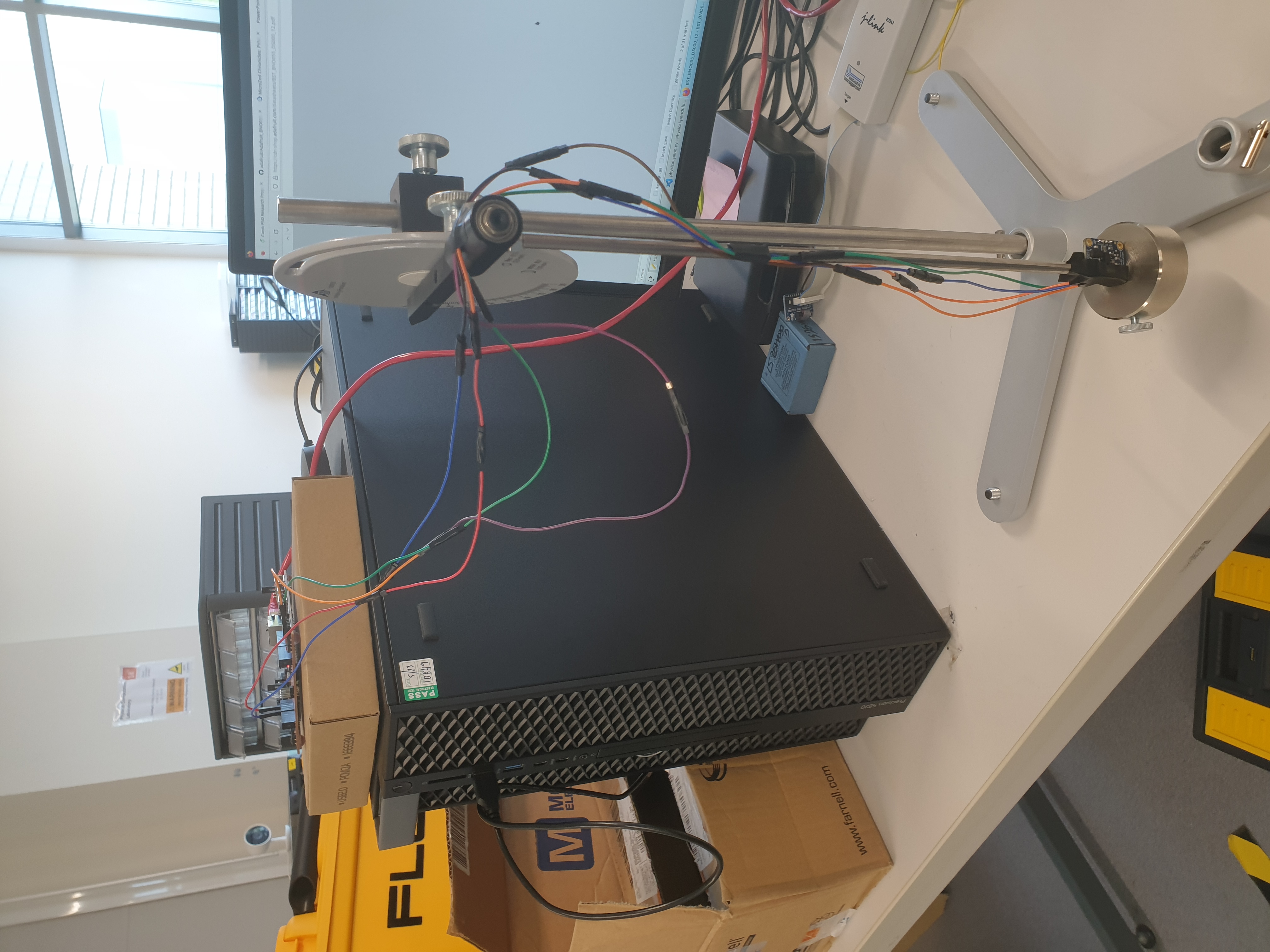}
    \caption{Experimental setup for the pendulum system. The BNO055~\cite{bosch:BNO055} is attached to the pendulum mass. The PYNQ-Z1 board (on top of the desktop computer) interfaces with the BNO055 through I2C~\cite{NXP:I2C}.}
    \label{fig:pend_exp_setup}
\end{figure}

We perform the experiment using a variable g pendulum with the oscillation plane adjusted at 0°, i.e. the plane of oscillation is perpendicular to the ground. We use the PYNQ-Z1 FPGA as the sensing platform, and measure angular displacement using the BNO055 absolute orientation sensor~\cite{bosch:BNO055}. The BNO055 contains an accelerometer, a gyroscope, and a magnetometer. It combines data from all three to calculate the absolute Euler angle orientation through a proprietary sensor fusion algorithm. It uses the Inter-Integrated Circuit (I2C) protocol~\cite{NXP:I2C} for communication. The PYNQ-Z1 processing system communicates with the sensor by accessing data that is captured by an AXI IIC IP block~\cite{Xilinx:AXI-IIC} on the FPGA programmable logic (PL) fabric. Figure~\ref{fig:AXI_IIC_block_design} shows the hardware block design. The PYNQ subsystem displays the sensor values through the Python API. Figure~\ref{fig:pend_exp_setup} shows the experimental setup.

\begin{figure}[ht]
    \centering
    \includegraphics[width=0.85\textwidth]{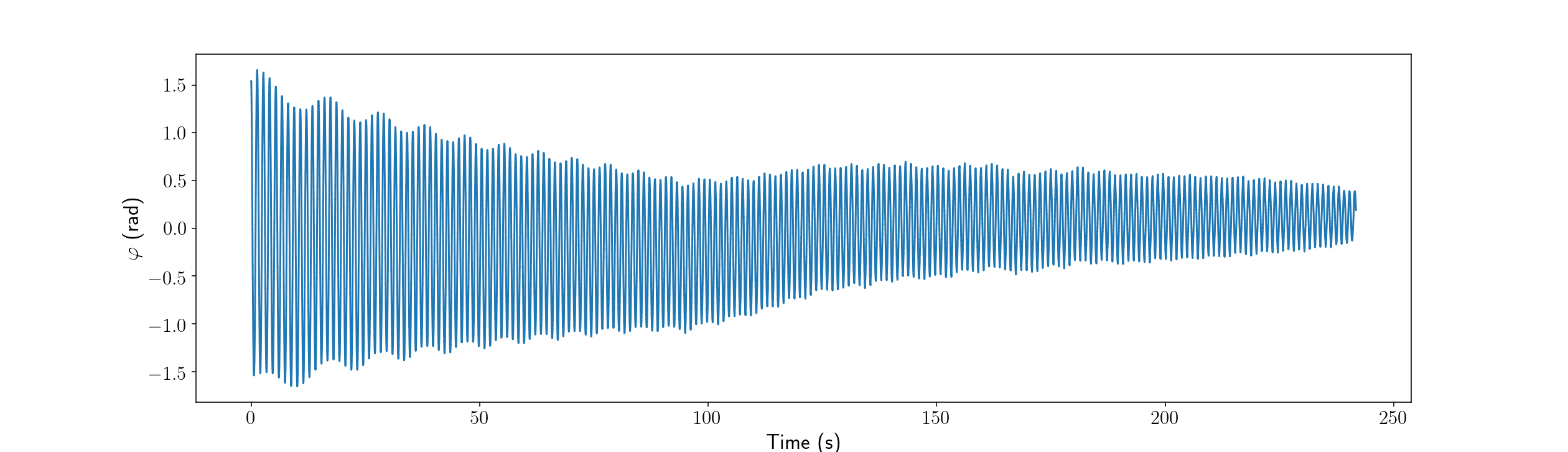}
    \caption{Pendulum oscillation data captured from the experimental setup shown in Figure~\ref{fig:pend_exp_setup}. The data has a much higher frequency than the numerically-generated solution in Figure~\ref{fig:theta_friction_sol}, although the sinusoidal nature is similar enough for making comparisons.}
    \label{fig:real_pend_data}
\end{figure}

Before capturing data, we ensure that the BNO055 is calibrated to filter out offsets with the three internal sensors. We do this by moving the BNO055 in random directions, and in a figure-eight motion. After this, we begin the experiment by first moving the pendulum mass towards the left until it is parallel to the ground. We start recording the data from the sensor and release the mass, allowing it oscillate freely. We record until the oscillation nearly stops. This was after 241 seconds. The data obtained still contains an offset, so we remove it by subtracting the midpoint value from the signal. Figure~\ref{fig:real_pend_data} shows the sensor data. The data shows a slight downwards trend which we attribute to the residual gyroscope drift which was not filtered by the fusion algorithm.

\subsection{Training setup}

\label{sec:real_pend_exp_training}

The training setup is similar to the one in Section~\ref{sec:ideal_pend_train_setup}. In this scenario we also solve an inverse problem. Rather than keeping the friction coefficient parameter $b$ constant, we assess whether the PINN can discover its value by setting it as a trainable parameter. We report on its value during PINN training runs. Our default configuration is an MLP architecture with 3 hidden layers and 32 neurons in each. We use Equation~\ref{eq:pend_PINN_loss} as our PINN loss component, with 8000 collocation points. We choose LBFGS as the optimizer, but this time we run it for as many iterations as is necessary and report this value. This is to overcome the spectral bias issue that we discuss in Section~\ref{sec:large_domain_training_pend}. Once again, we use sine as the activation function. We use a $\lambda_p$ value of 0.1 and a learning rate of 0.05.

\subsection{Results}

We evaluate training cases similar to the ones in Section~\ref{sec:ideal_pend_results}. First we outline a problem encountered when training over large domains, and find an appropriate domain size for the training cases. Then we evaluate network performances after 50000 iterations using linearly-spaced, random uniformly-distributed, and adjacent data.

\subsubsection{Training over a large domain}

\label{sec:large_domain_training_pend}

\begin{figure}[ht]
    \centering
    \includegraphics[width=0.75\textwidth]{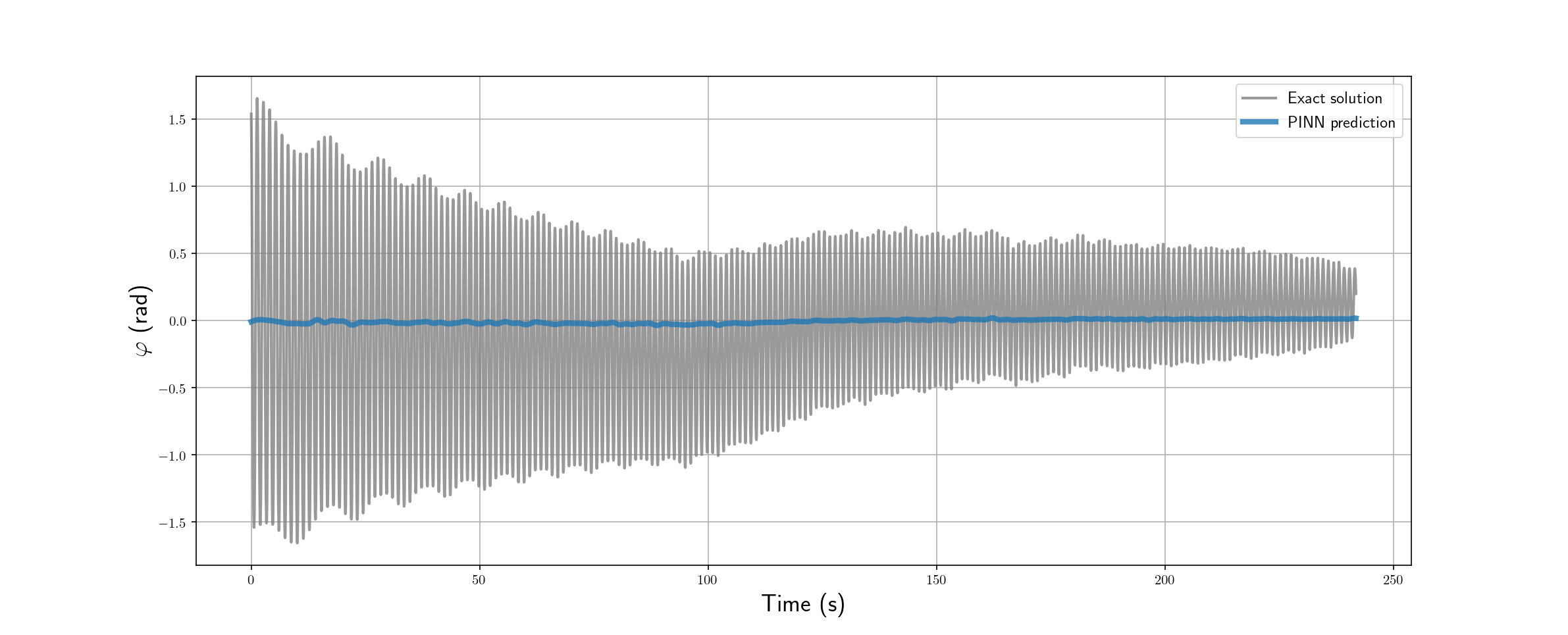}
    \caption{PINN predictions over the entire domain of the sampled training data. The PINN fails to arrive at a valid solution due to the difficulty of optimizing over a large domain.}
    \label{fig:real_pend_pinn_pred_failure}
\end{figure}

\begin{figure}[ht]
    \centering
    \includegraphics[width=0.75\textwidth]{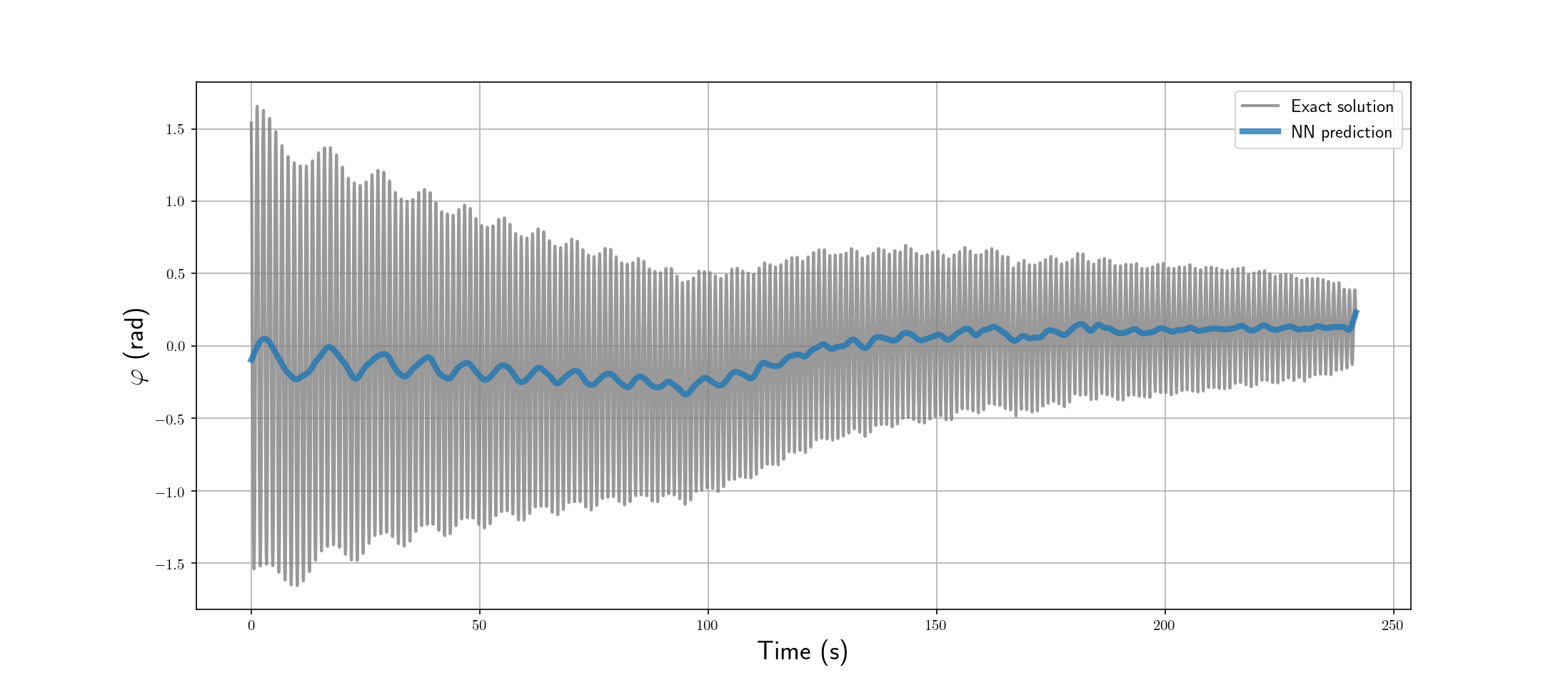}
    \caption{NN predictions over the entire domain of the sampled training data. Similarly to the PINN the NN also fails to converge, although it is more flexible in its predictive capability due to not being constrained by the physics loss term.}
    \label{fig:real_pend_nn_pred_failure}
\end{figure}

In the initial attempt to train the data, the networks failed to converge to an accurate solution when the data was sampled over the entire time domain of oscillation. Figures~\ref{fig:real_pend_pinn_pred_failure} and~\ref{fig:real_pend_nn_pred_failure} show this problem for the PINN and NN respectively. Both networks meet their termination conditions at just over 2000 iterations, as the optimization algorithm is not able to learn anything further. This relates to a common problem that NNs, and particularly PINNs, face. They are difficult to train for high-frequency features or solutions~\cite{Wang2021}. This is due to the spectral bias of NNs, where the rate of convergence for low-frequency loss components is much faster than that of high-frequency ones. Moseley et al. proposed FBPINNs as a solution to this problem~\cite{moseley2021finite}, but for our purposes we choose a simpler approach. Increasing the size of the domain is equivalent to increasing the solution's frequency. Therefore we gradually decrease the domain size to find an appropriately-sized problem that we can fix for the training cases. 

We keep the network structures fixed and vary the domain proportion of the data that we use for training. For all of the data proportions, we use linearly-spaced data with a spacing of 7 and 23 points for train and test datasets respectively. In contrast to the train and test points, we do not constrain the collocation points to be within the domain proportion that we evaluate. This yielded the best results, and is reasonable choice given that collocation points can be evaluated outside of the problem scope if the governing equation is known. Additionally we set both the gradient and function value/parameter tolerance parameters (\verb|tolerance_grad| and \verb|tolerance_change|) to 0, thereby eliminating the termination condition and allowing LBFGS to train indefinitely until the maximum number of iterations. This allowed the optimizer to overcome a training rut where it mistakenly assumes that there is nothing left to learn. We run the optimizer for as many iterations as required and report the findings in Table~\ref{tab:real_pend_domain_variation}.

\begin{table}[ht!]
\centering
\begin{tabular}{|c|c|c|c|c|c|c|}
\hline
\textbf{Domain Proportion} & \multicolumn{1}{l|}{\textbf{Time (s)}} & $\mathbf{N_d}$  & $\mathbf{N_t}$ & \textbf{NN}                                            & \textbf{PINN}                                          & $\mathbf{b}$ \\ \hline
\textbf{0.1}               & \textbf{24.33}                         & \textbf{1429}  & \textbf{435}  & \begin{tabular}[c]{@{}c@{}}0.0227\\ 30000\end{tabular} & \begin{tabular}[c]{@{}c@{}}0.1095\\ 30000\end{tabular} & 0.0587     \\ \hline
\textbf{0.2}               & \textbf{48.62}                         & \textbf{2858}  & \textbf{870}  & \begin{tabular}[c]{@{}c@{}}0.0259\\ 29895\end{tabular} & \begin{tabular}[c]{@{}c@{}}0.1482\\ 55000\end{tabular} & 0.0324     \\ \hline
\textbf{0.4}               & \textbf{97.37}                         & \textbf{5715}  & \textbf{1740} & \begin{tabular}[c]{@{}c@{}}0.0324\\ 27930\end{tabular} & \begin{tabular}[c]{@{}c@{}}0.2811\\ 67005\end{tabular} & 0.0212     \\ \hline
\textbf{0.6}               & \textbf{145.90}                        & \textbf{8572}  & \textbf{2609} & \begin{tabular}[c]{@{}c@{}}0.0414\\ 35805\end{tabular} & \begin{tabular}[c]{@{}c@{}}0.7250\\ 240\end{tabular}   & 0.0002     \\ \hline
\textbf{0.8}               & \textbf{194.33}                        & \textbf{11429} & \textbf{3479} & \begin{tabular}[c]{@{}c@{}}0.0500\\ 39615\end{tabular} & \begin{tabular}[c]{@{}c@{}}0.6528\\ 135\end{tabular}   & 0.0000     \\ \hline
\end{tabular}
\caption{RMSE values based on varying sizes of the domain. The time column is the size of the domain in seconds. $\mathbf{N_d}$ and $\mathbf{N_t}$ are the number of train and test points respectively. $\mathbf{b}$ is the learned value of the friction coefficient. For the NN and PINN entries we report the RMSE on the first line and the iteration number on the second line. We stop the training early if we observe the RMSE value remaining constant for an extended number of iterations.}
\label{tab:real_pend_domain_variation}
\end{table}

\begin{figure}[ht]
    \centering
    \includegraphics[width=\textwidth]{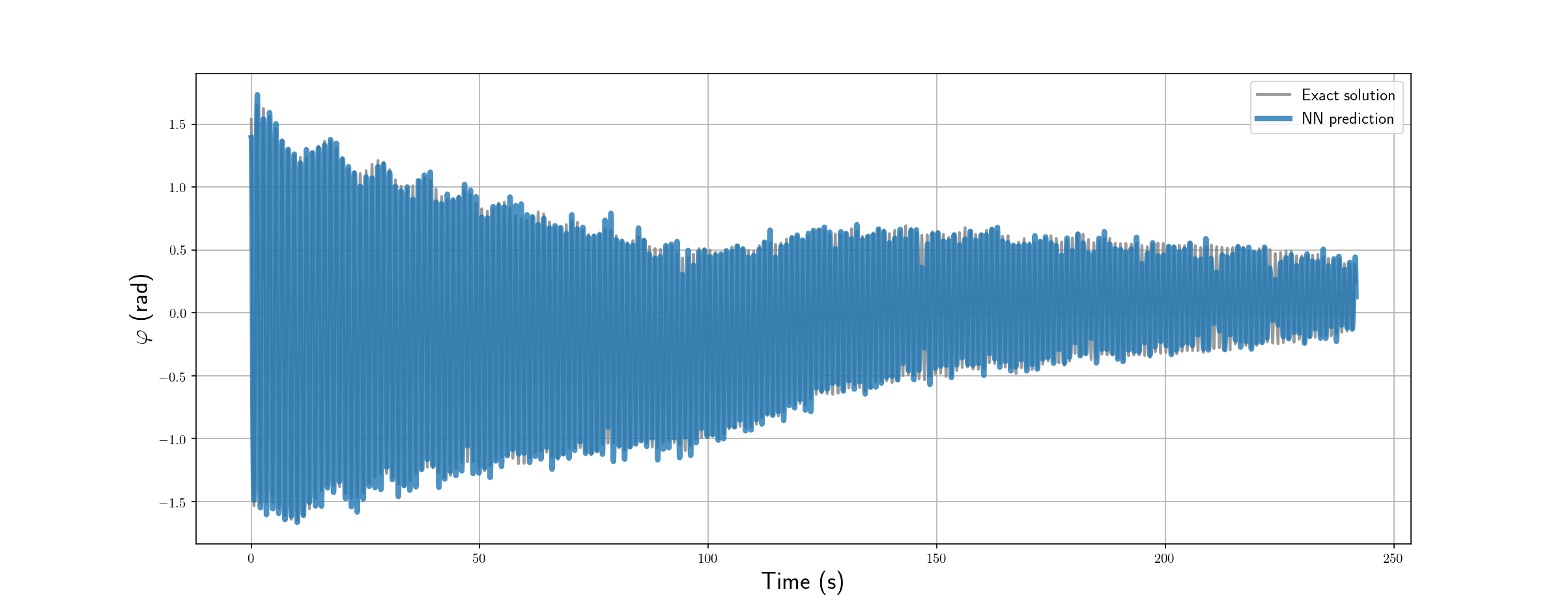}
    \caption{NN predictions over the entire domain of the sampled training data, after 34380 iterations. The main difference from the predictions shown in Figure~\ref{fig:real_pend_nn_pred_failure} is that we do not enforce any termination conditions, and instead allow the training to run indefinitely.}
    \label{fig:real_pend_nn_pred_whole_domain_success}
\end{figure}

The results show that the NN is eventually able to overcome the spectral bias issue for all domain proportions if it is allowed to run indefinitely. We re-ran the training case with the entire domain for the NN, but with indefinite training, and found that it is able to converge to an accurate solution after 34380 iterations with an RMSE of $0.0538$. Figure~\ref{fig:real_pend_nn_pred_whole_domain_success} shows this solution. The PINN on the other hand does not perform as well, and is only able to converge to a reasonable solution for 20\% of the domain size.

After a careful investigation of this problem, we found that the cause of it is the large $N_d$ values for each training case. PINNs can often make more accurate predictions if they are provided with less training data. This is because of the data loss term $\pazocal{L}_d$ dominating over the physics loss term $\pazocal{L}_p$, causing it to become less flexible at adapting to the underlying physics of the problem. Therefore we re-ran the last three PINN evaluations in Table~\ref{tab:real_pend_domain_variation} with less data. The tables in~\ref{tab:pinn_preds_less_data_domains} show the results based on a variation of the number of training points. By comparing the RMSE values in Table~\ref{tab:real_pend_domain_variation} with the ones in Table~\ref{tab:pinn_preds_less_data_domains}, we observe that using less training points allows the PINN to make more accurate predictions after training. More expressive architectures and an extensive hyperparameter grid search would be required to find models that show significant increases in accuracy. For our purposes, we fix the domain size at 20\% to allow for training flexibility for the test cases.

\begin{table}[ht]
\centering
\begin{subtable}[t]{0.4\linewidth}
\centering
\begin{tabular}{|c|c|c|c|}
\hline
\textbf{$\mathbf{N_d}$} & \textbf{RMSE}               & \textbf{Iters}             & $\mathbf{b}$                  \\ \hline
\textbf{223}  & \multicolumn{1}{l|}{0.1923} & \multicolumn{1}{l|}{75000} & \multicolumn{1}{l|}{0.0180} \\ \hline
\textbf{334}  & 0.1967                      & 74910                      & 0.0214                      \\ \hline
\textbf{500}  & 0.2801                      & 74775                      & 0.0185                      \\ \hline
\end{tabular}
\caption{Domain proportion = \textbf{0.4}}
\label{tab:pinn_preds_less_data_domain0.4}
\end{subtable}
\begin{subtable}[t]{0.4\linewidth}
\centering
\begin{tabular}{|c|c|c|c|}
\hline
\textbf{$\mathbf{N_d}$} & \textbf{RMSE}               & \textbf{Iters}             & \textbf{b}                  \\ \hline
\textbf{250}  & \multicolumn{1}{l|}{0.5598} & \multicolumn{1}{l|}{70575} & \multicolumn{1}{l|}{0.0308} \\ \hline
\textbf{500}  & 0.3257                      & 75000                      & 0.0307                      \\ \hline
\textbf{1000} & 0.3340                      & 75000                      & 0.0238                      \\ \hline
\end{tabular}
\caption{Domain proportion = \textbf{0.6}}
\label{tab:pinn_preds_less_data_domain0.6}
\end{subtable}
\begin{subtable}[t]{0.4\linewidth}
\centering
\begin{tabular}{|c|c|c|c|}
\hline
$\mathbf{N_d}$ & \textbf{RMSE}               & \textbf{Iters}             & \textbf{b}                  \\ \hline
\textbf{334}  & \multicolumn{1}{l|}{0.5111} & \multicolumn{1}{l|}{65069} & \multicolumn{1}{l|}{0.0212} \\ \hline
\textbf{667}  & 0.3672                      & 74804                      & 0.0302                      \\ \hline
\textbf{889}  & 0.3999                      & 75000                      & 0.0471                      \\ \hline
\end{tabular}
\caption{Domain proportion = \textbf{0.8}}
\label{tab:pinn_preds_less_data_domain0.8}
\end{subtable}
\caption{PINN prediction RMSE values for the last three domain proportions shown in Table~\ref{tab:real_pend_domain_variation}, but with less data. Decreasing the amount of data enables more accurate PINN models.}
\label{tab:pinn_preds_less_data_domains}
\end{table}

\subsubsection{Linearly-spaced data}

\begin{figure}[ht]
    \centering
    \includegraphics[width=\textwidth]{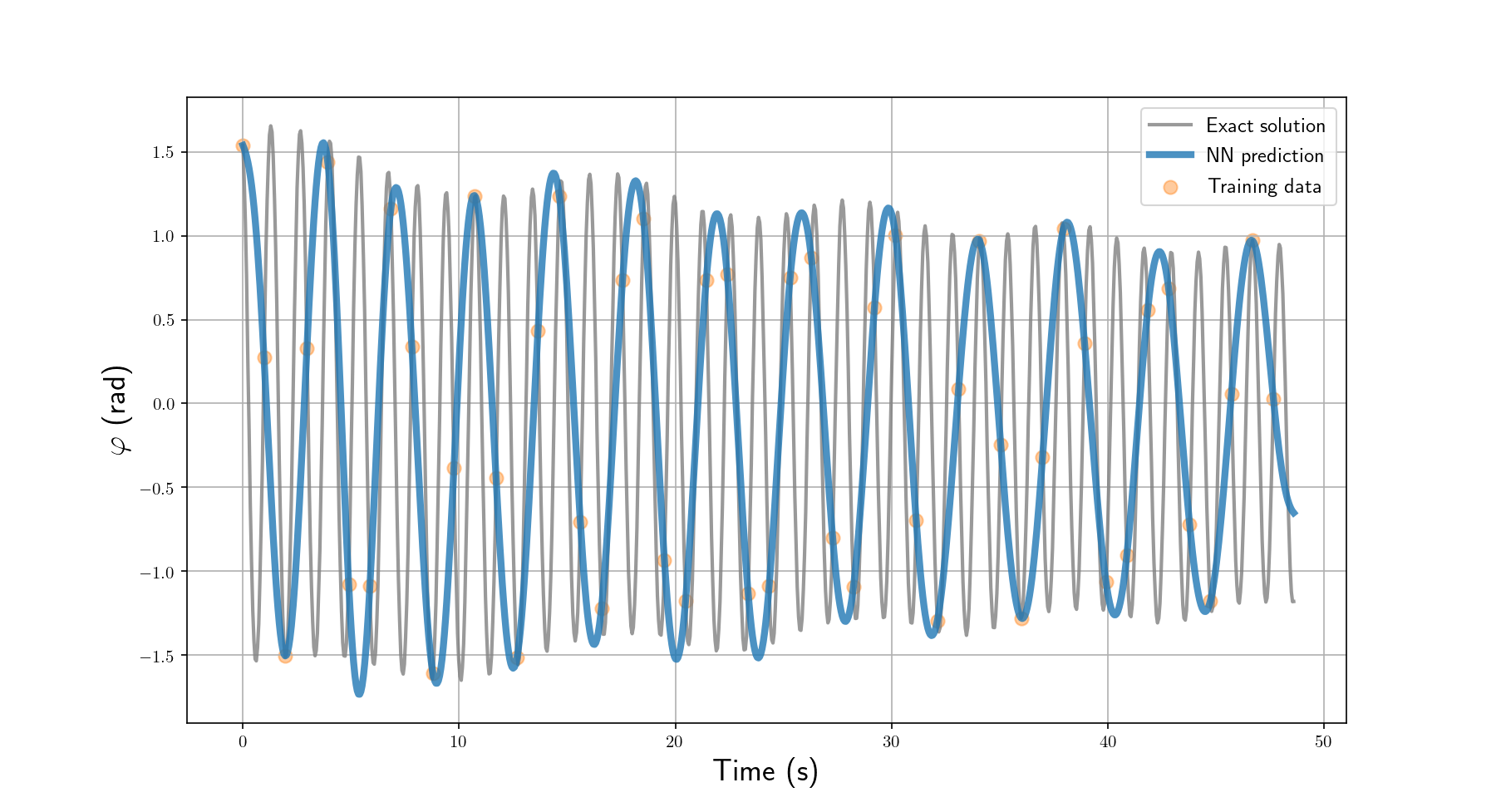}
     \caption{NN predictions based on training with 50 linearly-spaced points. The NN solution misses the majority of the sinusoids as it is only able to fit data.}
    \label{fig:real_pend_nn_linspace50}
\end{figure}

\begin{figure}[ht]
    \centering
    \includegraphics[width=\textwidth]{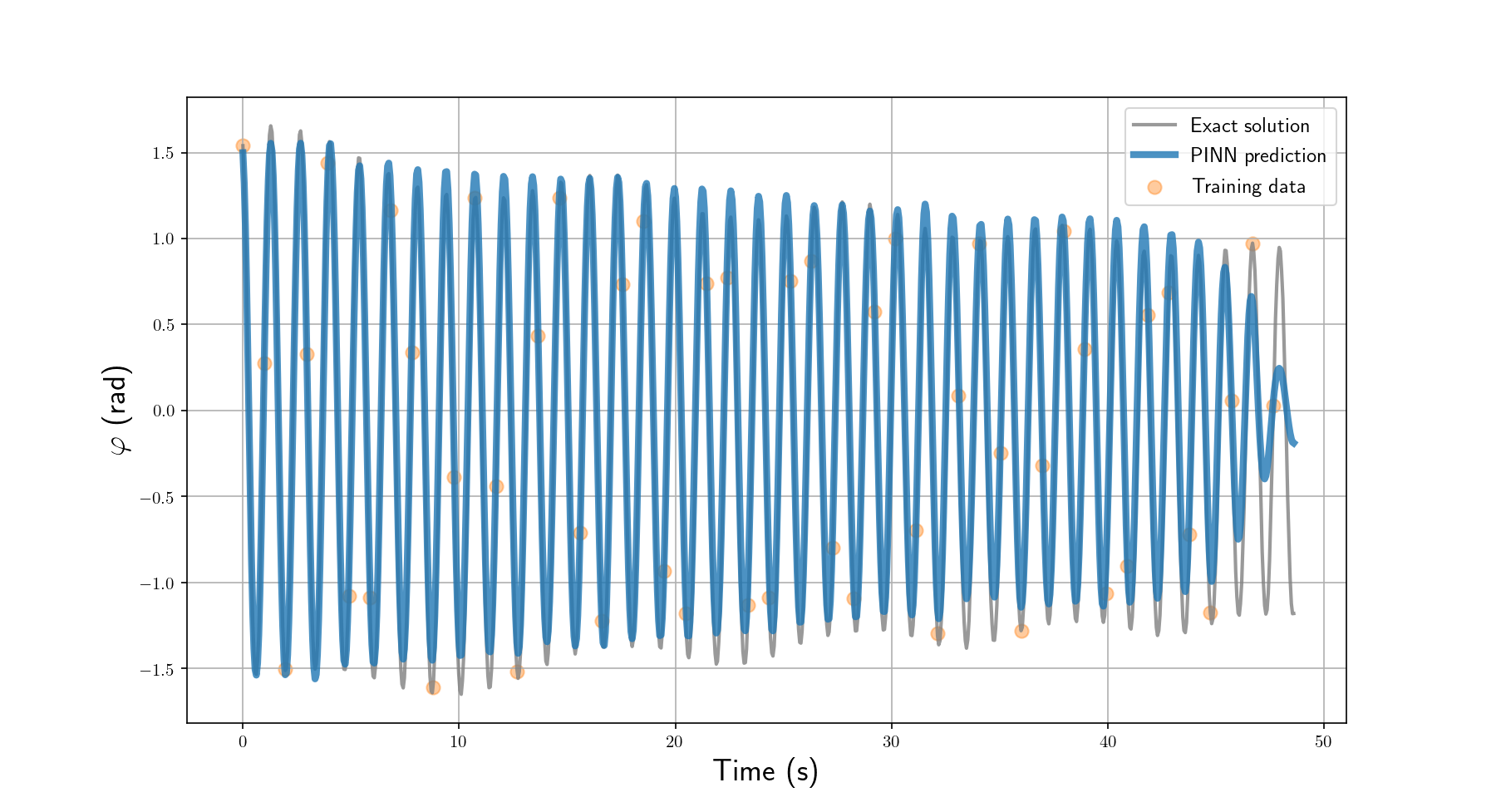}
    \caption{PINN predictions based on training with 50 linearly-spaced points. In contrast to the NN prediction in Figure~\ref{fig:real_pend_nn_linspace50}, the PINN is able to capture the sinusoids correctly due to physics loss term.}
    \label{fig:real_pend_pinn_linspace50}
\end{figure}

Table~\ref{tab:real_pend_linspace} shows the training results based on a variation of the number of linearly-spaced training points. The PINN maintains its accuracy even with the reduction $N_d$, while the NN suffers considerably after 167 points. Figure~\ref{fig:real_pend_Nd_variation_loss_graph} shows this trend visualized, where we see that the NN predictions fail for less data points and the PINN predictions stay relatively consistent. A closer inspection into the predicted solutions in Figures~\ref{fig:real_pend_nn_linspace50} and~\ref{fig:real_pend_pinn_linspace50}, allows us to see that similar to the ideal case predictions in Section~\ref{sec:ideal_data_linspace}, the PINN is able to regularise the solution according to physics whereas the NN is only able to fit the data points.

\begin{table}[ht]
\centering
\begin{subtable}{0.45\linewidth}
\centering
\begin{tabular}{|c|c|c|c|}
\hline
$\mathbf{N_d}$ & \textbf{NN}                 & \textbf{PINN}               & \textbf{b}                  \\ \hline
\textbf{1000} & \multicolumn{1}{l|}{0.0245} & \multicolumn{1}{l|}{0.1357} & \multicolumn{1}{l|}{0.0346} \\ \hline
\textbf{500}  & 0.0231                      & 0.1427                      & 0.0284                      \\ \hline
\textbf{334}  & 0.0253                      & 0.1636                      & 0.0042                      \\ \hline
\textbf{250}  & 0.2142                      & 0.1344                      & 0.0288                      \\ \hline
\textbf{200}  & 0.1110                      & \multicolumn{1}{l|}{0.1381} & \multicolumn{1}{l|}{0.0336} \\ \hline
\textbf{167}  & \multicolumn{1}{l|}{0.0408} & \multicolumn{1}{l|}{0.1772} & \multicolumn{1}{l|}{0.0374} \\ \hline
\textbf{143}  & \multicolumn{1}{l|}{0.7083} & \multicolumn{1}{l|}{0.3675} & \multicolumn{1}{l|}{0.0468} \\ \hline
\textbf{125}  & \multicolumn{1}{l|}{0.8337} & \multicolumn{1}{l|}{0.1718} & 0.0362                      \\ \hline
\textbf{100}  & \multicolumn{1}{l|}{1.3689} & \multicolumn{1}{l|}{0.1661} & \multicolumn{1}{l|}{0.0334} \\ \hline
\textbf{84}   & 1.3930                      & 0.1838                      & 0.0342                      \\ \hline
\textbf{67}   & 2.1644                      & 0.2325                      & 0.0350                      \\ \hline
\textbf{50}   & 1.3206                      & 0.1828                      & 0.0350                      \\ \hline
\end{tabular}
\caption{Linearly-spaced points.}
\label{tab:real_pend_linspace}
\end{subtable}
\begin{subtable}{0.45\linewidth}
\centering
\begin{tabular}{|c|c|c|c|}
\hline
$\mathbf{N_d}$ & \textbf{NN}                 & \textbf{PINN}               & \textbf{b}                  \\ \hline
\textbf{1000} & \multicolumn{1}{l|}{0.0319} & \multicolumn{1}{l|}{0.1394} & \multicolumn{1}{l|}{0.0292} \\ \hline
\textbf{500}  & 0.0512                      & 0.1428                      & 0.0273                      \\ \hline
\textbf{334}  & 0.1596                      & 0.1491                      & 0.0293                      \\ \hline
\textbf{250}  & 0.1012                      & 0.1514                      & 0.0237                      \\ \hline
\textbf{200}  & 0.6762                      & \multicolumn{1}{l|}{0.1724} & \multicolumn{1}{l|}{0.0204} \\ \hline
\textbf{167}  & \multicolumn{1}{l|}{1.0134} & \multicolumn{1}{l|}{0.2131} & \multicolumn{1}{l|}{0.0198} \\ \hline
\textbf{143}  & \multicolumn{1}{l|}{0.7937} & \multicolumn{1}{l|}{0.1942} & \multicolumn{1}{l|}{0.0182} \\ \hline
\textbf{125}  & 1.3762                      & \multicolumn{1}{l|}{0.2154} & 0.0157                      \\ \hline
\textbf{100}  & \multicolumn{1}{l|}{2.3499} & \multicolumn{1}{l|}{0.2020} & \multicolumn{1}{l|}{0.0165} \\ \hline
\textbf{84}   & 2.0139                      & 0.2046                      & 0.0200                      \\ \hline
\textbf{67}   & 3.3333                      & 0.3644                      & 0.0142                      \\ \hline
\textbf{50}   & 2.9213                      & 0.2221                      & 0.0170                      \\ \hline
\end{tabular}
\caption{Uniformly-distributed points.}
\label{tab:real_pend_random-uniformly-distributed}
\end{subtable}
\caption{RMSE values for variations of numbers of training points. The PINN predictions for both cases stay relatively consistent, whereas the NN predictions fail as the number of points decreases.}
\label{tab:real_pend_random_points}
\end{table}

\subsubsection{Uniformly-distributed random data}

\begin{figure}[ht]
    \centering
    \includegraphics[width=\textwidth]{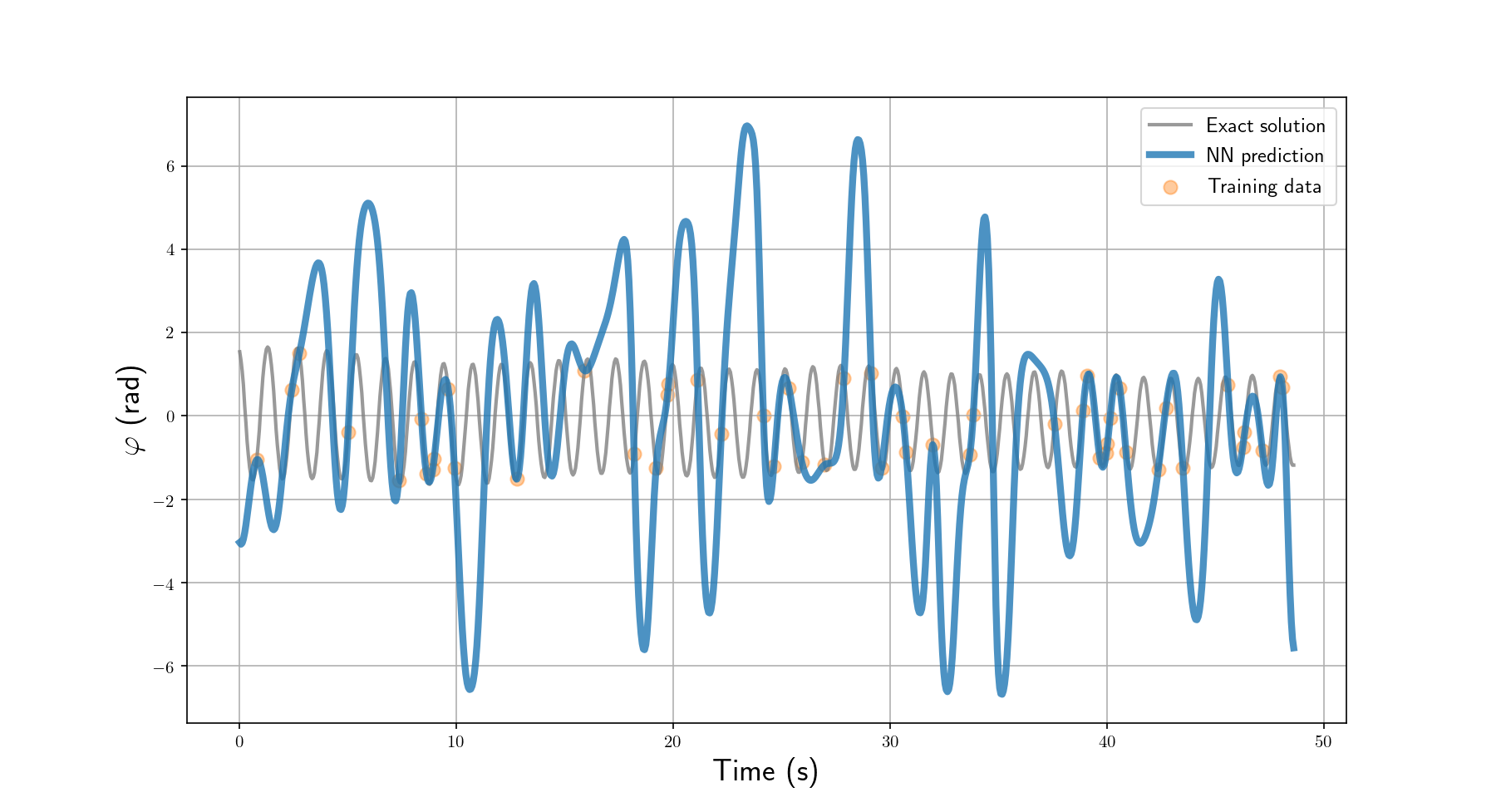}
    \caption{NN predictions based on training with 50 uniformly-distributed points. The NN fails to make reasonable predictions in areas with no training points.}
    \label{fig:real_pend_nn_uniform-dist_points50}
\end{figure}

\begin{figure}[ht]
    \centering
    \includegraphics[width=\textwidth]{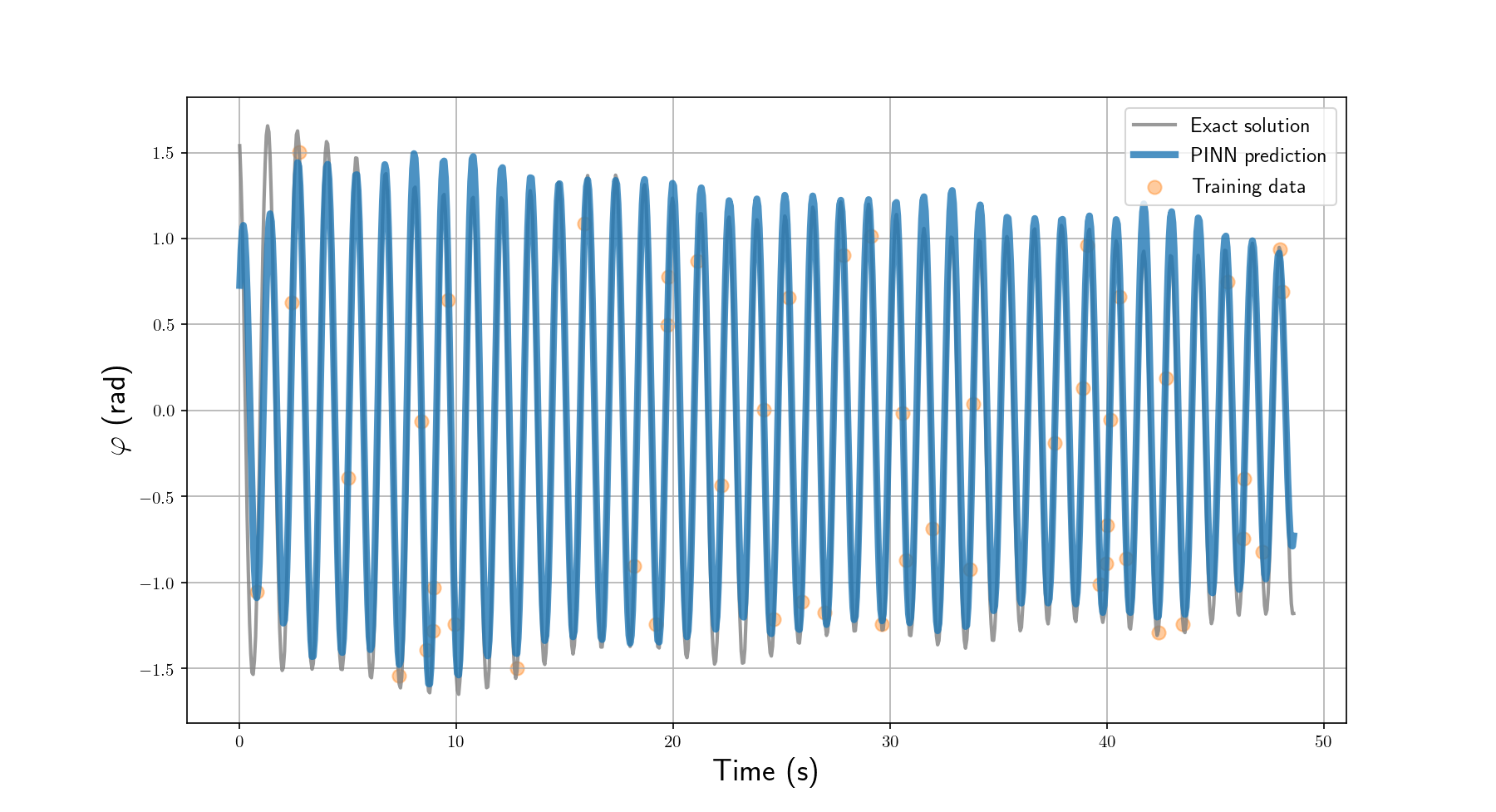}
    \caption{PINN predictions based on training with 50 uniformly-distributed points. The PINN maintains the trend of making valid predictions.}
    \label{fig:real_pend_pinn_uniform-dist_points50}
\end{figure}

\begin{figure}[ht]
    \centering
    \includegraphics[width=\textwidth]{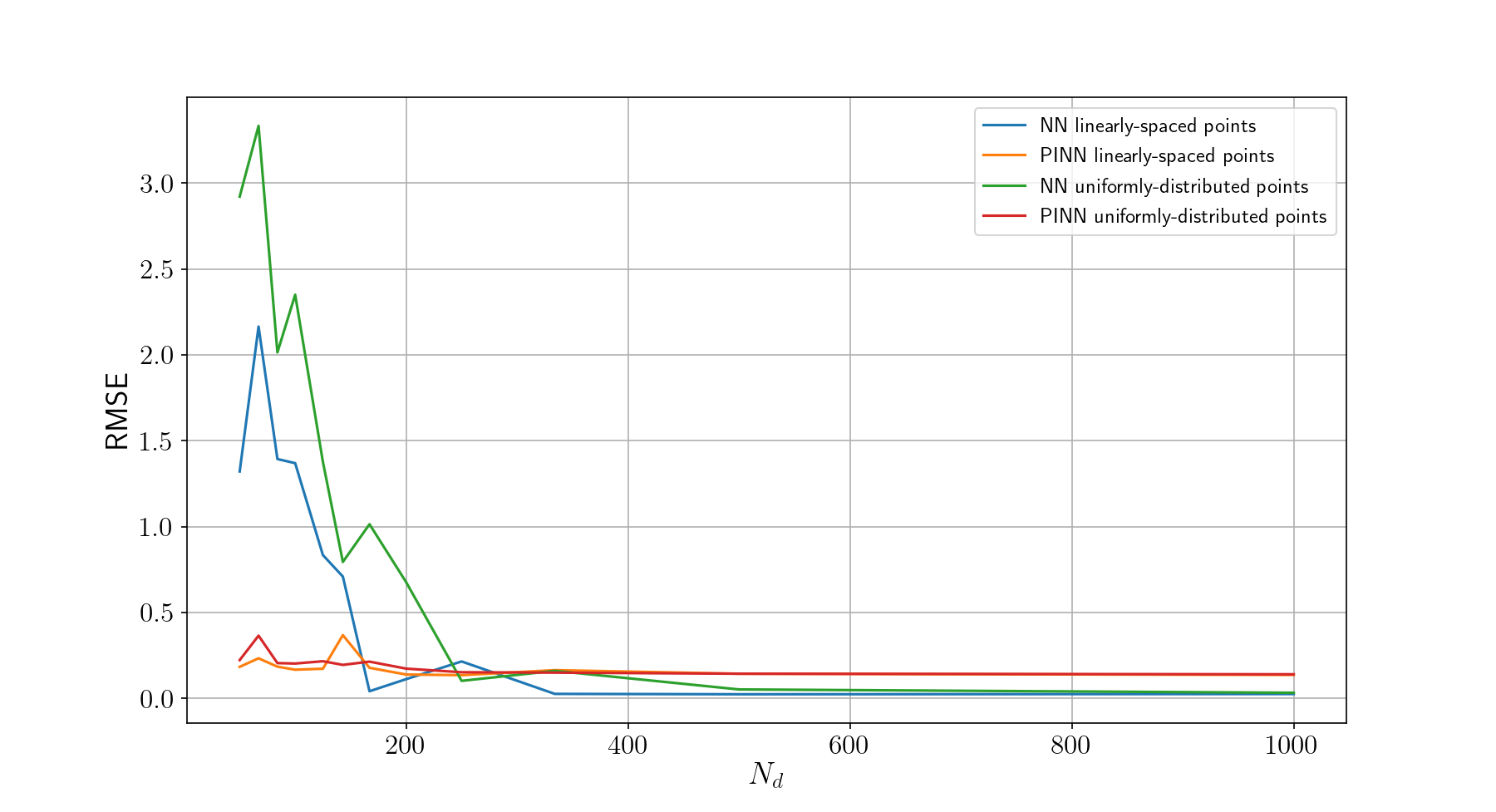}
    \caption{NN and PINN RMSE values for linearly-spaced and uniformly-distributed data, reported in Table~\ref{tab:real_pend_random_points}. The PINN maintains a constant accuracy irrespective of the number of training points, while the NN fails as the points get less.}
\label{fig:real_pend_Nd_variation_loss_graph}
\end{figure}

Table~\ref{tab:real_pend_random-uniformly-distributed} shows the training results based on a variation of the number of uniformly-distributed training points. Similar to the case with linearly-spaced points, the PINN maintains its accuracy regardless of $N_d$ whereas the NN does not. However the NN suffers more in the case of uniformly-distributed points than it does for linearly-spaced points, as we show in Figure~\ref{fig:real_pend_Nd_variation_loss_graph}. By comparing Figures~\ref{fig:real_pend_nn_uniform-dist_points50} and~\ref{fig:real_pend_pinn_uniform-dist_points50}, we observe the continuing trend of NNs fitting data points in contrast with PINNs that are able to regularise based on the governing equation.

\subsubsection{Adjacent points}

\begin{figure}[ht]
    \centering
    \includegraphics[width=\textwidth]{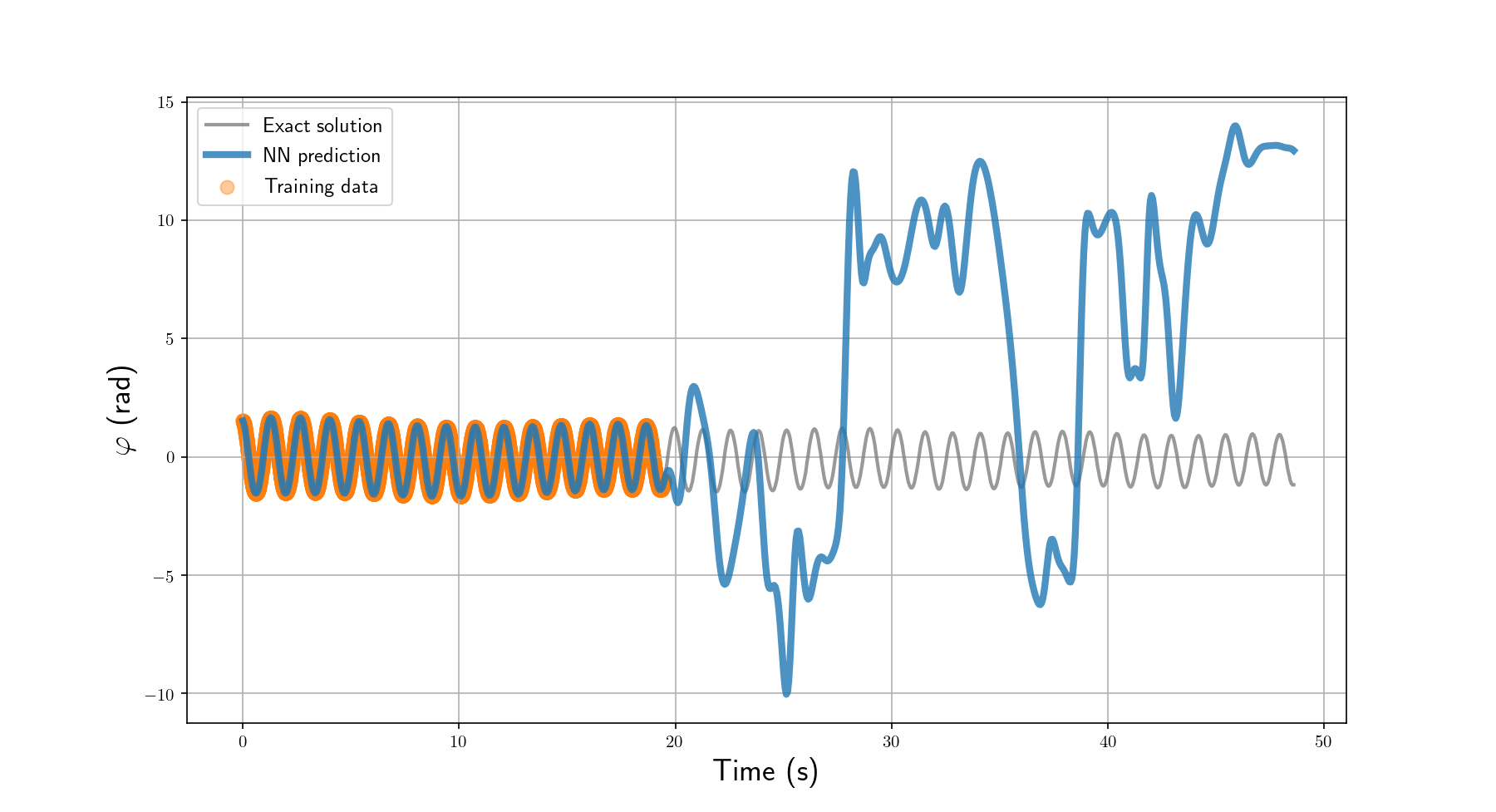}
    \caption{NN predictions based on adjacent points that comprise 40\% of the problem domain. The predictions become unstable outside of the training data region.}
\label{fig:real_pend_nn_extrap0.4}
\end{figure}

\begin{figure}[ht]
    \centering
    \includegraphics[width=\textwidth]{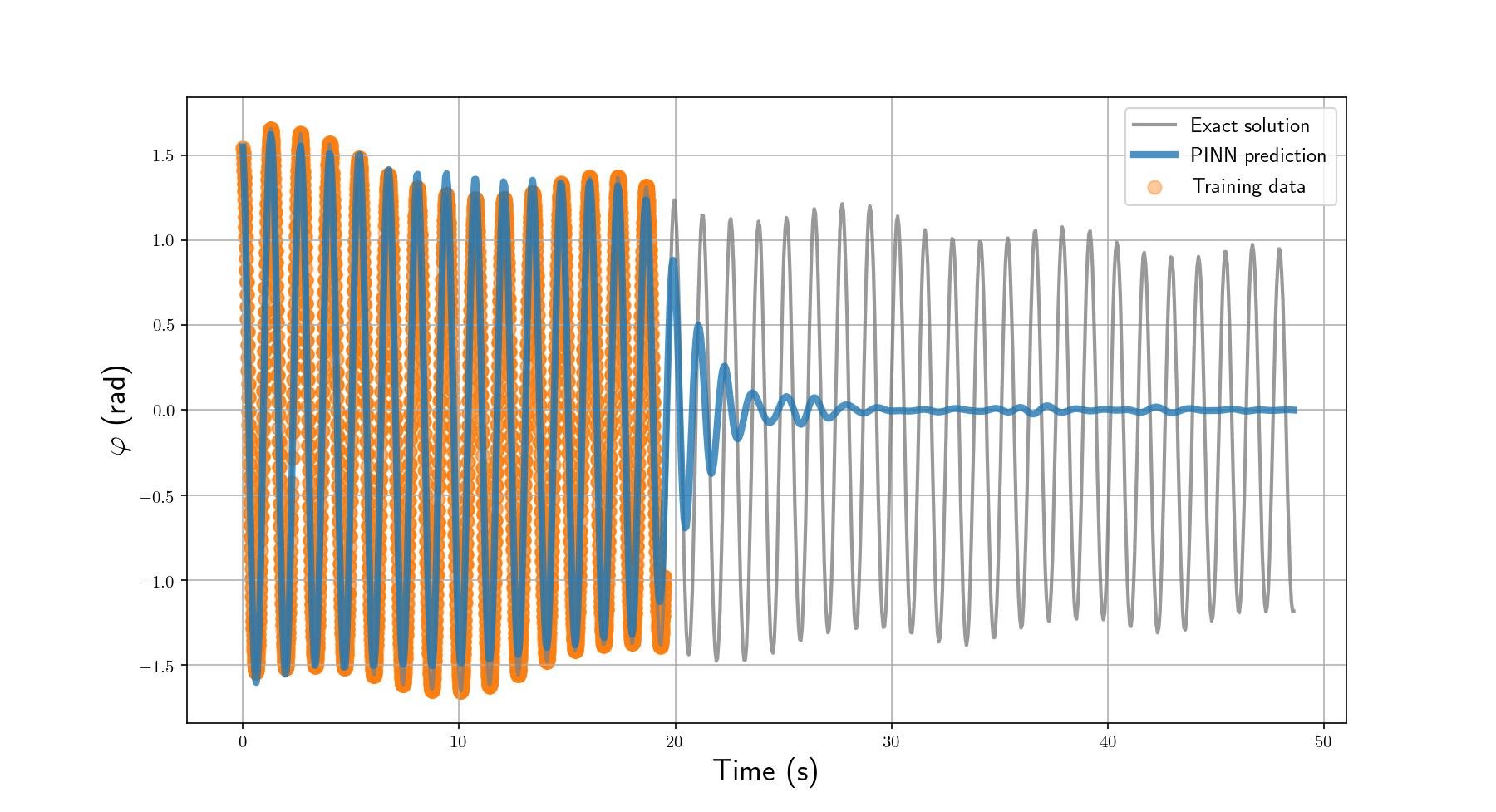}
    \caption{PINN predictions based on adjacent points that comprise 40\% of the problem domain. The predictions maintain stability just outside of the training data, but fail to extrapolate for the rest of the domain.}
\label{fig:real_pend_pinn_extrap0.4}
\end{figure}

Table~\ref{tab:real_pend_adj_points} shows the results for training based on adjacent points, taken as percentages of the problem domain. In contrast to the previous evaluations, here both the PINN and the NN fail at predicting the solution outside of the training data, even though the NN suffers more significantly. Figure~\ref{fig:real_pend_pinn_extrap0.4} shows that the PINN predictions maintain a semblance of the physical behaviour just outside of the training points, but then fall to 0 after that. The NN predictions on the other hand, shown in Figure~\ref{fig:real_pend_nn_extrap0.4}, maintain no semblance of the governing physics and are in ranges that are entirely outside of physical plausibility.

\begin{table}[ht]
\centering
\begin{tabular}{|c|c|c|c|}
\hline
\textbf{Train Percent} & \textbf{NN}                 & \textbf{PINN}               & \textbf{b}                  \\ \hline
\textbf{80}            & \multicolumn{1}{l|}{2.3288} & \multicolumn{1}{l|}{0.3474} & \multicolumn{1}{l|}{0.0381} \\ \hline
\textbf{60}            & 4.6108                      & 0.5218                      & 0.0514                      \\ \hline
\textbf{40}            & 6.3657                      & 0.6498                      & 0.0653                      \\ \hline
\textbf{20}            & 6.2091                      & 0.8083                      & 0.1065                      \\ \hline
\end{tabular}
\caption{RMSE values for percentages of adjacent points starting from $t = 0$. Both the PINN and NN fail at predicting accurate solutions but the NN fails harder.}
\label{tab:real_pend_adj_points}
\end{table}

\section{Closing Remarks}

In this chapter we have analysed the predictive performance and trainability of PINNs against standard NNs for the prediction of a pendulum's oscillation angle, in an ideal simulated scenario and using real sensor data from an experiment. We have shown that in most cases, for both the simulated and real system, the PINN is able to regularise the the solution according to physical principles resulting in accurate predictions in low-data scenarios. This includes extrapolating the solution outside of the training data in the simulated case, since the numerical solution directly adheres to the governing equation. We hypothesize that, if given a more accurate description of the governing equation for the real system, PINNs should also be able to extrapolate outside of the training domain for the real system as well. In contrast, standard NNs fail to predict an accurate solution in the absence of sufficient data, which highlights a flaw in uninformed deep learning approaches where the NN is treated as a black box. The pendulum system has served as a simple but effective illustrative example for the effectiveness of incorporating physics knowledge into deep learning for physical systems.

%% file: Chapter4/chapter4.tex

\chapter{Predicting the Surface Temperatures Across a Metal Block During Heating}

\label{chapter:heat_diffusion}

\ifpdf
    \graphicspath{{Chapter4/Figs/Raster/}{Chapter4/Figs/PDF/}{Chapter4/Figs/}}
\else
    \graphicspath{{Chapter4/Figs/Vector/}{Chapter4/Figs/}}
\fi

\section{Introduction}

In this chapter, we study the performance of PINNs for a slightly more complicated dynamical system --- heat diffusion across the 2D surface of a metal block. The setup that we use for our experiment is irregular and does not necessarily adhere to a perfect physical model or governing equation. Additionally, we use a non-uniform scrap metal block that we do not have ground-truth reference values for its physical parameters (thermal conduction, density, emissivity, etc.), so we use best-guess values for the coefficients. The sensor that we use is inexpensive and highly susceptible to noise. Therefore instead of running an idealized simulation as a benchmark, we immediately start by analysing the experimental data. Therefore, we are evaluating PINNs on real data in a situation that is dominated by significant amounts of noise, and where the exact physics is unknown. First we go over the dynamics of heat diffusion by showing a derivation of the heat equation. Then we outline our experimental procedure for data collection, denoising, and training. Finally we end the chapter with some closing insights based on the results.

\section{Heat Diffusion Dynamics}

The heat equation~\cite{ahtt5p_heat_conduction}, also referred to as the heat diffusion or heat conduction equation, is commonly one of the first PDEs that students are introduced to~\cite{farlow1993partial_laplace}. It provides a physical interpretation for the dynamics of heat transfer across space through the process of conduction. We outline its derivation which we borrow jointly from Strauss~\cite{strauss2008partial} and a PDEs course handout from Stanford University~\cite{Levandosky2003}.

First we consider a region $D\in \mathbb{R}^n$ where $n$ is the number of dimensions. Let $x = [x_1,..., x_n]^T$ be a spatial vector in $\mathbb{R}^n$, and let $u(x, \, t)$ be the temperature at point $x$ and time $t$. Additionally, let $c$ be the specific heat of the material of region $D$ and $\rho$ its density. We express $H(t)$, the total amount of heat in calories contained in $D$ as follows:

\begin{equation}
    H(t) = \int_D \, c \, \rho \, u(x, \, t) \, dx \nonumber
\end{equation}

By considering the change in heat we get the following (note the time derivative of $u$):

\begin{equation}
    \label{eq:rate_heat_transfer_initial}
    \frac{dH}{dt} = \int_D \, c \, \rho \, u_t(x, \, t) \, dx 
\end{equation}

Fourier's law states that the rate of heat transfer is proportional to the negative temperature gradient, meaning that heat can only flow from hot to cold regions at a rate proportional to the thermal conductivity $k$. Mathematically, this is expressed as follows:

\begin{equation}
    \label{eq:rate_heat_transfer_fouriers_law}
    \frac{dH}{dt} = \int_{\partial D} \, k \, \nabla \, u \cdot \mathbf{\hat{n}} \, dS 
\end{equation}

$\partial D$ is the boundary of $D$, $\mathbf{\hat{n}}$ is the outward normal unit vector to $\partial D$, and $dS$ is the surface measure over $\partial D$. By equating Equations~\ref{eq:rate_heat_transfer_initial} and~\ref{eq:rate_heat_transfer_fouriers_law} we obtain the following:

\begin{equation}
    \label{eq:rate_heat_transfer_equated}
    \int_D \, c \, \rho \, u_t(x, \, t) \, dx = \int_{\partial D} \, k \, \nabla \, u \cdot \mathbf{\hat{n}} \, dS 
\end{equation}

The Divergence theorem states that the volume integral over an enclosed volume is equal to the surface integral over the boundary of the volume. For a vector field $F$ this is represented as follows:

\begin{equation}
    \int_{\partial D} \, F \cdot \mathbf{\hat{n}} \, dS = \int_D \, \nabla \cdot F \, dx \nonumber
\end{equation}

Therefore, we simplify Equation~\ref{eq:rate_heat_transfer_equated} to get:

\begin{equation}
    \int_D \, c \, \rho \, u_t(x, \, t) \, dx = \int_D \nabla \cdot (k \, \nabla \, u) \, dx \nonumber
\end{equation}

By further simplifying, we obtain the following PDE:

\begin{equation}
    c \, \rho \, u_t = \nabla \cdot (k \, \nabla u) \nonumber
\end{equation}

Since $c$, $\rho$, and $k$ are constants, by simplifying once more we get the heat equation:

\begin{equation}
\label{eq:heat_equation}
    u_t = \alpha \, \Delta u
\end{equation}

\section{Block Heating Experiment}

\label{sec:heating_diff_exp}

\subsection{Experimental setup}

\label{sec:heat_diff_exp_setup}

\begin{figure}[ht]
    \centering
    \includegraphics[width=0.83\textwidth]{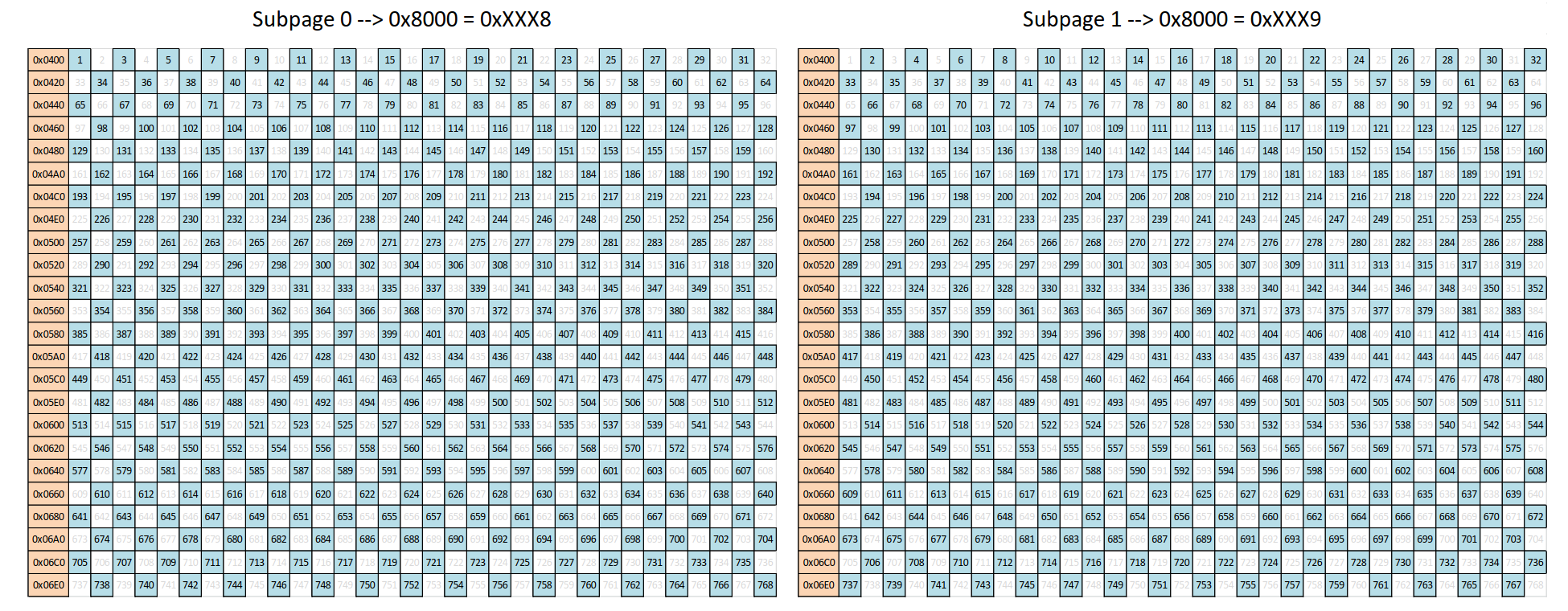}
    \caption{MLX90640 pixel RAM chess reading pattern configuration, borrowed from the datasheet~\cite{Melexis:MLX90640}. The highlighted cells correspond to a subpage and we read one with each I2C transaction. The subpages get updated with new data after each read.}
    \label{fig:chess_reading_pattern_MLX90640}
\end{figure}

In devising a setup for the heat diffusion experiment, we needed to find a way to conveniently collect data without requiring complicated equipment. Therefore, we use scrap metal aluminium alloy block which we heat using a soldering iron. We chose to use an aluminium block because we needed a metal that had a medium level of conductivity so that the observed temperature gradients can be apparent over time. Using a metal that was too thermally conductive, like copper for example, would result in temperature gradients that are not very pronounced.

\begin{figure}[ht]
    \centering
    \includegraphics[scale=0.06, angle=-90]{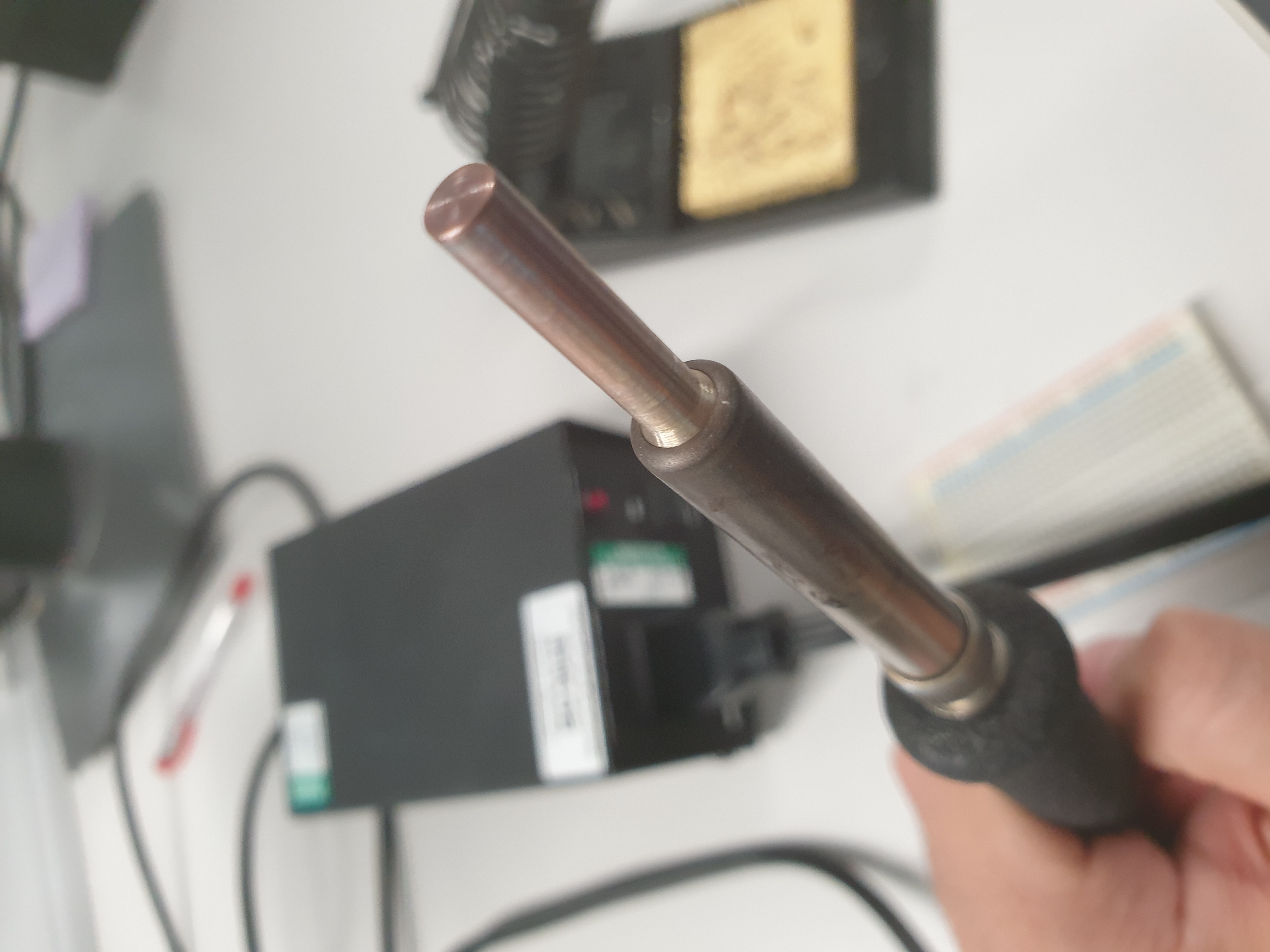}
    \caption{Custom copper tip to improve surface contact for conduction.}
    \label{fig:copper_tip_img}
\end{figure}

\begin{figure}[ht]
    \centering
    \includegraphics[width=\textwidth]{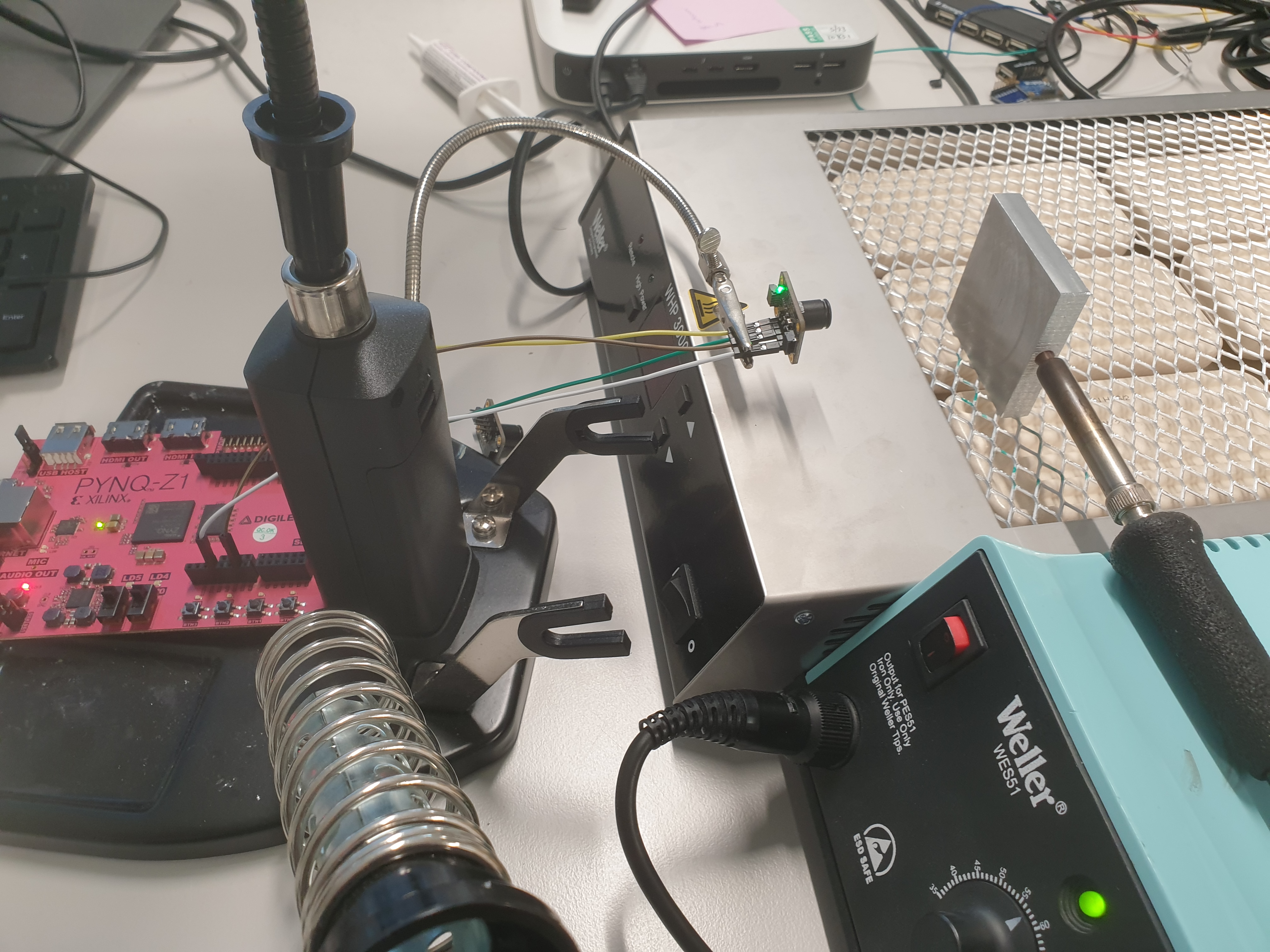}
    \caption{Block heating experimental setup. The MLX90640 is held with an alligator clip attached to a flexible helping hand. The custom solder tip is inserted into the block from the side into a hole so that it fits in place during heating. The sensor is directly connected to the FPGA.}
    \label{fig:block_heating_exp_img}
\end{figure}

We use the PYNQ-Z1 FPGA as the sensing platform, and record surface temperatures across the block using the MLX90640 infra-red (IR) thermal camera~\cite{Melexis:MLX90640}. The MLX90640 has a resolution of 32x24 pixels, a field-of-view (FOV) of 55°x35°, and a temperature range of -40°C -- 300°C. The pixels are split into two subpages within the RAM of the sensor, for the odd and even pixels. These pixels are arranged in a chess-like pattern as Figure~\ref{fig:chess_reading_pattern_MLX90640} shows, and we read the RAM twice and compile the subpages together to get a valid frame. The FPGA issues bulk I2C commands to read the RAM all at once rather than individual I2C reads for each pixel, as we have found that this method is faster. We use the same AXI IIC IP block design shown in Figure~\ref{fig:AXI_IIC_block_design} to communicate with the sensor. We implement a custom driver for interfacing with the MLX90640 through the Python PYNQ API, based off of the manufacturer's device driver library~\cite{MLX90640_library}. The main difference in our implementation is that we obtain the raw frames during the heating and only perform the conversion routines after the experiment is over and we have collected all of the data. The conversion routines require the specification of an emissivity parameter, and based on our aluminium block we use an estimated value of $0.05$ for this. For the heat source we use a WES51 soldering iron, but we replace the conical nib with a custom cylindrical copper tip for better surface contact for the conduction. Figure~\ref{fig:copper_tip_img} shows the custom tip and Figure~\ref{fig:block_heating_exp_img} shows the entire experimental setup.

Before we start the experiment, we configure the sensor by setting the analogue-to-digital converter (ADC) resolution to 19 bits, and the frame refresh rate to 64 Hz. Next we ensure that the camera is positioned correctly so that it covers the entire area of the block. To do this we implement real-time thermal imaging so that we can observe what the camera is capturing. We heat the block slightly so that it is visually apparent on the thermal imager. Using this setup, we move the camera and the block accordingly until both are positioned in place with the entire block surface appearing in frame. Then we leave the block to cool down and after that we turn the temperature up to approximately 298°C. We insert the iron into the hole once the temperature has stabilized and begin recording.

We record the raw sensor readings for 15 minutes. After that we convert the readings into temperature measurement values using conversion routines specified by the sensor manufacturer. We save the readings and then move them onto the workstation to process the data.

\subsection{Data analysis}

\begin{figure}[ht]
    \centering
    \includegraphics[width=\textwidth]{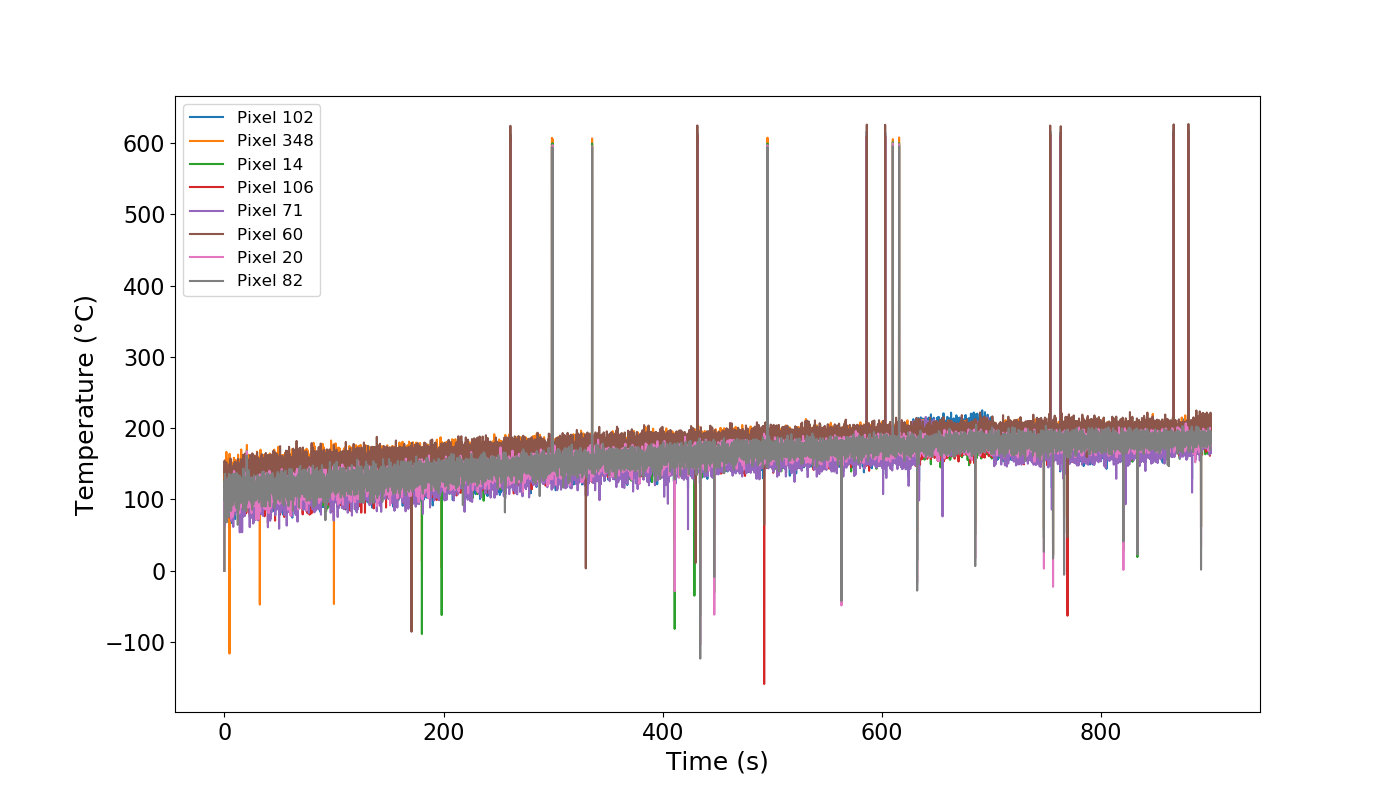}
    \caption{Converted temperature measurements over time for 8 randomly-selected pixels. The measurements are noisy and are also dominated by noise spikes.}
    \label{fig:raw_random_pixels}
\end{figure}

\begin{figure}[ht]
\centering
\begin{subfigure}{0.3\textwidth}
    \centering
    \includegraphics[width=\textwidth]{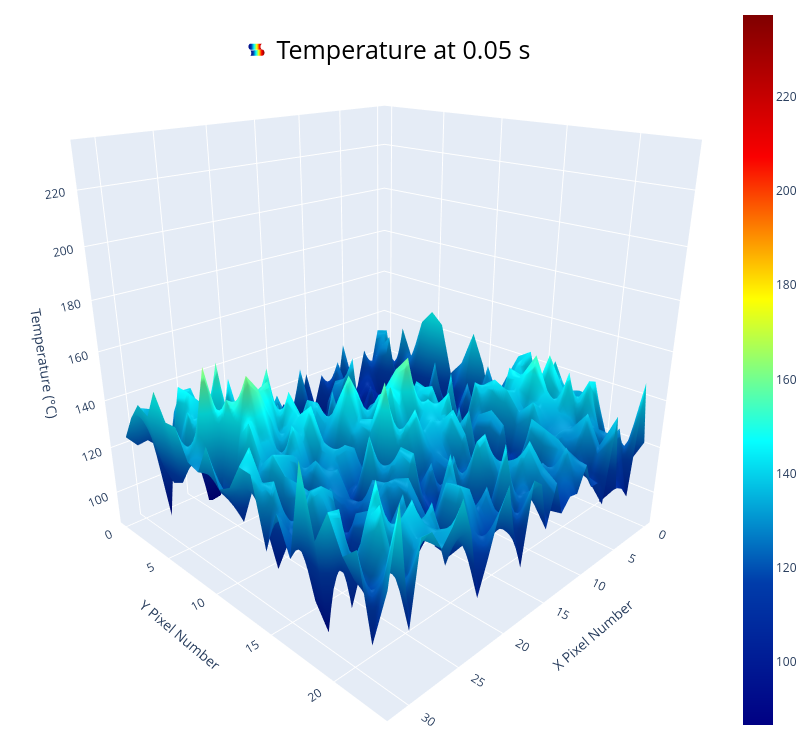}
    \caption{$0.05 \, s$.}
    \label{fig:raw_thermal_3D_t=0.05}
\end{subfigure}
\hspace{5em}
\begin{subfigure}{0.3\textwidth}
    \centering
    \includegraphics[width=\textwidth]{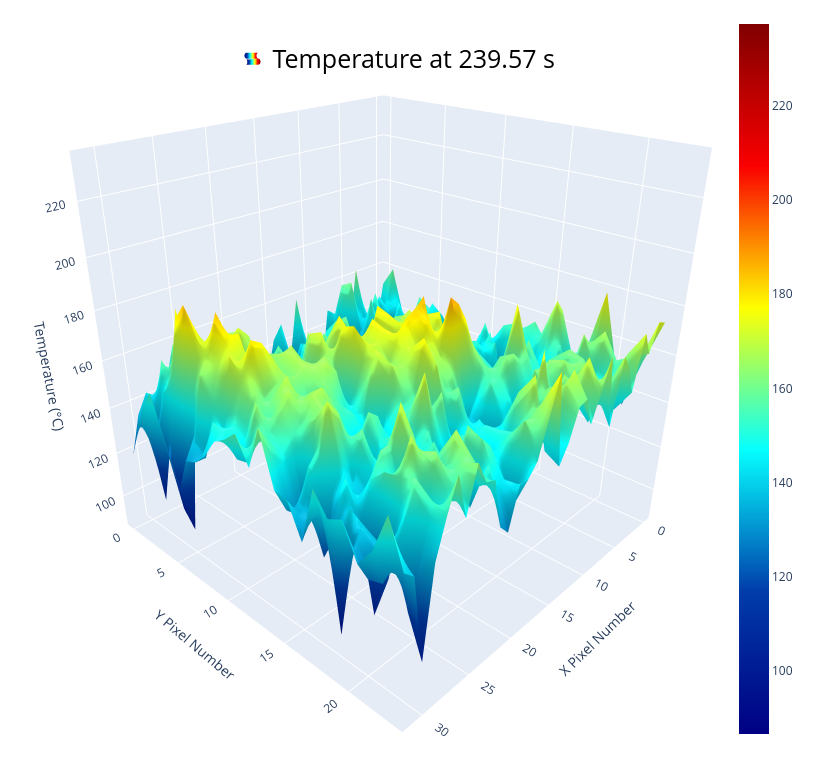}
    \caption{$239.57 \, s$.}
    \label{fig:raw_thermal_3D_t=239.57}
\end{subfigure}

\begin{subfigure}{0.3\textwidth}
    \centering
    \includegraphics[width=\textwidth]{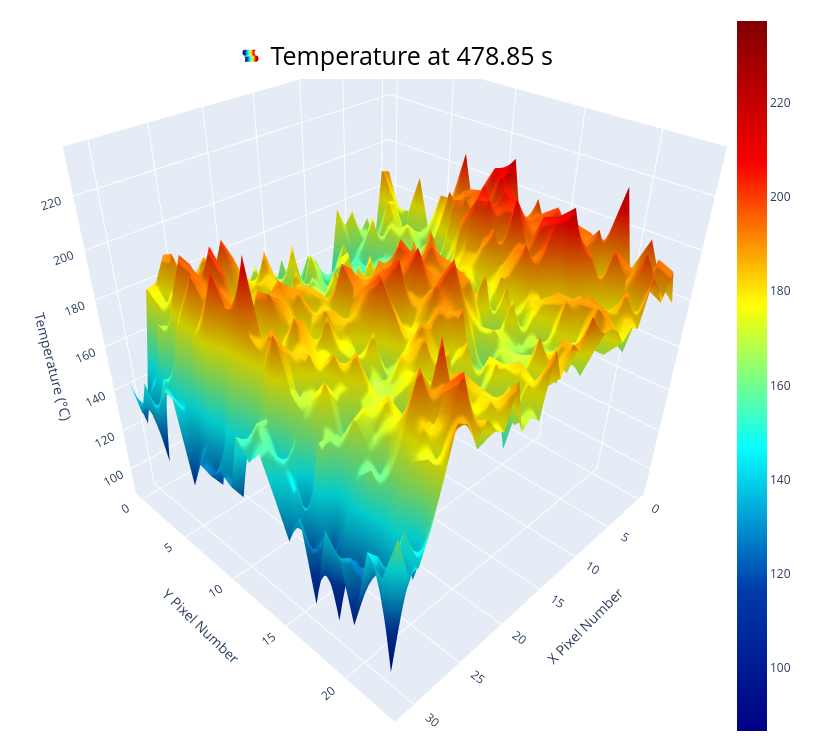}
    \caption{$478.85 \, s$.}
    \label{fig:raw_thermal_3D_t=478.85}
\end{subfigure}
\hspace{5em}
\begin{subfigure}{0.3\textwidth}
    \centering
    \includegraphics[width=\textwidth]{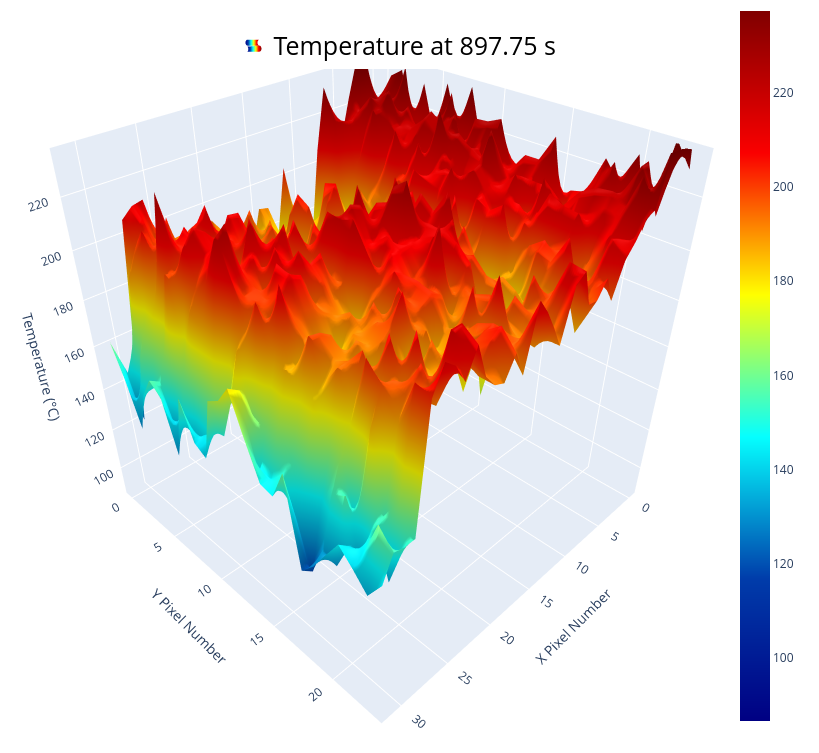}
    \caption{$897.75 \, s$.}
    \label{fig:raw_thermal_3D_t=897.75}
\end{subfigure}
\caption{3D plots of the heating profile at different instances in time. The plots show valid temperature gradients over time.}
\label{fig:raw_thermal_3D_temps}
\end{figure}

\begin{figure}[ht!]
    \centering
    \includegraphics[width=\textwidth]{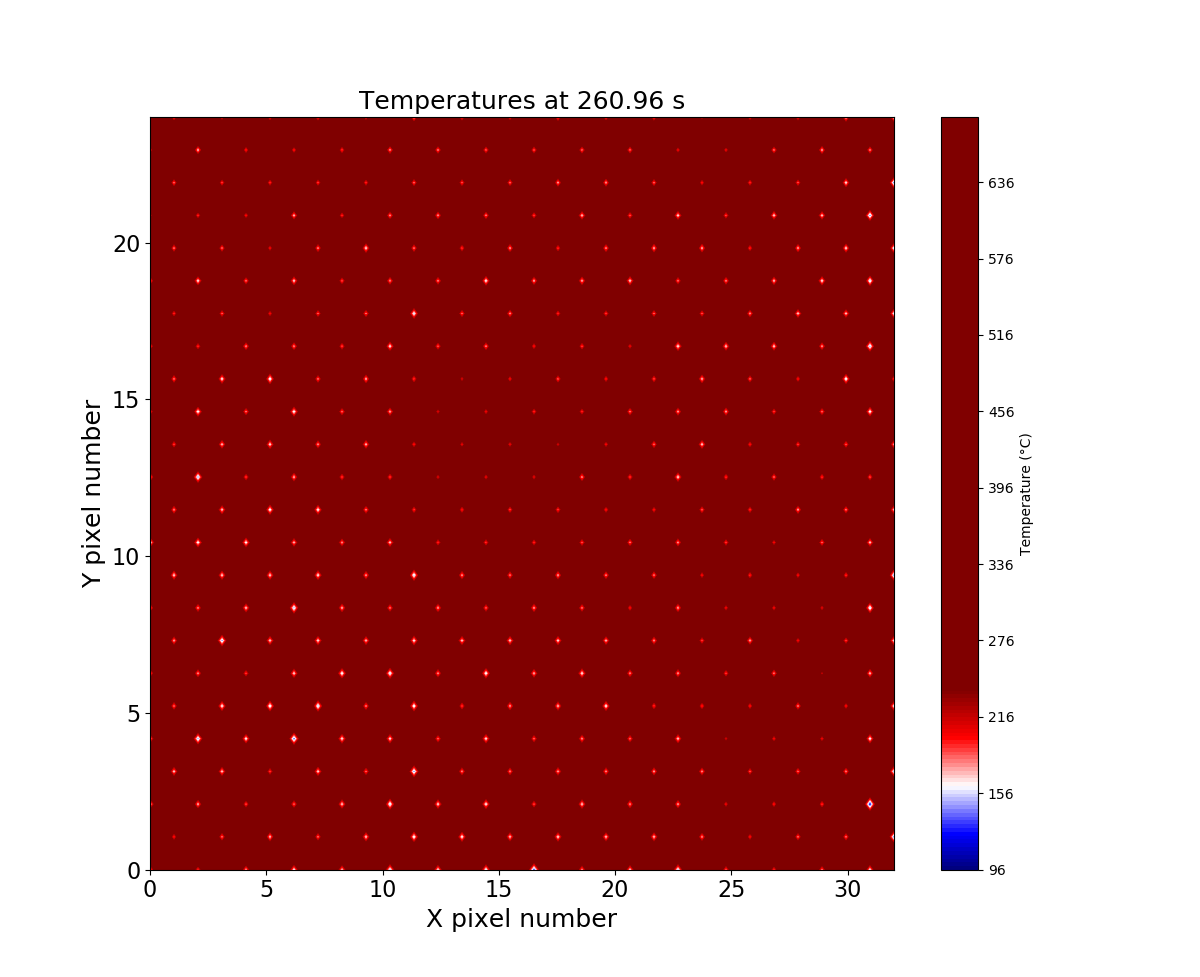}
    \caption{Frame visualisation of one of the temperature spikes shown in Figure~\ref{fig:raw_random_pixels}. These correspond to instances when the camera fails to capture valid frames.}
    \label{fig:spike_frame_contourf}
\end{figure}

Over the duration of the heating, the data exhibits temperature gradients that can be used for training. Figure~\ref{fig:raw_thermal_3D_temps} shows a 3D plot of the heating landscape for raw data at different instances in time, plotted using the Plotly graphing library~\cite{plotly}. Figure~\ref{fig:raw_random_pixels} shows the temperature measurements after the sensor conversion routines for 8 randomly-selected pixels. The temperature measurements are initially very noisy and dominated by high-amplitude spikes. These spikes usually correspond to instances when the sensor data fails to update with new data in time, resulting in the sharp dotted pattern that  Figure~\ref{fig:spike_frame_contourf} shows.

The existence of high amplitude spikes, as well as the high fluctuations in the temperature signal will cause training difficulties in the optimization problem that we formulate. Therefore, it was necessary for us to implement a denoising strategy to eliminate the spikes and smooth out the data over time. Later on, we investigate the training results with and without denoising.

\subsection{Denoising strategy}

\subsubsection{Spike filtering}

\begin{figure}[ht]
    \centering
    \includegraphics[width=\textwidth]{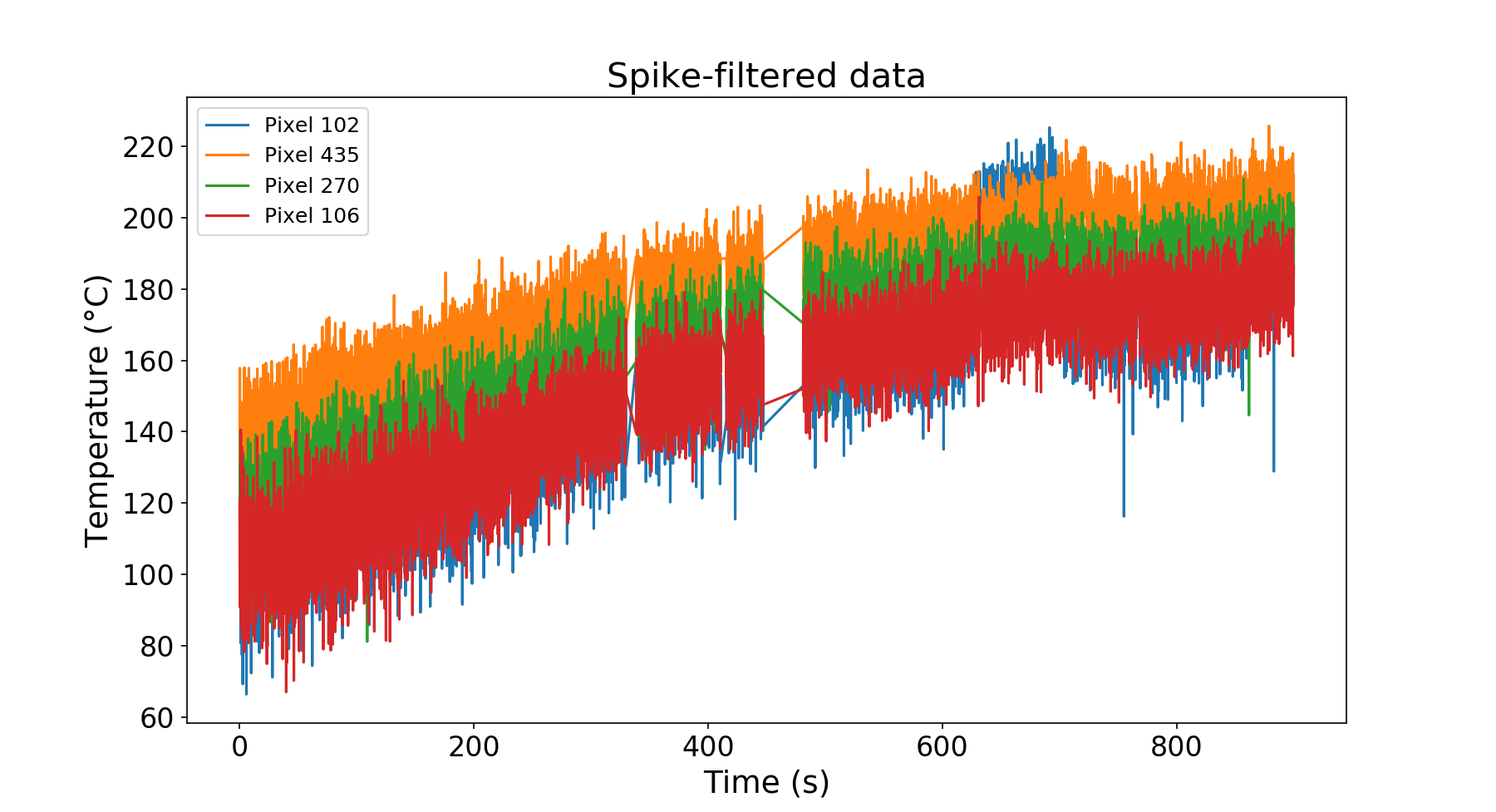}
    \caption{Spike-filtered temperature time-series for four randomly-chosen pixels. The regions near 340 and 470 seconds displayed high concentrations of spikes across the pixels so were filtered out completely.}
    \label{fig:spike_filtered_data}
\end{figure}

The first step was to filter out the spikes in the the data. The main issue that we faced here was that different pixels exhibit spikes at different instances in time. Therefore, we adopt a strategy where we remove the entire frame from the data if any given pixel exhibits spiky behaviour. A better alternative to this would be to replace spike pixels with the average of their neighbouring pixels, however this introduces an additional level of complexity given that neighbouring pixels in both space and time may also be spiky. For our purposes, our filtering method has acceptable performance. Algorithm~\ref{alg:spike_filtering} outlines the spike filtering algorithm. We iterate over all of the pixels and their respective measurements over time, and compare their values against their neighbours, adding the spike value indices to the \verb|spiky_indices| array. We delete the spiky indices from the data to obtain the spike-filtered data. We use a threshold value of 100, and based on this we find that the spikes comprise $8.53$\% of the data. Figure~\ref{fig:spike_filtered_data} shows a plot of the spike-filtered data for 4 randomly chosen pixels. We notice that two regions of the data have been filtered out completely as they correspond to durations with high spike concentratons across many pixels. Additionally, we notice that a few downwards spikes are still present. These will be smoothed out with our next denoising step.

\begin{algorithm}
    \caption{Spike filtering method}
    \label{alg:spike_filtering}
    \textbf{\textit{get\_spike\_indices}}(\verb|data|, \verb|threshold|):
    
    \verb|spike_indices| $\gets [\,]$ 

    \ForEach{$\mathtt{pixel\_measurements} \in \mathtt{data}$}{
        $\mathtt{current\_valid\_measurement} = \mathtt{pixel\_measurements}[0]$\;
        \ForEach{$\mathtt{measurement} \in \mathtt{pixel\_measurements}$}{
            \If{$|\mathtt{current\_valid\_measurement} - \mathtt{measurement}| > \mathtt{threshold}$}{
                $\mathtt{spike\_indices.append(measurement)}$
            }
            \Else{
                $\mathtt{current\_valid\_measurement} = \mathtt{measurement}$
            }
        }
    }
    \Return $\mathtt{sort(unique\_values(spike\_indices))}$
\end{algorithm}

\subsubsection{Data smoothing}

\begin{figure}[ht]
    \centering
    \includegraphics[width=\textwidth]{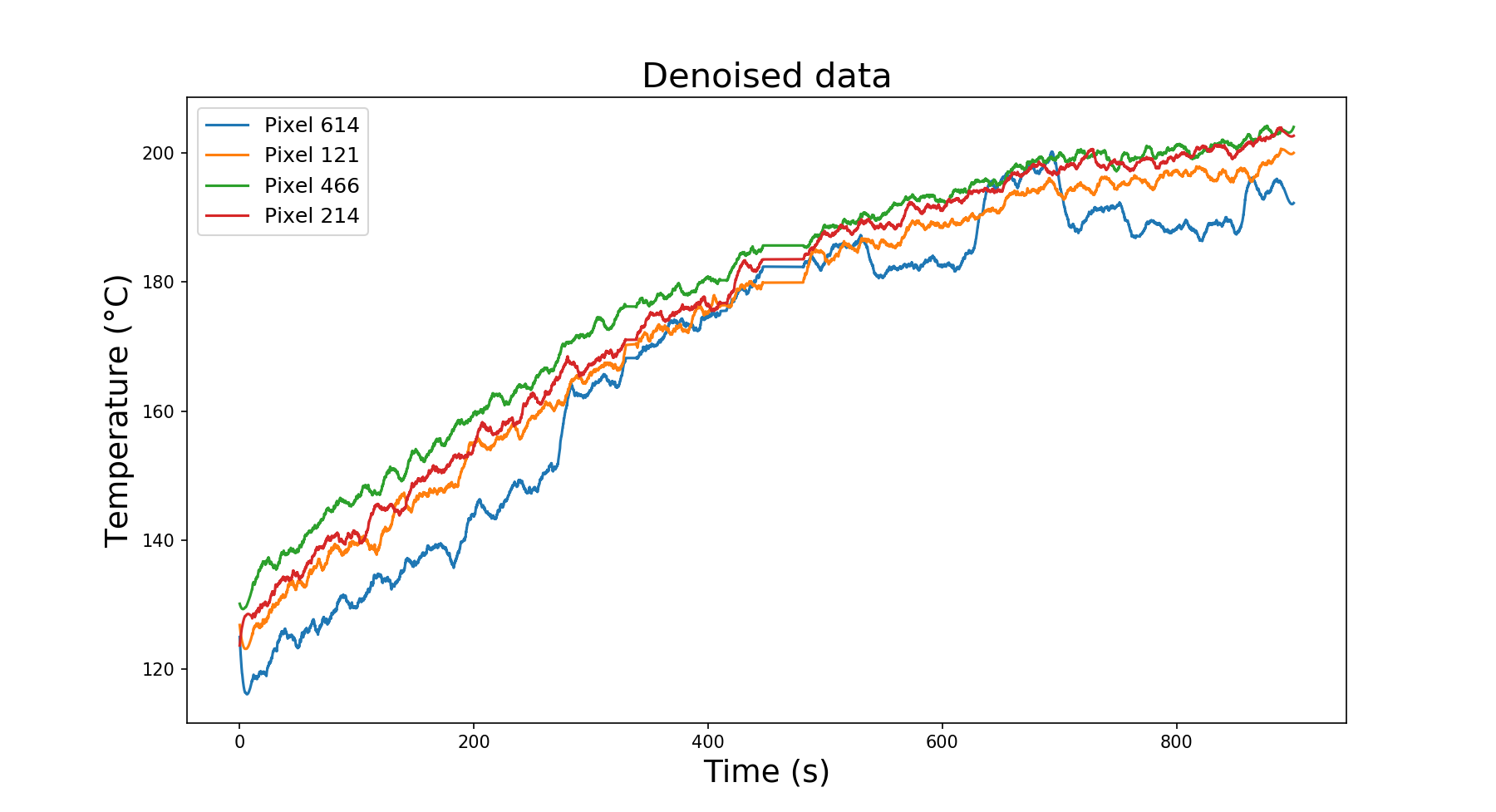}
    \caption{Data for 4 random pixels after applying the Savitzky-Golay filter~\cite{savgol1964}. The data retains small amounts of noise, although most of it has been smoothed out.}
    \label{fig:denoised_data}
\end{figure}

After filtering out spikes, the data still exhibits fluctuations in temperatures that cannot be physically possible. Therefore we perform an additional step to smooth out the data fluctuations and capture the true behaviour of the temperature signals. For this we apply a Savitzky-Golay filter~\cite{savgol1964} on the time-series for each pixel, using the SciPy library~\cite{scipy2020}. The Savitzky-Golay filter is a data-smoothing algorithm based on fitting low-order polynomials using the linear least squares method.  It is a simple and effective method that is suitable for our purposes since the temperatures measurements maintain a relatively linear upwards trend. We use a polynomial order of 3, and a window size of 400. Figure~\ref{fig:denoised_data} shows the measurements after smoothing with the filter.

\subsection{Training setup}

In this section we keep the following parameters constant unless otherwise specified:

\begin{itemize}
    \item Architecture: 2-layer MLP with 64 and 32 units.
    \item Training set: Denoised data.
    \item Optimizer: LBFGS.
    \item Activation function: Hyperbolic tangent (Tanh).
    \item Learning rate: 0.01.
    \item $\lambda_p$: 0.5.
    \item $N_p$: 32000 points corresponding to 500 frames.
\end{itemize}

\subsubsection{PINN loss function}

The loss function we use is based on a substitution of the 2D version of Equation~\ref{eq:heat_equation} in Equation~\ref{eq:PINN_loss}. Additionally we solve an inverse problem to discover the $\alpha$ parameter, although we assume that the value might differ slightly for the $x$ and $y$ dimensions so we use the coefficients $\alpha$ and $\beta$ for $x$ and $y$ respectively. Thus, the result is Equation~\ref{eq:heat_eq_PINN}.

\begin{equation}
    \label{eq:heat_eq_PINN}
    \pazocal{L} = \frac{1}{N_d}\sum_{i=1}^{N_d}|f_{u} - u_d^i|^2 + \frac{\lambda_p}{N_p}\sum_{i=1}^{N_p}|u_t^i - \alpha \, u_{xx}^i - \beta \, u_{yy}^i|^2
\end{equation}
We use an initial value of $10.0$ for $\alpha$ and $\beta$.

\subsubsection{Initial training difficulties}

Our initial attempt at training involved using LBFGS with full batch training, similar to the approach we took in Chapter~\ref{chapter:pendulum}, but with 3 dimensions for the $x$, $y$, and $t$ coordinates. The current LBFGS Pytorch implementation only supports full-batch training. The full dataset collected from our experiment consists of $18,798$ frames where each frame consists of 768 2-byte pixels, and the denoised data similarly consists of $17,195$ frames. Each pixel is associated with a 3-dimensional spatio-temporal coordinate point. This gives us a total of $18,798 \times 768 \times 2 = 28,873,728$ bytes ($\approx 28.9$ MB) for the label set and $18,798 \times 768 \times 3 \times 4 = 173,242,368$ bytes ($\approx 173.2$ MB) for the feature set. Given that we need to train a large model for this problem, in addition to needing to compute large Jacobian and Hessian matrices, we began to face memory issues with full-batch training on our 4 GB GPU.

\begin{figure}[ht]
    \centering
    \includegraphics[width=\textwidth]{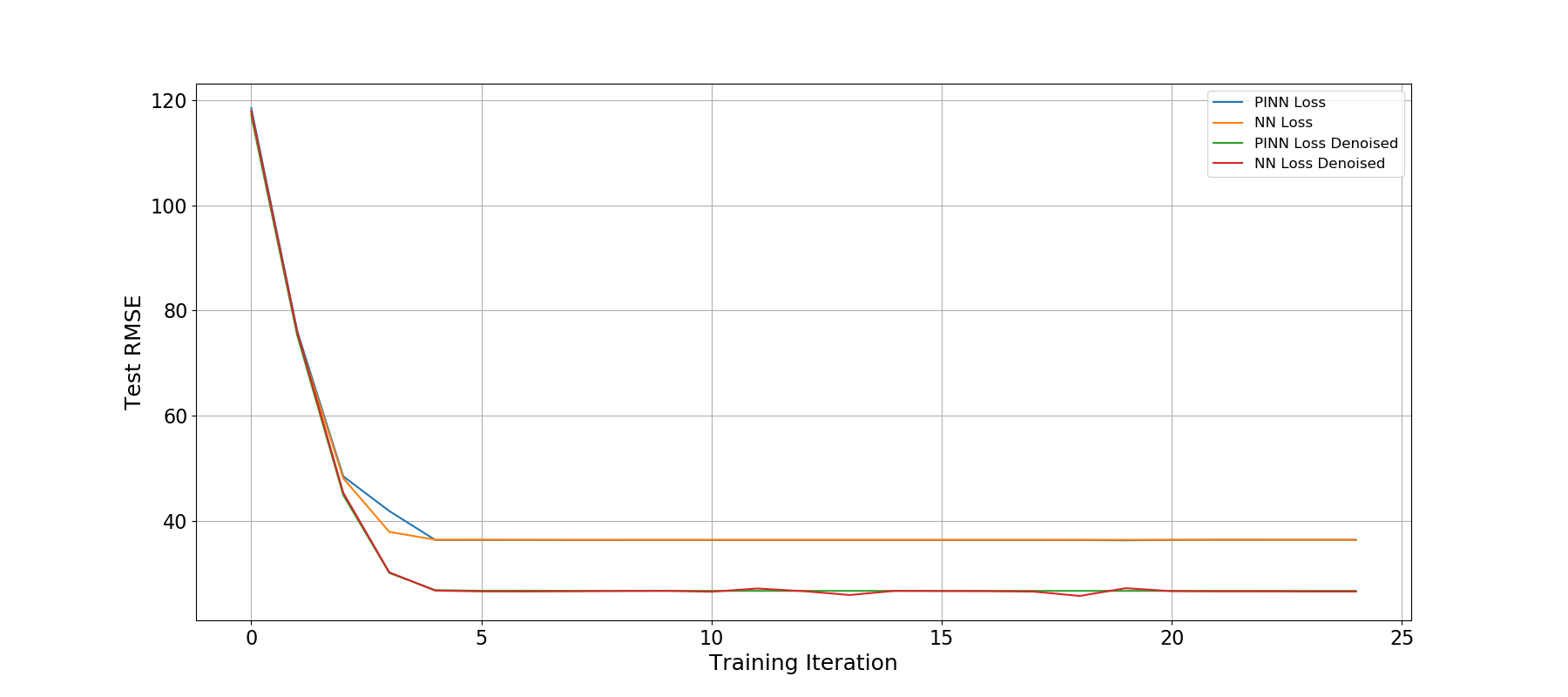}
    \caption{RMSE graph for a 2-layer 32-unit PINN and NN trained with Adam~\cite{Kingma2014}. The training data is shown based on the raw and denoised data, and is sampled from the full 768-pixel frames. Denoising certainly helps achieve better training results, although there is a negligible difference between PINN and NN training performance.}
    \label{fig:heat_diff_loss_graph_Adam_full_frame}
\end{figure}

The bigger issue however, is the difficulty posed on the optimization problem since our problem involves a 3-dimensional cuboid with millions of points. Therefore we attempted to use Adam~\cite{Kingma2014} instead of LBFGS since it supports mini-batch training, and we down-sample our data in time. Unfortunately, it was still the case that the training failed to converge to a satisfactory solution even with many variations on the down-sampled training data, the network architectures, and the training hyperparameters. Figure~\ref{fig:heat_diff_loss_graph_Adam_full_frame} shows example training evaluations with a 2-layer 32-unit fully-connected architecture with $57,775$ points randomly chosen throughout the domain, and with a batch size of 4096. We find that in most cases using a simpler architecture performs better than a complex one. The predictions are better after denoising, although in all cases we could not get an RMSE lower than 26.

Therefore, we reduce the size our problem by reducing the size of the training frame.

\subsubsection{Frame size reduction}

\begin{figure}[ht]
\centering
\begin{subfigure}{0.49\textwidth}
    \centering
    \includegraphics[width=\textwidth]{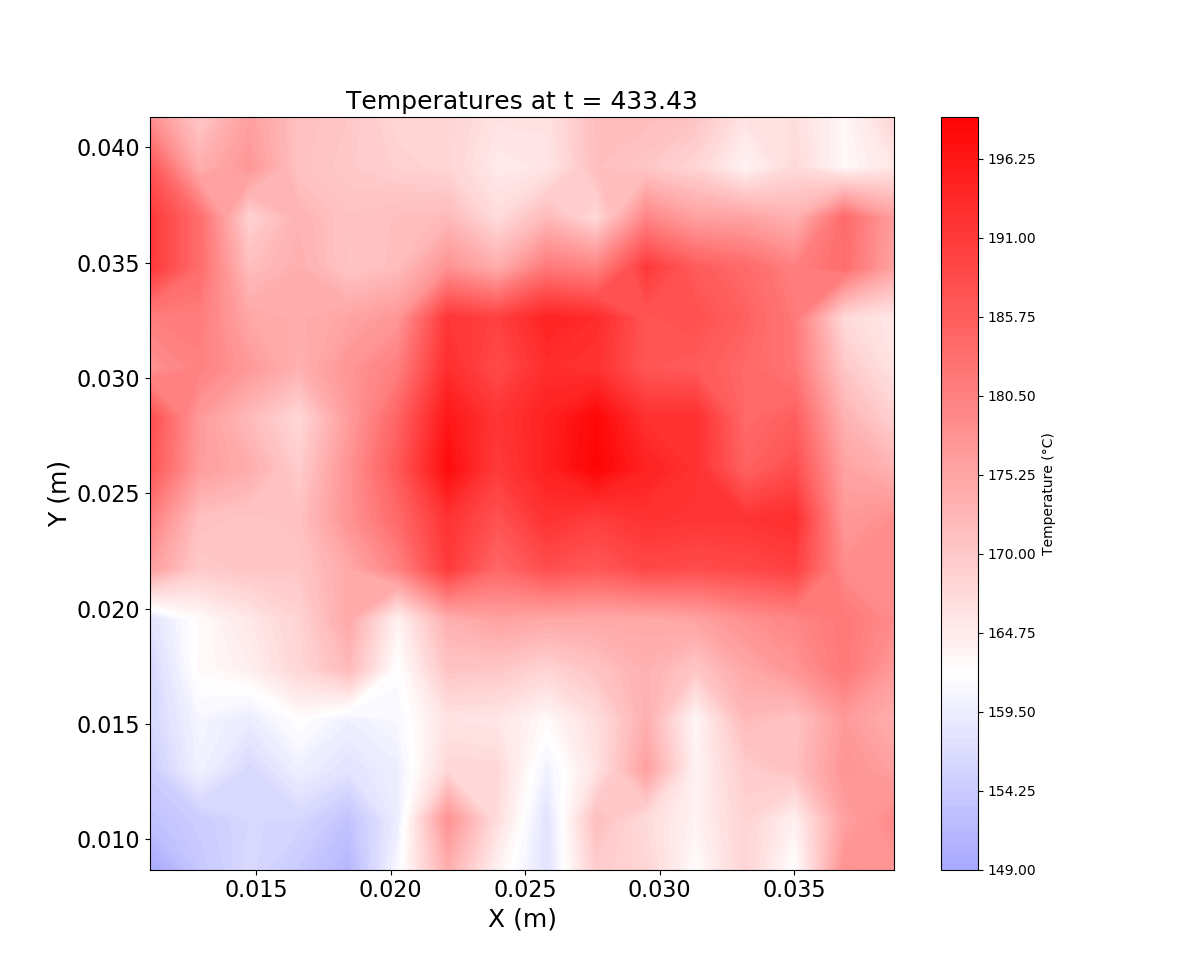}
    \caption{16x16.}
    \label{fig:example_reduced_frame16}
\end{subfigure}
\begin{subfigure}{0.49\textwidth}
    \centering
    \includegraphics[width=\textwidth]{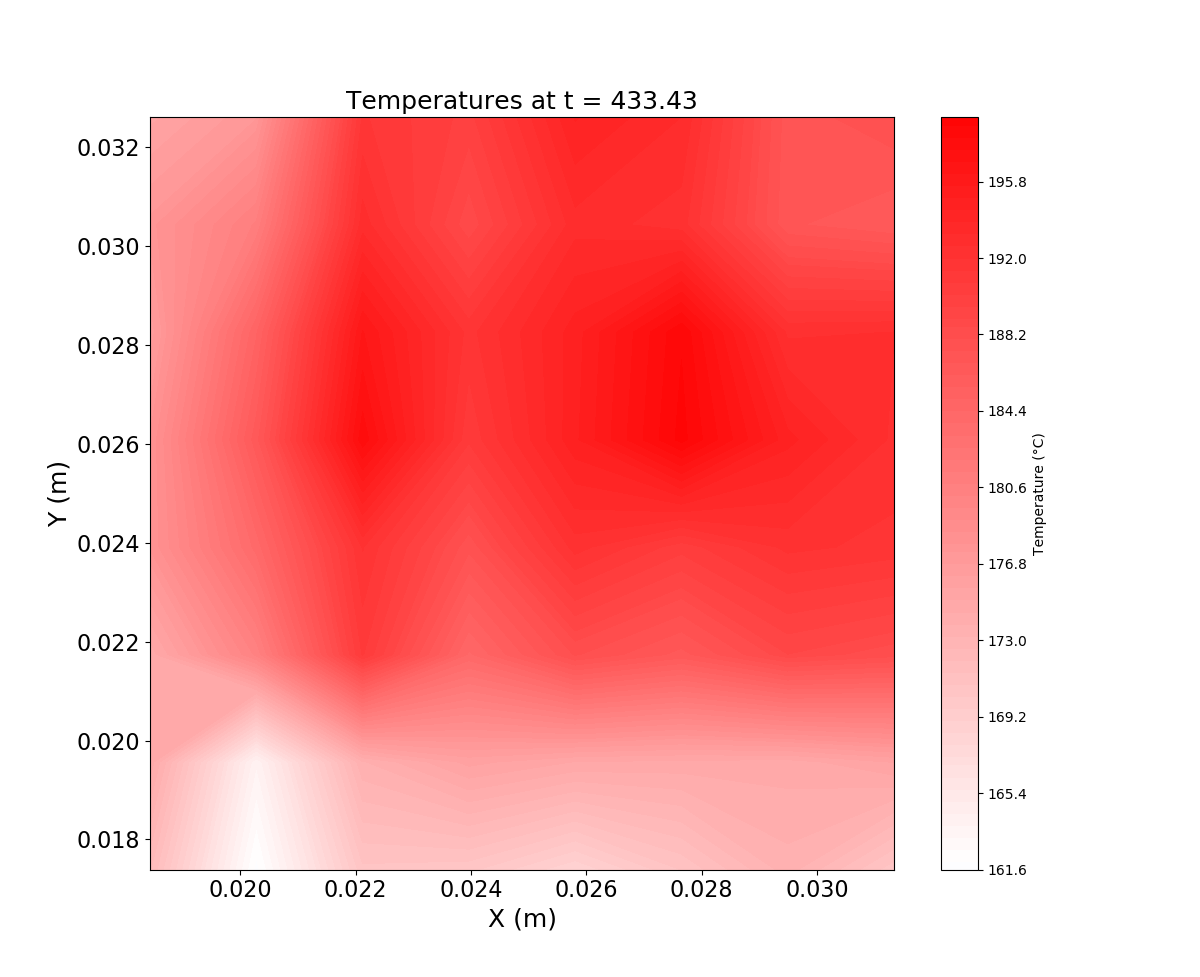}
    \caption{8x8.}
    \label{fig:example_reduced_frame8}
\end{subfigure}

\begin{subfigure}{0.49\textwidth}
    \centering
    \includegraphics[width=\textwidth]{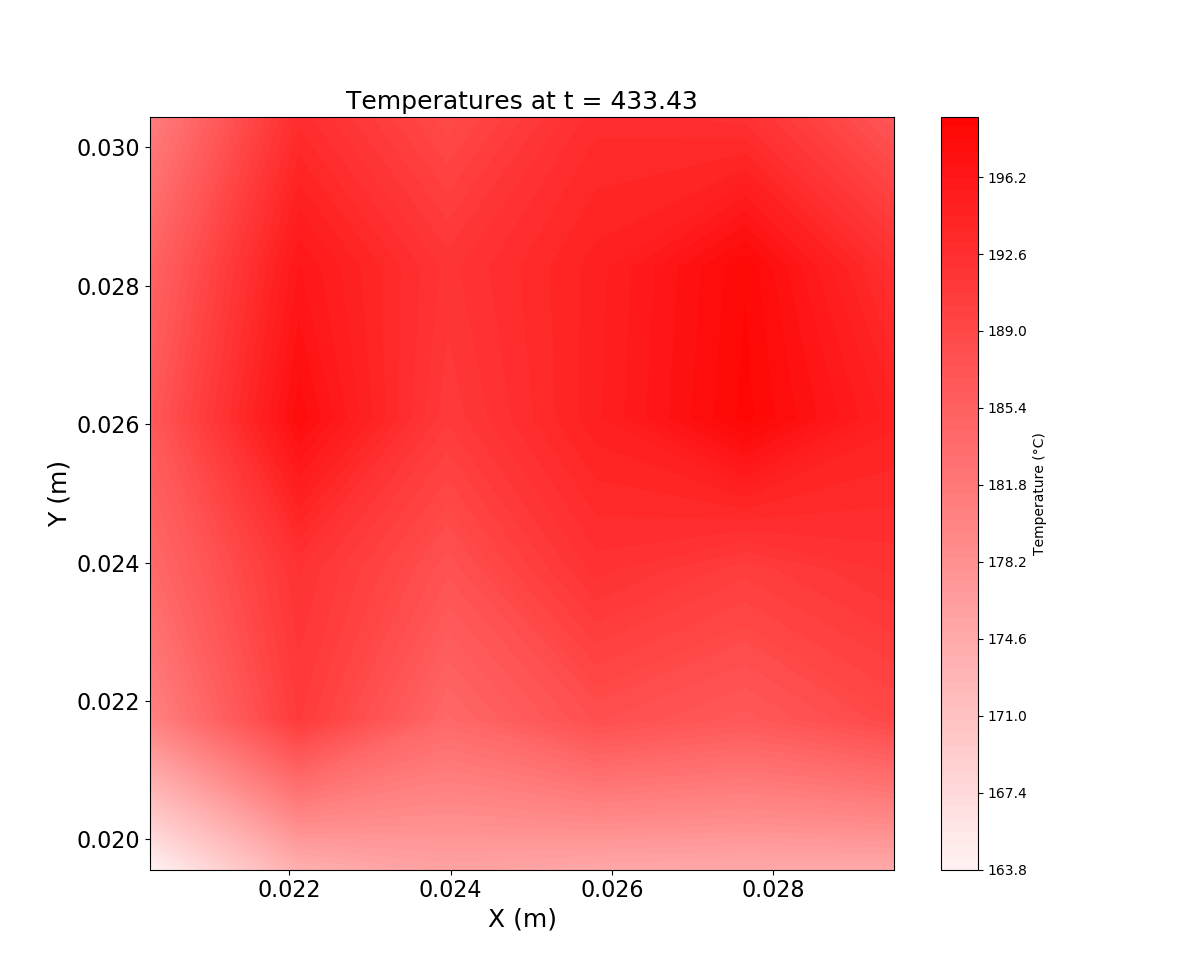}
    \caption{6x6.}
    \label{fig:example_reduced_frame6}
\end{subfigure}
\begin{subfigure}{0.49\textwidth}
    \centering
    \includegraphics[width=\textwidth]{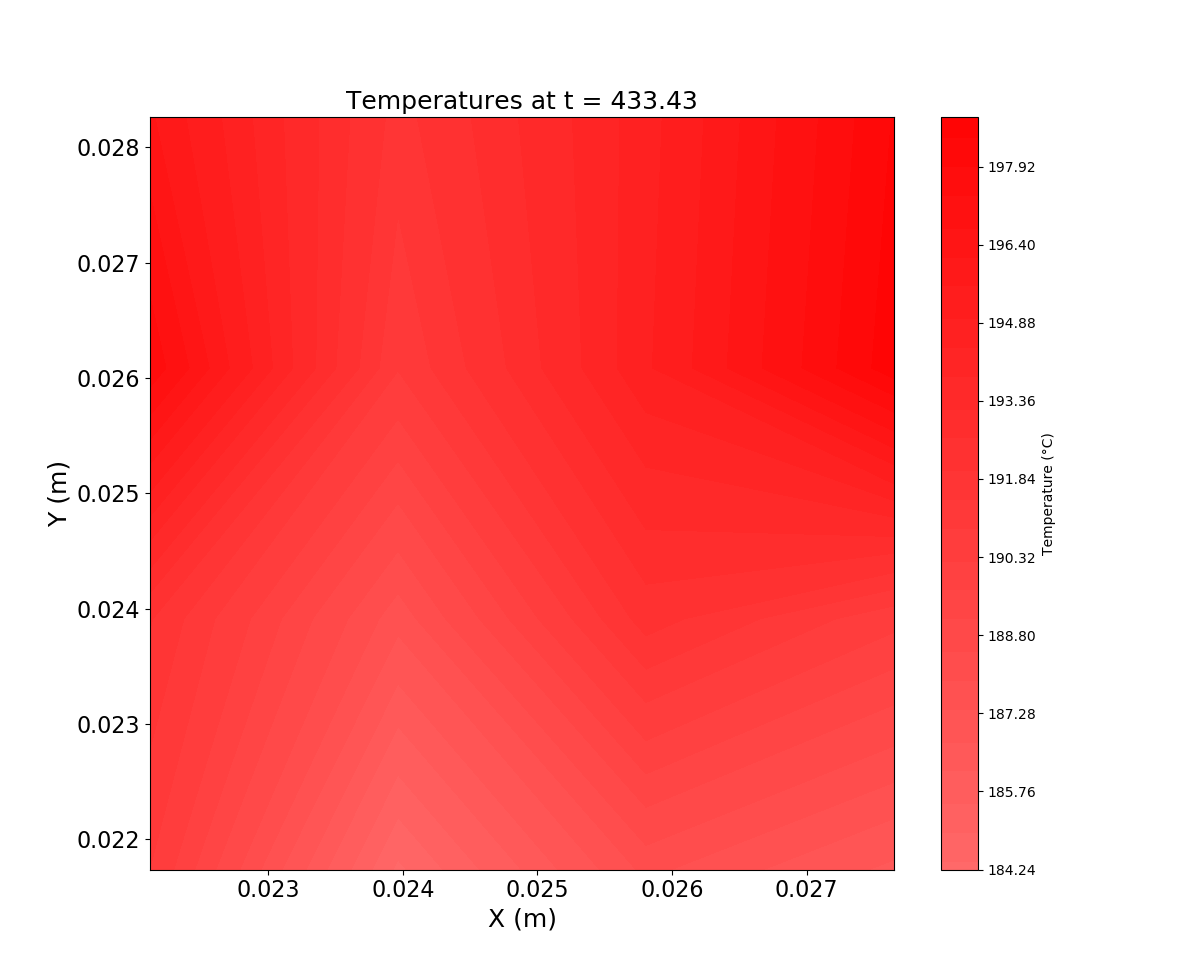}
    \caption{4x4.}
    \label{fig:example_reduced_frame4}
\end{subfigure}
\caption{Reduced size frames at $t = 433.43$. Smaller frames are easier to train with than larger ones.}
\label{fig:example_reduced_frame}
\end{figure}

\begin{figure}[ht]
    \centering
    \includegraphics[width=\textwidth]{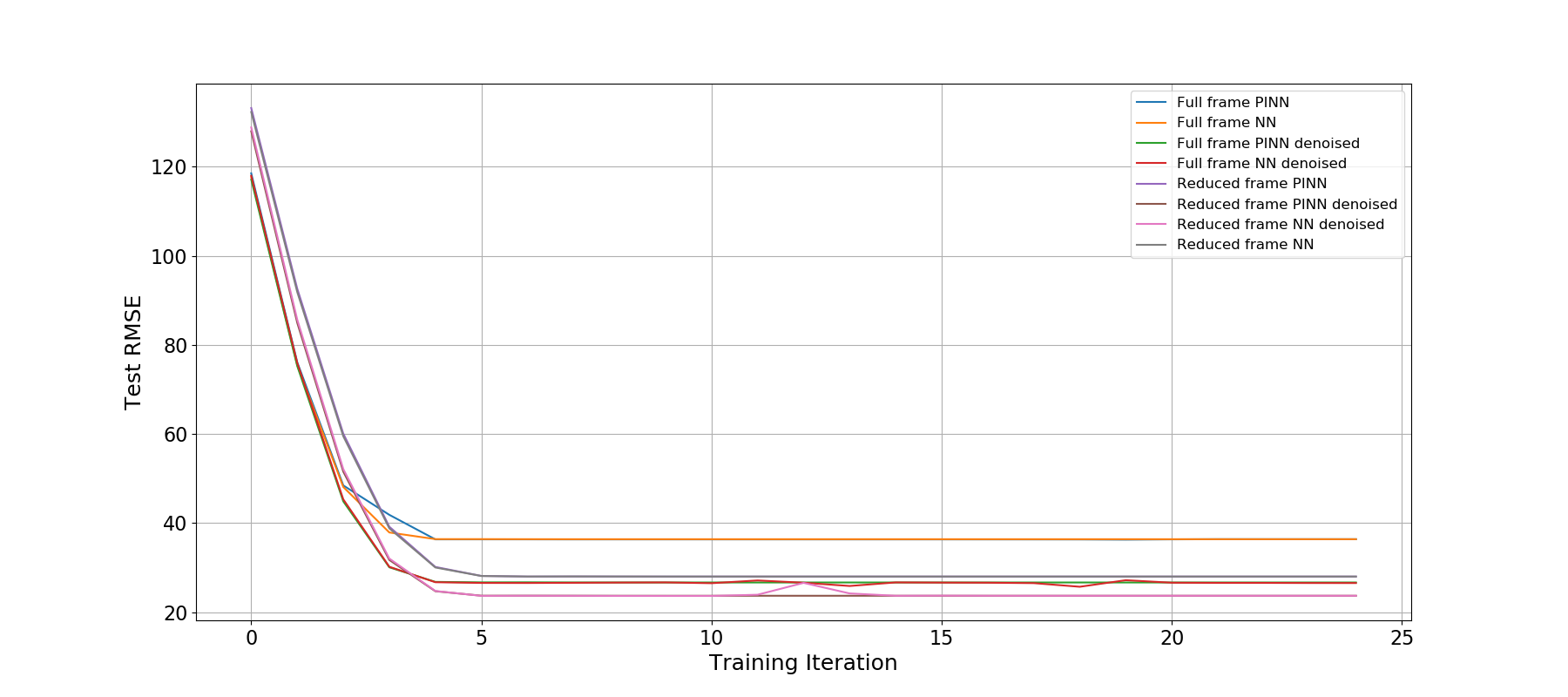}
    \caption{Reduced frame vs full frame training comparison. The minimum RMSE is $23.67$. Both denoising and reducing the frame size improve the training performance, although the training accuracy remains to be satisfactory.}
    \label{fig:full_vs_reduced_frame_training_comparison}
\end{figure}

\begin{figure}[ht]
    \centering
    \includegraphics[width=\textwidth]{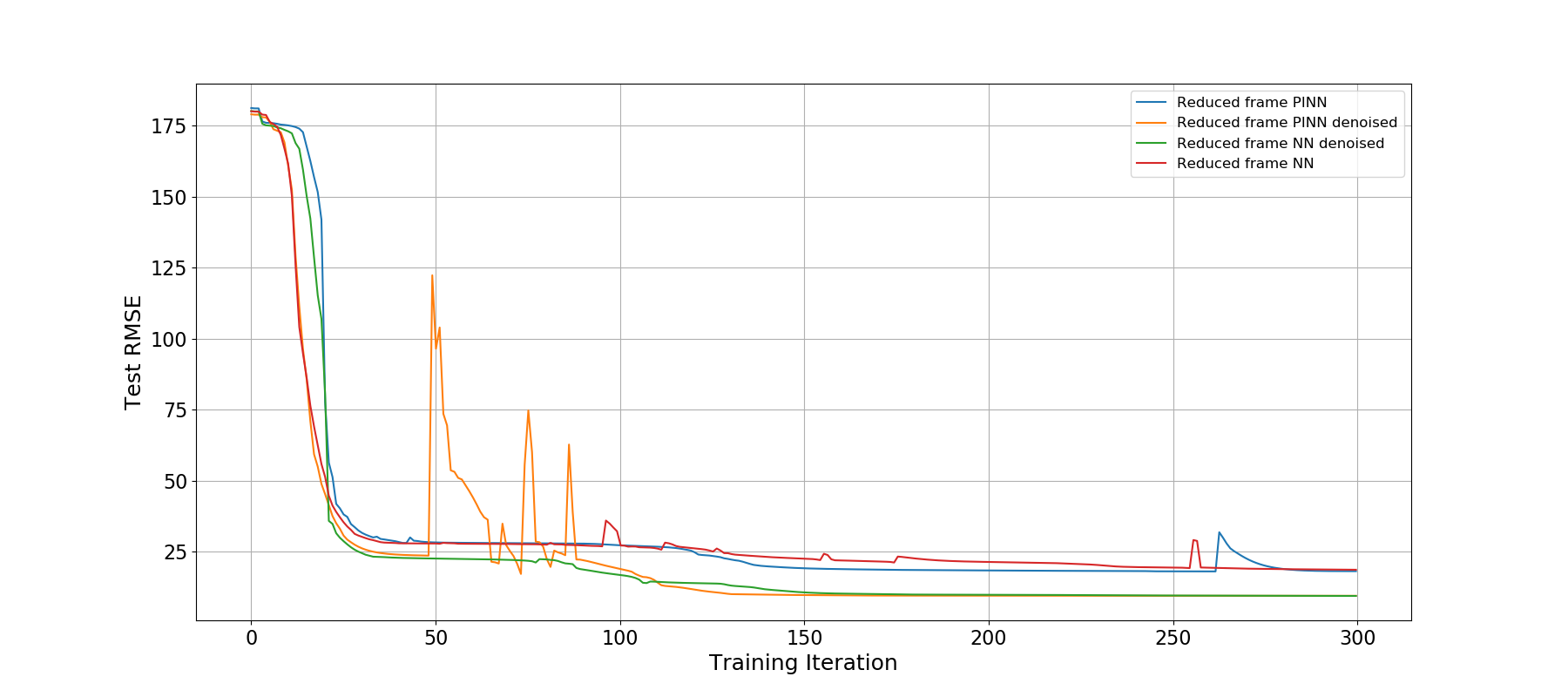}
    \caption{LBFGS training evaluation for a 3-layer network with 64-32-32 units. The minimum RMSE is $9.42$. Using LBFGS improves training performance considerably.}
    \label{fig:reduced_frame_lbfgs_loss}
\end{figure}

We reduce our problem size so that we focus on a central square of the frame. Figure~\ref{fig:example_reduced_frame} shows contour plots for different reduced-size frames after 433.43 seconds of heating. After frame reduction the memory taken up by our training data has reduced, and we are now also able to use LBFGS with full-batch training. Figure~\ref{fig:full_vs_reduced_frame_training_comparison} shows the \textbf{8x8} reduced frame training cases compared against the full-frame ones using Adam, and Figure~\ref{fig:reduced_frame_lbfgs_loss} shows a similar evaluation but only for the \textbf{8x8} reduced frame using LBFGS. For LBFGS we used a 3-layer network with 64 units in the first layer as the 2-layer architecture caused training instabilities. Additionally we trained the LBFGS case for 500 iterations as the network was still able to learn more, as opposed to the Adam case where the RMSE plateaued after 20 iterations. The best case RMSE for Figure~\ref{fig:full_vs_reduced_frame_training_comparison} was $23.67$ while the best case RMSE for Figure~\ref{fig:reduced_frame_lbfgs_loss} was $9.42$. We can see that LBFGS performs much better than Adam so we fix it for the upcoming evaluations. Additionally we will only be training with the denoised data.

\subsection{Results}

\subsubsection{Frame size variation}

Here we investigate the training performance based varying frame size. We train with 819 linearly-spaced frames and test with 441 linearly-spaced frames. We train until convergence and report on the number of iterations needed. Table~\ref{tab:frame_size_variation} shows the training results for different frame dimensions. We notice a predictable pattern in the PINN RMSE column in that the values decrease with a decreasing frame size. The NN RMSEs do not adhere to this pattern as strictly, as we can see with the RMSE for the \textbf{6x6} pixel frame RMSE which is greater than that of the \textbf{10x10} frame.

\begin{table}[ht]
\centering
\begin{tabular}{|c|c|c|c|c|}
\hline
\textbf{Frame Dimensions} & \textbf{NN}                                           & \textbf{PINN}                                           & $\mathbf{\alpha}$ & $\mathbf{\beta}$ \\ \hline
\textbf{16x16}      & \begin{tabular}[c]{@{}c@{}}9.894\\ 5000\end{tabular} & \begin{tabular}[c]{@{}c@{}}24.894\\ 1000\end{tabular}  & 9.9626         & 9.9845        \\ \hline
\textbf{10x10}      & \begin{tabular}[c]{@{}c@{}}8.524\\ 5000\end{tabular} & \begin{tabular}[c]{@{}c@{}}10.433\\ 4999\end{tabular} & 9.9445         & 9.9942        \\ \hline
\textbf{8x8}        & \begin{tabular}[c]{@{}c@{}}9.822\\ 976\end{tabular}   & \begin{tabular}[c]{@{}c@{}}9.420\\ 4971\end{tabular}    & 10.0266        & 9.9754        \\ \hline
\textbf{6x6}        & \begin{tabular}[c]{@{}c@{}}10.804\\ 1998\end{tabular} & \begin{tabular}[c]{@{}c@{}}8.084\\ 4352\end{tabular}    & 9.9166         & 9.9319        \\ \hline
\textbf{4x4}        & \begin{tabular}[c]{@{}c@{}}4.293\\ 4938\end{tabular} & \begin{tabular}[c]{@{}c@{}}4.291\\ 5000\end{tabular}   & 9.9486         & 9.9153        \\ \hline
\end{tabular}
\caption{RMSE values based on varying the frame size. The first line in the NN and PINN entries corresponds to the RMSE, and the second line corresponds to the iteration number.}
\label{tab:frame_size_variation}
\end{table}

\subsubsection{Linearly-spaced frames}

We fix the frame size to be \textbf{8x8} pixels and investigate predictive performance for a variation of the number of linearly-spaced frames. Table~\ref{tab:heat_diffusion_linspace_data} shows the results, and the second and third rows of Figure~\ref{fig:heat_diff_pred_comparison_visualised} show visualisations of the predictions. Contrary to our expectation, the PINN results do not appear to be better than the NN results as the amount of data decreases, except for the 36-frame and 6-frame cases. Additionally, we generally see from Figure~\ref{fig:heat_diff_pred_comparison_visualised} that the PINNs failed to capture spatial temperature gradients compared to NNs, as they made near constant temperature predictions for their entire frames. This may be because the $\alpha$ and $\beta$ parameters are not optimized, or because of the fact that the data does not strongly adhere to the physics. This is most likely due to the large amounts of noise in the system which influences what the physics looks like for the observed data. It may be possible to get results for the PINN that are better than the NN by tweaking the hyperparameters since the optimization process is stochastic in nature.

\begin{table}[ht]
\centering
\centering
\begin{tabular}{|c|c|c|c|c|}
\hline
\textbf{\begin{tabular}[c]{@{}c@{}}$\mathbf{N_{fr}}$\\ $\mathbf{N_d}$\end{tabular}} & \textbf{NN}                                           & \textbf{PINN}                                         & $\mathbf{\alpha}$ & $\mathbf{\beta}$ \\ \hline
\textbf{\begin{tabular}[c]{@{}c@{}}3439\\ 220096\end{tabular}}         & \begin{tabular}[c]{@{}c@{}}8.344\\ 5000\end{tabular}  & \begin{tabular}[c]{@{}c@{}}9.537\\ 5000\end{tabular}  & 10.0382        & 9.8948        \\ \hline
\textbf{\begin{tabular}[c]{@{}c@{}}1147\\ 73408\end{tabular}}          & \begin{tabular}[c]{@{}c@{}}8.701\\ 5000\end{tabular}  & \begin{tabular}[c]{@{}c@{}}17.526\\ 1901\end{tabular} & 9.9853         & 9.9753        \\ \hline
\textbf{\begin{tabular}[c]{@{}c@{}}574\\ 36736\end{tabular}}           & \begin{tabular}[c]{@{}c@{}}9.805\\ 5000\end{tabular}  & \begin{tabular}[c]{@{}c@{}}9.426\\ 4987\end{tabular}  & 10.0397        & 10.0631       \\ \hline
\textbf{\begin{tabular}[c]{@{}c@{}}287\\ 18368\end{tabular}}           & \begin{tabular}[c]{@{}c@{}}9.272\\ 4999\end{tabular}  & \begin{tabular}[c]{@{}c@{}}9.462\\ 4994\end{tabular}  & 9.9517         & 9.9112        \\ \hline
\textbf{\begin{tabular}[c]{@{}c@{}}144\\ 9216\end{tabular}}            & \begin{tabular}[c]{@{}c@{}}8.952\\ 5000\end{tabular}  & \begin{tabular}[c]{@{}c@{}}9.421\\ 5000\end{tabular}  & 9.9574         & 9.8677        \\ \hline
\textbf{\begin{tabular}[c]{@{}c@{}}72\\ 4608\end{tabular}}             & \begin{tabular}[c]{@{}c@{}}8.828\\ 2898\end{tabular}  & \begin{tabular}[c]{@{}c@{}}9.443\\ 5000\end{tabular}  & 10.0699        & 9.7978        \\ \hline
\textbf{\begin{tabular}[c]{@{}c@{}}36\\ 2304\end{tabular}}             & \begin{tabular}[c]{@{}c@{}}11.381\\ 4388\end{tabular} & \begin{tabular}[c]{@{}c@{}}9.436\\ 5000\end{tabular} & 9.7256         & 9.8428        \\ \hline
\textbf{\begin{tabular}[c]{@{}c@{}}28\\ 1792\end{tabular}}             & \begin{tabular}[c]{@{}c@{}}9.579\\ 1943\end{tabular}  & \begin{tabular}[c]{@{}c@{}}23.606\\ 461\end{tabular}  & 10.0837        & 9.9646        \\ \hline
\textbf{\begin{tabular}[c]{@{}c@{}}18\\ 1152\end{tabular}}             & \begin{tabular}[c]{@{}c@{}}9.508\\ 1859\end{tabular}  & \begin{tabular}[c]{@{}c@{}}15.905\\ 819\end{tabular}  & 9.9908         & 9.8933        \\ \hline
\textbf{\begin{tabular}[c]{@{}c@{}}13\\ 832\end{tabular}}              & \begin{tabular}[c]{@{}c@{}}7.915\\ 5000\end{tabular}  & \begin{tabular}[c]{@{}c@{}}9.578\\ 5000\end{tabular}  & 9.7833         & 9.8016        \\ \hline
\textbf{\begin{tabular}[c]{@{}c@{}}11\\ 704\end{tabular}}              & \begin{tabular}[c]{@{}c@{}}9.683\\ 1633\end{tabular}  & \begin{tabular}[c]{@{}c@{}}14.848\\ 4820\end{tabular} & 9.8494         & 9.9277        \\ \hline
\textbf{\begin{tabular}[c]{@{}c@{}}6\\ 384\end{tabular}}               & \begin{tabular}[c]{@{}c@{}}13.394\\ 823\end{tabular}  & \begin{tabular}[c]{@{}c@{}}11.281\\ 3547\end{tabular} & 9.9824         & 9.7317        \\ \hline
\end{tabular}
\caption{RMSE values based on a variation of the number of linearly-spaced frames. $\mathbf{N_{fr}}$ is the number of frames and $\mathbf{N_d}$ is the corresponding number of points.}
\label{tab:heat_diffusion_linspace_data}
\end{table}

\subsubsection{Uniformly-distributed points}

\begin{figure}
    \centering
    \begin{subfigure}{0.25\textwidth}
        \includegraphics[width=\textwidth]{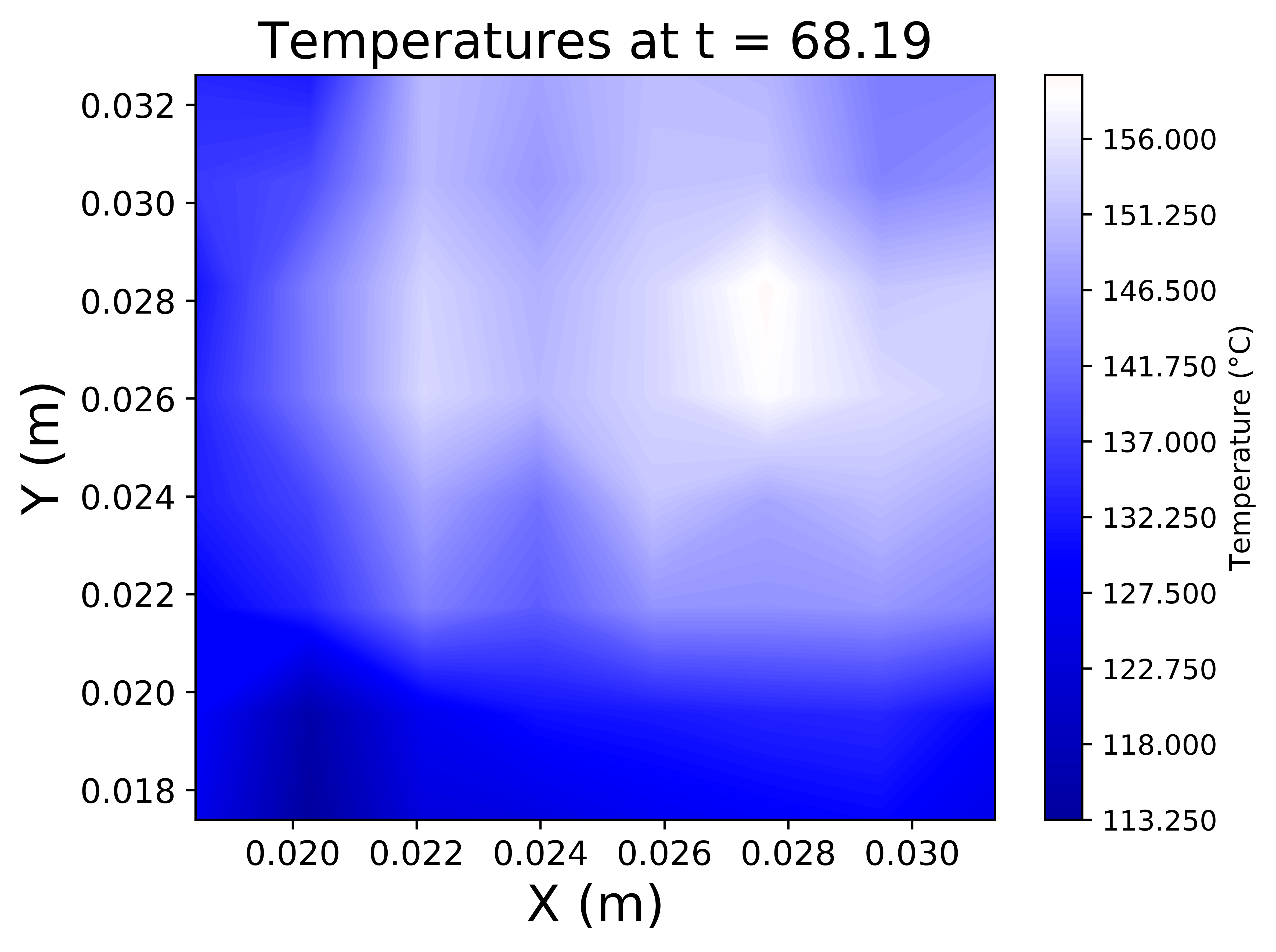}
        \caption{Test data, i = 35.}
        \label{fig:test_data_idx35}
    \end{subfigure}
    \begin{subfigure}{0.25\textwidth}
        \includegraphics[width=\textwidth]{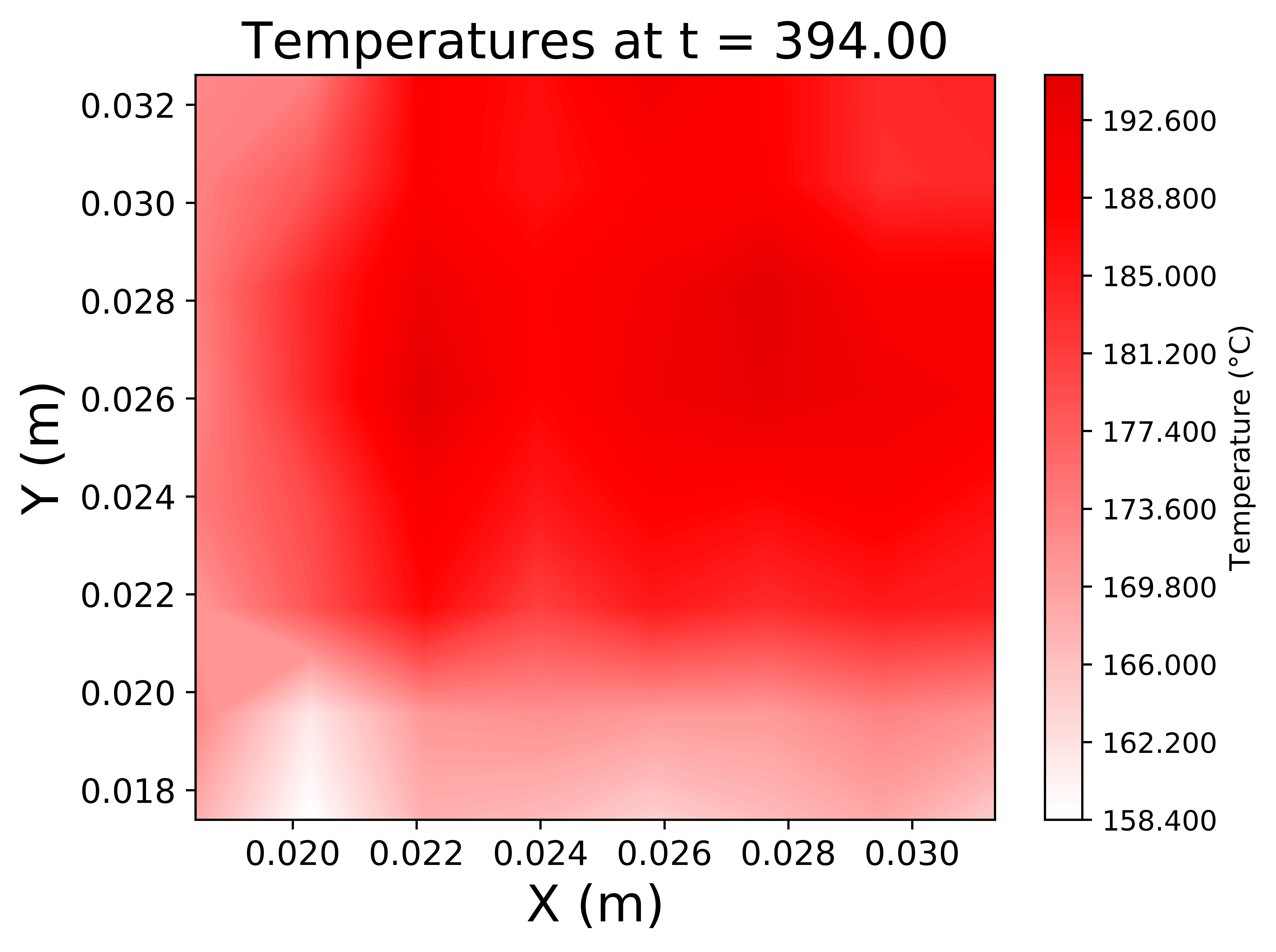}
        \caption{Test data, i = 200.}
        \label{fig:test_data_idx200}
    \end{subfigure}
    \begin{subfigure}{0.25\textwidth}
        \includegraphics[width=\textwidth]{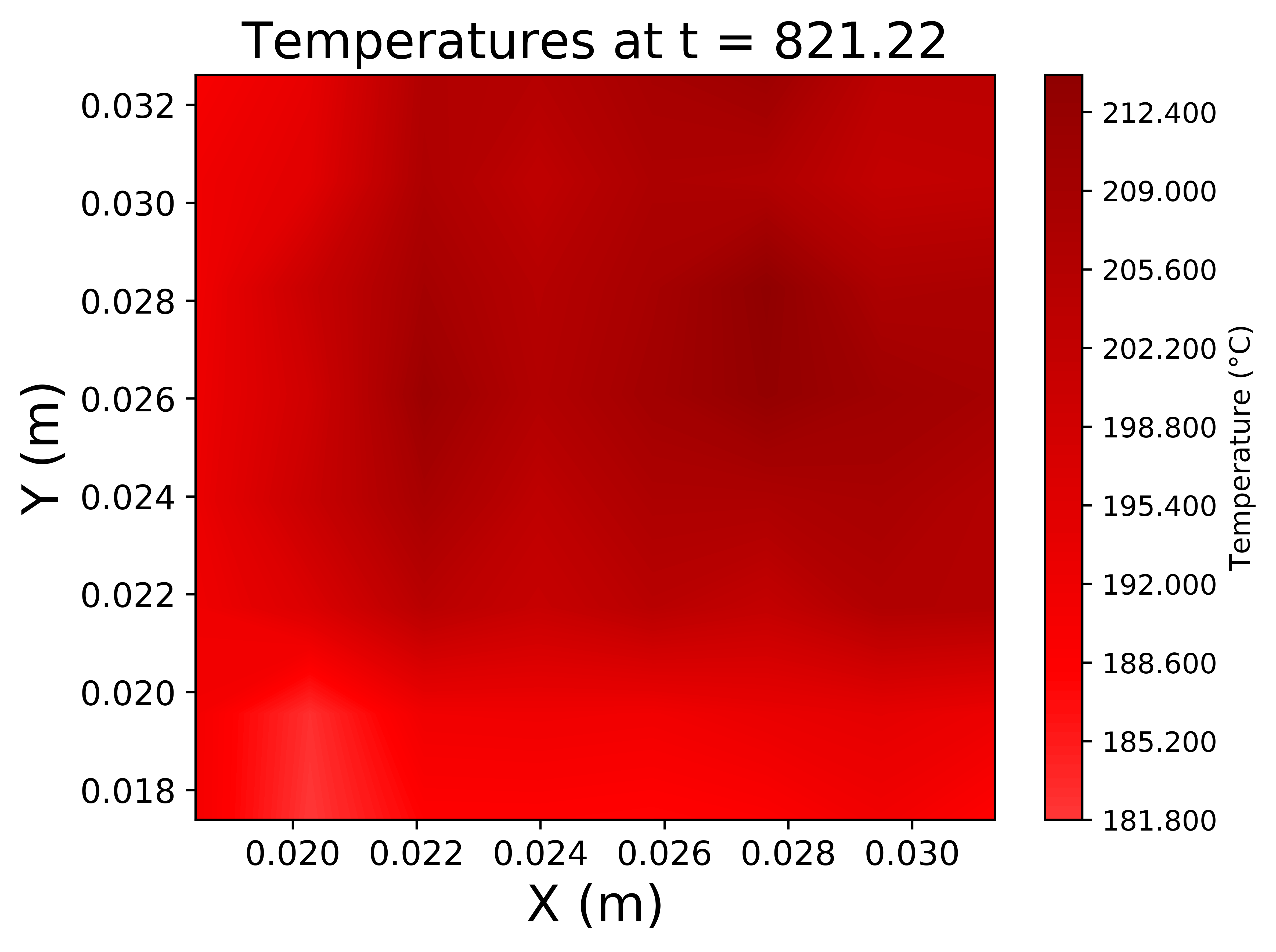}
        \caption{Test data, i = 400.}
        \label{fig:test_data_idx400}
    \end{subfigure}

    \begin{subfigure}{0.25\textwidth}
        \includegraphics[width=\textwidth]{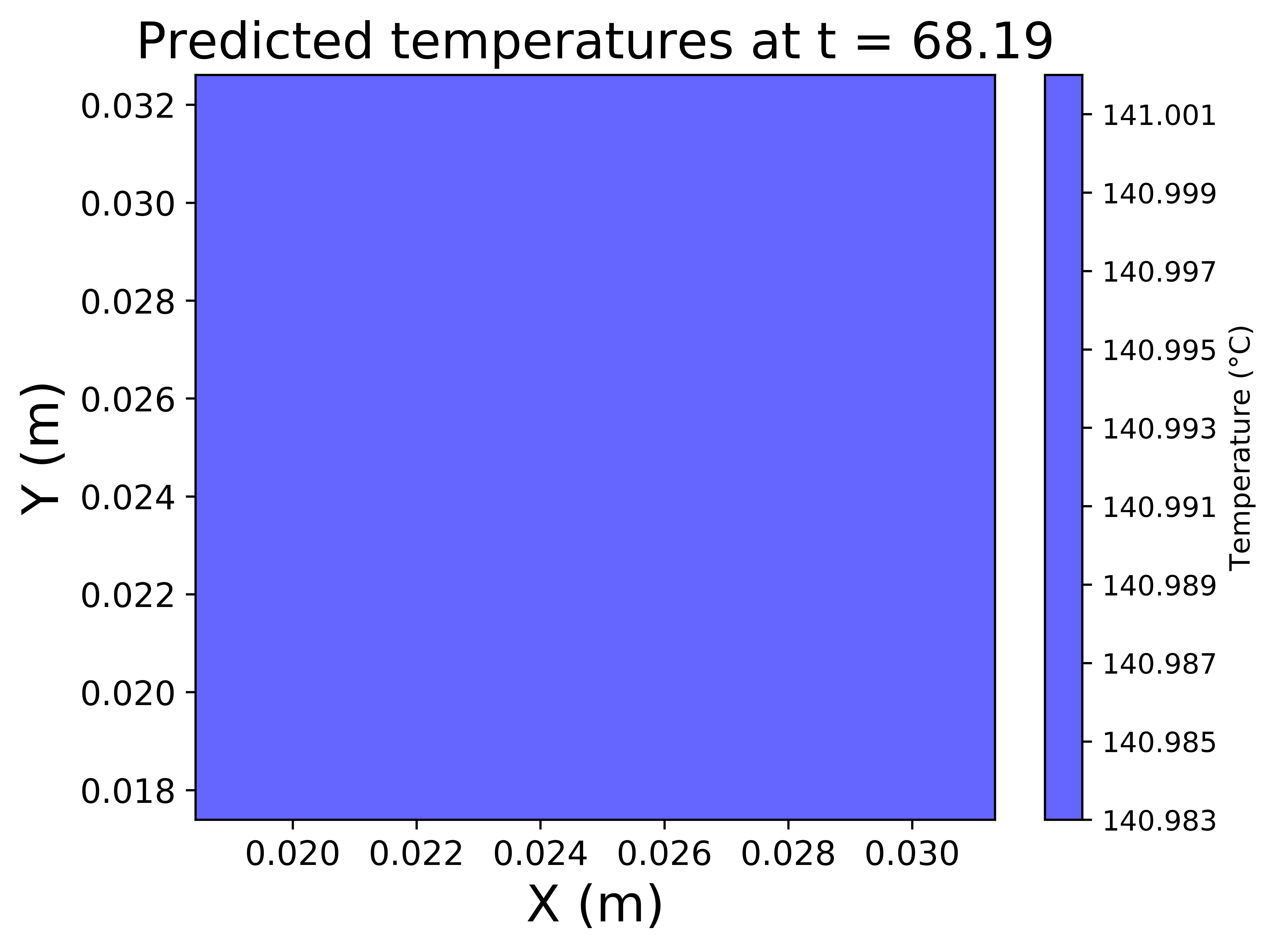}
        \caption{LS, NN,  i = 35.}
        \label{fig:linspace_nn_idx35}
    \end{subfigure}
    \begin{subfigure}{0.25\textwidth}
        \includegraphics[width=\textwidth]{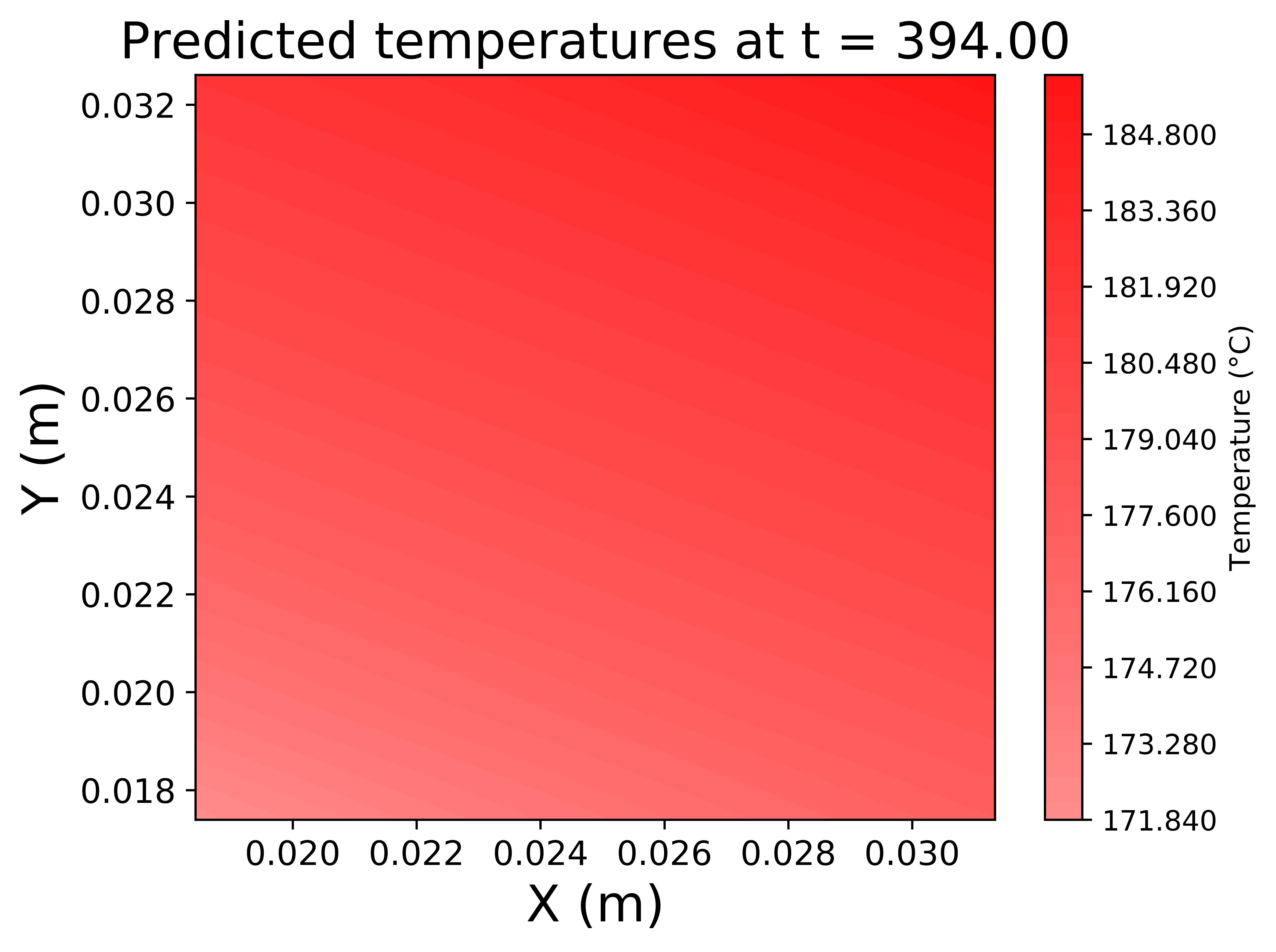}
        \caption{LS, NN,  i = 200.}
        \label{fig:linspace_nn_idx200}
    \end{subfigure}
    \begin{subfigure}{0.25\textwidth}
        \includegraphics[width=\textwidth]{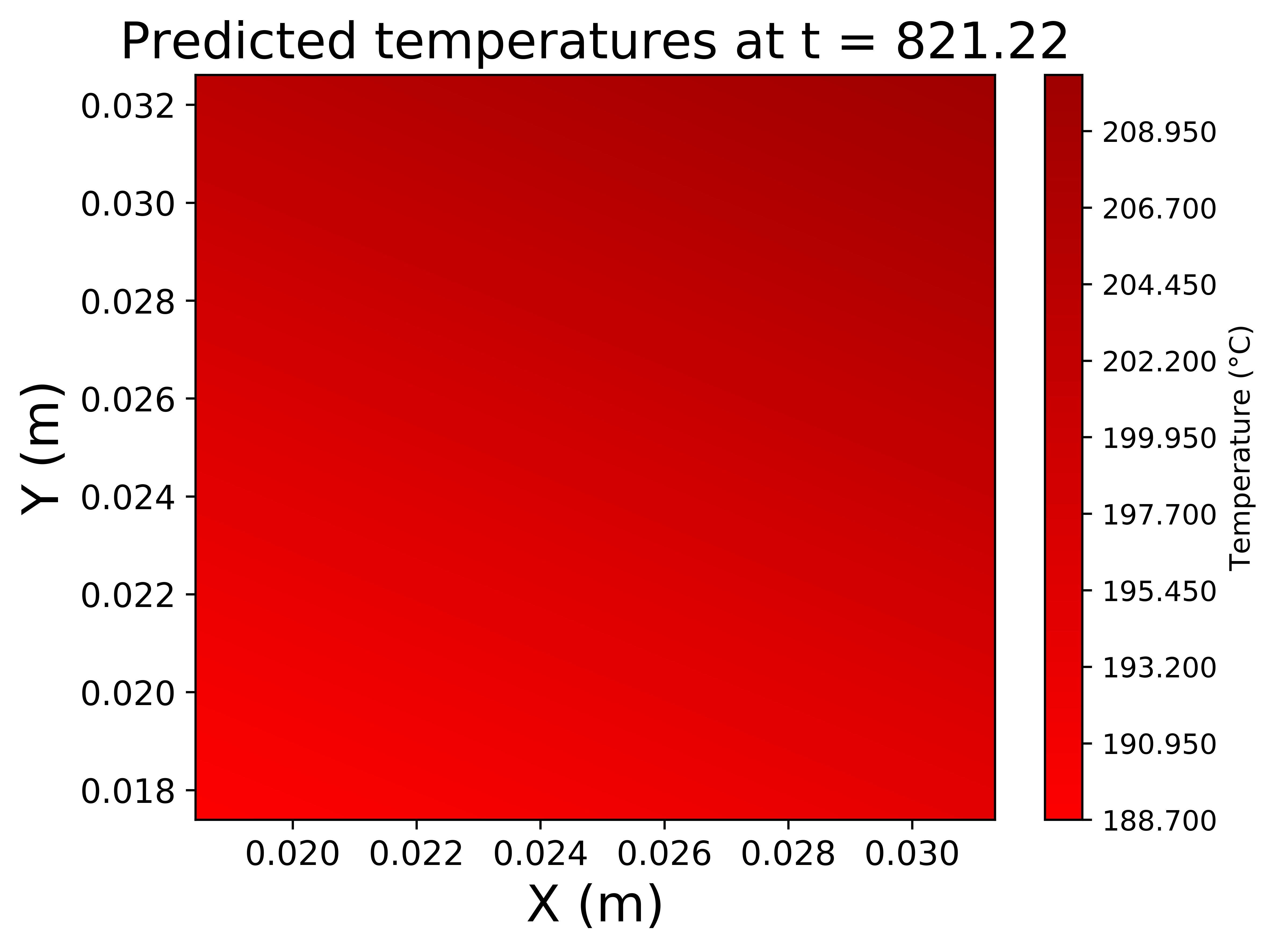}
        \caption{LS, NN,  i = 400.}
        \label{fig:linspace_nn_idx400}
    \end{subfigure}

    \begin{subfigure}{0.25\textwidth}
        \includegraphics[width=\textwidth]{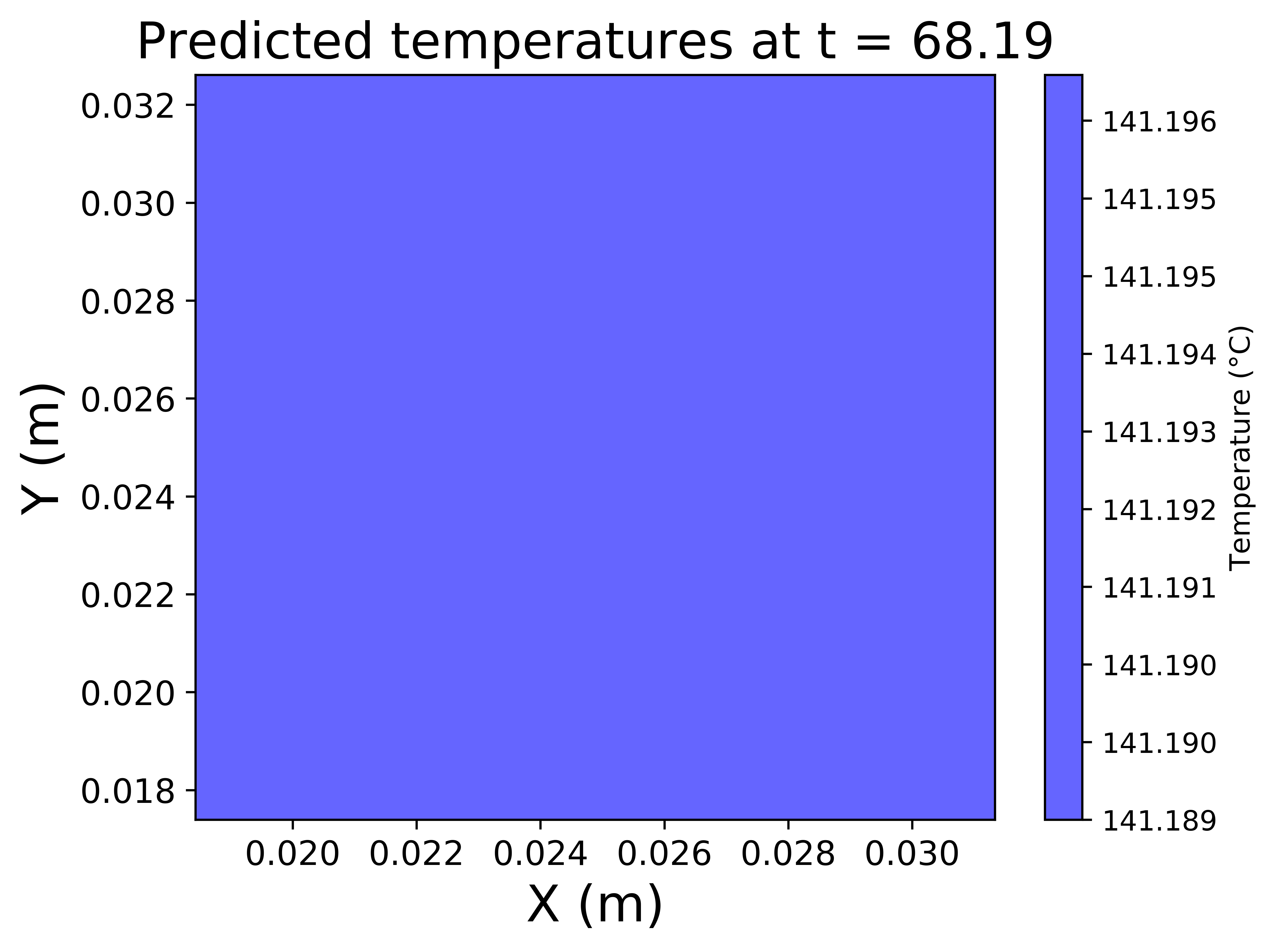}
        \caption{LS, PINN,  i = 35.}
        \label{fig:linspace_pinn_idx35}
    \end{subfigure}
    \begin{subfigure}{0.25\textwidth}
        \includegraphics[width=\textwidth]{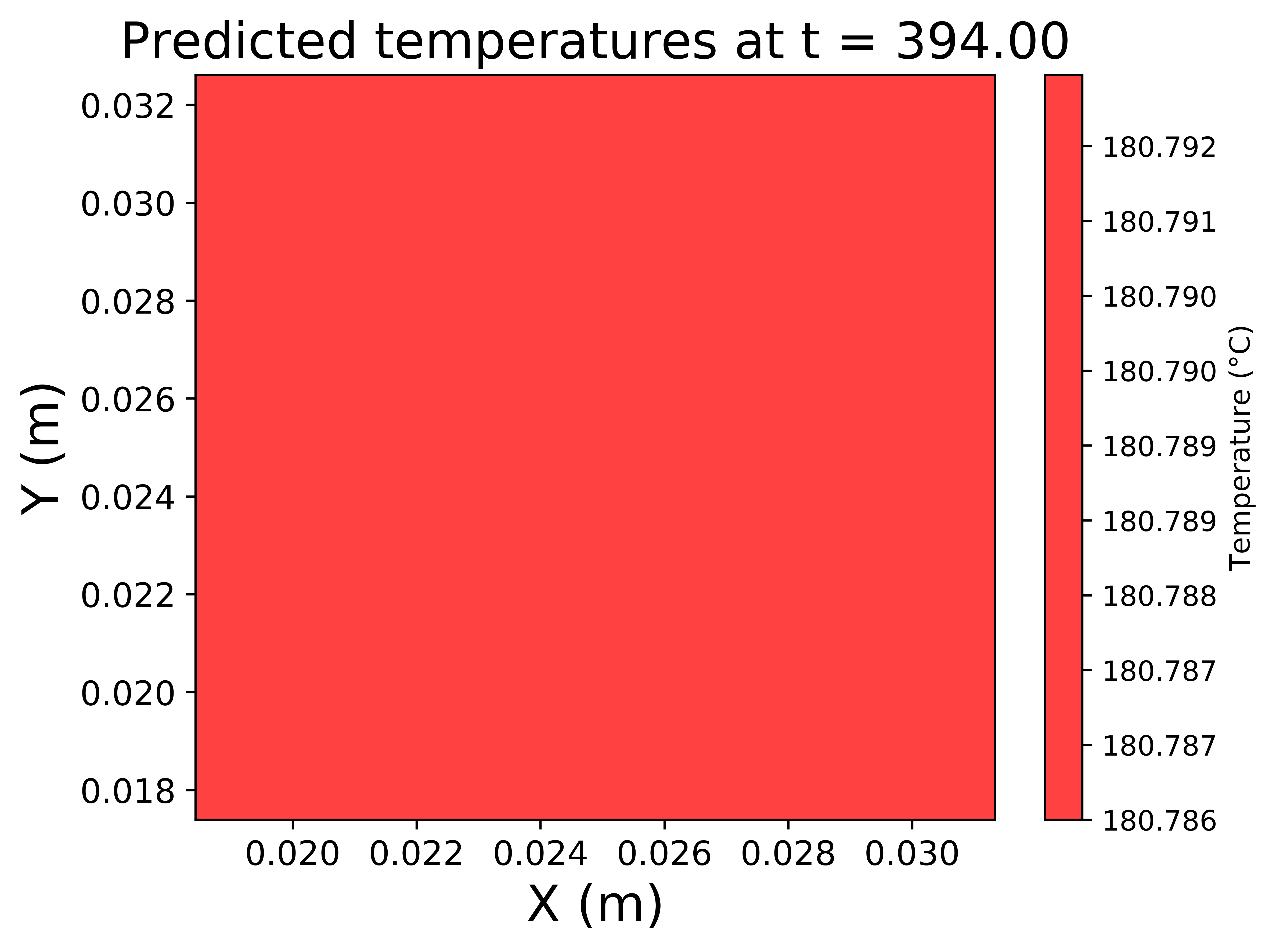}
        \caption{LS, PINN,  i = 200.}
        \label{fig:linspace_pinn_idx200}
    \end{subfigure}
    \begin{subfigure}{0.25\textwidth}
        \includegraphics[width=\textwidth]{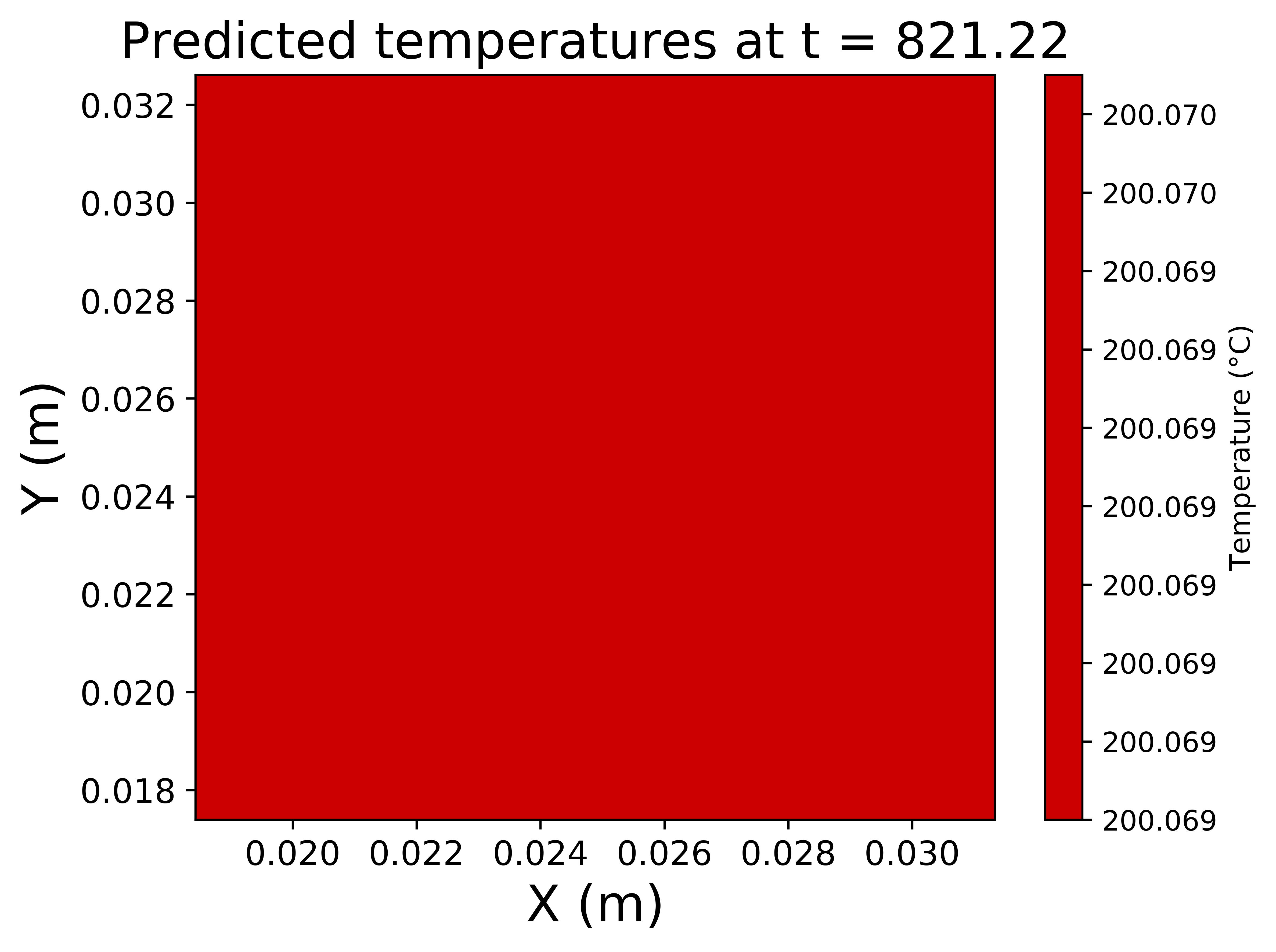}
        \caption{LS, PINN,  i = 400.}
        \label{fig:linspace_pinn_idx400}
    \end{subfigure}

    \begin{subfigure}{0.25\textwidth}
        \includegraphics[width=\textwidth]{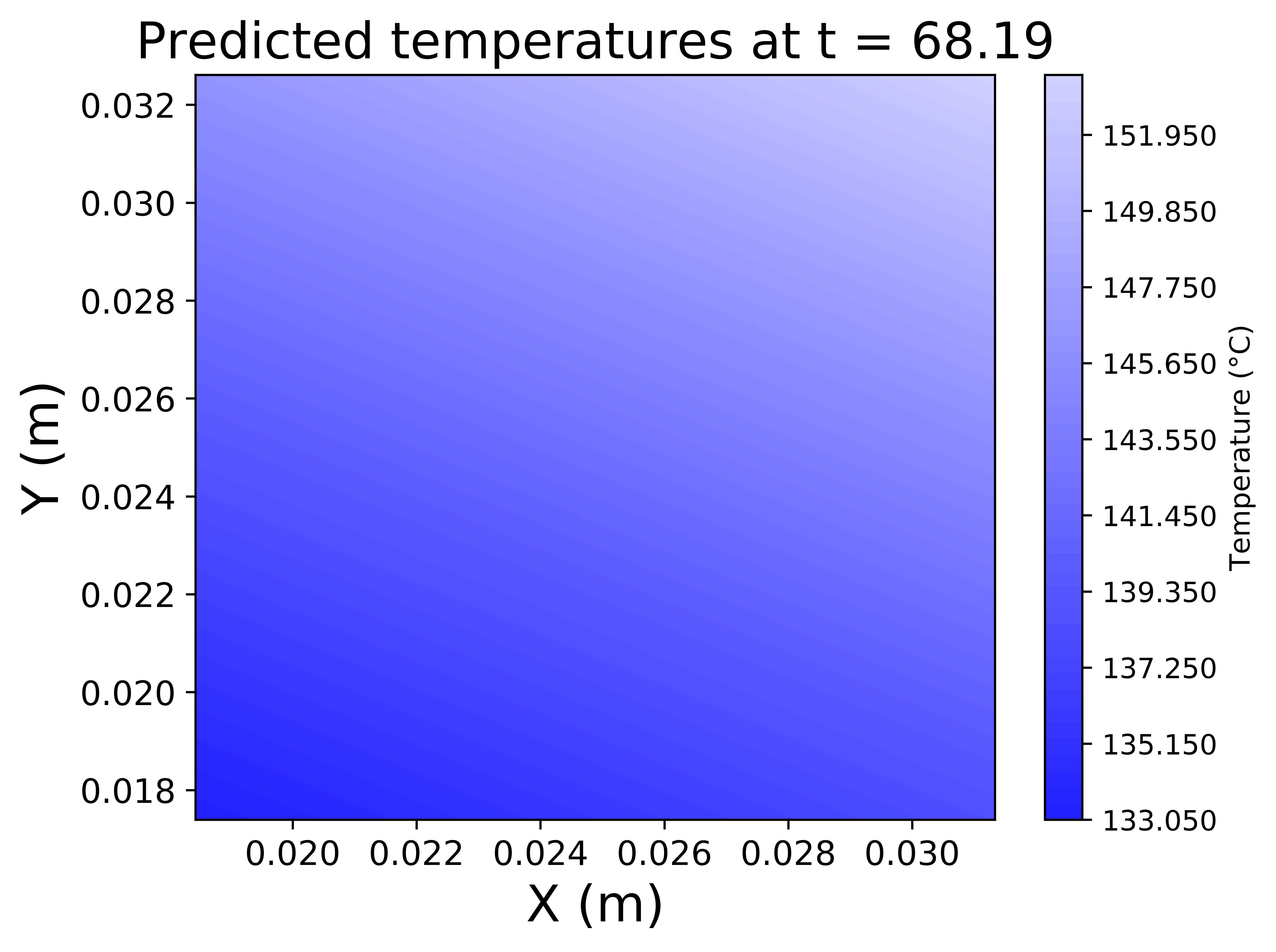}
        \caption{UD, NN,  i = 35.}
        \label{fig:uniform-dist_nn_idx35}
    \end{subfigure}
    \begin{subfigure}{0.25\textwidth}
        \includegraphics[width=\textwidth]{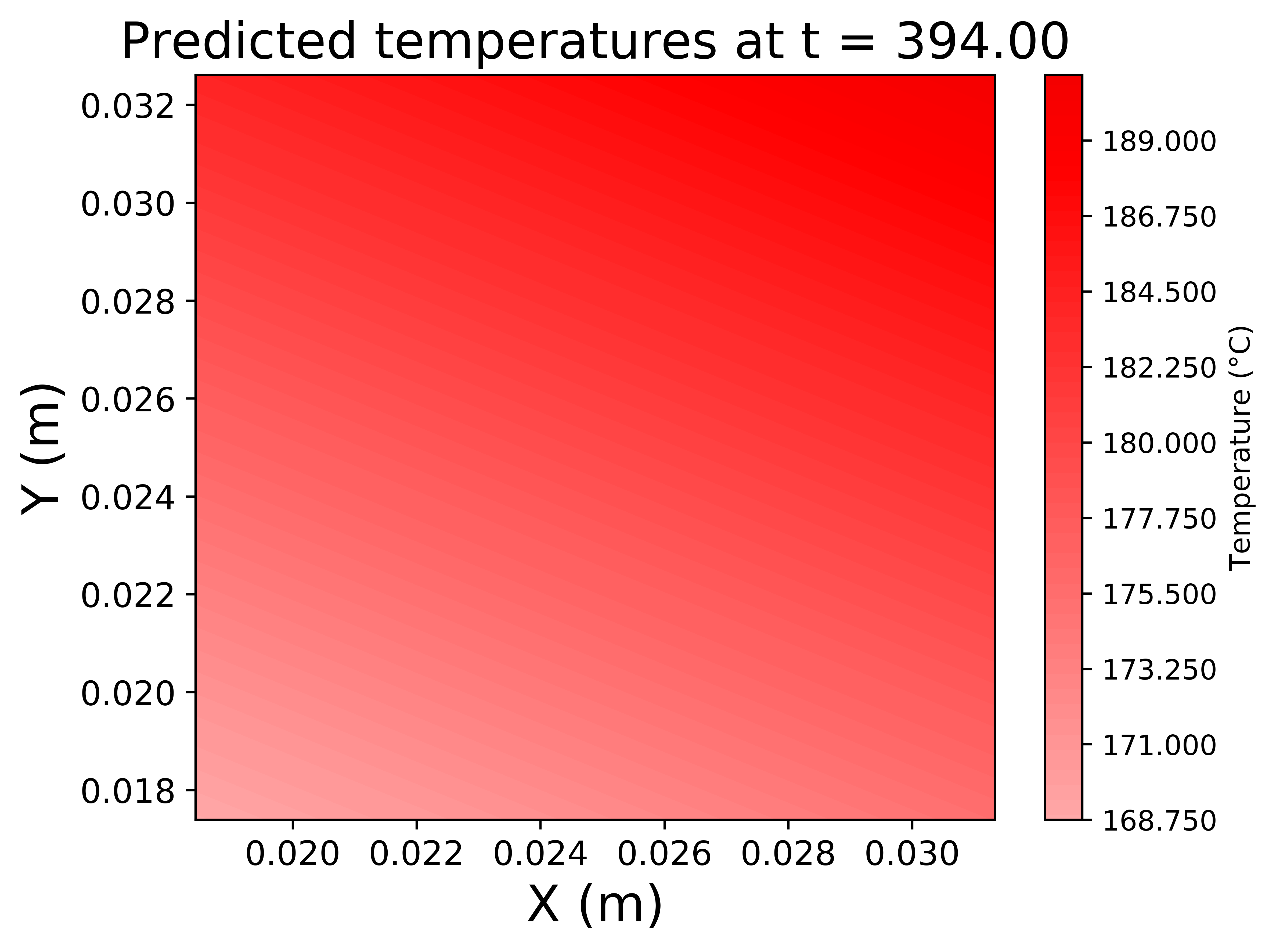}
        \caption{UD, NN,  i = 200.}
        \label{fig:uniform-dist_nn_idx200}
    \end{subfigure}
    \begin{subfigure}{0.25\textwidth}
        \includegraphics[width=\textwidth]{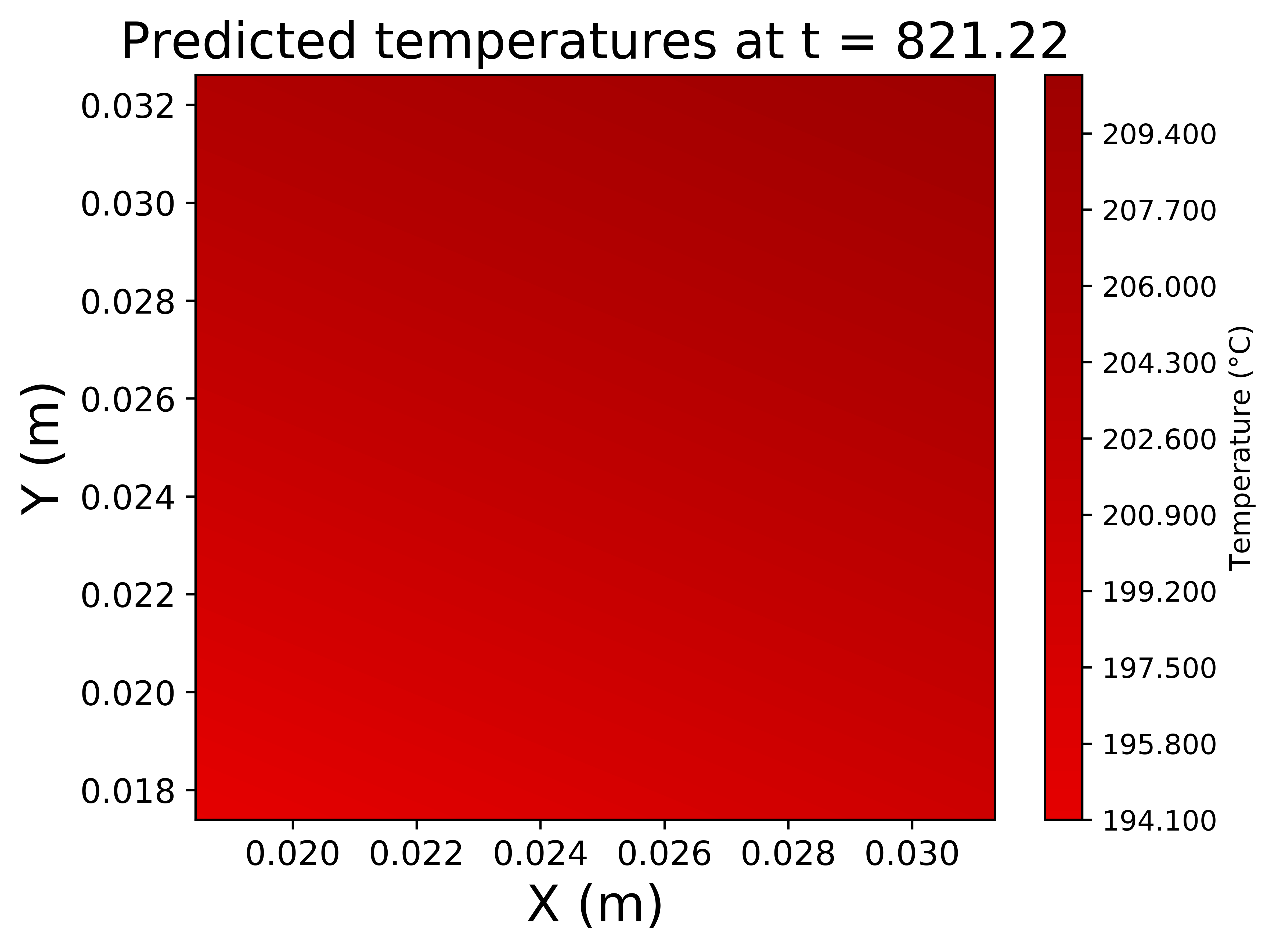}
        \caption{UD, NN,  i = 400.}
        \label{fig:uniform-dist_nn_idx400}
    \end{subfigure}

    \begin{subfigure}{0.25\textwidth}
        \includegraphics[width=\textwidth]{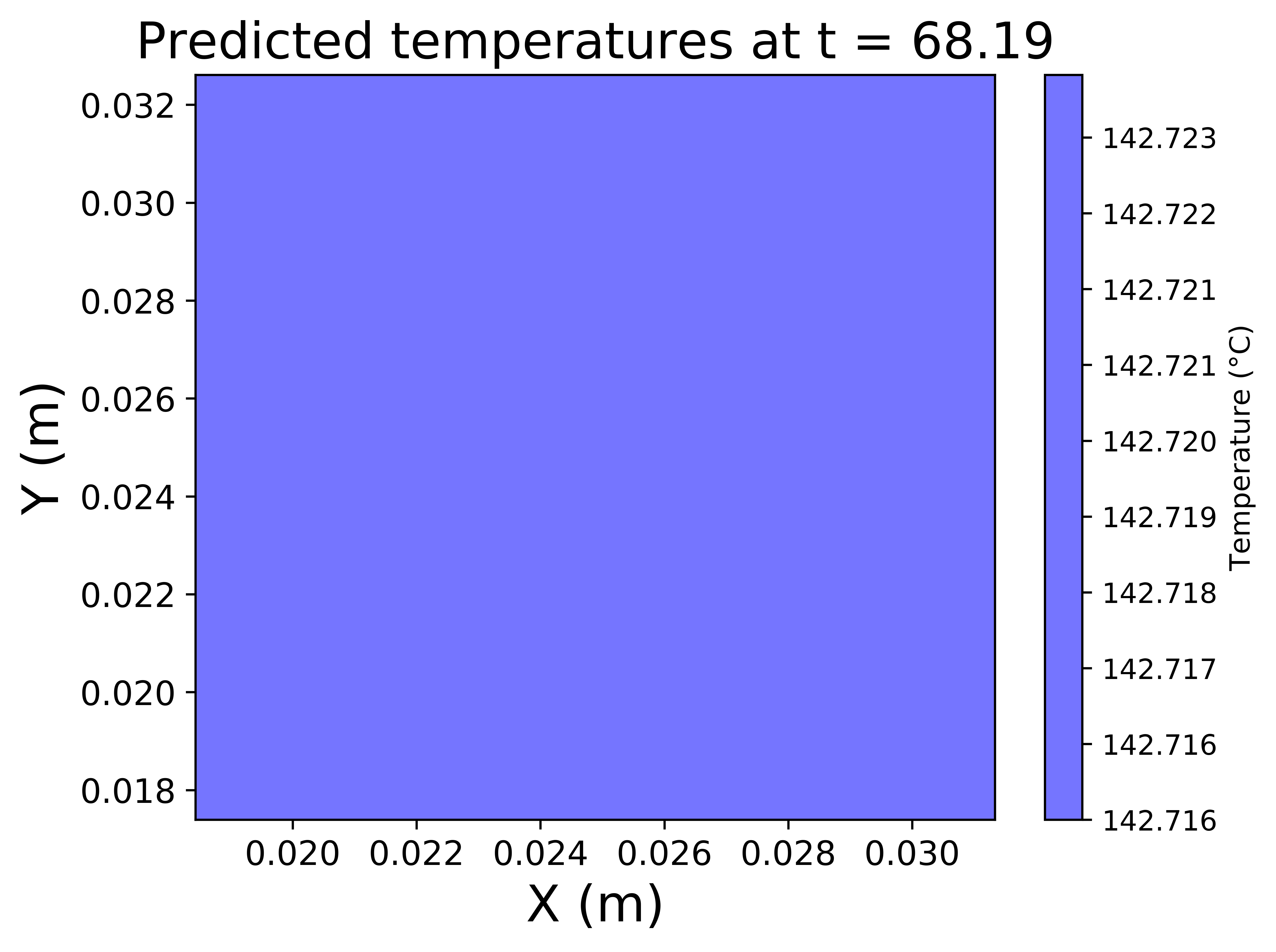}
        \caption{UD, PINN,  i = 35.}
        \label{fig:uniform-dist_pinn_idx35}
    \end{subfigure}
    \begin{subfigure}{0.25\textwidth}
        \includegraphics[width=\textwidth]{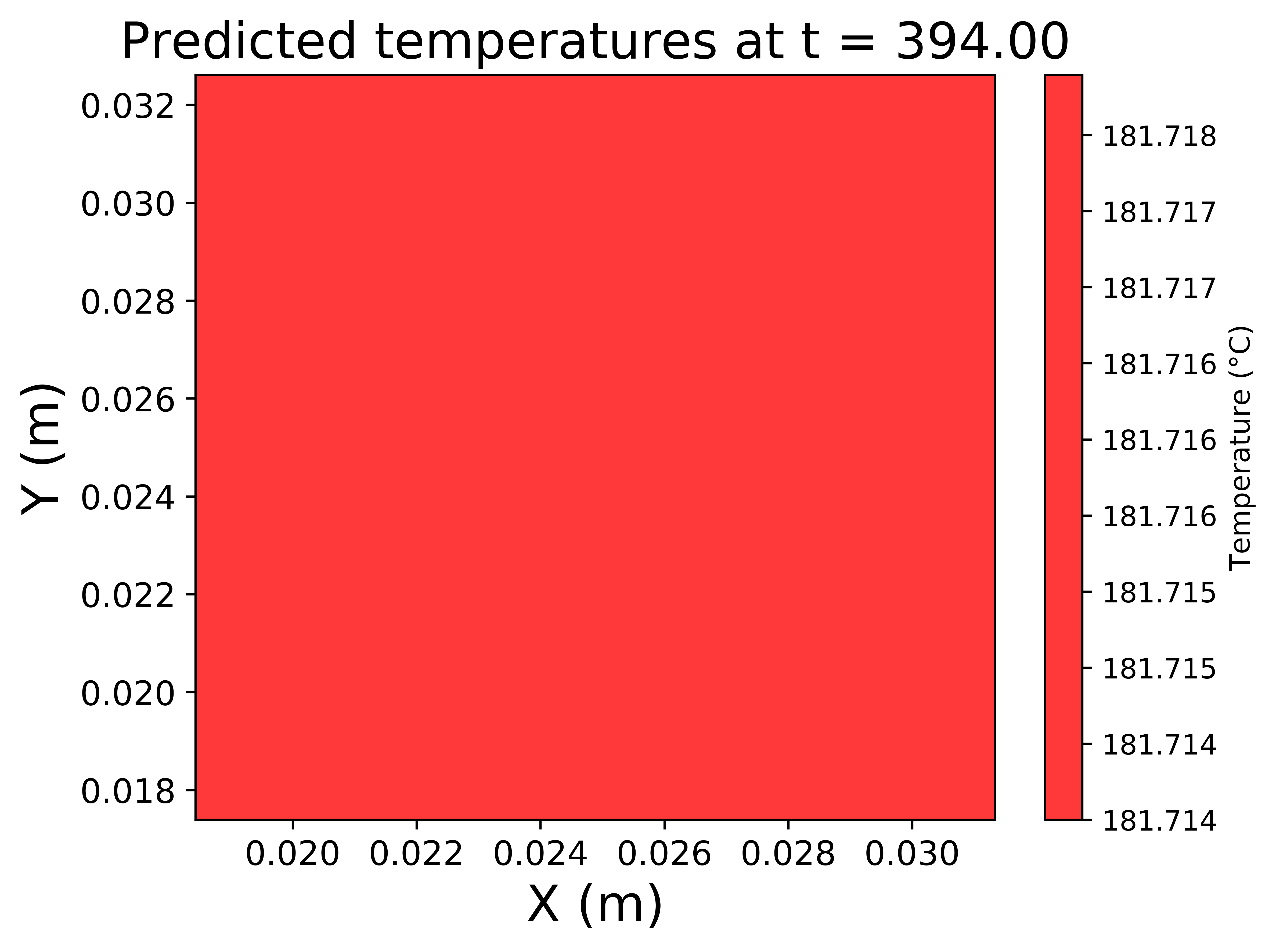}
        \caption{UD, PINN,  i = 200.}
        \label{fig:uniform-dist_pinn_idx200}
    \end{subfigure}
    \begin{subfigure}{0.25\textwidth}
        \includegraphics[width=\textwidth]{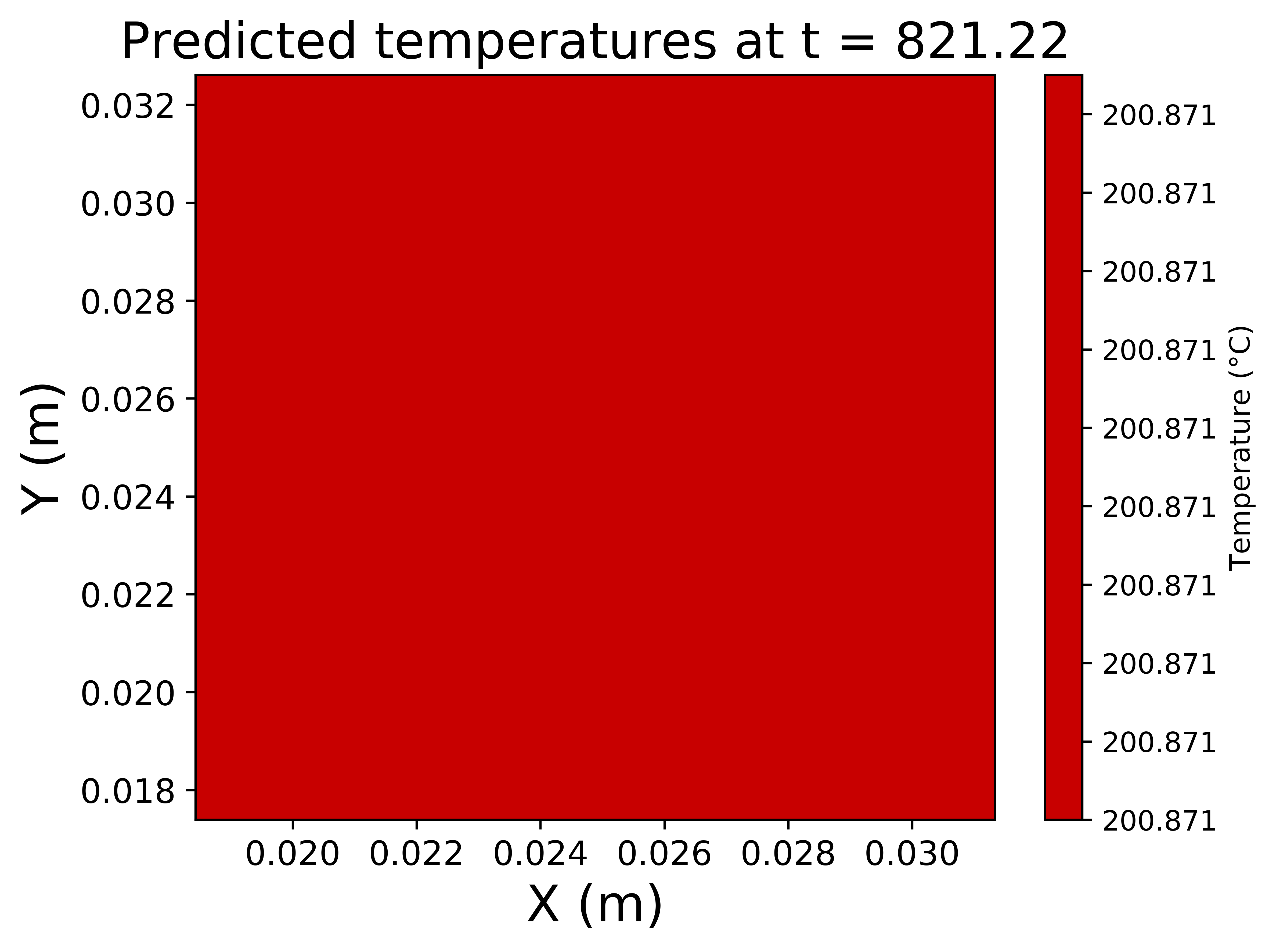}
        \caption{UD, PINN,  i = 400.}
        \label{fig:uniform-dist_pinn_idx400}
    \end{subfigure}
    \caption{Comparison between the test data and the predictions of NNs and PINNs after training with 832 points. LS denotes linearly-spaced data and UD denotes uniformly-distributed random data. i refers to the time sample index. The NNs capture slight heating gradients in space, whereas the PINNs predict almost constant temperatures for specific frames.}
    \label{fig:heat_diff_pred_comparison_visualised}
\end{figure}

Similarly to the case for linearly-spaced frames, we vary the number of training points within the domain but this time based on a random choice of uniformly-distributed points. Table~\ref{tab:heat_diffusion_uniform_dist_data} shows the training performance for these evaluations, and the fourth and fifth rows of Figure~\ref{fig:heat_diff_pred_comparison_visualised} show visualisations for their predictions. With the exception of a few outlier training cases, such as $\mathbf{N_d} = 73408$ and $\mathbf{N_d} = 384$, the PINN and NN have similar RMSEs. In general, the concentration of points at different spatiotemporal regions would affect how well the networks predict for those regions. However, based on the constant RMSE trend at values close to 9 irrespective of $\mathbf{N_d}$, it may be the case that for this problem the networks are converging to similar solutions regardless of the training data. The similarity between the predicted solutions in the second and fourth rows of Figure~\ref{fig:heat_diff_pred_comparison_visualised} further support this idea. 

\begin{table}[ht]
\centering
\begin{tabular}{|c|c|c|c|c|}
\hline
$\mathbf{N_d}$ & \textbf{NN}                                           & \textbf{PINN}                                         & $\mathbf{\alpha}$      & $\mathbf{\beta}$       \\ \hline
\textbf{220096}    & \begin{tabular}[c]{@{}c@{}}9.475\\ 3557\end{tabular}  & \begin{tabular}[c]{@{}c@{}}9.429\\ 2542\end{tabular}  & 9.9781              & 9.9761             \\ \hline
\textbf{73408}     & \begin{tabular}[c]{@{}c@{}}7.016\\ 3500\end{tabular}  & \begin{tabular}[c]{@{}c@{}}15.436\\ 4562\end{tabular} & -4e5        & -8e5       \\ \hline
\textbf{36736}     & \begin{tabular}[c]{@{}c@{}}9.2324\\ 5000\end{tabular} & \begin{tabular}[c]{@{}c@{}}9.5814\\ 5000\end{tabular} & 9.9888              & 9.7793             \\ \hline
\textbf{18368}     & \begin{tabular}[c]{@{}c@{}}8.792\\ 4341\end{tabular}  & \begin{tabular}[c]{@{}c@{}}10.384\\ 5000\end{tabular} & 10.0844             & 9.8509             \\ \hline
\textbf{9216}      & \begin{tabular}[c]{@{}c@{}}8.453\\ 2605\end{tabular}  & \begin{tabular}[c]{@{}c@{}}9.848\\ 5000\end{tabular}  & 9.8036              & 9.8602             \\ \hline
\textbf{4608}      & \begin{tabular}[c]{@{}c@{}}14.030\\ 1163\end{tabular} & \begin{tabular}[c]{@{}c@{}}14.599\\ 4506\end{tabular} & 9.9838              & 9.9897             \\ \hline
\textbf{2304}      & \begin{tabular}[c]{@{}c@{}}9.482\\ 4468\end{tabular}  & \begin{tabular}[c]{@{}c@{}}10.517\\ 1311\end{tabular} & 9.9933              & 10.0076            \\ \hline
\textbf{1792}      & \begin{tabular}[c]{@{}c@{}}20.813\\ 155\end{tabular}  & \begin{tabular}[c]{@{}c@{}}23.072\\ 5000\end{tabular} & -1e13 & -9e12 \\ \hline
\textbf{1152}      & \begin{tabular}[c]{@{}c@{}}11.198\\ 2666\end{tabular} & \begin{tabular}[c]{@{}c@{}}9.756\\ 5000\end{tabular}  & 9.8414              & 9.8285             \\ \hline
\textbf{832}       & \begin{tabular}[c]{@{}c@{}}7.819\\ 5000\end{tabular}  & \begin{tabular}[c]{@{}c@{}}9.467\\ 3139\end{tabular}  & 10.0405             & 9.932            \\ \hline
\textbf{704}       & \begin{tabular}[c]{@{}c@{}}9.241\\ 5000\end{tabular}  & \begin{tabular}[c]{@{}c@{}}9.470\\ 2458\end{tabular}  & 10.1434             & 10.0899            \\ \hline
\textbf{384}       & \begin{tabular}[c]{@{}c@{}}9.637\\ 3564\end{tabular}  & \begin{tabular}[c]{@{}c@{}}33.971\\ 225\end{tabular}  & 10.0501             & 10.0402            \\ \hline
\end{tabular}
\caption{RMSE values based on a variation of the number of uniformly-distributed points $\mathbf{N_d}$.}
\label{tab:heat_diffusion_uniform_dist_data}
\end{table}

\section{Closing Remarks}

In this chapter we have investigated and compared the predictive performance and trainability of PINNs against standard NNs for the reconstruction of the thermal diffusion regime of a block as it is being heated, based on an experiment that we set up. We also looked into issues related to sensor denoising and its effect on training. Unfortunately we have not been able to reconstruct a sufficiently accurate solution with PINNs nor with NNs, although we have found ways of achieving better accuracy. These include denoising the data, using a smaller frame size to reduce the difficulty of the optimization problem, and using the LBFGS optimizer instead of Adam. We hypothesize that, similarly to what we discussed in Section~\ref{sec:large_domain_training_pend}, the training domain in time is too large, resulting in stiffness in the optimization due to the difficulty of traversing the loss landscape. This is especially the case when we are trying to reconstruct 2D temperature grids rather than single angle measurements as in Section~\ref{sec:real_pend_exp_training}. In Section~\ref{sec:large_domain_training_pend} we solved this issue by reducing the domain of our problem to make the optimization easier. This was applicable for the case of the pendulum since it had a predictable sinusoidal pattern in a single dimension. The complexity of the 2D heat diffusion problem on the other hand is something that we want to fully investigate for the entire duration of our experiment, so we are interested in solving it for the entire domain.

One possible approach to this is to use sequence-to-sequence training, as was proposed by Krishnapriyan et al.~\cite{krishnapriyan2021}. Sequence-to-sequence training involves splitting the time domain into time-steps and training on the consecutive time-steps, one after the other. Krishnapriyan et al. show that using it enables them to achieve lower losses in the order of 1-2 magnitudes for a simulation of a 1D reaction-diffusion problem. We believe that applying it to our problem could allow us to arrive at accurate solutions for both NNs and PINNs. Additionally, by observing the coefficient values in Tables~\ref{tab:frame_size_variation},~\ref{tab:heat_diffusion_linspace_data}, and~\ref{tab:heat_diffusion_uniform_dist_data} we notice that the $\alpha$ and $\beta$ coefficients do not change much from their initial value. It it generally the case that inversion problems suffer from ill-posedness of optimization, require large numbers of network forward passes, and are highly susceptible to noise~\cite{moseley2022a}. Instead of posing their optimization as an inverse problem, Krishnapriyan et al. introduce a curriculum learning approach where the PINN is initially trained on small coefficient values, and then gradually retrained with larger coefficients~\cite{krishnapriyan2021}. This may be a promising approach to look into for our problem, since the abundance of noise in our setup may have affected the optimal values for the coefficients. A curriculum learning approach may be promising in terms of searching the solution space of the coefficients.

%% file: Chapter5/chapter5.tex
\chapter{Parallel Hardware and Time-coherent Sensing}

\label{chapter:hardware}

\ifpdf
    \graphicspath{{Chapter5/Figs/Raster/}{Chapter5/Figs/PDF/}{Chapter5/Figs/}}
\else
    \graphicspath{{Chapter5/Figs/Vector/}{Chapter5/Figs/}}
\fi

\section{Introduction}

This chapter studies sensing issues related to deployment within hardware for real-time inference, and focuses particularly on time coherence. Time coherence is the degree to which two or more $n$-dimensional data inputs that occur at the same time instance are captured at with minimal latency between them, for an arbitrary value of $n$. The inputs to deployed predictive models will commonly arrive from digital interfaces of sensors, and in many cases they could be multi-dimensional inputs arriving from many different sensors.  Common embedded microcontrollers face difficulties in maintaining time coherence for many sensors due to the sequential nature of their programs. FPGAs on the other hand are inherently parallel, and so are an appropriate choice for this type of sensing architecture as they are able to sample independently from parallel interfaces. Based on this, we present an experiment to shed light on the issues of time and space coherence for digital sensing applications.

\section{Parallel Capture Heating Experiment}

\label{sec:parallel_heat_exp}

One of the issues that came up in the heat diffusion investigation in Chapter~\ref{chapter:heat_diffusion}, was the noisiness of the data. Therefore we perform a similar block heating experiment but this time with multiple thermal cameras. Specifically, we are interested in understanding the aleatoric uncertainty by comparing the data from the different cameras, as well as looking into the viability of time-coherent sensing.

\subsection{Experimental setup}

\begin{figure}[ht]
    \centering
    \includegraphics[width=\textwidth]{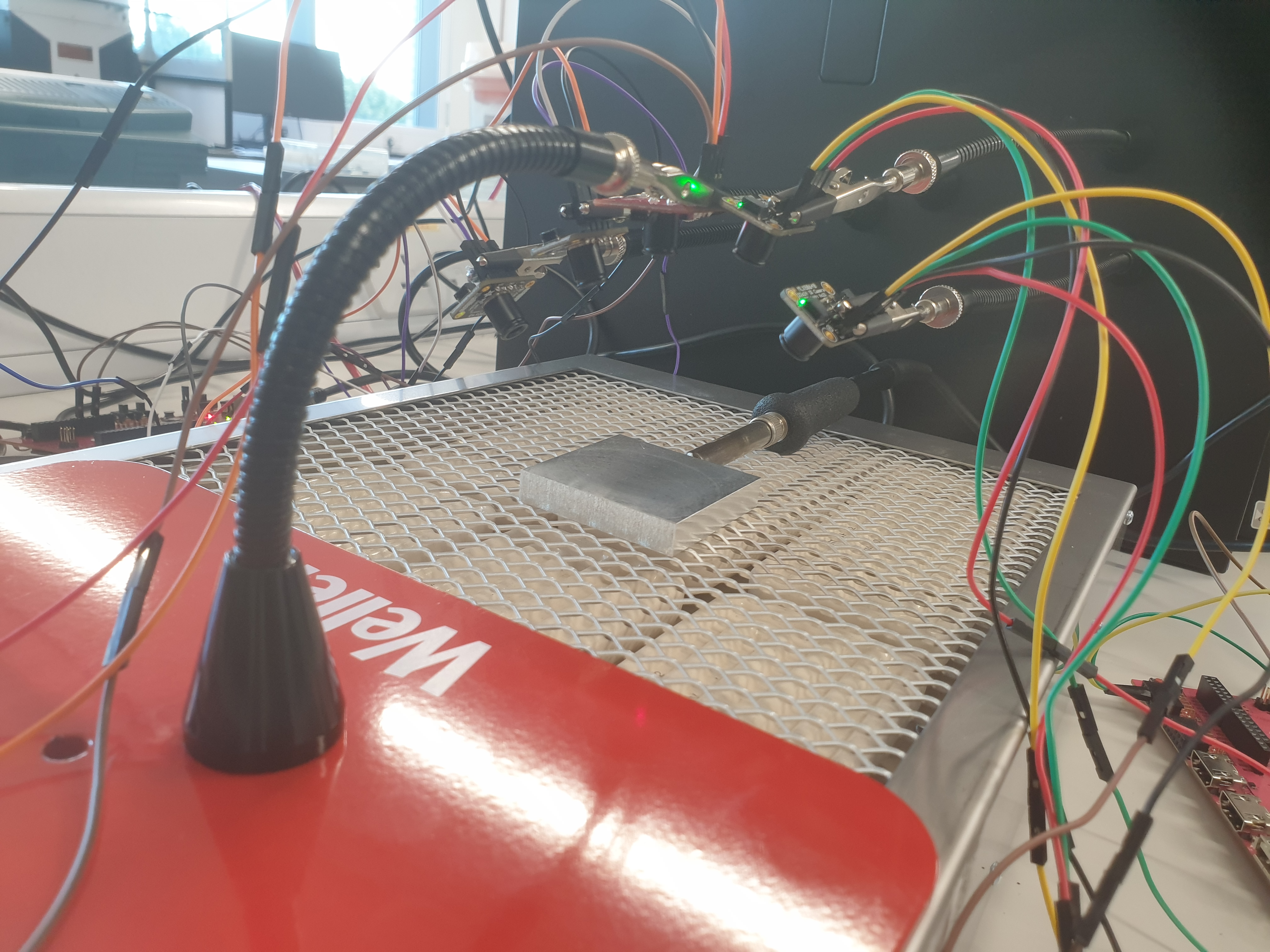}
    \caption{Experimental setup for the parallel heating experiment. We use 5 magnetic alligator clamps to hold 5 MLX90640 thermal cameras, which are connected to 5 PYNQ-Z1 FPGA boards. The cameras are pointed at the block so that the block surface takes up the most area in the camera FOVs.}
\label{fig:parallel_heating_exp_side_view}
\end{figure}

The setup for this experiment is similar to the one outlined in Section~\ref{sec:heat_diff_exp_setup}, but with 5 thermal cameras instead of one.  Figure~\ref{fig:parallel_heating_exp_side_view} shows the experimental setup.

We use 5 PYNQ-Z1 boards to interface with the 5 cameras. An ideal setup would involve 5 different AXI IIC blocks on one FPGA where the data from each block is offloaded to a BRAM. The BRAM would be read from in a single time instance to ensure time-coherence. Unfortunately however, due to project time constraints we settled on using 5 FPGAs instead. The PYNQ boards require an ethernet connection to access the Jupyter server that runs the Python subsystem. For the single camera setup we would connect the PYNQ board directly to the computer's ethernet port. However, since we have 5 boards for this experiment we need 5 ethernet ports. Therefore we use a network switch to connect the boards together.

The cameras are held in place with magnetic alligator clamp holders, and each camera is angled so that it covers as much of the block's surface as possible. We also apply thermal compound to increase the amount of conduction through surface contact.

We start the thermal data capture programs on each FPGA with as minimal human delay as possible, insert the soldering iron into the block once it is at just under 298°C, and begin the experiment.

\subsection{Data alignment}

\label{sec:data_alignment}

\begin{figure}[ht]
    \centering
    \includegraphics[width=\textwidth]{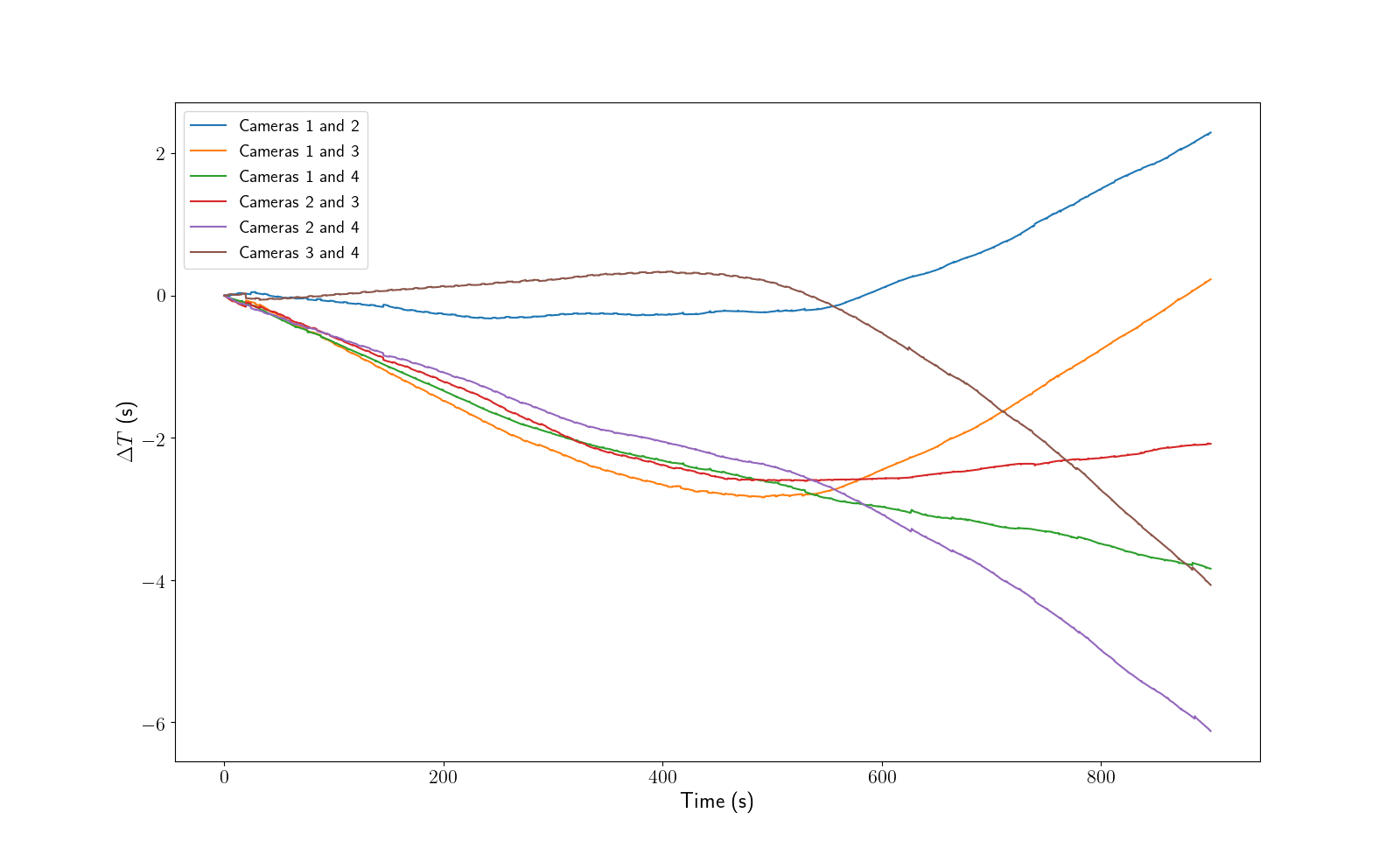}
    \caption{Time sample difference over the experiment duration for different cameras. The time coherence between each camera reduces over time, and in the worst case the difference is 6 seconds (cameras 2 and 4).}
\label{fig:time_diff_plot}
\end{figure}

After performing the experiment, we found that one the cameras had unfortunately failed to capture any data. This was the third camera which is positioned vertically above the block in Figure~\ref{fig:parallel_heating_exp_side_view}. Therefore we make do with data captured from 4 cameras. The following sections discuss issues with data alignment in both time and space.

\subsubsection{Time}

During the experiment, we captured timestamps for each measurement for all of the cameras using the Python \verb|time.time()| function which returns the time elapsed in seconds since the 1st of January 1970. Therefore, all of the measurements have an absolute reference of time that we can compare against. First, we subtract the first timestamp for each of the cameras from their respective timestamp arrays so that time starts from $0$ for each of them. To compare the time-coherence of the data between different cameras, we subtract their timestamps from each other and observe the time difference. Figure~\ref{fig:time_diff_plot} shows a plot of the differences. We can see that over time, the $\Delta \, T$ values between the cameras increases, indicating that the data misalignment is increasing. In the worst case, $\Delta T$ between cameras 2 and 4 drifts to a $6$ second difference by the end of the experiment duration, which is significant.

To ensure data validity, the frames for all of the cameras would have to be aligned in time, but this unfortunately is not the case with the data. There are two issues that are apparent here. The first is that the cameras have different levels of delays between each other, so a proper reference point would have to be found which is not a trivial matter. The second is that between any two given cameras the amount of delay is not constant, nor is it constant across the measurements of any single camera. Therefore, shifting the measurements for the cameras by fixed amounts would still not lead to correct alignment. The proper course of action in this case is to ensure that samples are taken from one device from 5 independant interfaces using dedicated AXI IIC blocks.

\subsubsection{Space}

\begin{figure}[ht]
    \centering
    \begin{subfigure}{0.49\textwidth}
        \includegraphics[width=\textwidth]{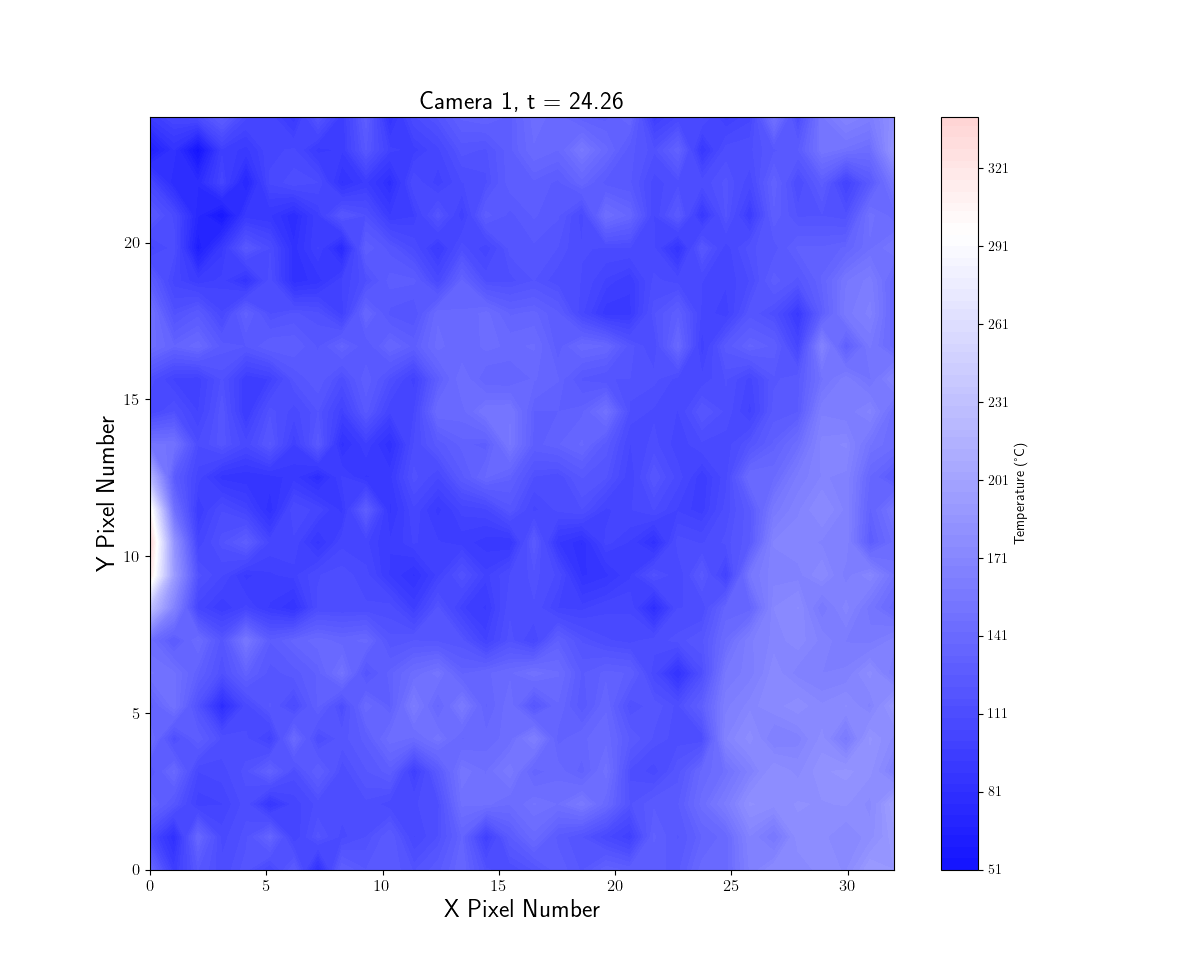}
        \caption{Camera 1.}
        \label{fig:parallel_heat_experiment_camera1_idx500}
    \end{subfigure}
    \begin{subfigure}{0.49\textwidth}
        \includegraphics[width=\textwidth]{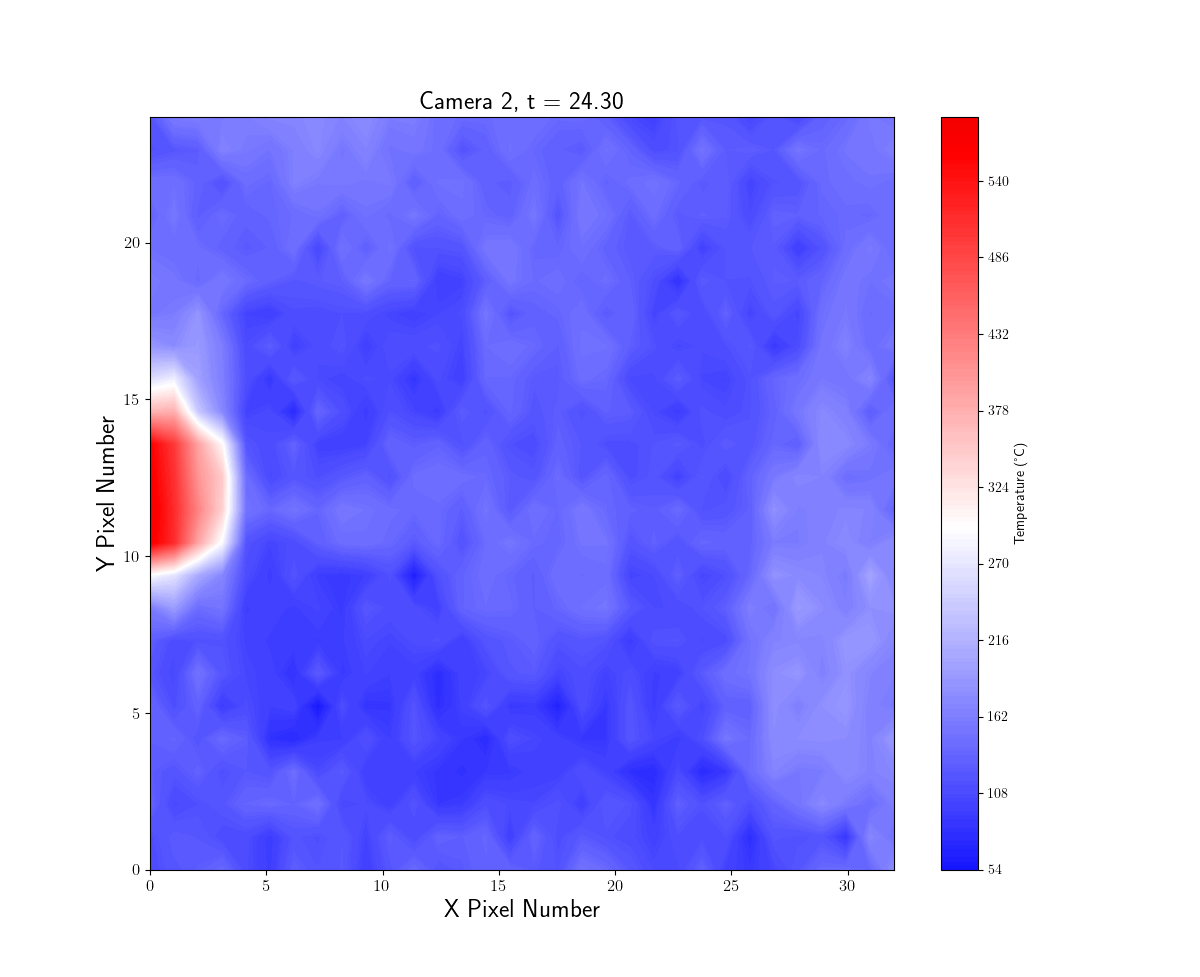}
        \caption{Camera 2.}
        \label{fig:parallel_heat_experiment_camera2_idx500}
    \end{subfigure}

    \begin{subfigure}{0.49\textwidth}
        \includegraphics[width=\textwidth]{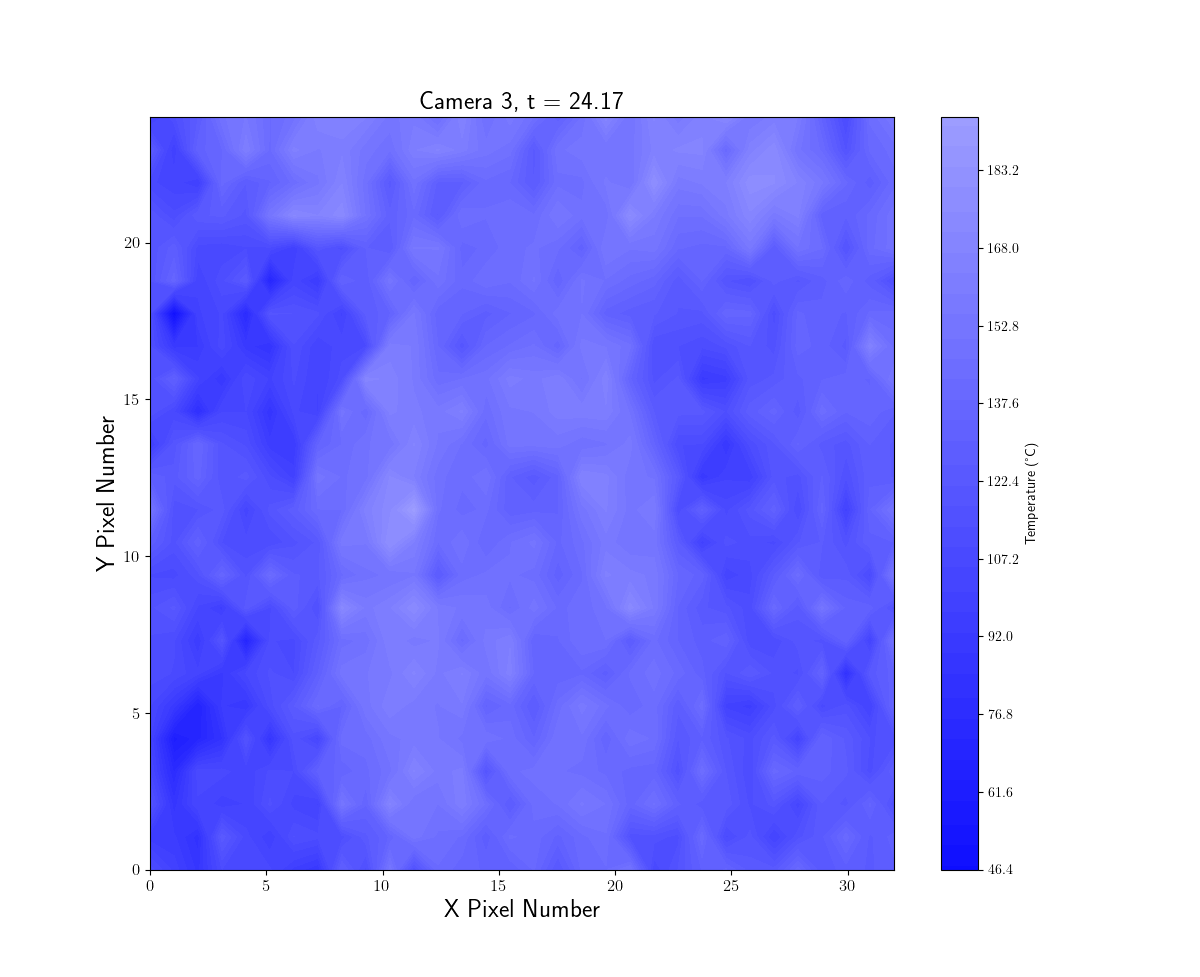}
        \caption{Camera 3.}
        \label{fig:parallel_heat_experiment_camera3_idx500}
    \end{subfigure}
    \begin{subfigure}{0.49\textwidth}
        \includegraphics[width=\textwidth]{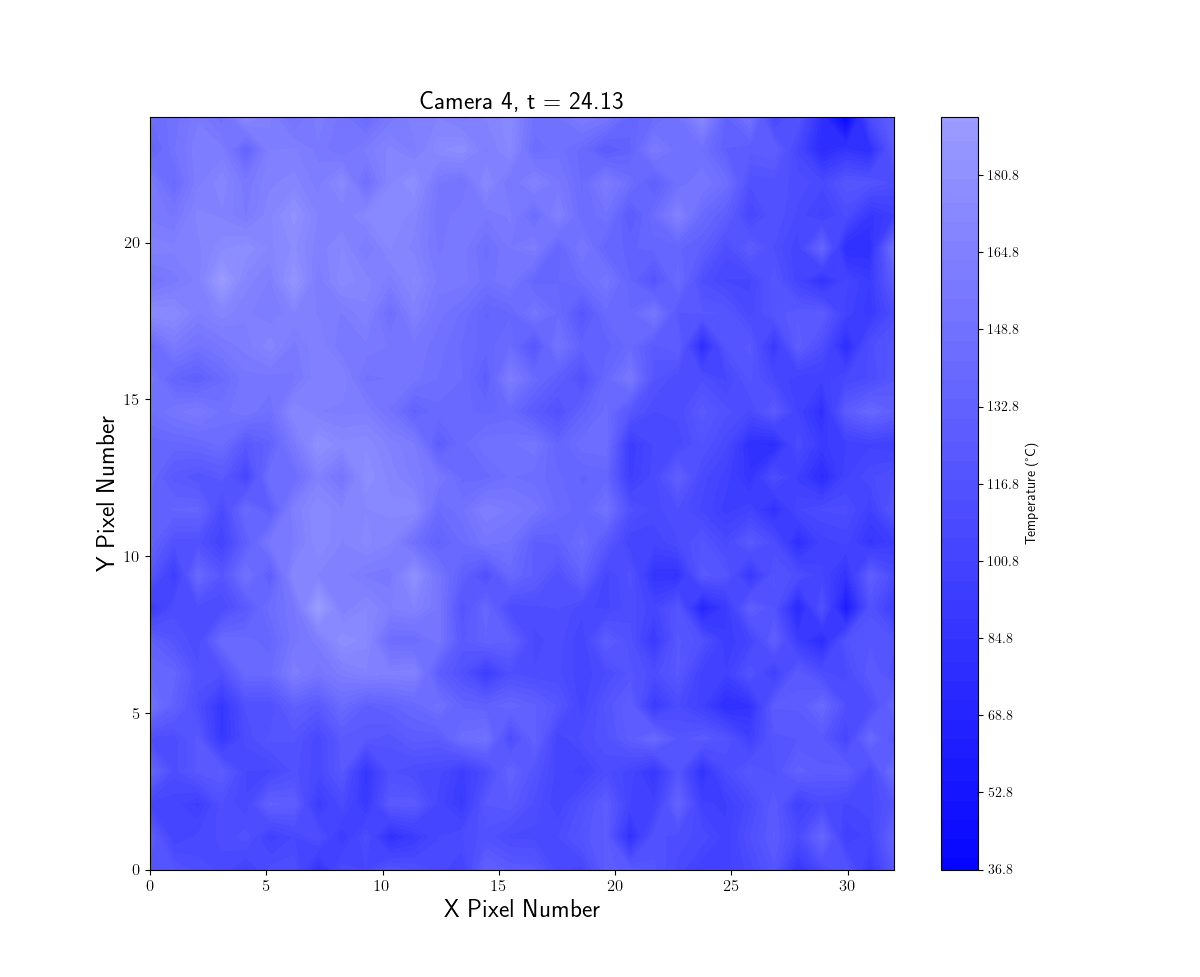}
        \caption{Camera 4.}
        \label{fig:parallel_heat_experiment_camera4_idx500}
    \end{subfigure}
    \caption{Frame visualisations at time sample 500.}
    \label{fig:parallel_heat_diff_idx500}
\end{figure}

\begin{figure}[ht]
    \begin{subfigure}{0.49\textwidth}
        \includegraphics[width=\textwidth]{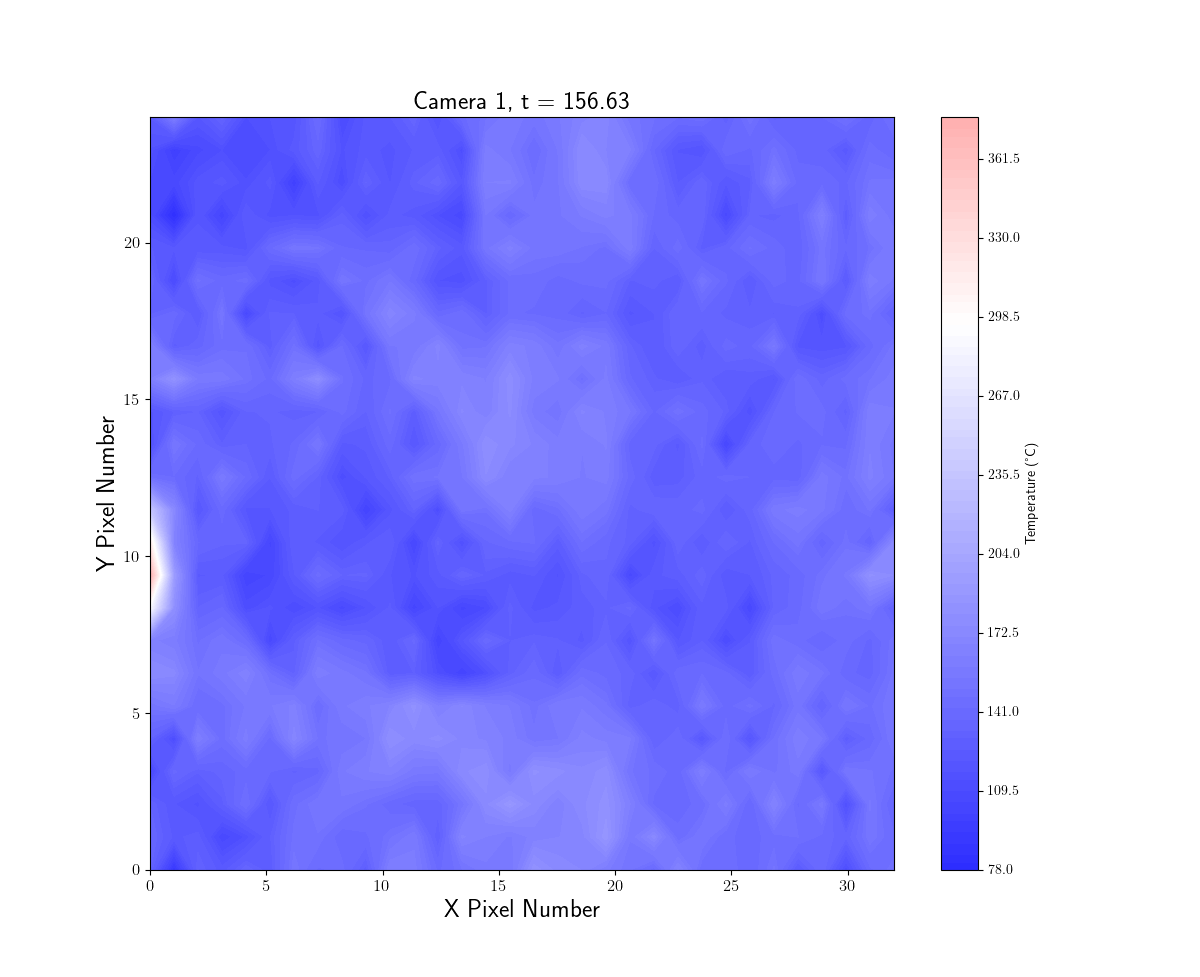}
        \caption{Camera 1.}
        \label{fig:parallel_heat_experiment_camera1_idx3250}
    \end{subfigure}
    \begin{subfigure}{0.49\textwidth}
        \includegraphics[width=\textwidth]{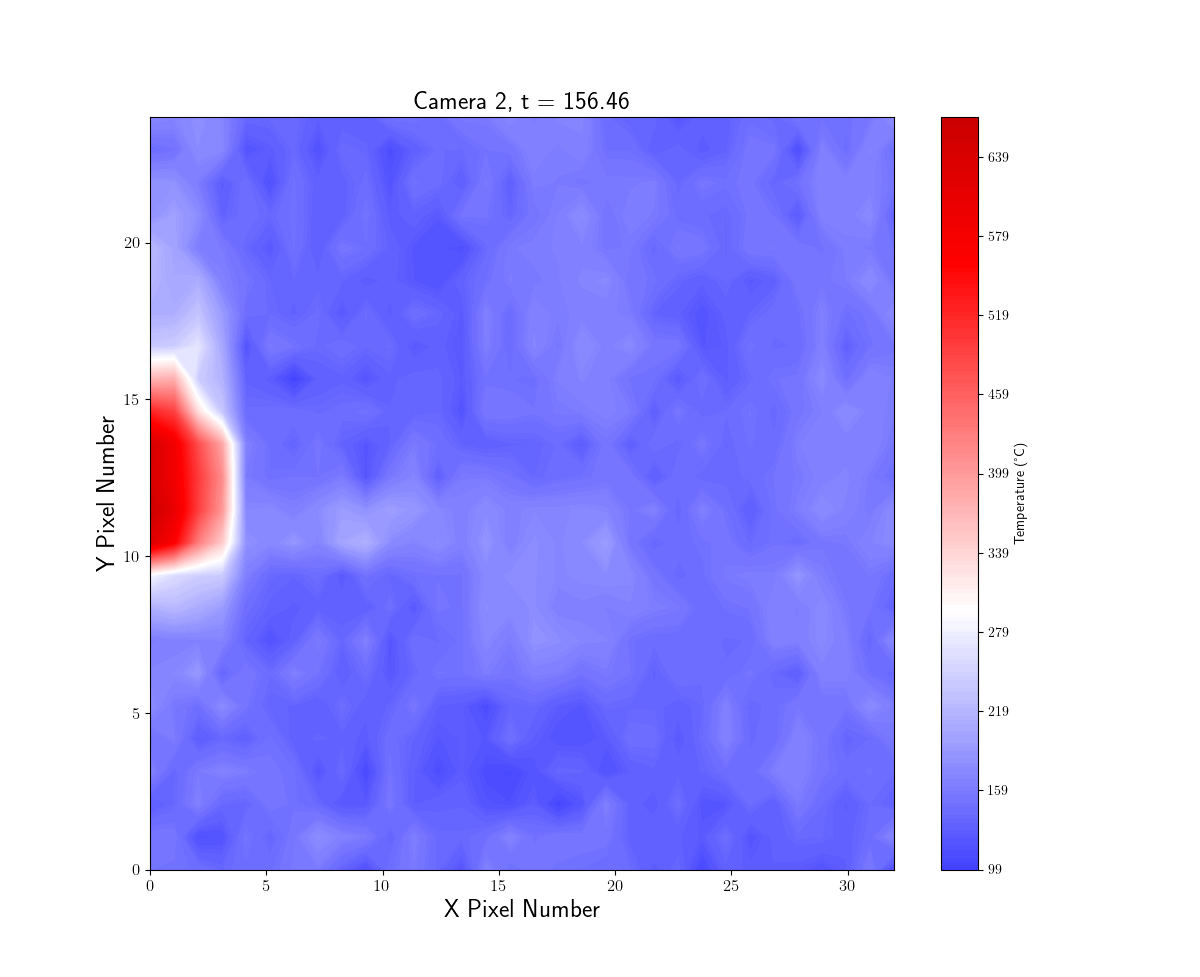}
        \caption{Camera 2.}
        \label{fig:parallel_heat_experiment_camera2_idx3250}
    \end{subfigure}

    \begin{subfigure}{0.49\textwidth}
        \includegraphics[width=\textwidth]{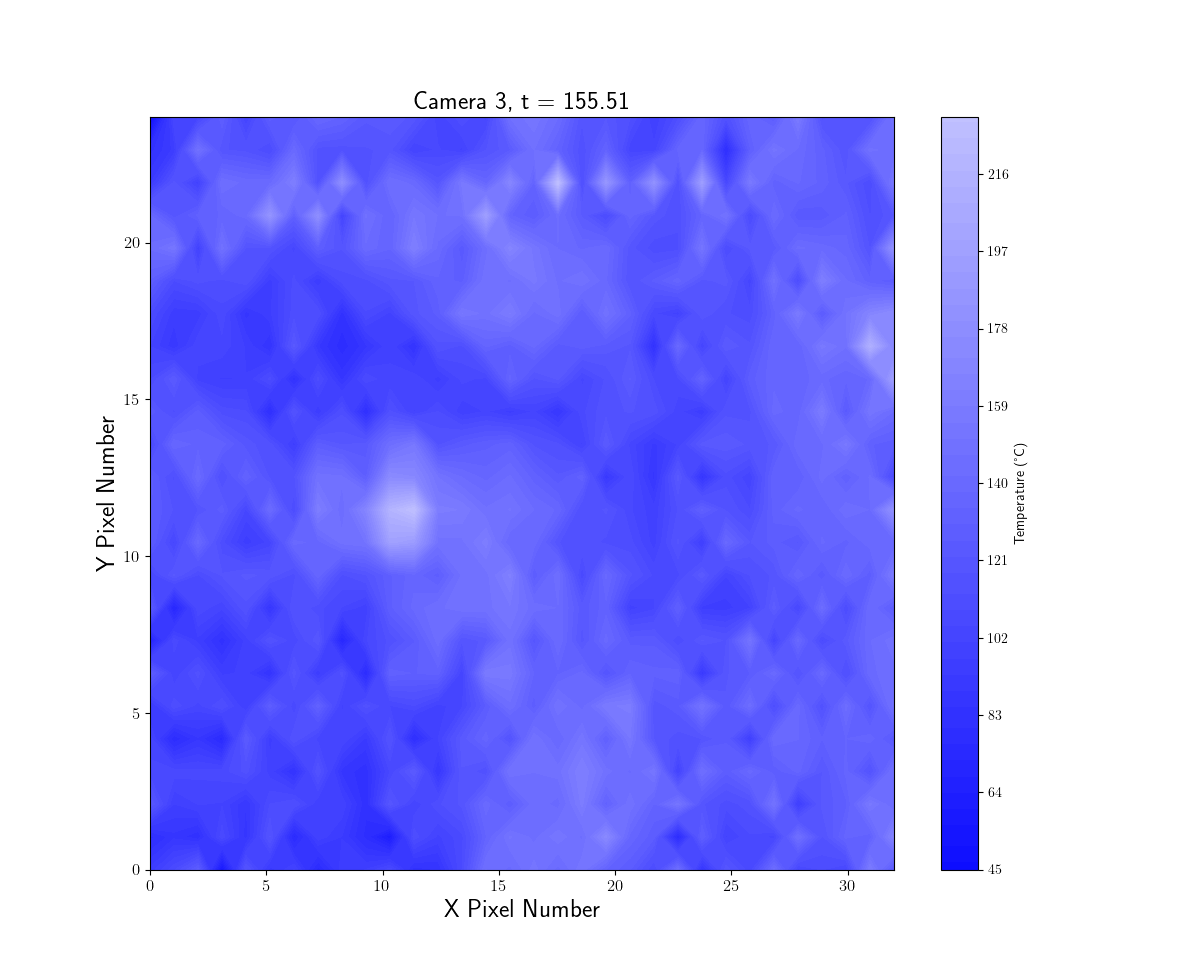}
        \caption{Camera 3.}
        \label{fig:parallel_heat_experiment_camera3_idx3250}
    \end{subfigure}
    \begin{subfigure}{0.49\textwidth}
        \includegraphics[width=\textwidth]{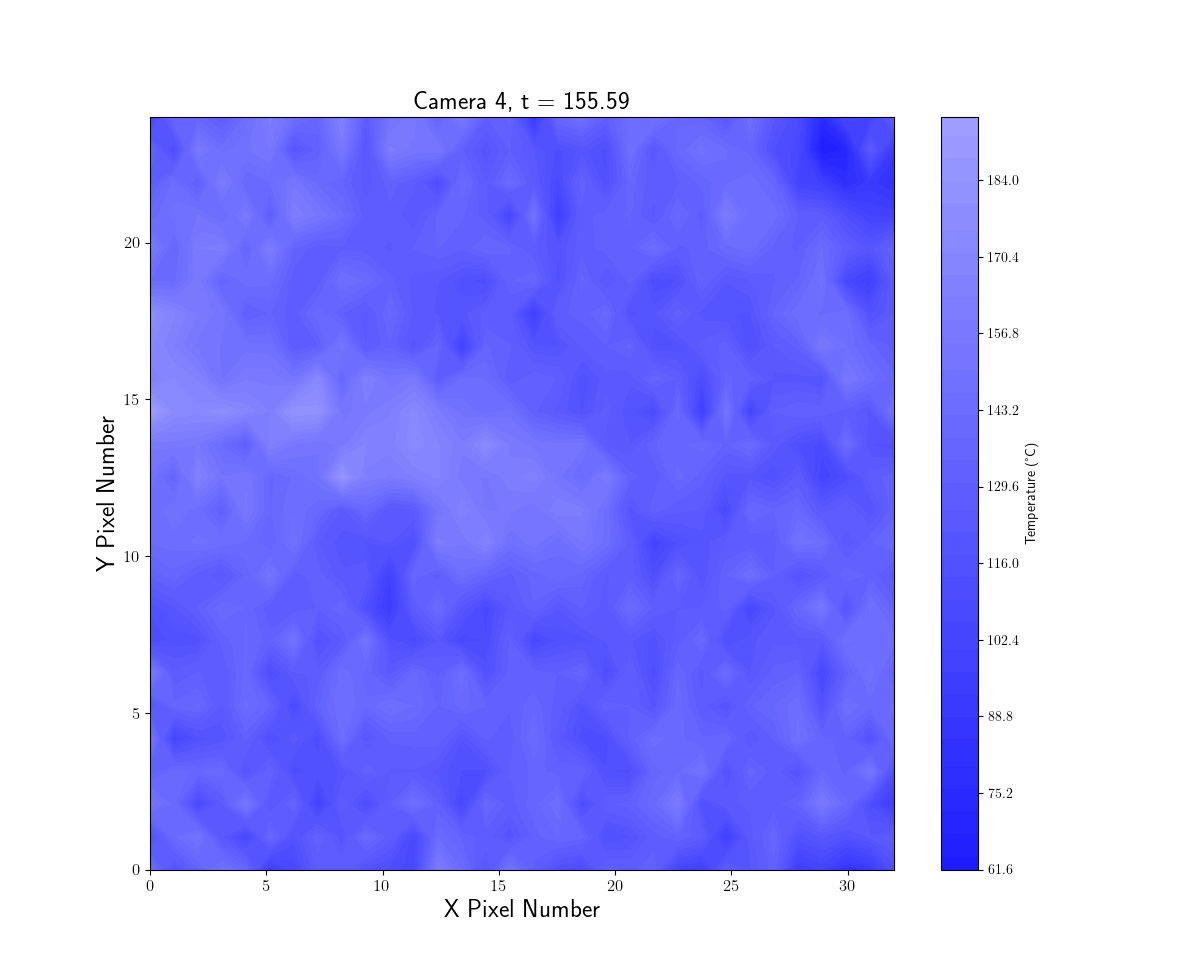}
        \caption{Camera 4.}
        \label{fig:parallel_heat_experiment_camera4_idx3250}
    \end{subfigure}
    \caption{Frame visualisations at time sample 3250.}
    \label{fig:parallel_heat_diff_idx3250}
\end{figure}

\begin{figure}[ht]
    \begin{subfigure}{0.49\textwidth}
        \includegraphics[width=\textwidth]{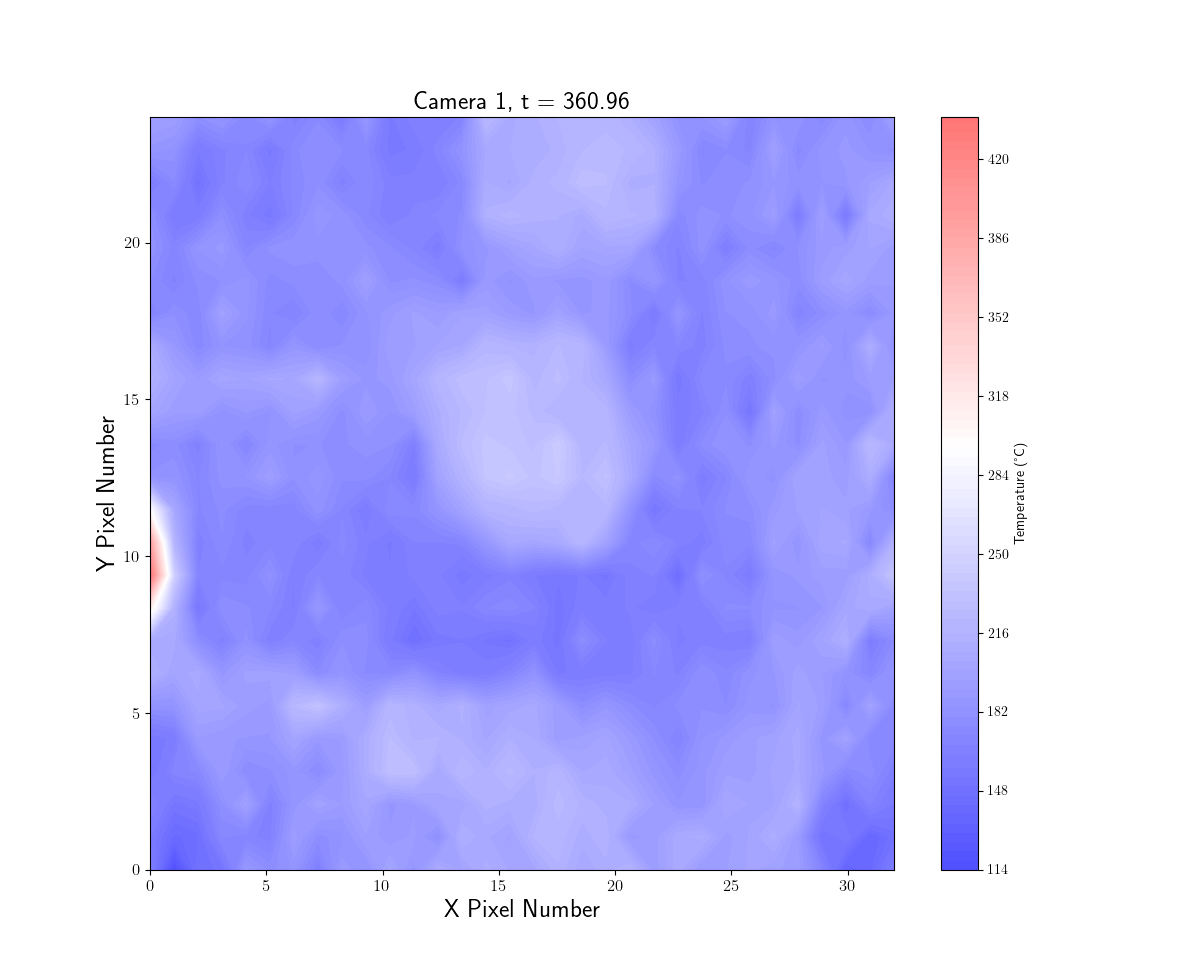}
        \caption{Camera 1.}
        \label{fig:parallel_heat_experiment_camera1_idx7500}
    \end{subfigure}
    \begin{subfigure}{0.49\textwidth}
        \includegraphics[width=\textwidth]{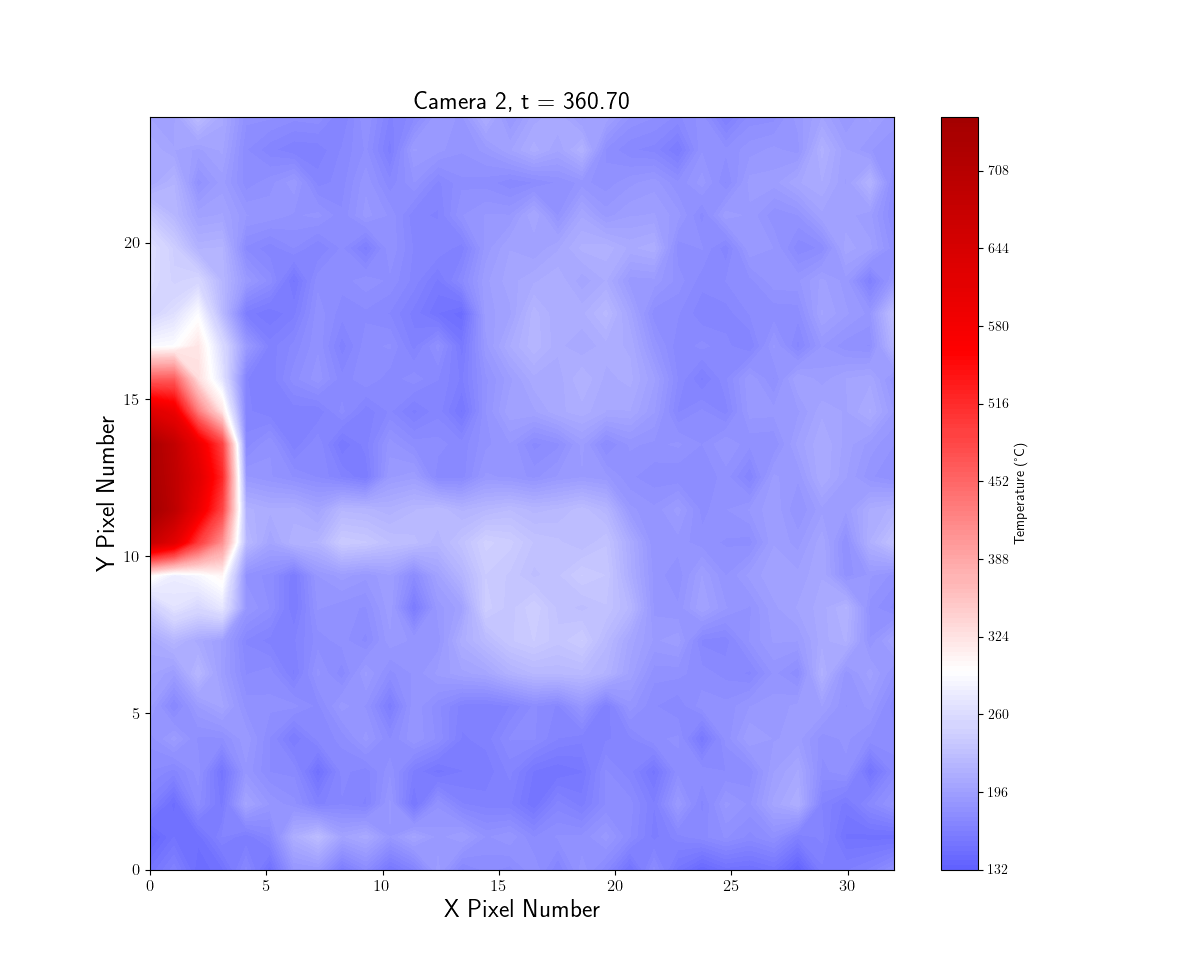}
        \caption{Camera 2.}
        \label{fig:parallel_heat_experiment_camera2_idx7500}
    \end{subfigure}

    \begin{subfigure}{0.49\textwidth}
        \includegraphics[width=\textwidth]{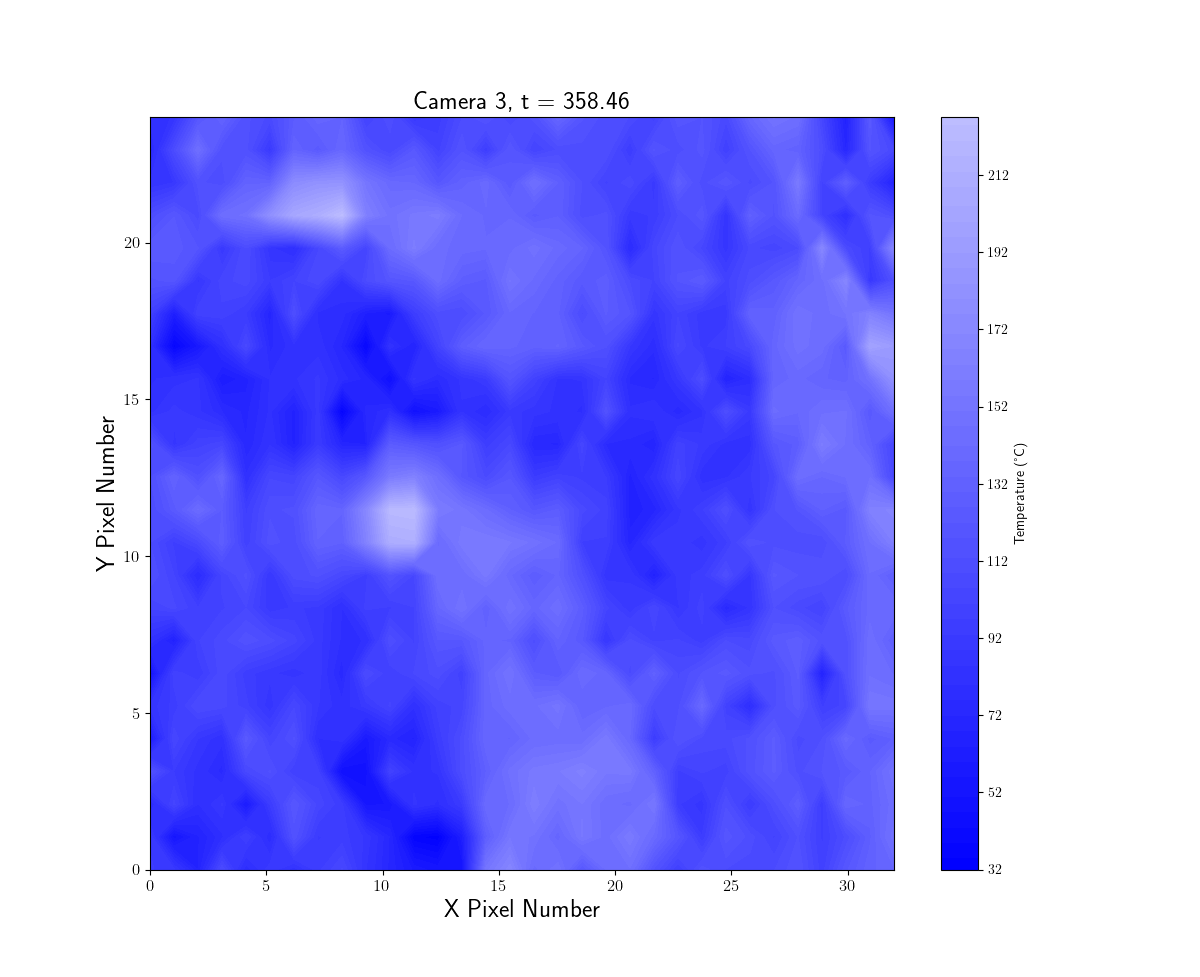}
        \caption{Camera 3.}
        \label{fig:parallel_heat_experiment_camera3_idx7500}
    \end{subfigure}
    \begin{subfigure}{0.49\textwidth}
        \includegraphics[width=\textwidth]{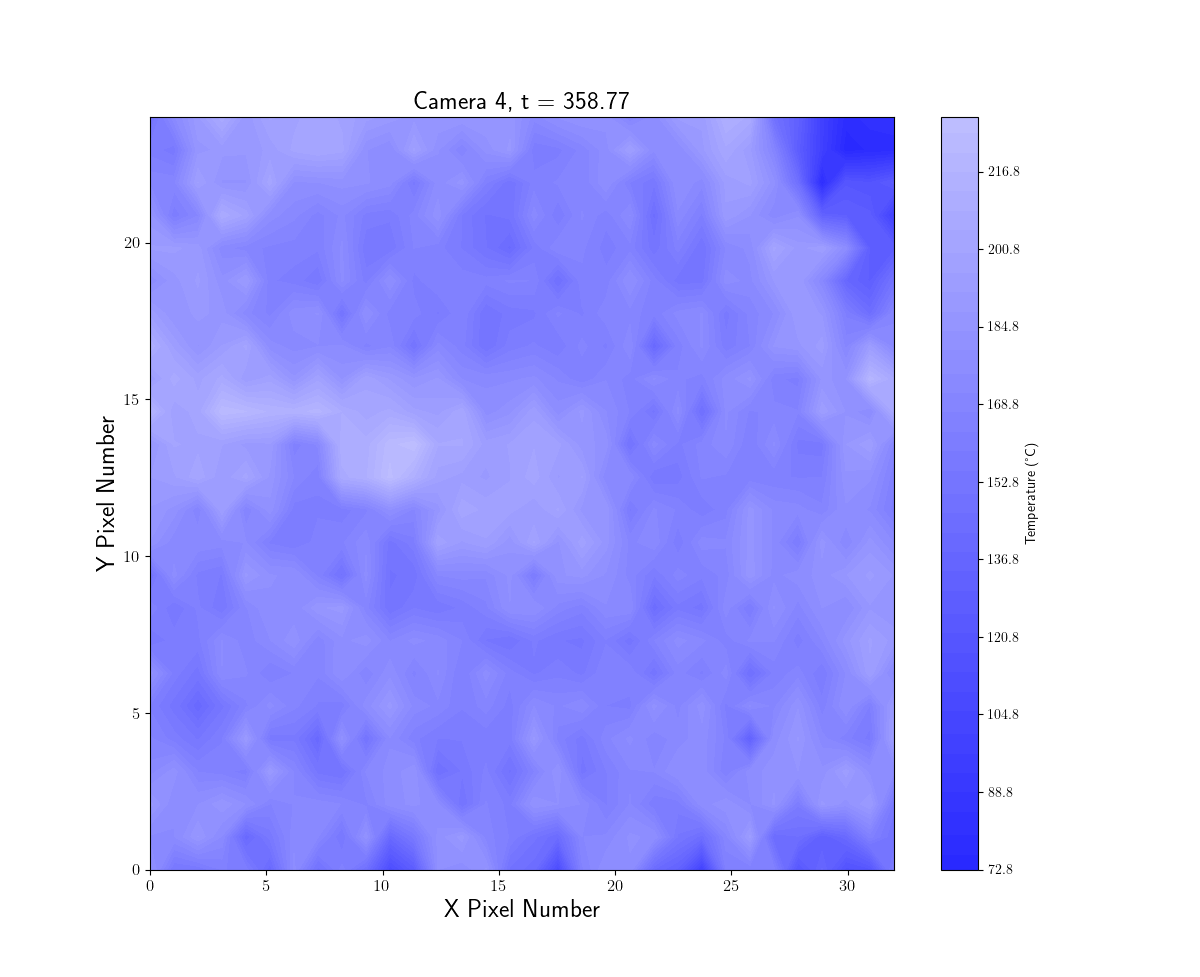}
        \caption{Camera 4.}
        \label{fig:parallel_heat_experiment_camera4_idx7500}
    \end{subfigure}
    \caption{Frame visualisations at time sample 7500.}
    \label{fig:parallel_heat_diff_idx7500}
\end{figure}

\begin{figure}[ht]
    \begin{subfigure}{0.49\textwidth}
        \includegraphics[width=\textwidth]{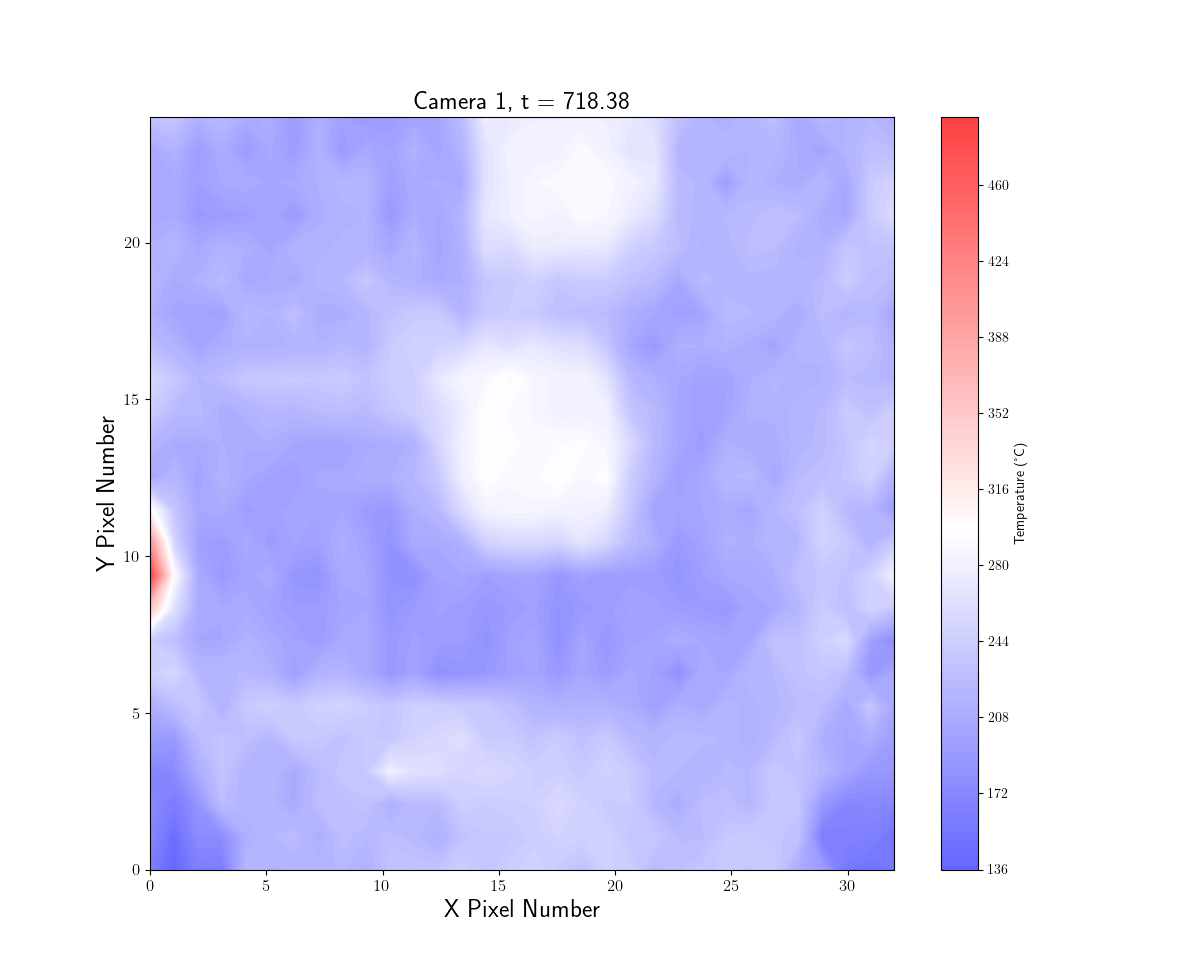}
        \caption{Camera 1.}
        \label{fig:parallel_heat_experiment_camera1_idx15000}
    \end{subfigure}
    \begin{subfigure}{0.49\textwidth}
        \includegraphics[width=\textwidth]{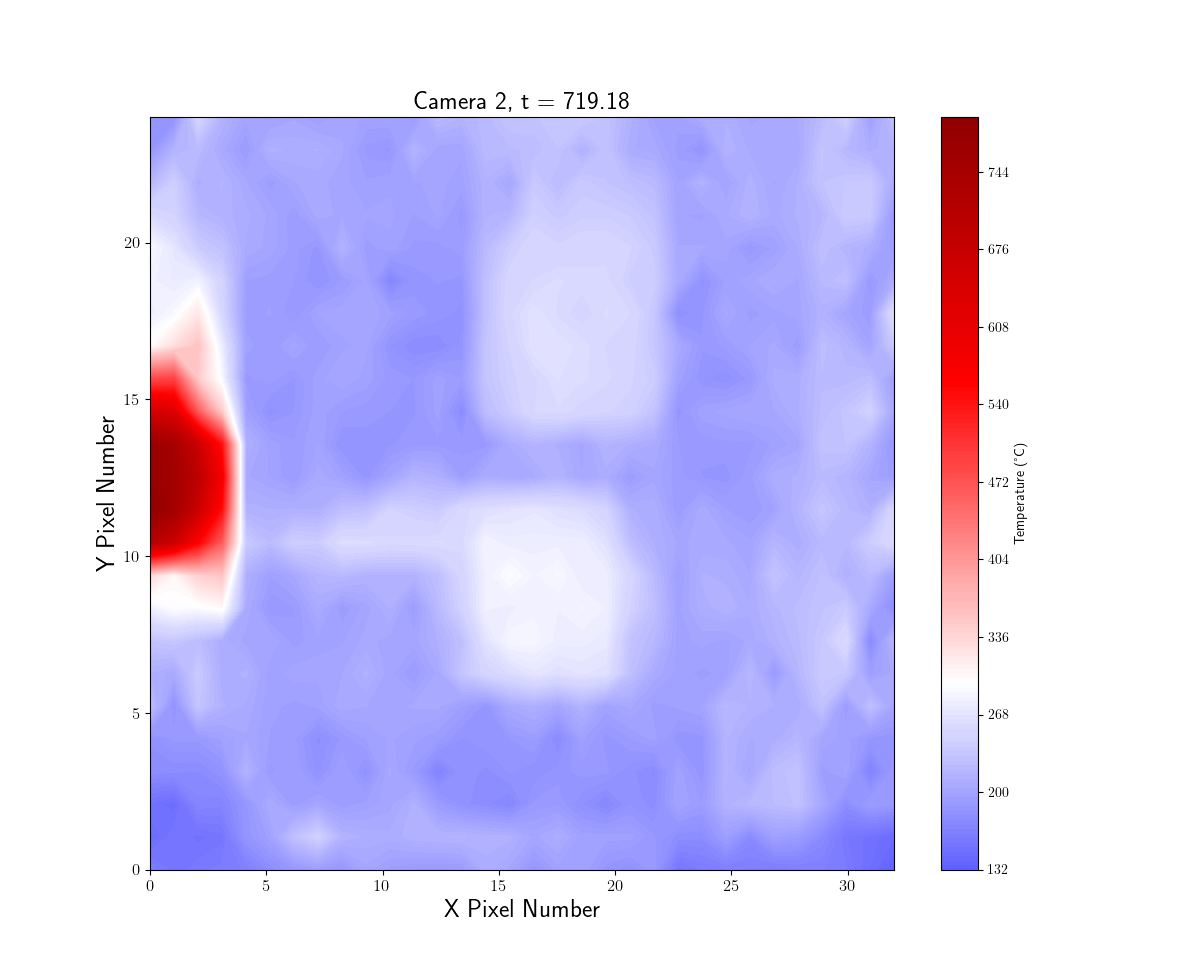}
        \caption{Camera 2.}
        \label{fig:parallel_heat_experiment_camera2_idx15000}
    \end{subfigure}

    \begin{subfigure}{0.49\textwidth}
        \includegraphics[width=\textwidth]{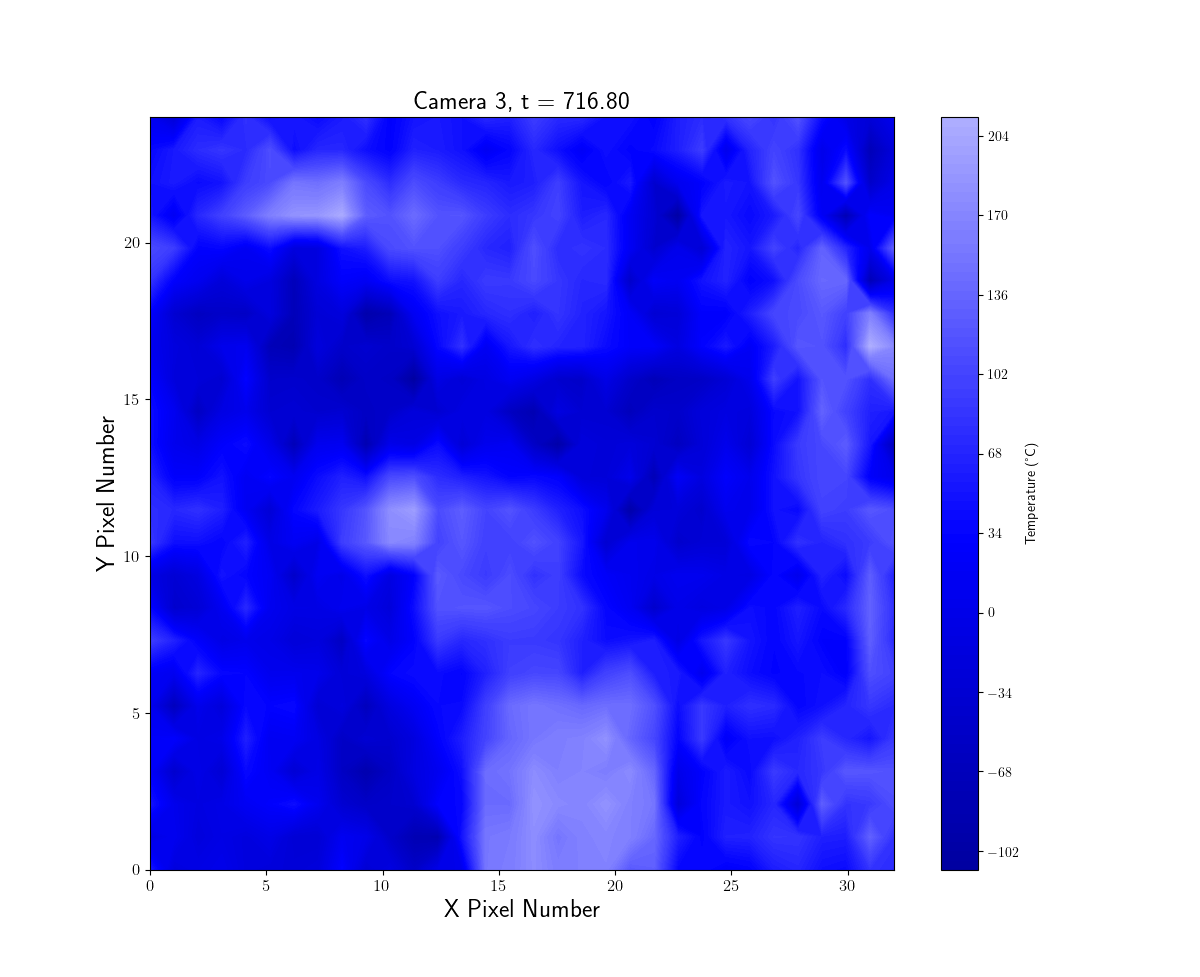}
        \caption{Camera 3.}
        \label{fig:parallel_heat_experiment_camera3_idx15000}
    \end{subfigure}
    \begin{subfigure}{0.49\textwidth}
        \includegraphics[width=\textwidth]{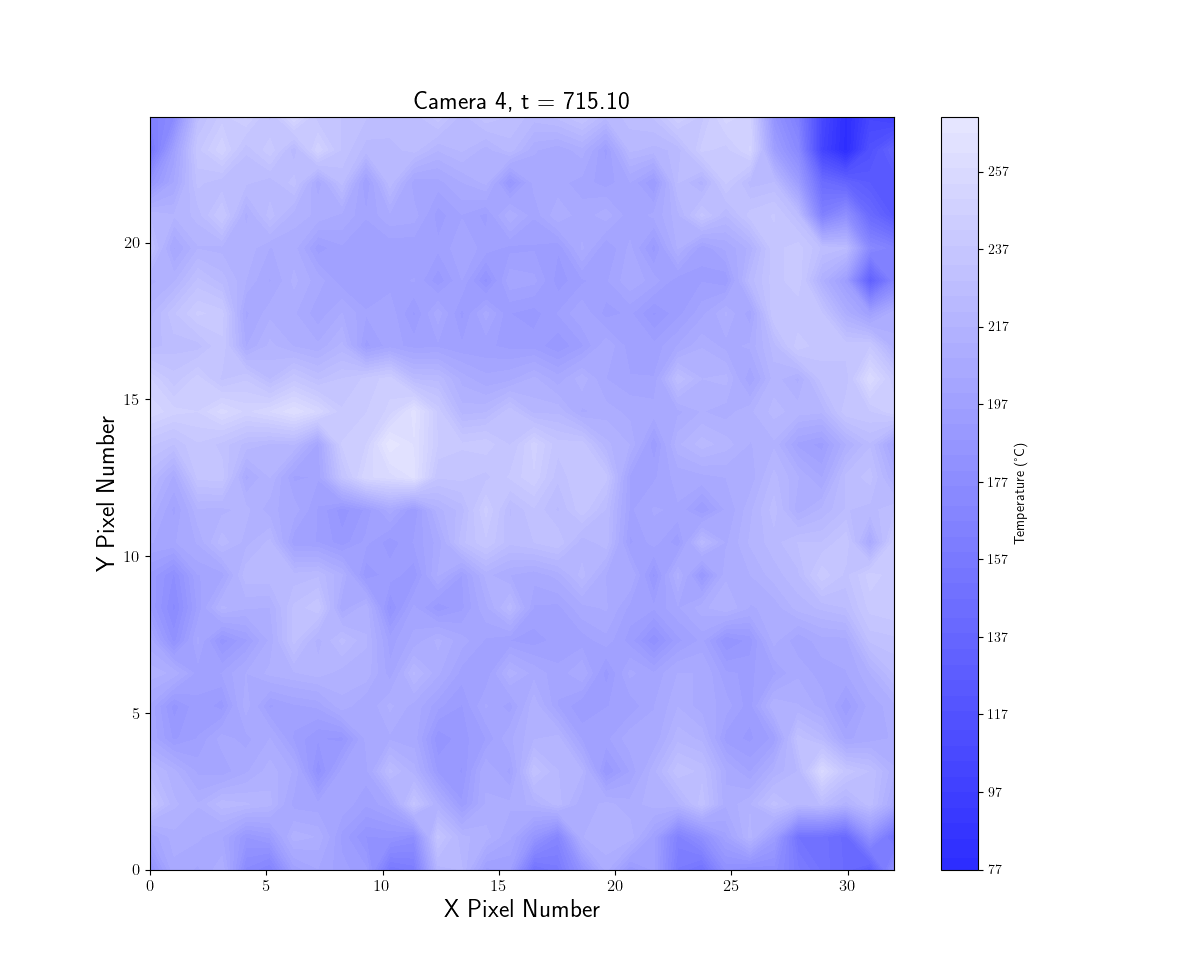}
        \caption{Camera 4.}
        \label{fig:parallel_heat_experiment_camera4_idx15000}
    \end{subfigure}
    \caption{Frame visualisations at time sample 15000.}
    \label{fig:parallel_heat_diff_idx15000}
\end{figure}

From right to left in Figure~\ref{fig:parallel_heating_exp_side_view}, the cameras are at angles 49.0°, 71.5°, 90.0°, 114.0°, and 144.0°. Since the vertical camera (third camera) did not work, we will call the next two cameras, from right to left, cameras 3 and 4. Figures~\ref{fig:parallel_heat_diff_idx500},~\ref{fig:parallel_heat_diff_idx3250},~\ref{fig:parallel_heat_diff_idx7500}, and~\ref{fig:parallel_heat_diff_idx15000} show the raw frames obtained from the four cameras at different time samples. These samples are delayed in time for the different cameras based on the sample differences that Figure~\ref{fig:time_diff_plot} shows. One of the first things we notice from the camera data, is the spatial misalignment of the frame views between the different cameras. Camera 1's view is positioned slightly to the right of camera 2's, as we can see from the solder end which appears in frame. Cameras 3 and 4 on the other hand do not have the solder end in the frame. We also have reason to believe that camera 3's lens is not at the same orientation as that of the rest of the cameras, despite it being positioned in the correct way in the experimental setup. This is because the hot part resulting from the solder is seen appearing vertically between $X$ pixels 15 and 20 (see Figure~\ref{fig:parallel_heat_experiment_camera3_idx15000}), rather than horizontally similar to the rest of the cameras. Also, camera 3 might be faulty since its recorded temperatures are not in the same range as the other cameras (almost 50 -- 100°C less).

In an ideal case where the cameras are aligned in time, we would align the data in space so that the frames from each camera represent the same area. We would do this by shifting their perspectives so they all point vertically downwards. Since the data is not aligned in time nor in space, it would not make sense to do this so we instead compare the data from cameras 1 and 2 from  $t=0 \, s$ to $t=600 \,$ since these cameras are closest in space, and have the least time delay between them up to 600 seconds.

\begin{figure}[ht]
    \centering
    \includegraphics[width=\textwidth]{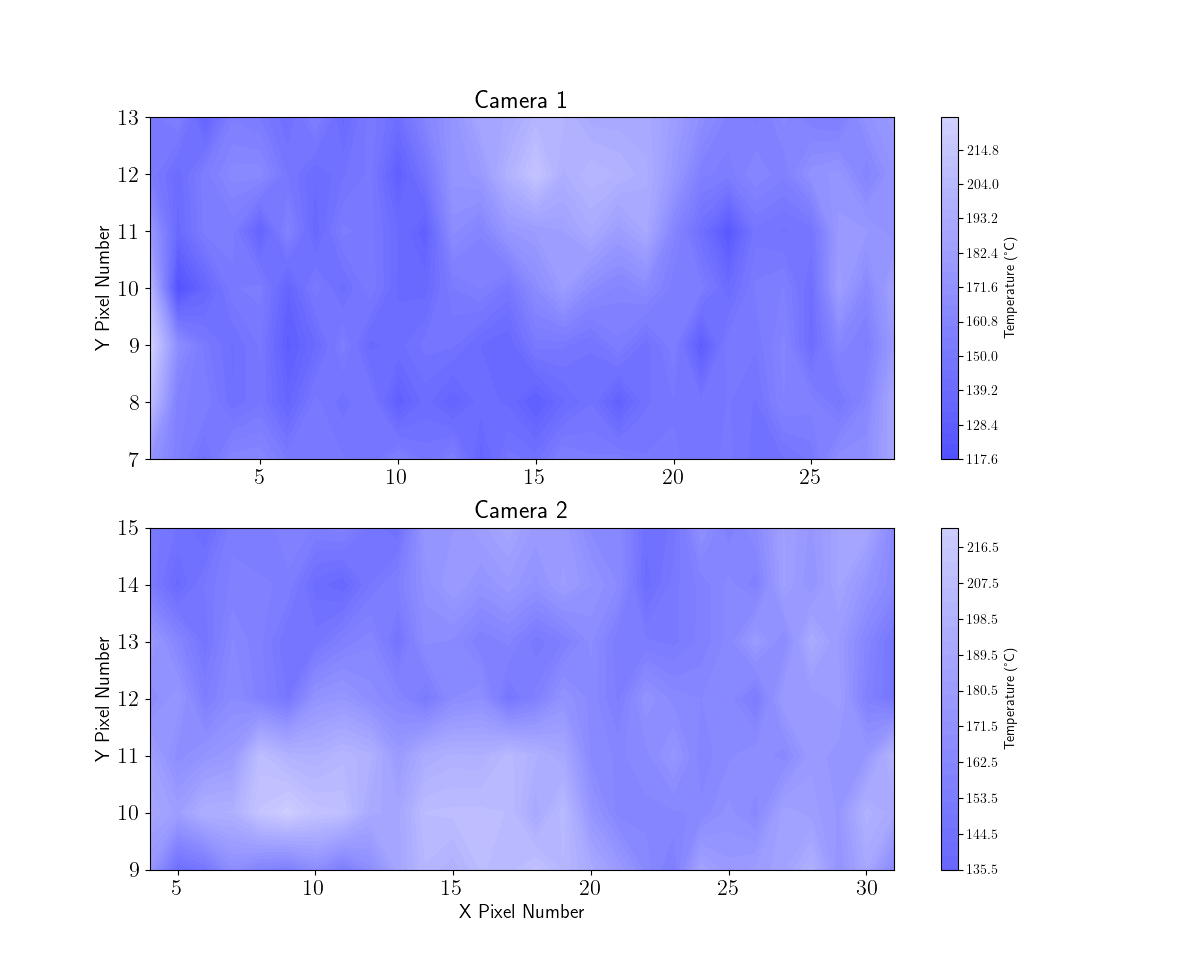}
    \caption{A comparison of the rectangular patch which we attribute as focusing in on the same area between the two cameras. The temperature ranges and the temperature grid are visually similar.}
\label{fig:rectangular_patch_comparison}
\end{figure}

\begin{figure}
    \centering
    \begin{subfigure}{0.49\textwidth}
    \centering
    \includegraphics[width=\textwidth]{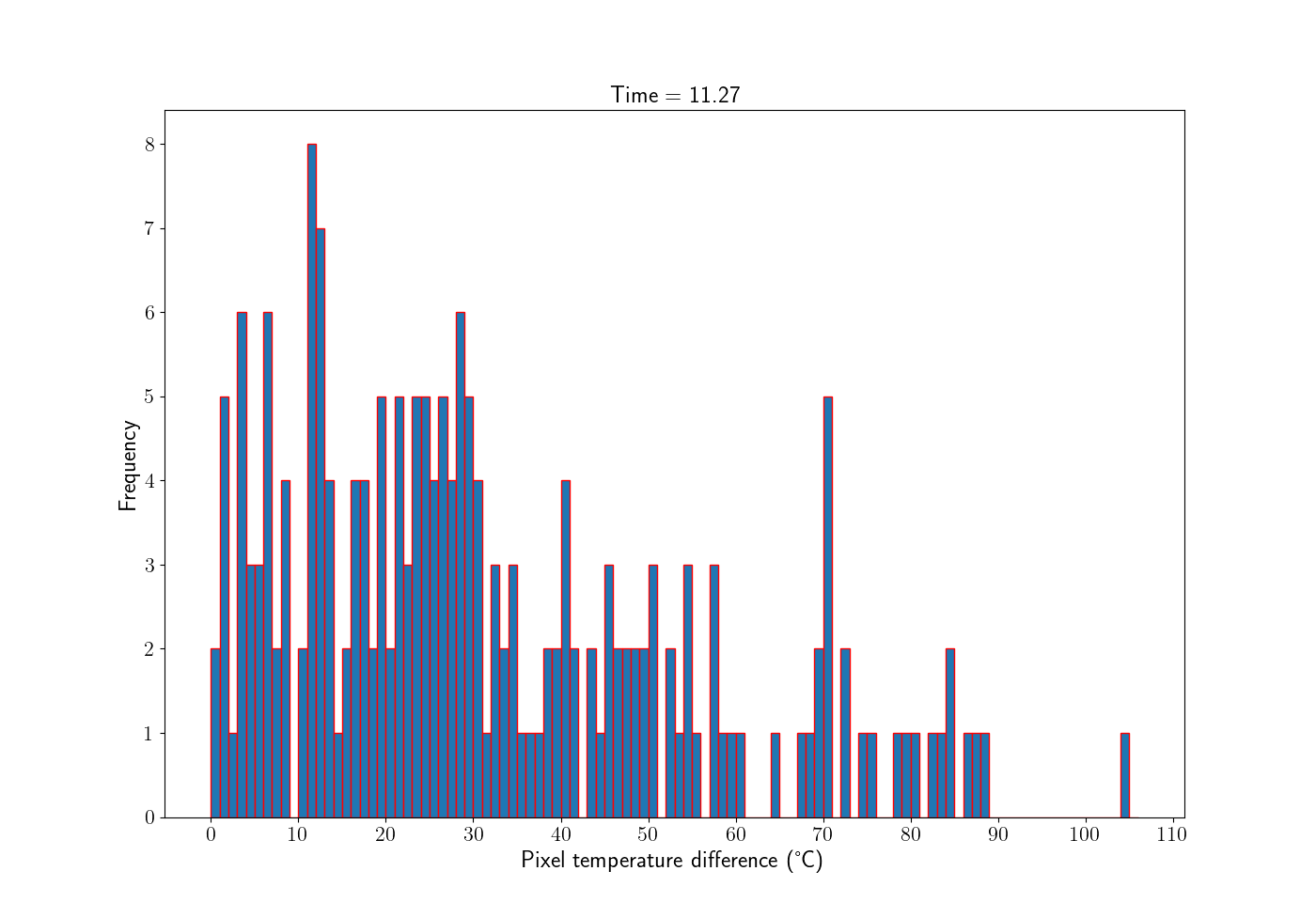}
    \caption{Time = 11.27 s.}
    \label{fig:hist230}
    \end{subfigure}
    \begin{subfigure}{0.49\textwidth}
    \centering
    \includegraphics[width=\textwidth]{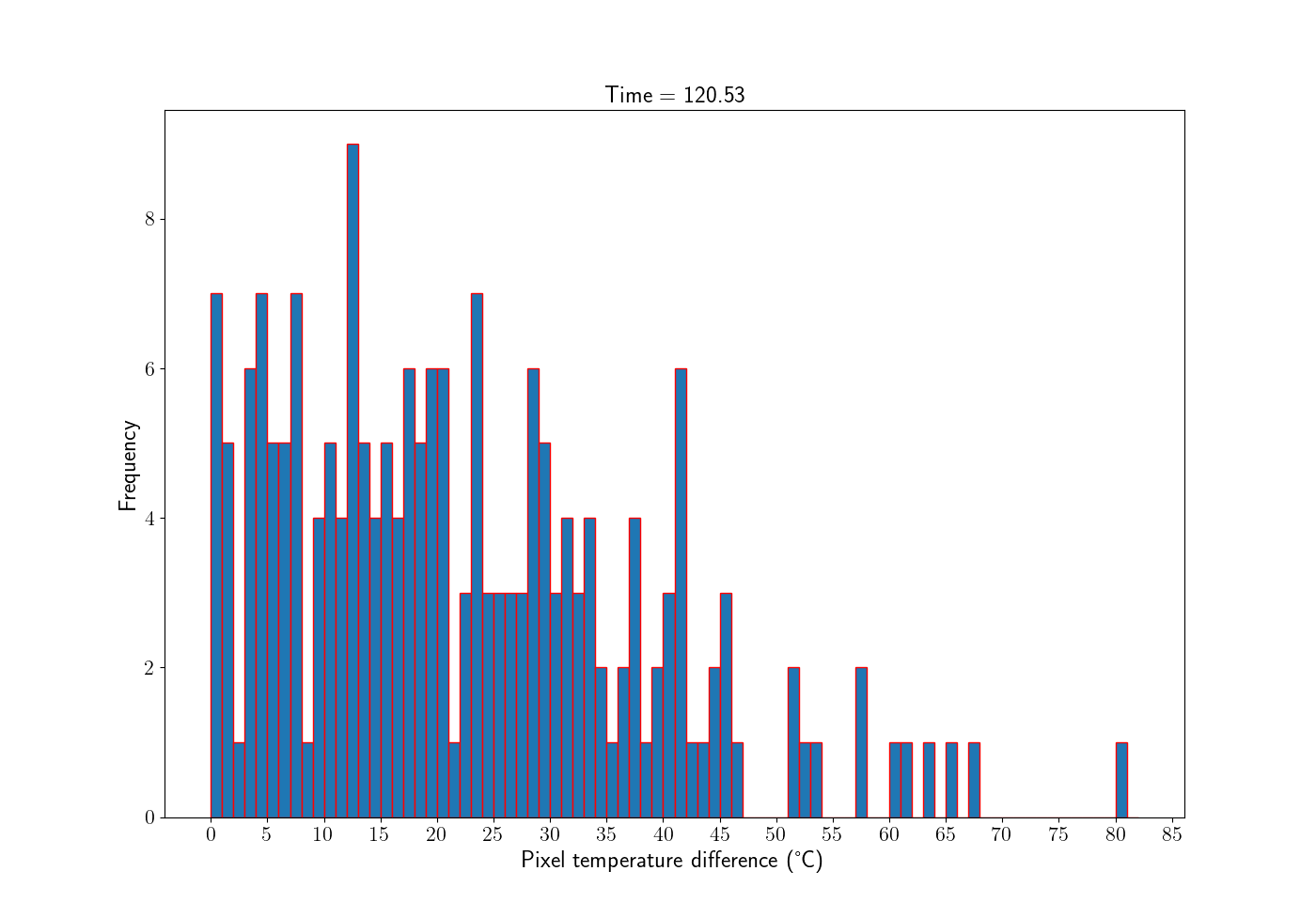}
    \caption{Time = 120.53 s.}
    \label{fig:hist2500}
    \end{subfigure}

    \begin{subfigure}{0.49\textwidth}
    \centering
    \includegraphics[width=\textwidth]{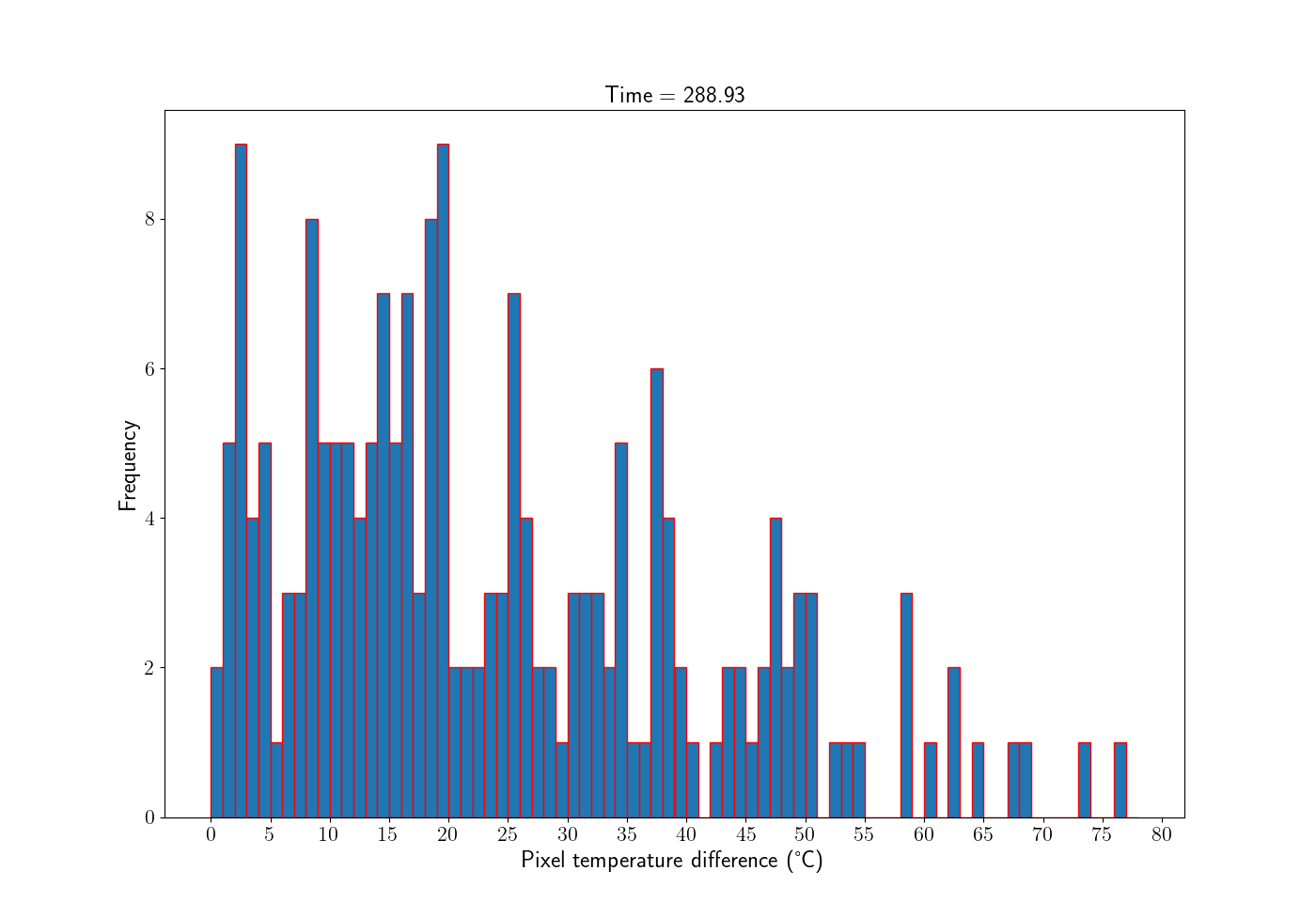}
    \caption{Time = 288.93 s.}
    \label{fig:hist6000}
    \end{subfigure}
    \begin{subfigure}{0.49\textwidth}
    \centering
    \includegraphics[width=\textwidth]{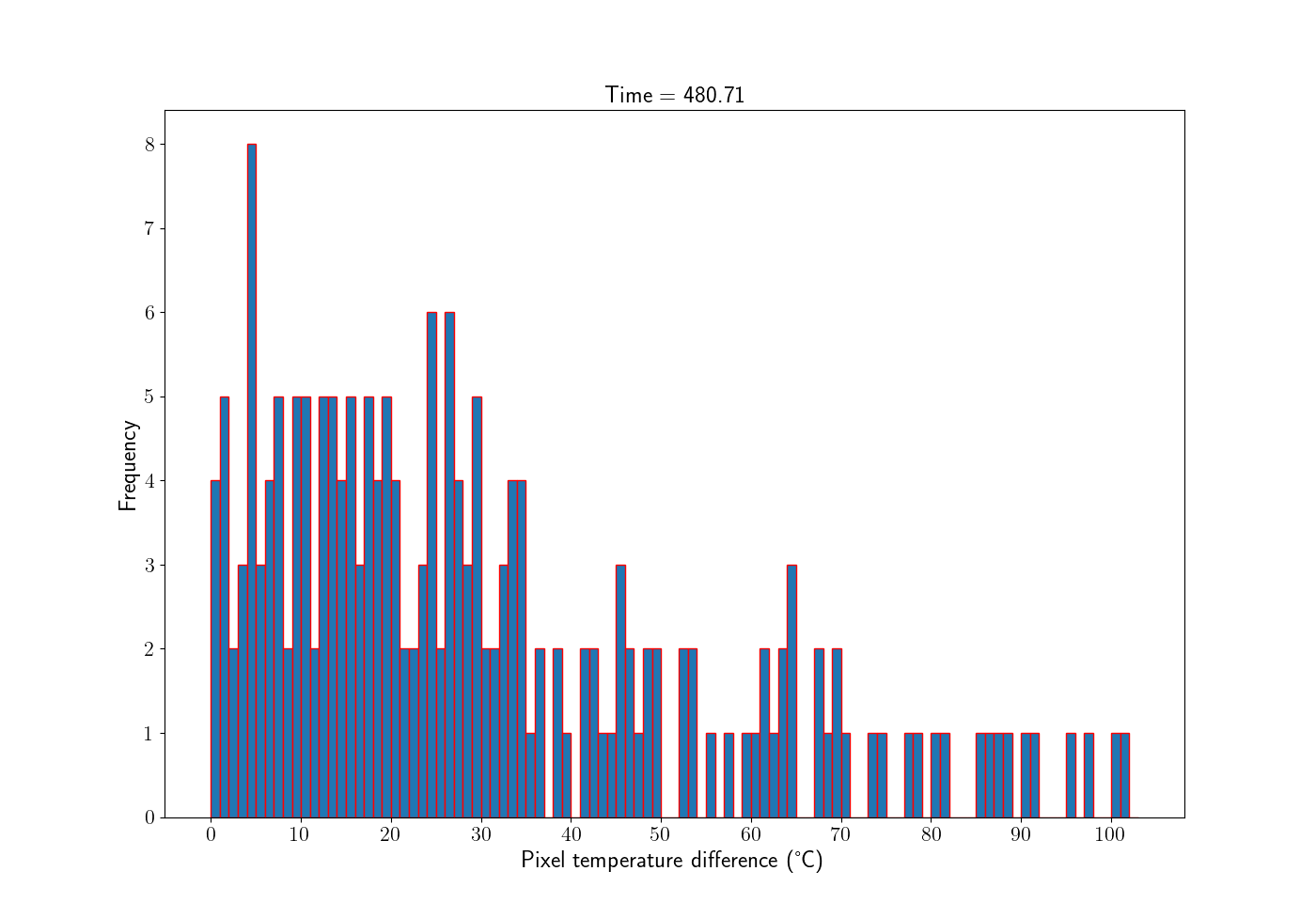}
    \caption{Time = 480.71 s.}
    \label{fig:hist10000}
    \end{subfigure}

    \begin{subfigure}{0.49\textwidth}
    \centering
    \includegraphics[width=\textwidth]{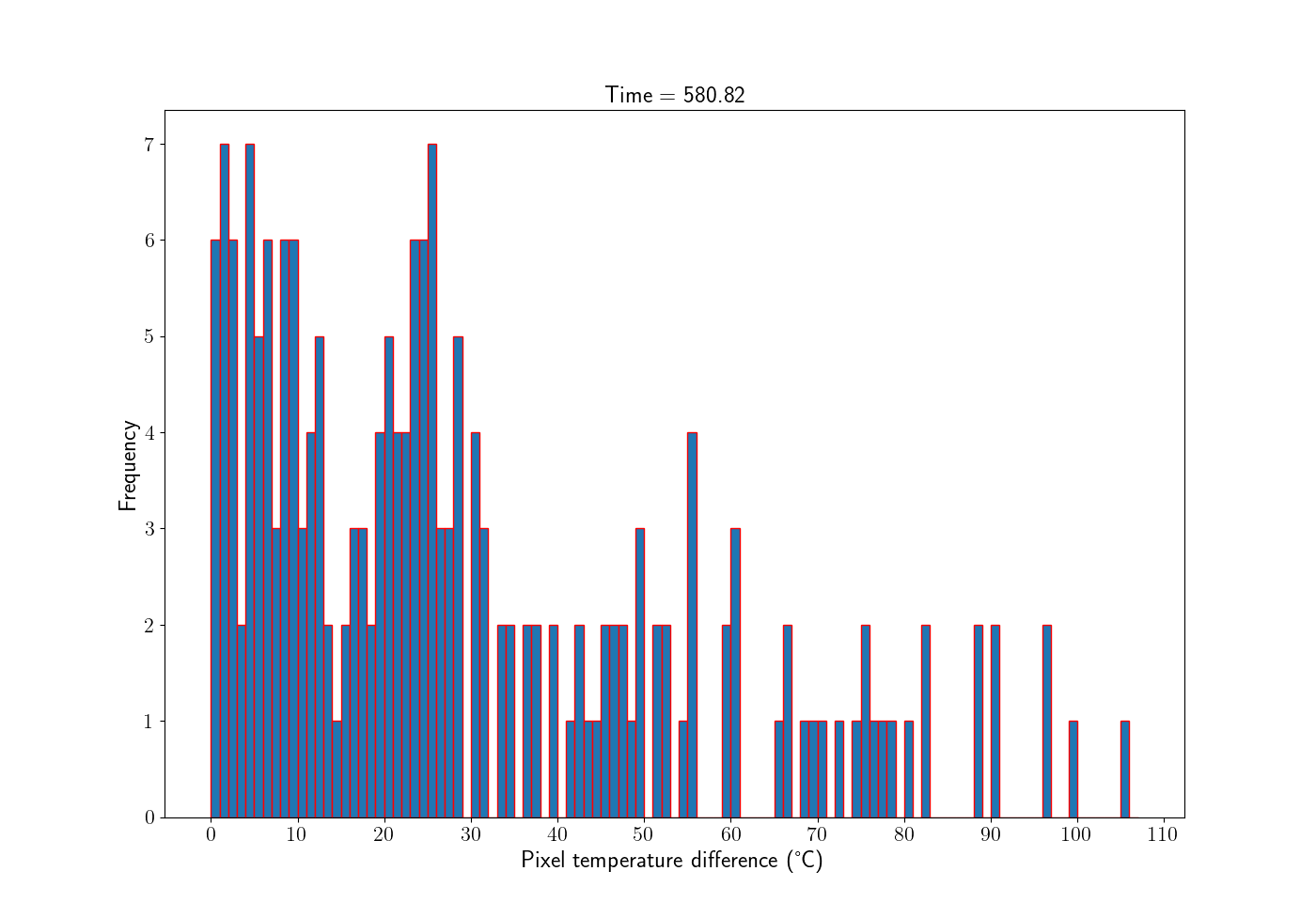}
    \caption{Time = 580.82 s.}
    \label{fig:hist12100}
    \end{subfigure}
    \caption{Histogram plots for the temperature difference between the pixels in the rectangular patch which corresponds to the same area in two different cameras (1 and 2). We can see that, even though a significant number of the differences are near 0, the majority of them are to the right of the graph and have large errors.}
    \label{fig:hist_diff}
\end{figure}

We compare the rectangular patch just after the solder end for the two cameras. For camera 1, the patch has corners with pixel coordinates $(1, \, 7), (29, \, 7), (1, \, 14), (29, \, 14)$. For camera 2, the corner coordinates are $(4, \, 9), (32, \, 9), (4, \, 16), (32, \, 16)$. Figure~\ref{fig:rectangular_patch_comparison} shows a comparison of the patch for the two cameras at $t = 240.85 \, s$. Figure~\ref{fig:hist_diff} shows histograms for the differences between the pixels within the rectangular patch at five instances in time. The expectation was that the temperature differences would be near 0 for the early part of the heating process, and then they would increase due to the gradual loss of time coherence. The histogram plots show that this is not the case, as there are large temperature differences that are consistent throughout the heating process. 

\section{Closing Remarks}

This chapter studied issues related to coherent sensing in both time and space. To do this we designed an experiment where the aim was to sample from 5 different thermal cameras in parallel and compare their data. The data turned out to be incoherent in both time and space, and we show that this incoherence is not easily amendable.

For the time incoherence, different cameras have different levels of delays between them. Additionally, the sampling rates turned out to be inconsistent across the heating duration for any given camera. In repeating this experiment a fixed delay would need to be introduced between each consecutive sample, and the samples would need to be aligned based on the absolute time reference provided by the \verb|time.time()| function.

For the space incoherence, the cameras were aligned at different angles with slight deviations in their horizontal and vertical offsets. This has affected which areas of the block they were able to capture, and to some extent the range of the temperatures that were captured. A structured setup based on FOV calculations would help in reducing errors caused by spatial alignment. Additionally, the large deviations in the temperature ranges that were captured by the cameras could be resolved by calibrating each of them.

%% file: Chapter6/chapter6.tex
\chapter{Discussion and Future Work}

\label{chapter:discussion}

\ifpdf
    \graphicspath{{Chapter6/Figs/Raster/}{Chapter6/Figs/PDF/}{Chapter6/Figs/}}
\else
    \graphicspath{{Chapter6/Figs/Vector/}{Chapter6/Figs/}}
\fi

\section{Discussion}

In Section~\ref{sec:intro}, we introduced the central research questions that motivated our work in this dissertation. The first question assessed the viability of physics-informed models as predictive bases for experimental data captured from real-world systems. Chapters~\ref{chapter:pendulum} and~\ref{chapter:heat_diffusion} addressed this, with a study of the performance of PINNs on two different physical systems: a simple 1D nonlinear pendulum, and a more complicated 2D heat diffusion system. 

For the pendulum many of the training cases, for both the idealized system and the experimental data, have proven that PINNs outperform standard NNs when it comes to regularizing differential equation solutions for sparse, noisy, and low-density data regions. This puts forward a strong case for encoding known information about system dynamics in deep learning as opposed to treating NNs as black boxes, especially for experimental data. The nature of experimental data is that it is dominated by aleatoric and epistemic uncertainty, and using PINNs or other techniques that encode physics information could be a promising strategy for taking these uncertainties into account. 

Unfortunately, training for the heat diffusion system did not fare as well as it did for the pendulum system. This was the case for both the NN and PINN cases. The reason for this ties back to the explanation in Section~\ref{sec:large_domain_training_pend} --- the optimization problem is too difficult to solve on the entire domain of the heating process. This is especially the case for the PINN, since the loss landscape is more difficult to traverse with the second-order derivative terms. Table~\ref{tab:frame_size_variation} supports this point, since the RMSE values improve once the spatial domain is reduced. A reduction in the temporal domain may also be necessary. Additionally, it may be the case that the data collected from the experiment is not adherent to the physics to a sufficient degree. This would explain the PINN frame predictions in Figure~\ref{fig:heat_diff_pred_comparison_visualised}, where the model predicts near constant temperatures throughout entire frames for specific time instances. It may be valuable to study the thermal and sensor noise models better to find a way to incorporate them into the training.

The second question focused on the feasibility of deployment of physics-informed models in physical setups, and the issues that might be faced in attempting to do so. To this end we used inexpensive sensors for our experiments and an FPGA as our embedded platform, and presented a review of issues related to sensor time coherence and spatial alignment in Chapter~\ref{chapter:hardware}.

The hardware design shown in Figure~\ref{fig:AXI_IIC_block_design} uses a single AXI IIC block with a 1000 KHz I2C clock frequency. The experiment in Section~\ref{sec:parallel_heat_exp} uses this design in the 5 FPGAs to sense in parallel. While this may be a reasonable setup for independant parallel sensing, as shown in Section~\ref{sec:data_alignment} it fails to retain time coherence in practise. A better approach that retains it, is to use an FPGA design with 5 AXI IIC blocks on a single board with the same I2C clock. The FPGA design should include a double buffer BRAM, where in one clock cycle the first BRAM stores the sensor data and the second BRAM outputs the data through the AXI interface. In the next cycle the BRAM roles are reversed --- the AXI interface receives the previously stored data from the first BRAM and the second BRAM stores the sensor data. This approach ensures that time coherence would be retained.

Spatial data alignment is a more complicated issue to tackle. The first step is to ensure that all of the cameras are calibrated properly so that they measure temperatures accurately and within the same ranges. After that, the cameras should be positioned correctly based on proper alignment of their FOVs with the block's surface. Once this is done, post-processing frame transformations will be required to ensure that the camera views all point vertically downwards on the surface of the camera. These steps will ensure that the pixels from each camera are aligned with each other, after which their comparison will be valid.

\section{Outlook}

\label{sec:outlook}

\begin{figure}
    \centering
    \includegraphics[width=0.9\textwidth]{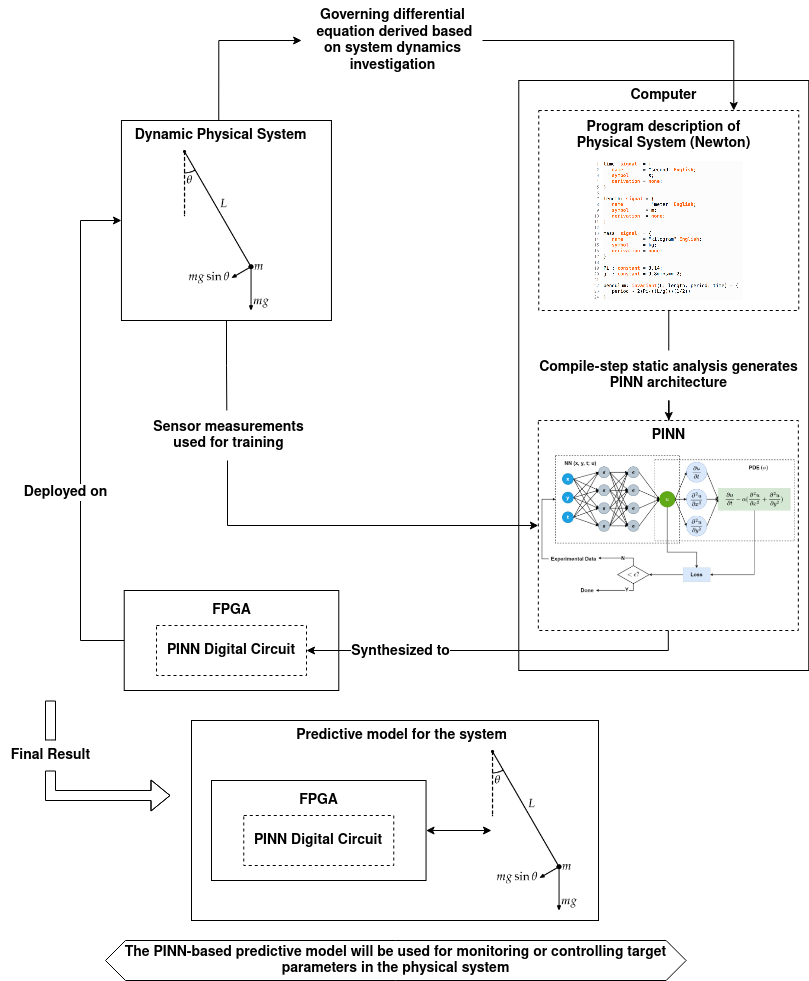}
    \caption{Architectural diagram for our proposed system. A user encodes the differential equation for a system using a description language such as Newton~\cite{lim2018}. A back-end compiler performs static analysis on the Newton description to generate a PINN architecture, which can be trained offline using experimental measurements taken from the system. The trained PINN can then be synthesized onto an FPGA using high-level synthesis (HLS) tools. Finally, the user can then integrate the FPGA with the synthesized model into the system for real-time inference, and by extension control.}
    \label{fig:mphil_outline_diagram}
\end{figure}

The motivation behind this work is to develop methods for encoding dynamics for the deployment of robust physics-aware models within real physical systems, for real-time inference and by extension model predictive control (MPC). Figure~\ref{fig:mphil_outline_diagram} shows an architectural diagram for the system that our methods work towards. ML models are being deployed in a wide variety of modern technologies such as autonomous vehicles~\cite{Schwarting2018}, biosensing~\cite{moin2021}, patient health monitoring~\cite{davoudi2019}, and smart manufacturing~\cite{Ahmad2021}. Most of the models used in these applications are context-agnostic and are unaware of the dynamics of the environments that they exist within. The development of the systems that follow the architecture in Figure~\ref{fig:mphil_outline_diagram} would introduce domain knowledge into models that make safety-critical decisions under uncertain and changing environments. This would increase their robustness and adaptability and would also give us confidence in their decisions, since we know that they are based on widely understood physical principles. Extending domain-specific languages such as Newton~\cite{lim2018} with features for encoding dynamics, coupled with the reconfigurability of FPGAs, would enable generalization of deployment across a wide variety of physical systems with different dynamical behaviours and with adaptive computer architectures. 

\section{Future Work}

The work presented in this dissertation is the initial step towards the development of the proposed system discussed in Section~\ref{sec:outlook}. In this section we list the future work to be performed based on the insights gained and issues faced from the work of this dissertation.

\subsection{Resolve optimization difficulties}

The trained networks for the heat diffusion system did not perform to a satisfactory degree. As mentioned previously, this is due to the difficulties faced with predicting 2D temperature frames across a large time domain. One method that shows promise is to use FBPINNs proposed by Moseley et al.~\cite{moseley2021finite}, or similar domain decomposition machine learning techniques. This would simplify the loss landscape by constraining it to small sub-domains, which are much easier to traverse and find minima within.

Additionally it would be valuable to run a numerical simulation for the heating regime, similar to Section~\ref{sec:ideal_pendulum}, and to train using it. This would provide us with a benchmark so that we can see how much the network is being impacted by the amount of noise in the system.

\subsection{Resolve coherent sensing issues}

Section~\ref{sec:data_alignment} showed the difficulties faced in attempting to sample sensor data coherently in time, and aligning that data in space. In repeating the parallel heating experiment, extra care will be taken to ensure that the sensors are working, calibrated, and operating within the same ranges. The sensing regime will make better use of the absolute time reference by beginning the experiment and heating after a specific time instance, and by sampling after a fixed time delay that is consistent between all of the sensors.

It may also be worthwhile to repeat the experiment with different thermal cameras --- ones that are less susceptible to noise. Possible candidates for this include the AMG8833~\cite{Panasonic:AMG8833} and the FLIR Lepton~\cite{FLIR:Lepton}.

\subsection{Study alternative PIML approaches}

This dissertation focused on PINNs as the candidate architecture for encoding differential equations, however other models might also be promising. These include neural ODEs~\cite{Chen2018}, deep operator networks~\cite{Lu2021}, or physics-constrained Gaussian processes~\cite{Raissi2018}. A thorough investigation into the application of different models would result in a deeper understanding of the interplay between physics and machine learning for real physical systems.

\subsection{Investigate deployment}

An idea that is central to the themes of this dissertation is the deployment of physics-informed model in real systems. As discussed in Section~\ref{sec:FINN}, the FINN~\cite{Blott2018} framework provides a workflow for quantized NNs by exploring the design space of network parameter precision, and deploying accelerator designs using the FINN compiler. One of the major advantages of FINN is that it supports NN implementations on the PYNQ-Z1 board. A prototype version of one of the PINNs that we trained in Chapters~\ref{chapter:pendulum} and~\ref{chapter:heat_diffusion} can be deployed as a proof-of-concept for PINN implementations in hardware. For the pendulum system, the issue is that currently there is no support for a quantized sine activation function. For the heat diffusion system, a sufficiently accurate model must be trained first.

One other issue with deploying PINNs is that a specific PINN is trained only on a specified set of initial and boundary conditions. A new instance of a PINN would have to be trained for a different set of conditions, so a solution must be found that accounts for a wide variety of initial and boundary conditions in a real-world setting.

\subsection{Extend Newton with dynamics constructs}

Newton~\cite{lim2018} currently supports constructs for defining signals, units, and invariant relationships between signals. Extending it with constructs for defining governing equations for dynamics would be a step forward towards the generalized deployment of physics-aware models within hardware.

\subsection{Implement MPC}

The advent of physics-informed MPC is an attractive prospect, especially in real-time, safety-critical or high-uncertainty systems. Arnold and King have shown the possibility of using PINNs for MPC for a simulated application based on the Burgers equation~\cite{Arnold2021}, however further work is required to verify the applicability of these methods for actuation within a real system.

%% file: Chapter7/chapter7.tex
\chapter{Conclusion}

\label{chapter:conclusion}

\ifpdf
    \graphicspath{{Chapter7/Figs/Raster/}{Chapter7/Figs/PDF/}{Chapter7/Figs/}}
\else
    \graphicspath{{Chapter7/Figs/Vector/}{Chapter7/Figs/}}
\fi

This dissertation has investigated two central motivating questions. The first is whether or not the encoding of differential equations in machine learning improves predictive performance for data collected from real physical systems. The second relates to the viability of deploying physics-informed models within physical systems for real-time inference. To answer the first question, we studied the performance of physics-informed neural networks for two different systems: a simple nonlinear pendulum, and 2D heat diffusion across the surface of a metal block.

For the first system, we found that the inclusion of the physics loss term based on the system's governing equation, helped in regularizing the solution according to the underlying physics. This resulted in accurate predictions of the exact solution for both the ideal numerical solution and for the experimental data, for very few training points. In the best case, the PINN achieved an $18 \times$ accuracy improvement over an equivalent NN for 10 linearly-spaced training points for the ideal data, and an over $6 \times$ improvement for 10 uniformly-distributed random points. For the real data case, the PINN achieved accuracy improvements of $9.3 \times$ and $9.1 \times$ for 67 linearly-spaced and uniformly-distributed random points respectively. This proves the predictive performance benefits of encoding known physics into neural networks for both ideal and real data cases for a simple pendulum system.

For the heat diffusion system, we addressed challenges related to denoising thermal camera data and simplifying the optimization for a complex 2D system. We have shown that data denoising, frame size reduction, and optimization using LBFGS are ways to improve the accuracy of network predictions. The PINN and NN showed similar RMSE values, and we were unable to obtain satisfactory accuracy for both despite the improvements. This was because of the difficulty in the underlying optimization problem when it exists within a large domain.

To answer the second question, we design an experiment involving 5 thermal cameras to investigate issues related to parallel sensing. We identify two important issues in the context of using the thermal cameras: time coherence and spatial data alignment.

We end the dissertation with a discussion of the results, an outlook based on the motivation, and future work.

%% file: Appendix1/appendix1.tex

%% file: Appendix2/appendix2.tex


%% file: thesis.bbl
\begin{thebibliography}{}

\bibitem[Abadi et~al., 2016]{tensorflow2016}
Abadi, M., Barham, P., Chen, J., Chen, Z., Davis, A., Dean, J., Devin, M.,
  Ghemawat, S., Irving, G., Isard, M., Kudlur, M., Levenberg, J., Monga, R.,
  Moore, S., Murray, D.~G., Steiner, B., Tucker, P., Vasudevan, V., Warden, P.,
  Wicke, M., Yu, Y., and Zheng, X. (2016).
\newblock Tensorflow: A system for large-scale machine learning.
\newblock In {\em 12th USENIX Symposium on Operating Systems Design and
  Implementation (OSDI 16)}, pages 265--283.

\bibitem[Ablowitz and Prinari, 2008]{Ablowitz2008NLS}
Ablowitz, M. and Prinari, B. (2008).
\newblock {N}onlinear {S}chrodinger systems: continuous and discrete.
\newblock {\em Scholarpedia}, 3(8):5561.
\newblock revision \#137230.

\bibitem[Abowd et~al., 1999]{Abowd1999}
Abowd, G.~D., Dey, A.~K., Brown, P.~J., Davies, N., Smith, M., and Steggles, P.
  (1999).
\newblock Towards a better understanding of context and context-awareness.
\newblock In Gellersen, H.-W., editor, {\em Handheld and Ubiquitous Computing},
  pages 304--307, Berlin, Heidelberg. Springer Berlin Heidelberg.

\bibitem[Abramowitz and Stegun, 1948]{abramowitz1948handbook}
Abramowitz, M. and Stegun, I.~A. (1948).
\newblock {\em Handbook of mathematical functions with formulas, graphs, and
  mathematical tables}, volume~55.
\newblock US Government printing office.

\bibitem[Ahmad and Rahimi, 2022]{Ahmad2021}
Ahmad, H.~M. and Rahimi, A. (2022).
\newblock Deep learning methods for object detection in smart manufacturing: A
  survey.
\newblock {\em Journal of Manufacturing Systems}, 64:181--196.

\bibitem[Arnold and King, 2021]{Arnold2021}
Arnold, F. and King, R. (2021).
\newblock State–space modeling for control based on physics-informed neural
  networks.
\newblock {\em Engineering Applications of Artificial Intelligence},
  101:104195.

\bibitem[Arroyo~Leon et~al., 1999]{Arroyo1999}
Arroyo~Leon, M., Ruiz~Castro, A., and Leal~Ascencio, R. (1999).
\newblock An artificial neural network on a field programmable gate array as a
  virtual sensor.
\newblock In {\em Proceedings of the Third International Workshop on Design of
  Mixed-Mode Integrated Circuits and Applications (Cat. No.99EX303)}, pages
  114--117.

\bibitem[Arzani et~al., 2021]{arzani2021}
Arzani, A., Wang, J.-X., and D'Souza, R.~M. (2021).
\newblock {Uncovering near-wall blood flow from sparse data with
  physics-informed neural networks}.
\newblock {\em Physics of Fluids}, 33(7).
\newblock 071905.

\bibitem[Bade and Hutchings, 1994]{bade1994}
Bade, S. and Hutchings, B. (1994).
\newblock Fpga-based stochastic neural networks-implementation.
\newblock In {\em Proceedings of IEEE Workshop on FPGA's for Custom Computing
  Machines}, pages 189--198.

\bibitem[Bai et~al., 2019]{bai2019}
Bai, J., Lu, F., Zhang, K., et~al. (2019).
\newblock Onnx: Open neural network exchange.
\newblock \url{https://github.com/onnx/onnx}.

\bibitem[Baker et~al., 2019]{baker_workshop2019}
Baker, N., Alexander, F., Bremer, T., Hagberg, A., Kevrekidis, Y., Najm, H.,
  Parashar, M., Patra, A., Sethian, J., Wild, S., Willcox, K., and Lee, S.
  (2019).
\newblock Workshop report on basic research needs for scientific machine
  learning: Core technologies for artificial intelligence.
\newblock {\em U.S. Department of Energy Office of Scientific and Technical
  Information}.

\bibitem[Bel{\'e}ndez et~al., 2007]{belendez2007exact}
Bel{\'e}ndez, A., Pascual, C., M{\'e}ndez, D., Bel{\'e}ndez, T., and Neipp, C.
  (2007).
\newblock Exact solution for the nonlinear pendulum.
\newblock {\em Revista brasileira de ensino de f{\'\i}sica}, 29:645--648.

\bibitem[Bhustali, 2021]{Bhustali2021}
Bhustali, P. (2021).
\newblock Physics-informed-neural-networks.
\newblock
  \url{https://github.com/omniscientoctopus/Physics-Informed-Neural-Networks}.

\bibitem[{BIPM}, 2019]{si-brochure}
{BIPM} (2019).
\newblock {\em Le Syst\`eme international d'unit\'es / The International System
  of Units (`The {SI} Brochure')}.
\newblock Bureau international des poids et mesures, ninth edition.

\bibitem[Bishop, 2006]{Bishop2006}
Bishop, C.~M. (2006).
\newblock {\em Pattern Recognition and Machine Learning (Information Science
  and Statistics)}.
\newblock Springer-Verlag, Berlin, Heidelberg.

\bibitem[Blott et~al., 2018]{Blott2018}
Blott, M., Preu\ss{}er, T.~B., Fraser, N.~J., Gambardella, G., O’brien, K.,
  Umuroglu, Y., Leeser, M., and Vissers, K. (2018).
\newblock Finn-r: An end-to-end deep-learning framework for fast exploration of
  quantized neural networks.
\newblock {\em ACM Trans. Reconfigurable Technol. Syst.}, 11(3).

\bibitem[Boyce et~al., 2017]{boyce2017}
Boyce, W., DiPrima, R., and Meade, D. (2017).
\newblock {\em Elementary Differential Equations and Boundary Value Problems}.
\newblock Wiley.

\bibitem[Buckingham, 1914]{buckingham1914}
Buckingham, E. (1914).
\newblock On physically similar systems; illustrations of the use of
  dimensional equations.
\newblock {\em Phys. Rev.}, 4:345--376.

\bibitem[Burgers, 1948]{burgers1948}
Burgers, J. (1948).
\newblock A mathematical model illustrating the theory of turbulence.
\newblock In {Von Mises}, R. and {Von Kármán}, T., editors, {\em Advances in
  Applied Mechanics}, volume~1, pages 171--199. Elsevier.

\bibitem[Cai et~al., 2021a]{cai2021physics}
Cai, S., Mao, Z., Wang, Z., Yin, M., and Karniadakis, G.~E. (2021a).
\newblock Physics-informed neural networks (pinns) for fluid mechanics: A
  review.
\newblock {\em Acta Mechanica Sinica}, 37(12):1727--1738.

\bibitem[Cai et~al., 2021b]{cai2021heat_transfer_pinns}
Cai, S., Wang, Z., Wang, S., Perdikaris, P., and Karniadakis, G.~E. (2021b).
\newblock {Physics-Informed Neural Networks for Heat Transfer Problems}.
\newblock {\em Journal of Heat Transfer}, 143(6):060801.

\bibitem[Chen et~al., 2018]{Chen2018}
Chen, R. T.~Q., Rubanova, Y., Bettencourt, J., and Duvenaud, D.~K. (2018).
\newblock Neural ordinary differential equations.
\newblock In Bengio, S., Wallach, H., Larochelle, H., Grauman, K.,
  Cesa-Bianchi, N., and Garnett, R., editors, {\em Advances in Neural
  Information Processing Systems}, volume~31. Curran Associates, Inc.

\bibitem[Cloutier et~al., 1996]{cloutier1996}
Cloutier, J., Cosatto, E., Pigeon, S., Boyer, F., and Simard, P. (1996).
\newblock Vip: an fpga-based processor for image processing and neural
  networks.
\newblock In {\em Proceedings of Fifth International Conference on
  Microelectronics for Neural Networks}, pages 330--336.

\bibitem[Cox and Blanz, 1992]{cox1992_ganglion}
Cox, C. and Blanz, W. (1992).
\newblock Ganglion-a fast field-programmable gate array implementation of a
  connectionist classifier.
\newblock {\em IEEE Journal of Solid-State Circuits}, 27(3):288--299.

\bibitem[Crank, 1975]{crank1975mathematics_diffusion}
Crank, J. (1975).
\newblock {\em The Mathematics of Diffusion}.
\newblock Oxford science publications. Clarendon Press.

\bibitem[Cromer, 1981]{Cromer1981}
Cromer, A. (1981).
\newblock {Stable solutions using the Euler approximation}.
\newblock {\em American Journal of Physics}, 49(5):455--459.

\bibitem[Cuomo et~al., 2022]{cuomo2022scientific}
Cuomo, S., Di~Cola, V.~S., Giampaolo, F., Rozza, G., Raissi, M., and Piccialli,
  F. (2022).
\newblock Scientific machine learning through physics--informed neural
  networks: where we are and what’s next.
\newblock {\em Journal of Scientific Computing}, 92(3):88.

\bibitem[Dahmen, 2015]{dahmen2015pendulums}
Dahmen, S.~R. (2015).
\newblock On pendulums and air resistance: the mathematics and physics of denis
  diderot.
\newblock {\em The European Physical Journal H}, 40:337--373.

\bibitem[Davoudi et~al., 2019]{davoudi2019}
Davoudi, A., Malhotra, K.~R., Shickel, B., Siegel, S., Williams, S., Ruppert,
  M., Bihorac, E., Ozrazgat-Baslanti, T., Tighe, P.~J., Bihorac, A., et~al.
  (2019).
\newblock Intelligent icu for autonomous patient monitoring using pervasive
  sensing and deep learning.
\newblock {\em Scientific reports}, 9(1):8020.

\bibitem[Denil et~al., 2013]{Denil2013}
Denil, M., Shakibi, B., Dinh, L., Ranzato, M., and de~Freitas, N. (2013).
\newblock Predicting parameters in deep learning.
\newblock In {\em Proceedings of the 26th International Conference on Neural
  Information Processing Systems - Volume 2}, NIPS'13, page 2148–2156, Red
  Hook, NY, USA. Curran Associates Inc.

\bibitem[Denton et~al., 2014]{denton2014}
Denton, E.~L., Zaremba, W., Bruna, J., LeCun, Y., and Fergus, R. (2014).
\newblock Exploiting linear structure within convolutional networks for
  efficient evaluation.
\newblock {\em Advances in neural information processing systems}, 27.

\bibitem[Ding et~al., 2023]{Ding2023}
Ding, Y., Wu, J., Gao, Y., Wang, M., and So, H. K.-H. (2023).
\newblock Model-platform optimized deep neural network accelerator generation
  through mixed-integer geometric programming.
\newblock In {\em 2023 IEEE 31st Annual International Symposium on
  Field-Programmable Custom Computing Machines (FCCM)}, pages 83--93.

\bibitem[Eeckhout, 2017]{eeckhout2017moore}
Eeckhout, L. (2017).
\newblock Is moore’s law slowing down? what’s next?
\newblock {\em IEEE Micro}, 37(04):4--5.

\bibitem[El-Maksoud et~al., 2021]{AbdElkMaksoud2021}
El-Maksoud, A. J.~A., Ebbed, M., Khalil, A.~H., and Mostafa, H. (2021).
\newblock Power efficient design of high-performance convolutional neural
  networks hardware accelerator on fpga: A case study with googlenet.
\newblock {\em IEEE Access}, 9:151897--151911.

\bibitem[Evans, 2010]{evans2010partial}
Evans, L. (2010).
\newblock {\em Partial Differential Equations}.
\newblock Graduate studies in mathematics. American Mathematical Society.

\bibitem[Farlow, 1993a]{farlow1993partial_wave}
Farlow, S. (1993a).
\newblock {\em Partial Differential Equations for Scientists and Engineers}.
\newblock Dover books on advanced mathematics. Dover Publications.

\bibitem[Farlow, 1993b]{farlow1993partial_laplace}
Farlow, S. (1993b).
\newblock {\em Partial Differential Equations for Scientists and Engineers}.
\newblock Dover books on advanced mathematics. Dover Publications.

\bibitem[Ferrer et~al., 2004]{Ferrer2004}
Ferrer, D., Gonzalez, R., Fleitas, R., Acle, J., and Canetti, R. (2004).
\newblock Neurofpga-implementing artificial neural networks on programmable
  logic devices.
\newblock In {\em Proceedings Design, Automation and Test in Europe Conference
  and Exhibition}, volume~3, pages 218--223 Vol.3.

\bibitem[{FLIR}, 2018]{FLIR:Lepton}
{FLIR} (2018).
\newblock {\em FLIR LEPTON® Engineering Datasheet}.

\bibitem[Guan et~al., 2017]{Guan2017}
Guan, Y., Liang, H., Xu, N., Wang, W., Shi, S., Chen, X., Sun, G., Zhang, W.,
  and Cong, J. (2017).
\newblock Fp-dnn: An automated framework for mapping deep neural networks onto
  fpgas with rtl-hls hybrid templates.
\newblock In {\em 2017 IEEE 25th Annual International Symposium on
  Field-Programmable Custom Computing Machines (FCCM)}, pages 152--159.

\bibitem[Guo et~al., 2018]{angel-eye_Guo2018}
Guo, K., Sui, L., Qiu, J., Yu, J., Wang, J., Yao, S., Han, S., Wang, Y., and
  Yang, H. (2018).
\newblock Angel-eye: A complete design flow for mapping cnn onto embedded fpga.
\newblock {\em IEEE Transactions on Computer-Aided Design of Integrated
  Circuits and Systems}, 37(1):35--47.

\bibitem[Guo et~al., 2019]{Guo2019}
Guo, K., Zeng, S., Yu, J., Wang, Y., and Yang, H. (2019).
\newblock [dl] a survey of fpga-based neural network inference accelerators.
\newblock {\em ACM Trans. Reconfigurable Technol. Syst.}, 12(1).

\bibitem[Han et~al., 2015]{Song2015}
Han, S., Pool, J., Tran, J., and Dally, W. (2015).
\newblock Learning both weights and connections for efficient neural network.
\newblock In Cortes, C., Lawrence, N., Lee, D., Sugiyama, M., and Garnett, R.,
  editors, {\em Advances in Neural Information Processing Systems}, volume~28.
  Curran Associates, Inc.

\bibitem[Hao et~al., 2023]{hao2023physicsinformed}
Hao, Z., Liu, S., Zhang, Y., Ying, C., Feng, Y., Su, H., and Zhu, J. (2023).
\newblock Physics-informed machine learning: A survey on problems, methods and
  applications.

\bibitem[Hubara et~al., 2016]{Hubara2016}
Hubara, I., Courbariaux, M., Soudry, D., El-Yaniv, R., and Bengio, Y. (2016).
\newblock Binarized neural networks.
\newblock In Lee, D., Sugiyama, M., Luxburg, U., Guyon, I., and Garnett, R.,
  editors, {\em Advances in Neural Information Processing Systems}, volume~29.
  Curran Associates, Inc.

\bibitem[Jia et~al., 2022]{Jia2022}
Jia, X., Zhang, Y., Liu, G., Yang, X., Zhang, T., Zheng, J., Xu, D., Wang, H.,
  Zheng, R., Pareek, S., Tian, L., Xie, D., Luo, H., and Shan, Y. (2022).
\newblock Xvdpu: A high performance cnn accelerator on the versal platform
  powered by the ai engine.
\newblock In {\em 2022 32nd International Conference on Field-Programmable
  Logic and Applications (FPL)}, pages 01--09.

\bibitem[Jiang et~al., 2022]{jiang2022}
Jiang, X., Wang, D., Fan, Q., Zhang, M., Lu, C., and Lau, A. P.~T. (2022).
\newblock Physics-informed neural network for nonlinear dynamics in fiber
  optics.
\newblock {\em Laser \& Photonics Reviews}, 16(9):2100483.

\bibitem[Jin et~al., 2021]{jin2021}
Jin, X., Cai, S., Li, H., and Karniadakis, G.~E. (2021).
\newblock Nsfnets (navier-stokes flow nets): Physics-informed neural networks
  for the incompressible navier-stokes equations.
\newblock {\em Journal of Computational Physics}, 426:109951.

\bibitem[Jouppi et~al., 2017]{Jouppi2017}
Jouppi, N.~P., Young, C., Patil, N., Patterson, D., Agrawal, G., Bajwa, R.,
  Bates, S., Bhatia, S., Boden, N., Borchers, A., Boyle, R., Cantin, P.-l.,
  Chao, C., Clark, C., Coriell, J., Daley, M., Dau, M., Dean, J., Gelb, B.,
  Ghaemmaghami, T.~V., Gottipati, R., Gulland, W., Hagmann, R., Ho, C.~R.,
  Hogberg, D., Hu, J., Hundt, R., Hurt, D., Ibarz, J., Jaffey, A., Jaworski,
  A., Kaplan, A., Khaitan, H., Killebrew, D., Koch, A., Kumar, N., Lacy, S.,
  Laudon, J., Law, J., Le, D., Leary, C., Liu, Z., Lucke, K., Lundin, A.,
  MacKean, G., Maggiore, A., Mahony, M., Miller, K., Nagarajan, R.,
  Narayanaswami, R., Ni, R., Nix, K., Norrie, T., Omernick, M., Penukonda, N.,
  Phelps, A., Ross, J., Ross, M., Salek, A., Samadiani, E., Severn, C.,
  Sizikov, G., Snelham, M., Souter, J., Steinberg, D., Swing, A., Tan, M.,
  Thorson, G., Tian, B., Toma, H., Tuttle, E., Vasudevan, V., Walter, R., Wang,
  W., Wilcox, E., and Yoon, D.~H. (2017).
\newblock In-datacenter performance analysis of a tensor processing unit.
\newblock {\em SIGARCH Comput. Archit. News}, 45(2):1–12.

\bibitem[Karniadakis et~al., 2021]{karniadakis2021physics}
Karniadakis, G.~E., Kevrekidis, I.~G., Lu, L., Perdikaris, P., Wang, S., and
  Yang, L. (2021).
\newblock Physics-informed machine learning.
\newblock {\em Nature Reviews Physics}, 3(6):422--440.

\bibitem[Khan and Lowther, 2022]{khan2022_EM}
Khan, A. and Lowther, D.~A. (2022).
\newblock Physics informed neural networks for electromagnetic analysis.
\newblock {\em IEEE Transactions on Magnetics}, 58(9):1--4.

\bibitem[Kingma and Ba, 2014]{Kingma2014}
Kingma, D. and Ba, J. (2014).
\newblock Adam: A method for stochastic optimization.
\newblock {\em International Conference on Learning Representations}.

\bibitem[Kingma and Welling, 2022]{kingma2022autoencoding}
Kingma, D.~P. and Welling, M. (2022).
\newblock Auto-encoding variational bayes.

\bibitem[Korteweg and de~Vries, 1895]{Korteweg_de_vries1895}
Korteweg, D. D.~J. and de~Vries, D.~G. (1895).
\newblock Xli. on the change of form of long waves advancing in a rectangular
  canal, and on a new type of long stationary waves.
\newblock {\em The London, Edinburgh, and Dublin Philosophical Magazine and
  Journal of Science}, 39(240):422--443.

\bibitem[Krishnapriyan et~al., 2021]{krishnapriyan2021}
Krishnapriyan, A., Gholami, A., Zhe, S., Kirby, R., and Mahoney, M.~W. (2021).
\newblock Characterizing possible failure modes in physics-informed neural
  networks.
\newblock In Beygelzimer, A., Dauphin, Y., Liang, P., and Vaughan, J.~W.,
  editors, {\em Advances in Neural Information Processing Systems}.

\bibitem[Krizhevsky et~al., 2012]{alexnet2012}
Krizhevsky, A., Sutskever, I., and Hinton, G.~E. (2012).
\newblock Imagenet classification with deep convolutional neural networks.
\newblock In Pereira, F., Burges, C., Bottou, L., and Weinberger, K., editors,
  {\em Advances in Neural Information Processing Systems}, volume~25. Curran
  Associates, Inc.

\bibitem[Kuhn, 2009]{kuhn2009}
Kuhn, K.~J. (2009).
\newblock Cmos scaling beyond 32nm: Challenges and opportunities.
\newblock In {\em Proceedings of the 46th Annual Design Automation Conference},
  DAC '09, page 310–313, New York, NY, USA. Association for Computing
  Machinery.

\bibitem[Lagaris et~al., 1998]{Lagaris1998}
Lagaris, I., Likas, A., and Fotiadis, D. (1998).
\newblock Artificial neural networks for solving ordinary and partial
  differential equations.
\newblock {\em {IEEE} Transactions on Neural Networks}, 9(5):987--1000.

\bibitem[Leoni et~al., 2023]{ClarkDiLeoni2023}
Leoni, P. C.~D., Agarwal, K., Zaki, T.~A., Meneveau, C., and Katz, J. (2023).
\newblock Reconstructing turbulent velocity and pressure fields from
  under-resolved noisy particle tracks using physics-informed neural networks.
\newblock {\em Experiments in Fluids}, 64(5).

\bibitem[Levandosky, 2003]{Levandosky2003}
Levandosky, J. (2003).
\newblock Math 220b lecture notes.
\newblock \url{https://web.stanford.edu/class/math220b/handouts/HEATEQN.pdf}.
\newblock Accessed: 14/08/2023.

\bibitem[Li et~al., 2023]{li2023_LMD_pinn}
Li, S., Wang, G., Di, Y., Wang, L., Wang, H., and Zhou, Q. (2023).
\newblock A physics-informed neural network framework to predict 3d temperature
  field without labeled data in process of laser metal deposition.
\newblock {\em Engineering Applications of Artificial Intelligence},
  120:105908.

\bibitem[Liang et~al., 2018]{Liang2018}
Liang, S., Yin, S., Liu, L., Luk, W., and Wei, S. (2018).
\newblock Fp-bnn: Binarized neural network on fpga.
\newblock {\em Neurocomputing}, 275:1072--1086.

\bibitem[Lienhard and Lienhard, 2019]{ahtt5p_heat_conduction}
Lienhard, IV, J.~H. and Lienhard, V, J.~H. (2019).
\newblock {\em A Heat Transfer Textbook}.
\newblock Dover Publications, Mineola, NY, 5th edition.

\bibitem[Lim and Stanley{-}Marbell, 2018]{lim2018}
Lim, J. and Stanley{-}Marbell, P. (2018).
\newblock Newton: {A} language for describing physics.
\newblock {\em CoRR}, abs/1811.04626.

\bibitem[Liu and Nocedal, 1989]{liu1989lbfgs}
Liu, D.~C. and Nocedal, J. (1989).
\newblock On the limited memory bfgs method for large scale optimization.
\newblock {\em Mathematical programming}, 45(1-3):503--528.

\bibitem[Liu et~al., 2016]{Liu2016}
Liu, Z., Dou, Y., Jiang, J., and Xu, J. (2016).
\newblock Automatic code generation of convolutional neural networks in fpga
  implementation.
\newblock In {\em 2016 International Conference on Field-Programmable
  Technology (FPT)}, pages 61--68.

\bibitem[Lu et~al., 2021]{Lu2021}
Lu, L., Jin, P., Pang, G., Zhang, Z., and Karniadakis, G.~E. (2021).
\newblock Learning nonlinear operators via deeponet based on the universal
  approximation theorem of operators.
\newblock {\em Nature machine intelligence}, 3(3):218--229.

\bibitem[Ma et~al., 2017]{Ma2017}
Ma, Y., Cao, Y., Vrudhula, S., and Seo, J.-s. (2017).
\newblock Optimizing loop operation and dataflow in fpga acceleration of deep
  convolutional neural networks.
\newblock In {\em Proceedings of the 2017 ACM/SIGDA International Symposium on
  Field-Programmable Gate Arrays}, FPGA '17, page 45–54, New York, NY, USA.
  Association for Computing Machinery.

\bibitem[Meech and Stanley-Marbell, 2022]{Meech2022}
Meech, J.~T. and Stanley-Marbell, P. (2022).
\newblock An algorithm for sensor data uncertainty quantification.
\newblock {\em IEEE Sensors Letters}, 6(1):1--4.

\bibitem[Melexis, 2018]{MLX90640_library}
Melexis (2018).
\newblock mlx90640-library.
\newblock \url{https://github.com/melexis/mlx90640-library}.

\bibitem[{Melexis}, 2019]{Melexis:MLX90640}
{Melexis} (2019).
\newblock {\em MLX90640 32x24 IR array}.

\bibitem[Misyris et~al., 2020]{misyris2020}
Misyris, G.~S., Venzke, A., and Chatzivasileiadis, S. (2020).
\newblock Physics-informed neural networks for power systems.
\newblock In {\em 2020 IEEE Power \& Energy Society General Meeting (PESGM)},
  pages 1--5.

\bibitem[Moin et~al., 2021]{moin2021}
Moin, A., Zhou, A., Rahimi, A., Menon, A., Benatti, S., Alexandrov, G.,
  Tamakloe, S., Ting, J., Yamamoto, N., Khan, Y., et~al. (2021).
\newblock A wearable biosensing system with in-sensor adaptive machine learning
  for hand gesture recognition.
\newblock {\em Nature Electronics}, 4(1):54--63.

\bibitem[Moseley, 2021]{Moseley2021}
Moseley, B. (2021).
\newblock harmonic-oscillator-pinn.
\newblock \url{https://github.com/benmoseley/harmonic-oscillator-pinn}.

\bibitem[Moseley, 2022]{moseley2022a}
Moseley, B. (2022).
\newblock {\em Physics-informed machine learning: from concepts to real-world
  applications}.
\newblock PhD thesis, University of Oxford.

\bibitem[Moseley et~al., 2021]{moseley2021finite}
Moseley, B., Markham, A., and Nissen-Meyer, T. (2021).
\newblock Finite basis physics-informed neural networks (fbpinns): a scalable
  domain decomposition approach for solving differential equations.

\bibitem[Moss et~al., 2017]{moss2017}
Moss, D. J.~M., Nurvitadhi, E., Sim, J., Mishra, A., Marr, D., Subhaschandra,
  S., and Leong, P. H.~W. (2017).
\newblock High performance binary neural networks on the xeon+fpga™ platform.
\newblock In {\em 2017 27th International Conference on Field Programmable
  Logic and Applications (FPL)}, pages 1--4.

\bibitem[Nakahara et~al., 2017]{Nakahara2017}
Nakahara, H., Fujii, T., and Sato, S. (2017).
\newblock A fully connected layer elimination for a binarizec convolutional
  neural network on an fpga.
\newblock In {\em 2017 27th International Conference on Field Programmable
  Logic and Applications (FPL)}, pages 1--4.

\bibitem[{NXP Semiconductors}, 2021]{NXP:I2C}
{NXP Semiconductors} (2021).
\newblock {\em I2C-bus specification and user manual}.
\newblock Rev. 7.0.

\bibitem[{Panasonic}, 2017]{Panasonic:AMG8833}
{Panasonic} (2017).
\newblock {\em Infrared Array Sensor Grid-EYE (AMG88)}.

\bibitem[Pappalardo, 2023]{brevitas}
Pappalardo, A. (2023).
\newblock Xilinx/brevitas.

\bibitem[Paszke et~al., 2019]{pytorch2019}
Paszke, A., Gross, S., Massa, F., Lerer, A., Bradbury, J., Chanan, G., Killeen,
  T., Lin, Z., Gimelshein, N., Antiga, L., Desmaison, A., Kopf, A., Yang, E.,
  DeVito, Z., Raison, M., Tejani, A., Chilamkurthy, S., Steiner, B., Fang, L.,
  Bai, J., and Chintala, S. (2019).
\newblock Pytorch: An imperative style, high-performance deep learning library.
\newblock In {\em Advances in Neural Information Processing Systems 32}, pages
  8024--8035. Curran Associates, Inc.

\bibitem[Patterson, 2018]{patterson2018}
Patterson, D. (2018).
\newblock 50 years of computer architecture: From the mainframe cpu to the
  domain-specific tpu and the open risc-v instruction set.
\newblock In {\em 2018 IEEE International Solid - State Circuits Conference -
  (ISSCC)}, pages 27--31.

\bibitem[{Plotly Technologies Inc.}, 2015]{plotly}
{Plotly Technologies Inc.} (2015).
\newblock Collaborative data science.
\newblock \url{https://plot.ly}.

\bibitem[Qasaimeh et~al., 2019]{Qasaimeh2019}
Qasaimeh, M., Denolf, K., Lo, J., Vissers, K., Zambreno, J., and Jones, P.~H.
  (2019).
\newblock Comparing energy efficiency of cpu, gpu and fpga implementations for
  vision kernels.
\newblock In {\em 2019 IEEE International Conference on Embedded Software and
  Systems (ICESS)}, pages 1--8.

\bibitem[Qiu et~al., 2016]{Qiu2016}
Qiu, J., Wang, J., Yao, S., Guo, K., Li, B., Zhou, E., Yu, J., Tang, T., Xu,
  N., Song, S., Wang, Y., and Yang, H. (2016).
\newblock Going deeper with embedded fpga platform for convolutional neural
  network.
\newblock In {\em Proceedings of the 2016 ACM/SIGDA International Symposium on
  Field-Programmable Gate Arrays}, FPGA '16, page 26–35, New York, NY, USA.
  Association for Computing Machinery.

\bibitem[Raissi and Karniadakis, 2018]{Raissi2018}
Raissi, M. and Karniadakis, G.~E. (2018).
\newblock Hidden physics models: Machine learning of nonlinear partial
  differential equations.
\newblock {\em Journal of Computational Physics}, 357:125--141.

\bibitem[Raissi et~al., 2019]{raissi2019}
Raissi, M., Perdikaris, P., and Karniadakis, G. (2019).
\newblock Physics-informed neural networks: A deep learning framework for
  solving forward and inverse problems involving nonlinear partial differential
  equations.
\newblock {\em Journal of Computational Physics}, 378:686--707.

\bibitem[Rajkumar et~al., 2010]{Rajkumar2010}
Rajkumar, R.~R., Lee, I., Sha, L., and Stankovic, J. (2010).
\newblock Cyber-physical systems: The next computing revolution.
\newblock In {\em Proceedings of the 47th Design Automation Conference}, DAC
  '10, page 731–736, New York, NY, USA. Association for Computing Machinery.

\bibitem[{Robert Bosch GmbH}, 2021]{bosch:BNO055}
{Robert Bosch GmbH} (2021).
\newblock {\em BNO055: Intelligent 9-axis absolute orientation sensor}.
\newblock Rev. 1.8.

\bibitem[Rohrhofer et~al., 2021]{Rohrhofer2021}
Rohrhofer, F.~M., Posch, S., and Geiger, B.~C. (2021).
\newblock On the pareto front of physics-informed neural networks.
\newblock {\em CoRR}, abs/2105.00862.

\bibitem[Sahli~Costabal et~al., 2020]{sahli2020physics}
Sahli~Costabal, F., Yang, Y., Perdikaris, P., Hurtado, D.~E., and Kuhl, E.
  (2020).
\newblock Physics-informed neural networks for cardiac activation mapping.
\newblock {\em Frontiers in Physics}, 8:42.

\bibitem[Savitzky and Golay, 1964]{savgol1964}
Savitzky, A. and Golay, M. J.~E. (1964).
\newblock Smoothing and differentiation of data by simplified least squares
  procedures.
\newblock {\em Analytical Chemistry}, 36:1627--1639.

\bibitem[Schwarting et~al., 2018]{Schwarting2018}
Schwarting, W., Alonso-Mora, J., and Rus, D. (2018).
\newblock Planning and decision-making for autonomous vehicles.
\newblock {\em Annual Review of Control, Robotics, and Autonomous Systems},
  1(1):187--210.

\bibitem[Shen et~al., 2018]{Shen2018}
Shen, J., Huang, Y., Wang, Z., Qiao, Y., Wen, M., and Zhang, C. (2018).
\newblock Towards a uniform template-based architecture for accelerating 2d and
  3d cnns on fpga.
\newblock In {\em Proceedings of the 2018 ACM/SIGDA International Symposium on
  Field-Programmable Gate Arrays}, FPGA '18, page 97–106, New York, NY, USA.
  Association for Computing Machinery.

\bibitem[Simonyan and Zisserman, 2015]{simonyan_vgg16}
Simonyan, K. and Zisserman, A. (2015).
\newblock Very deep convolutional networks for large-scale image recognition.
\newblock In {\em International Conference on Learning Representations}.

\bibitem[Stefan, 1891]{stefan1891theorie}
Stefan, J. (1891).
\newblock {\"U}ber die theorie der eisbildung, insbesondere {\"u}ber die
  eisbildung im polarmeere.
\newblock {\em Annalen der Physik}, 278(2):269--286.

\bibitem[Strauss, 2008]{strauss2008partial}
Strauss, W. (2008).
\newblock {\em Partial Differential Equations: An Introduction}.
\newblock Wiley.

\bibitem[Suda et~al., 2016]{Suda2016}
Suda, N., Chandra, V., Dasika, G., Mohanty, A., Ma, Y., Vrudhula, S., Seo,
  J.-s., and Cao, Y. (2016).
\newblock Throughput-optimized opencl-based fpga accelerator for large-scale
  convolutional neural networks.
\newblock In {\em Proceedings of the 2016 ACM/SIGDA International Symposium on
  Field-Programmable Gate Arrays}, FPGA '16, page 16–25, New York, NY, USA.
  Association for Computing Machinery.

\bibitem[Szegedy et~al., 2015]{szegedy_googlenet2015}
Szegedy, C., Liu, W., Jia, Y., Sermanet, P., Reed, S., Anguelov, D., Erhan, D.,
  Vanhoucke, V., and Rabinovich, A. (2015).
\newblock Going deeper with convolutions.
\newblock In {\em 2015 IEEE Conference on Computer Vision and Pattern
  Recognition (CVPR)}, pages 1--9, Los Alamitos, CA, USA. IEEE Computer
  Society.

\bibitem[Tessier et~al., 2015]{tessier2015reconfigurable}
Tessier, R., Pocek, K., and DeHon, A. (2015).
\newblock Reconfigurable computing architectures.
\newblock {\em Proceedings of the IEEE}, 103(3):332--354.

\bibitem[Tsoutsouras et~al., 2022]{tsoutsouras2022}
Tsoutsouras, V., Kaparounakis, O., Samarakoon, C., Bilgin, B., Meech, J., Heck,
  J., and Stanley-Marbell, P. (2022).
\newblock The laplace microarchitecture for tracking data uncertainty.
\newblock {\em IEEE Micro}, 42(4):78--86.

\bibitem[van Daalen et~al., 1993]{van_daalen1993}
van Daalen, M., Jeavons, P., and Shawe-Taylor, J. (1993).
\newblock A stochastic neural architecture that exploits dynamically
  reconfigurable fpgas.
\newblock In {\em [1993] Proceedings IEEE Workshop on FPGAs for Custom
  Computing Machines}, pages 202--211.

\bibitem[Venieris and Bouganis, 2016]{fpgaConvNet_Venieris2016}
Venieris, S.~I. and Bouganis, C.-S. (2016).
\newblock fpgaconvnet: A framework for mapping convolutional neural networks on
  fpgas.
\newblock In {\em 2016 IEEE 24th Annual International Symposium on
  Field-Programmable Custom Computing Machines (FCCM)}, pages 40--47.

\bibitem[Vipin and Fahmy, 2018]{vipin2018}
Vipin, K. and Fahmy, S.~A. (2018).
\newblock Fpga dynamic and partial reconfiguration: A survey of architectures,
  methods, and applications.
\newblock {\em ACM Comput. Surv.}, 51(4).

\bibitem[Virtanen et~al., 2020]{scipy2020}
Virtanen, P., Gommers, R., Oliphant, T.~E., Haberland, M., Reddy, T.,
  Cournapeau, D., Burovski, E., Peterson, P., Weckesser, W., Bright, J., {van
  der Walt}, S.~J., Brett, M., Wilson, J., Millman, K.~J., Mayorov, N., Nelson,
  A. R.~J., Jones, E., Kern, R., Larson, E., Carey, C.~J., Polat, {\.I}., Feng,
  Y., Moore, E.~W., {VanderPlas}, J., Laxalde, D., Perktold, J., Cimrman, R.,
  Henriksen, I., Quintero, E.~A., Harris, C.~R., Archibald, A.~M., Ribeiro,
  A.~H., Pedregosa, F., {van Mulbregt}, P., and {SciPy 1.0 Contributors}
  (2020).
\newblock {{SciPy} 1.0: Fundamental Algorithms for Scientific Computing in
  Python}.
\newblock {\em Nature Methods}, 17:261--272.

\bibitem[Wang et~al., 2021]{Wang2021}
Wang, S., Wang, H., and Perdikaris, P. (2021).
\newblock On the eigenvector bias of fourier feature networks: From regression
  to solving multi-scale pdes with physics-informed neural networks.
\newblock {\em Computer Methods in Applied Mechanics and Engineering},
  384:113938.

\bibitem[Wang et~al., 2019]{wang2019}
Wang, Y., Willis, S., Tsoutsouras, V., and Stanley-Marbell, P. (2019).
\newblock Deriving equations from sensor data using dimensional function
  synthesis.
\newblock {\em ACM Trans. Embed. Comput. Syst.}, 18(5s).

\bibitem[Willink, 2013]{willink_2013}
Willink, R. (2013).
\newblock {\em Measurement Uncertainty and Probability}.
\newblock Cambridge University Press.

\bibitem[Xiao et~al., 2017]{Xiao2017}
Xiao, Q., Liang, Y., Lu, L., Yan, S., and Tai, Y.-W. (2017).
\newblock Exploring heterogeneous algorithms for accelerating deep
  convolutional neural networks on fpgas.
\newblock In {\em 2017 54th ACM/EDAC/IEEE Design Automation Conference (DAC)},
  pages 1--6.

\bibitem[{Xilinx}, 2021]{Xilinx:AXI-IIC}
{Xilinx} (2021).
\newblock {\em AXI IIC Bus Interface}.
\newblock v2.1.

\bibitem[Zhang et~al., 2016]{Zhang2016}
Zhang, C., Fang, Z., Zhou, P., Pan, P., and Cong, J. (2016).
\newblock Caffeine: Towards uniformed representation and acceleration for deep
  convolutional neural networks.
\newblock In {\em 2016 IEEE/ACM International Conference on Computer-Aided
  Design (ICCAD)}, pages 1--8.

\bibitem[Zhang et~al., 2022]{Zhang2022}
Zhang, L., Yan, X., and Ma, D. (2022).
\newblock A binarized neural network approach to accelerate in-vehicle network
  intrusion detection.
\newblock {\em IEEE Access}, 10:123505--123520.

\bibitem[Zhou et~al., 2017]{zhou2017incremental}
Zhou, A., Yao, A., Guo, Y., Xu, L., and Chen, Y. (2017).
\newblock Incremental network quantization: Towards lossless {CNN}s with
  low-precision weights.
\newblock In {\em International Conference on Learning Representations}.

\end{thebibliography}
